\documentclass{article} % For LaTeX2e
\usepackage{iclr2021_conference,times}

\usepackage{amsmath,amsfonts,bm}

\def\eqref#1{equation~\ref{#1}}
\def\1{\bm{1}}

\DeclareMathAlphabet{\mathsfit}{\encodingdefault}{\sfdefault}{m}{sl}
\SetMathAlphabet{\mathsfit}{bold}{\encodingdefault}{\sfdefault}{bx}{n}

\usepackage[hidelinks]{hyperref}
\usepackage{url}

\usepackage{graphicx}
\usepackage{amsmath}
\usepackage{amsfonts}
\usepackage{bbm}
\usepackage{multirow}
\usepackage{natbib}
\usepackage{algorithm}
\usepackage{algorithmic}
\usepackage{verbatim}
\usepackage{color}
\usepackage{booktabs}
\usepackage{subfigure}
\usepackage{wrapfig}

\usepackage{threeparttable}

\usepackage{array}
\newcolumntype{H}{>{\setbox0=\hbox\bgroup}c<{\egroup}@{}}
\newcommand{\citeyp}[1]{(\citeyear{#1})}

\title{DARTS-: Robustly Stepping out of Performance Collapse Without Indicators}

\iclrfinalcopy
\author{Xiangxiang Chu$^1$, Xiaoxing Wang$^{1,2}$\thanks{Work done as an intern at Meituan Inc.}, Bo Zhang$^1$, Shun Lu$^{1,3*}$,  Xiaolin Wei$^1$, Junchi Yan$^2$\thanks{Correspondent Author.} \\
	$^1$Meituan, $^2$Shanghai Jiao Tong University, $^3$University of Chinese Academy of Sciences\\
	\texttt{\{chuxiangxiang,zhangbo97,weixiaolin02\}@meituan.com} \\
	\texttt{\{figure1\_wxx,yanjunchi\}@sjtu.edu.cn} \\
    \texttt{lushun19@mails.ucas.ac.cn} \\
}

\begin{document}

\maketitle

\begin{abstract}

Despite the fast development of differentiable architecture search (DARTS), it suffers from long-standing performance instability, which extremely limits its application. Existing robustifying methods draw clues from the resulting deteriorated behavior instead of finding out its causing factor. Various indicators such as Hessian eigenvalues are proposed as a signal to stop searching before the performance collapses. However, these indicator-based methods tend to easily reject good architectures if the thresholds are inappropriately set, let alone the searching is intrinsically noisy. In this paper, we undertake a more subtle and direct approach to resolve the collapse. 
We first demonstrate that skip connections have a clear advantage over other candidate operations, where it can easily recover from a disadvantageous state and become dominant. We conjecture that this privilege is causing  degenerated performance. Therefore, we propose to factor out this benefit with an auxiliary skip connection, ensuring a fairer competition for all operations. We call this approach DARTS-.
Extensive experiments on various datasets verify that it can substantially improve robustness. Our code is available at \url{https://github.com/Meituan-AutoML/DARTS-}. %We achieve the state-of-the-art results on CIFAR-10 and ImageNet.

%Despite abundant variants of differentiable architecture search (DARTS \citep{liu2018darts}),  there is much less attention paid to its critical performance collapse issue, which greatly inhibits its use. Existing robustifying methods tend to draw clues from the outcome instead of finding out the causing factor. Various indicators such as Hessian eigenvalues are treated as a signal of performance collapse to stop searching when they reach a preset threshold.
%However,  these methods can easily reject good architectures if thresholds are inappropriately set, let alone the searching is intrinsically noisy in the meantime. 
%%However, it is common that the indicator fails to spot a good model. 
%In this paper, we undertake a more subtle and direct approach to resolve the collapse. 
%%First, we confirm the real cause of performance collapse for DARTS is that skip connections have an unfair advantage over other operations. 
%Firstly, we empirically demonstrate that it is a trend for skip connection to dominate the architecture importance in DARTS.
%Secondly, by factoring out this advantage with the help of an auxiliary skip connection, we expect all operations enter into a fairer competition, hence improving the credibility of searching results. It comes with no extra hyper-parameters, modifying only a line of DARTS's code. Finally, extensive experiments on various datasets verify that our approach can robustly step out of the collapse problem. %We achieve the state-of-the-art results on CIFAR-10 and ImageNet.
\end{abstract}

\section{Introduction}
Recent studies~\citep{zela2020understanding,liang2019darts,chu2019fair} have shown that one critical issue for differentiable architecture search \citep{liu2018darts} regarding the performance collapse due to superfluous skip connections. Accordingly, some empirical indicators for detecting the occurrence of collapse have been produced. R-DARTS \citep{zela2020understanding} shows that the loss landscape has more curvatures (characterized by higher Hessian eigenvalues w.r.t. architectural weights) when the derived architecture generalizes poorly. By regularizing for a lower Hessian eigenvalue, \cite{zela2020understanding,chen2020stabilizing} attempt to stabilize the search process. Meanwhile, by directly constraining the number of skip connections to a fixed number (typically 2), the collapse issue becomes less pronounced \citep{chen2019progressive,liang2019darts}.
%One of the most critical issues for differentiable architecture search \citep{liu2018darts} is that it suffers evident performance collapse due to superfluous skip connections \citep{zela2020understanding,liang2019darts,chu2019fair}.  Most recent work endeavors to design various indicators by studying the outcome when this collapse happens. For instance,  R-DARTS \citep{zela2020understanding} discovers that the loss landscape has more curvature (characterized by higher Hessian eigenvalues w.r.t. architectural weights) when the searched model generalizes poorly.  By regularizing for a lower Hessian eigenvalue, \citet{zela2020understanding,chen2020stabilizing} attempt to stabilize the searching process.  Meanwhile, by directly constraining the number of skip connections to a fixed number (typically 2), the collapse becomes less pronounced \citep{chen2019progressive,liang2019darts}.  
These indicator-based approaches have several main drawbacks. Firstly, robustness relies heavily on the quality of the indicator.  An imprecise indicator either inevitably accepts poor models or mistakenly reject good ones. Secondly, indicators impose strong priors by directly manipulating the inferred model, which is somewhat suspicious, akin to touching the test set. Thirdly, extra computing cost \citep{zela2020understanding} or careful tuning of hyper-parameters \citep{chen2019progressive,liang2019darts} are required. Therefore, it's natural to ask the following questions:

\begin{wrapfigure}{R}{0.5\columnwidth}
\vskip -5pt
	\centering
	\includegraphics[width=0.5\columnwidth]{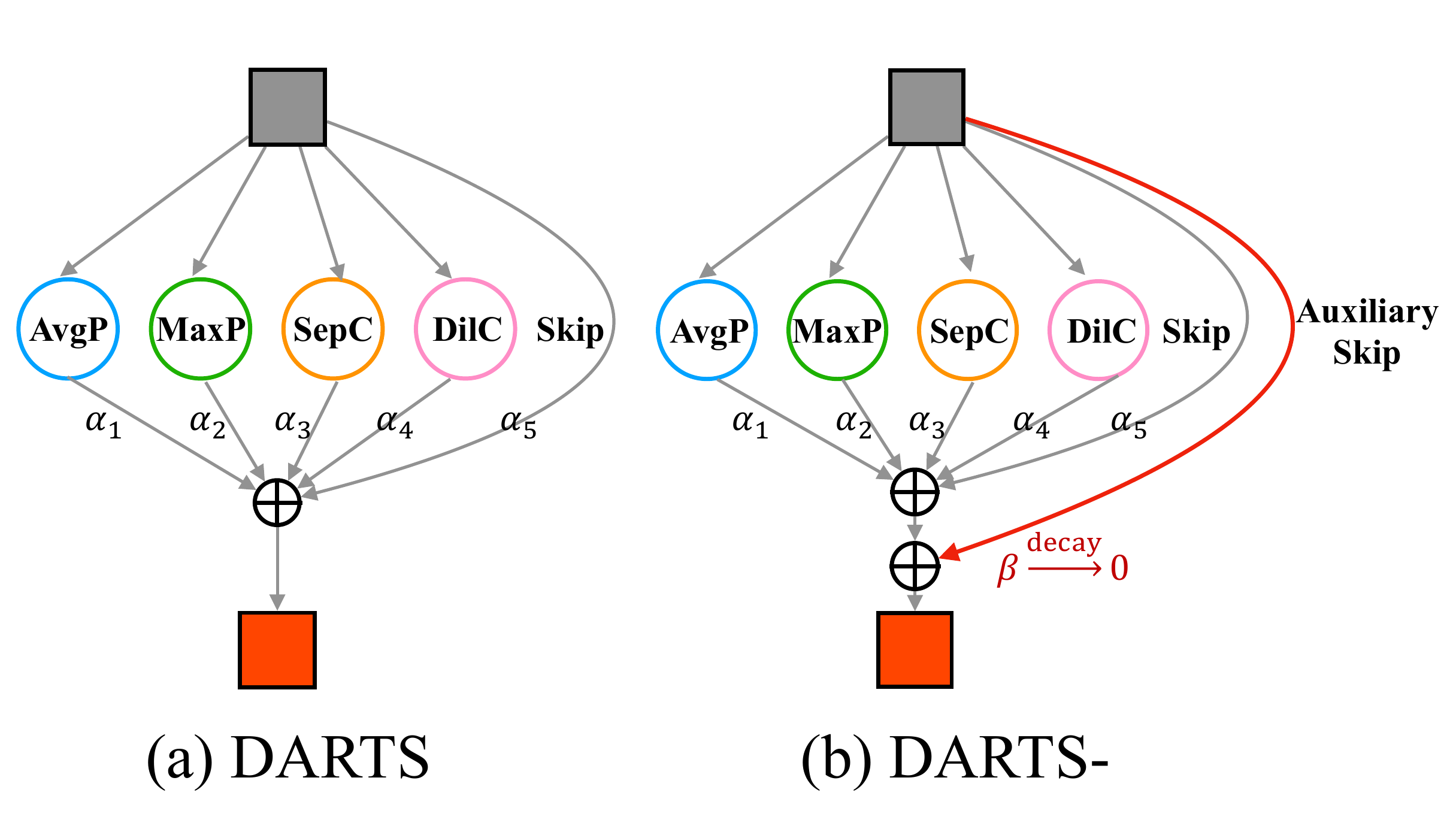}
	\vspace{-15pt}
	\caption{Schematic illustration about (a) DARTS and (b) the proposed DARTS-, featuring an auxiliary skip connection (thick red line) with a decay rate $\beta$ between every two nodes to remove the potential unfair advantage that leads to performance collapse.}
	\label{fig:darts-illustration}
	\vskip -15pt
\end{wrapfigure}

\begin{itemize}
	\item Can we resolve the collapse without handcrafted indicators and restrictions to interfere with the searching and/or discretization procedure?
	\item Is it possible to achieve robustness in DARTS without tuning extra hyper-parameters?
\end{itemize}

To answer the above questions, we propose an effective and efficient approach to stabilize DARTS. Our contributions can be summarized as follows:

\textbf{New Paradigm to Stabilize DARTS.} 
While empirically observing that current indicators ~\citep{zela2020understanding,chen2020stabilizing} can avoid performance collapse at a cost of reduced exploration coverage in the search space, we propose a novel \emph{indicator-free} approach to stabilize DARTS, referred to as DARTS-\footnote{We name it so as we undertake an inward way, as opposed to those outward ones who design new indicators, add extra cost and introduce new hyper-parameters.}, which involves an auxiliary skip connection (see Figure~\ref{fig:darts-illustration}) to remove the \emph{unfair advantage} \citep{chu2019fair} in the searching phase.

\textbf{Strong Robustness and Stabilization.} 
We conduct thorough experiments across seven search spaces and three datasets to demonstrate the effectiveness of our method. Specifically, our approach robustly obtains state-of-the-art results on 4 search space with 3$\times$ fewer search cost than R-DARTS \citep{zela2020understanding}, which requires four independent runs to report the final performance.

\textbf{Seamless Plug-in Combination with DARTS Variants.}  
We conduct experiments to demonstrate that our approach can work together seamlessly with other orthogonal DARTS variants by removing their handcrafted indicators without extra overhead. In particular, our approach is able to improve $0.8\%$ accuracy for P-DARTS, and $0.25\%$ accuracy for PC-DARTS on CIFAR-10 dataset.

%Not only can our method be used independently, but also it can work together seamlessly with other orthogonal DARTS variants. Their handcrafted collapse-addressing strategies can be effortlessly replaced by our indicator-free method without extra overhead.
%\begin{comment}
%\begin{itemize}
%	\item We propose the first indicator-free approach to stabilize DARTS, which utilizes an auxiliary skip connection (see Fig.~\ref{fig:darts-illustration}) to remove the unfair advantage \citep{chu2019fair} in the search process. We call our novel and simple approach \textbf{DARTS-}. We name it so as we undertake an inward way, as opposed to those outward ones who design new indicators, add extra cost and introduce new hyper-parameters.
%	\item  While empirically discovering that the well-known Hessian eigenvalue \citep{zela2020understanding,chen2020stabilizing} may be far from a perfect indicator of the collapse, we disapprove the need of indicator-based approach and suggest a new paradigm of stabilizing DARTS. 
%	\item We conduct thorough experiments on several search spaces and various datasets (CIFAR-10, CIFAR-100, SVHN, ImageNet) show that our method is quite effective. It robustly obtains state-of-the-art results with 3$\times$ fewer search cost than R-DARTS \citep{zela2020understanding}. 
%\end{itemize}
%\end{comment}

%Our source code and training logs are also provided for reproducibility.
%We provide the source code and training logs to make our work reproducible.

\section{Related Work}

\textbf{Neural architecture search and DARTS variants.} 
Over the years, researchers have sought to automatically discover neural architectures for various deep learning tasks to relieve human from the tedious effort, ranging from image classification \citep{zoph2017learning}, objection detection \citep{ghiasi2019fpn}, image segmentation \citep{liu2019auto} to machine translation \citep{so2019evolved} etc. Among many proposed approaches, Differentiable Architecture Search \citep{liu2018darts} features weight-sharing and resolves the searching problem via gradient descent, which is very efficient and easy to generalize. A short description of DARTS can be found in \ref{app:prelim}. Since then, many subsequent works have been dedicated to accelerating the process \citep{dong2019searching}, reducing memory cost \citep{xu2020pcdarts}, or fostering its ability such as hardware-awareness \citep{cai2018proxylessnas,wu2018fbnet}, finer granularity \citep{mei2020atomnas} and so on. However, regardless of these endeavors, a fundamental issue of DARTS over its searching performance collapse remains not properly solved, which extremely prohibits its application.
%, which asks for a thorough investigation, otherwise, other progress might be largely discounted.

%\textbf{Robustifying approaches to address the performance collapse in DARTS.}  
\textbf{Robustifying  DARTS.}  As DARTS \citep{liu2018darts} is known to be unstable as a result of performance collapse \citep{chu2019fair}, some recent works have devoted to resolving it by either designing indicators like Hessian eigenvalues for the collapse \citep{zela2020understanding} or adding perturbations to regularize such an indicator \citep{chen2020stabilizing}. Both methods rely heavily on the indicator's accurateness, i.e., to what extent does the indicator  correlate with the performance collapse? Other methods like Progressive DARTS \citep{chen2019progressive} and DARTS+ \citep{liang2019darts} employ a strong human prior, i.e., limiting the number of skip-connections to be a fixed value. Fair DARTS \citep{chu2019fair} argues that the collapse results from the \emph{unfair advantage} in an exclusive competitive environment, from which skip connections overly benefit to cause an abundant aggregation. To suppress such an advantage from overshooting, they convert the competition into collaboration where each operation is independent of others. It is however an indirect approach. SGAS \citep{li2019sgas}, instead, circumvents the problem with a greedy strategy where the unfair advantage can be prevented from taking effect. Nevertheless, potentially good operations might be pruned out too early because of greedy underestimation.

\section{DARTS-}

\begin{comment}
Following one-shot based NAS~ \citep{one-shot NAS}, DARTS~ \citep{liu2018darts} constructs neural networks by stacking normal and reduction cells and utilizes a directed acyclic graph (DAG) to represent the architecture of a cell. $N$ nodes, which represents a feature map, are contained in each cell in sequential order, and node $i$ connects with all the previous nodes in the same cell. We denote the edge from node $i$ to $j$ as $e_{i,j}$, which contains all the candidate operations in the search space $o_{i,j}, o\in \mathcal{O}$. Furthermore, DARTS~ \citep{liu2018darts} leads in the architecture parameters $\alpha_{i,j}^o$ to control the importance of different operations and connections. Consequently, the output of edge $e_{i,j}$, denoted as $\overline{o}_{i,j}$, is the weighted average of the output of operations $o_{i,j}$
\begin{equation}
\overline{o}_{i,j}(x_i) = \sum_{o\in \mathcal{O}}\frac{\exp(\alpha^o_{i,j})}{\sum_{o'\in \mathcal{O}}\exp(\alpha^{o'}_{i,j})} o_{i,j}(x_i)
\end{equation} 
where $x_i$ is the output of node $i$, and the output of node $j$ can be computed as:
\begin{equation}
x_j = \sum_{i<j} \overline{o}_{i,j}(x_i)
\end{equation} 
Neural architecture search can be modeled as a bilevel optimization problem as Eq.~\ref{eq:}, where $\omega$ is the network parameters and $\alpha$ is the architecture parameters:
\begin{align}
\min_\alpha  \quad &\mathcal{L}_{val}(\omega^*(\alpha), \alpha) \\
\text{s.t.} \quad &\omega^*(\alpha) = \text{argmin}_\omega \mathcal{L}_{train}(\omega, \alpha)
\end{align}
\end{comment}

\subsection{Motivation}
We start from the detailed analysis of the role of skip connections. Skip connections were proposed to construct a residual block in ResNet~ \citep{he2016deep}, which  significantly improves training stability.  It is even possible to deepen the network up to hundreds of layers without accuracy degradation by simply stacking them up. In contrast, stacking the plain blocks of VGG has degenerated performance when the network gets deeper. Besides, \cite{ren2015faster, wei2017boosting,tai2017image,li2018multi} also empirically demonstrate that deep residual network can achieve better performance on various tasks. 

From the gradient flow's perspective, skip connection is able to alleviate the gradient vanishing problem. %, which eases the training difficulty. 
Given a stack of $n$ residual blocks, the output of the $(i+1)^{\text{th}}$ residual block $X_{i+1}$ can be computed as
$X_{i+1} = f_{i+1}(X_{i}, W_{i+1}) + X_{i}$, 
where $f_{i+1}$ denotes the operations of the $(i+1)^{\text{th}}$ residual block with weights $W_{i+1}$. Suppose the loss function of the model is $\mathcal{L}$, and the gradient of $X_{i}$ % with skip connection 
can be obtained as follows ($\mathbbm{1}$ denotes a tensor whose items are all ones):
\begin{align}
\frac{\partial \mathcal{L}}{\partial X_{i}} &= \frac{\partial \mathcal{L}}{\partial X_{i+1}} \cdot \left(\frac{\partial f_{i+1}}{\partial X_{i}} + \mathbbm{1} \right)
=  \frac{\partial \mathcal{L}}{\partial X_{i+j}} \cdot \prod_{k=1}^{j} \left(\frac{\partial f_{i+k}}{\partial X_{i+k-1}} + \mathbbm{1} \right)
\end{align}
We observe that the gradient of shallow layers always includes the gradient of deep layers, which mitigates the gradient vanishing of $W_i$. Formally we have,
\begin{align}
\frac{\partial \mathcal{L}}{\partial W_{i}} 
%&= \frac{\partial \mathcal{L}}{\partial X_{i+1}} \cdot \frac{\partial f_{i}}{\partial W_{i}}\notag \\
&=  \frac{\partial \mathcal{L}}{\partial X_{i+j}} \cdot \prod_{k=1}^{j} \left(\frac{\partial f_{i+k}}{\partial X_{i+k-1}} + \mathbbm{1} \right) \cdot \frac{\partial f_{i}}{\partial W_{i}}
\end{align}
To analyze how skip connect affects the performance of residual networks, we introduce a trainable coefficient $\beta$ on all skip connections in ResNet. Therefore, the gradient of $X_{i}$ is converted to:
\begin{align}
\frac{\partial \mathcal{L}}{\partial X_{i}} &= \frac{\partial \mathcal{L}}{\partial X_{i+1}} \cdot \left(\frac{\partial f_{i+1}}{\partial X_{i}} + \mathbf{\beta} \right)
\end{align}
Once $\beta < 1$, gradients of deep layers gradually vanish during the back-propagation (BP) towards shallow layers. Here $\beta$ controls the memory of gradients in BP to stabilize the training procedure.

\begin{wrapfigure}{R}{0.4\columnwidth}
\vskip -0.1 in
	\centering
	\includegraphics[width=0.4\columnwidth]{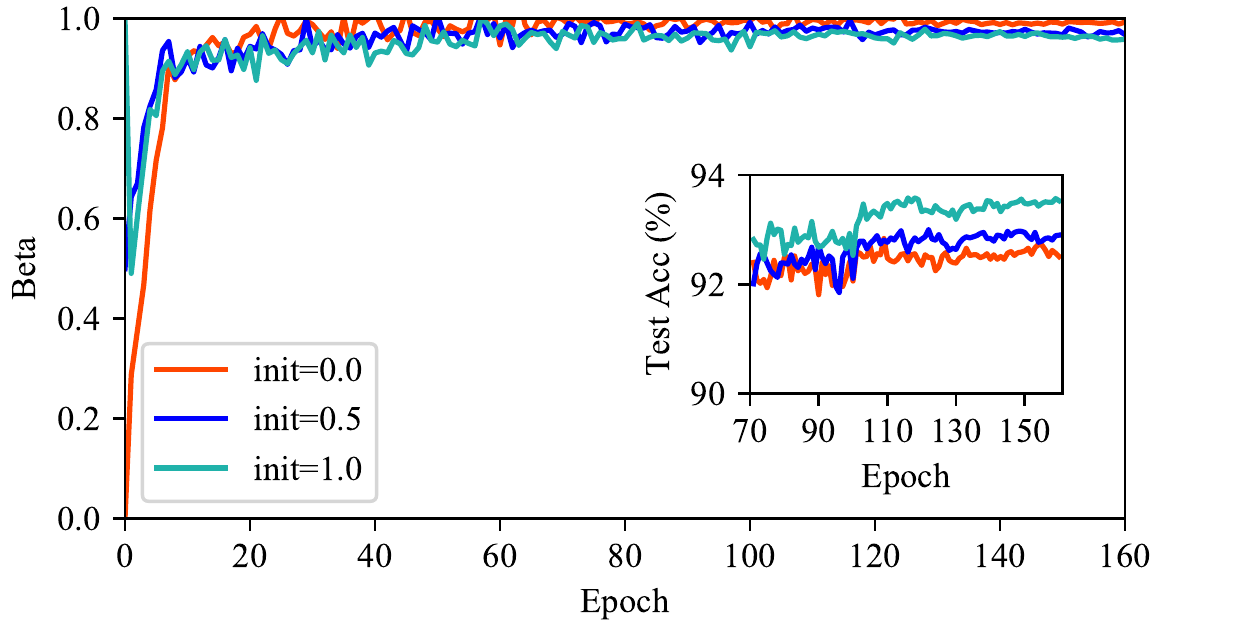} %figures/resnet_beta.pdfexample-image-duck
	\vspace{-15pt}
	\caption{Tendency of trainable coefficient $\beta$ (initialized with \{0, 0.5, 1\}) of the skip connection in ResNet50 and test accuracy (inset figure) vs. epochs. The residual structure is proved to learn a large $\beta$ to ease training in all three cases. All models are trained and tested on CIFAR-10.}
	\label{fig:resnet_beta}
\end{wrapfigure}
We conduct a confirmatory experiment on ResNet50 and show the result in Fig~\ref{fig:resnet_beta}. By initializing $\beta$ with $\{ 0, 0.5, 1.0 \}$, we can visualize the tendency of $\beta$ along with training epochs. We observe that $\beta$ converges towards $1$ after 40 epochs no matter the initialization, which demonstrates that the residual structure learns to push $\beta$ to a rather large value to alleviate gradient vanishing.

%Similarly in DARTS \citep{liu2018darts}, the trainable parameter $\beta_{skip}$ is involved in the searching phase to denote the importance of skip connection. It has been noticed that $\beta_{skip}$ is generally high, which is not desired as it leads to performance collapse. 
Similarly, DARTS \citep{liu2018darts} utilizes a trainable parameter $\beta_{skip}$ to denote the importance of skip connection.
However, In the search stage, $\beta_{skip}$ can generally increase and dominate the architecture parameters, and finally leads to performance collapse.
%The reason for a large $\beta_{skip}$ could differ: 
We analyze that a large $\beta_{skip}$ in DARTS could result from two aspects: On the one hand, as the supernet automatically learns to alleviate gradient vanishing, it pushes $\beta_{skip}$ to a proper large value; On the other hand, the skip connection is indeed an important connection for the target network, which should be selected in the discretization stage. As a consequence, the skip connection in DARTS plays two-fold roles: as \emph{an auxiliary connection to stabilize the supernet training}, and as \emph{a candidate operation to build the final network}. 
Inspired by the above observation and analysis, we propose to stabilize the search process by distinguishing the two roles of skip connection and handling the issue of gradient flow.

\subsection{Stepping out of the Performance Collapse}

To distinguish the two roles, we introduce an auxiliary skip connection between every two nodes in a cell, see Fig.~\ref{fig:darts-illustration} (b). 
On the one hand, the fixed auxiliary skip connection carries the function of stabilizing the supernet training, even when $\beta_{skip}$ is rather small. On the other hand, it also breaks the unfair advantage~\citep{chu2019fair} as the advantageous contribution from the residual block is factored out. 
Consequently, the learned architectural parameter $\beta_{skip}$ can be freed from the role of controlling the memory of gradients, and is more precisely aimed to represent the relative importance of skip connection as a candidate operation. In contrast to Eq.~\ref{eq:darts-node-softmax}, the output feature map of edge $e^{(i,j)}$ can now be obtained by Eq.~\ref{eq:ours-node-softmax}, where $\beta_o^{i,j} = \frac{\exp(\alpha_{o}^{(i,j)})}{\sum_{o' \in \mathcal{O}} \exp(\alpha_{o'}^{(i,j)})}$ denotes the normalized importance, and $\beta$ is a coefficient independent from the architecture parameters.

Moreover, to eliminate the impact of auxiliary connection on the discretization procedure, we propose to decrease $\beta$ to 0 in the search phase, and our method can be degenerated to standard DARTS at the end of the search.  Note that our method is not insensitive to the type of decay strategy, so we choose linear decay by default  for simplicity.
\begin{align}\label{eq:ours-node-softmax}
\bar{o}^{(i,j)}(x) &= \beta x + \sum_{o \in \mathcal{O}}\beta^{(i,j)}_o o(x) = \left(\beta+\beta^{(i,j)}_{skip}\right)x + \sum_{o \neq skip}\beta^{(i,j)}_o o(x) 
\end{align}
%On the other hand, {\color{red} show experiments to demonstrate that auxiliary skip connection can not affect the discretization stage: 1. there still skip connection selected; 2. adding auxiliary skip connection to the final model can hardly affect the ultimate performance}

We then analyze how the auxiliary skip connection handles the issue of gradient flow. Referring to the theorem of a recent work by ~\cite{zhou2020theory}, the convergence of network weight $\mathbf{W}$ in the supernet can heavily depend on $\beta_{skip}$. Specifically, suppose only three operations (none, skip connection, and convolution) are included in the search space and MSE loss is utilized as training loss, when architecture parameters $\beta^o_{i,j}$ are fixed to optimize $\mathbf{W}$ via gradient descent, training loss can decrease by ratio ($1-\eta \lambda/4$) at one step with probability at least $1-\delta$, where $\eta$ is the learning rate that should be bounded by $\delta$, and $\lambda$ follows Eq.~\ref{eq:theory_lambda}.
\begin{equation}
\lambda \propto \sum_{i=0}^{h-2} \left[ \left(\beta^{(i,h-1)}_{conv}\right)^2 \prod_{t=0}^{i-1}\left(\beta^{(t,i)}_{skip}\right)^2 \right]
\label{eq:theory_lambda}
\end{equation}
where $h$ is the number of layers of the supernet. From Eq.~\ref{eq:theory_lambda}, we observe that $\lambda$ relies much on $\beta_{skip}$ than $\beta_{conv}$, which indicates that the network weights $\mathbf{W}$ can converge faster with a large $\beta_{skip}$. 
% also expound a theorem from the view of convergence of network weights to demonstrate that $\alpha_{skip}$ can generally increase and dominate the architecture parameters in the search stage, and finally leads to performance collapse. Specifically, suppose only three operations (none, skip connection, and convolution) are included in the search space
However, by involving an auxiliary skip connection weighted as $\beta$, Eq.~\ref{eq:theory_lambda} can be refined as follows:
\begin{equation}
\lambda \propto \sum_{i=0}^{h-2} \left[ \left(\beta^{(i,h-1)}_{conv}\right)^2 \prod_{t=0}^{i-1}\left(\beta^{(t,i)}_{skip} + \beta \right)^2 \right]
\label{eq:theory_lambda_ours}
\end{equation}

\begin{wrapfigure}[17]{R}{0.52\textwidth}
\vspace{-22pt}
 \begin{minipage}{0.52\textwidth}
\begin{algorithm}[H]
	\caption{DARTS-}
	\label{alg:darts-minus}
	\begin{algorithmic}[1]
		\REQUIRE~~\\
		Network weights $w$; Architecture parameters $\alpha$;  \\
		Number of search epochs $E$; \\
		Decay strategy for $\beta_e, e\in\{ 1,2,..., E\}$.
                \ENSURE ~~\\
                Searched architecture parameters $\alpha$.
%		\STATE {\bfseries Input:} 
%		Network weights $w$; Architecture parameters $\alpha$;  Number of search epochs $E$; Decay strategy for $\beta(e), e\in\{ 1,2,..., E\}$.
		\STATE Construct a super-network by stacking cells in which there is an auxiliary skip connection between every two nodes of choice
%		\WHILE{\emph{not reach $E$}}
 		\FOR{each $e\in [1,E]$}
		\STATE Update weights $w$ by  $\nabla_{w}{\cal L}_{train}(w, \alpha, \beta_e)$
		\STATE Update parameters $\alpha$  by  $\nabla_{\alpha }{\cal L}_{val}(w, \alpha, \beta_e)$
		\ENDFOR
%		\ENDWHILE
		\STATE Derive the final architecture based on learned $\alpha$ from the best validation supernet.
	\end{algorithmic}
\end{algorithm}
\end{minipage}
%\vspace{-20pt}
\end{wrapfigure}
where $\beta \gg \beta_{skip}$ making $\lambda$ insensitive to $\beta_{skip}$, so that the convergence of network weights $\mathbf{W}$ depends more on $\beta_{conv}$. In the beginning of the search,  the common value for $\beta_{skip} $ is 0.15 while $\beta$ is 1.0. From the view of convergence theorem \citep{zhou2020theory}, the auxiliary skip connection alleviates the privilege of $\beta_{skip}$ and equalize the competition among architecture parameters. Even when $\beta$ gradually decays, the fair competition still holds since network weights $\mathbf{W}$ have been converged to an optimal point. Consequently, DARTS- is able to stabilize the search stage of DARTS. 

Extensive experiments are performed to demonstrate the efficiency of the proposed auxiliary skip connection, and we emphasize that our method is flexible to combine with other methods to further improve the stabilization and searching performance. The overall algorithm is given in Alg.~\ref{alg:darts-minus}.

\subsection{Relationship to Prior Work}
Our  method is aimed to address the performance collapse in differentiable neural architecture search. Most previous works \citep{zela2020understanding,chen2020stabilizing,liang2019darts} concentrate on developing various criteria or indicators characterizing the occurrence of collapse. Whereas, we don't study or rely on these indicators because they can mistakenly reject good models. Inspired by \cite{chu2019fair}, our method focuses on calibrating the biased searching process. The underlying philosophy is simple: if the biased process is rectified, the searching result will be better.  In summary, our method differs from others in two aspects: being process-oriented and indicator-free. Distinct from \cite{chu2019fair} that tweaks the competitive environment, our method can be viewed as one that breaks the unfair advantage. 
%PC-DARTS \citep{xu2020pcdarts}  proposes partial channel connections for architectural optimization to reduce memory cost where other channels remain unchanged, which is very akin to being handled by skip connection.  Although our auxiliary skip connection works on full channels, to save memory we can also apply only on the partial ones. 
Moreover, we don't introduce any handcrafted indicators to represent performance collapse, thus greatly reducing the burden of shifting to different tasks.

\section{Experiments} \label{sec:exp}

\subsection{Search Spaces and Training Settings}\label{sec: training setting}

For searching and evaluation in the standard DARTS space (we name it as \textbf{S0} for simplicity), we keep the same settings as in DARTS \citep{liu2018darts}. We follow R-DARTS \citep{zela2020understanding} for their proposed reduced spaces \textbf{S1- S4} (harder than S0). However, the inferred models are trained with two different settings from R-DARTS \citep{zela2020understanding} and SDARTS \citep{chen2020stabilizing}. The difference lies in the number of layers and initial channels for evaluation on CIFAR-100. R-DARTS sets 8 layers and 16 initial channels. Instead, SDARTS uses 20 and 36 respectively. For the proxyless searching on ImageNet, we instead search in MobileNetV2-like search space (we name it \textbf{S5}) proposed in FBNet \citep{wu2018fbnet}. We use the SGD optimizer for weight  and Adam ($\beta_1=0.5$ and $\beta_2=0.999$, 0.001 learning rate) for architecture parameters with the batch-size of 768. The initial learning rate is 0.045 and decayed to 0 within 30 epochs following the cosine schedule.  We also use L2 regularization with 1e-4.  It takes about 4.5 GPU days on Tesla V100.  More details are provided in the appendix.  We also use NAS-Bench-201 (\textbf{S6}) since DARTS performs severely bad. In total, we use 7 different search spaces to conduct the experiments, which involves three datasets.

\subsection{Searching Results}

\begin{wraptable}{r}{6.5cm}
\vspace{-20pt}
\caption{Comparison of searched CNN in the DARTS search space on two different datasets.}\smallskip
\centering
\resizebox{.45\columnwidth}{!}{
\smallskip\begin{tabular}{lrrr}
\toprule
\textbf{Dataset}  &  \textbf{DARTS}  &  \textbf{R-DARTS (L2)}  &  \textbf{Ours}\\
\midrule
C10 (S0)  &  2.91$\pm$0.25  &  2.95$\pm$0.21  & \textbf{2.63$\pm$0.07}\\
%\hline
C100 (S0)  &  20.58$\pm$0.44  &  18.01$\pm$0.26 &  \textbf{17.51$\pm$0.25} \\
%\hline
%SVHN  &  2.46$\pm$0.09  & 2.17$\pm$0.09 &  2.31$\pm$0.08\\
\bottomrule
\end{tabular}
}
\label{tab:CNN-standard-space}
\vspace{-10pt}
\end{wraptable}
\textbf{CIFAR-10 and CIFAR-100.}
Following the settings of R-DARTS \citep{zela2020understanding}, we obtain an average top-1 accuracy of  $97.36\%$  on CIFAR-10, as shown in Table~\ref{tab:comparison-cifar-imagenet}. Moreover, our method is very robust since out of six independent runs the searching results are quite stable. The best cells found on CIFAR-10 ($97.5\%$) are shown in Figure~\ref{fig:c10_best_cell} (\ref{app:fig-geno}). Results on CIFAR-100 are presented in Table~\ref{tab:comparison-cifar100} (see \ref{app:train}).
Moreover, our method has a much lower searching cost (\textbf{3$\times$ less}) than R-DARTS \citep{zela2020understanding}, where four independent searches with different regularization settings are needed to generate the best architecture. In other words, its robustness comes from the cost of more $CO_2$ emissions.

\begin{table}[tb!]
\setlength{\tabcolsep}{1pt}
	\begin{center}
		\caption{Comparison of the state-of-the-art models on CIFAR-10 (left) and ImageNet (right). On CIFAR-10 dataset, our average result is obtained on 5 independently searched models to assure the robustness. For ImageNet, networks in the top block are directly searched on ImageNet; the middle indicates that  architectures are searched on CIFAR-10 and then transferred to ImageNet; the bottom indicates models have SE and Swish. We search in S0 for CIFAR-10 and S5 for ImageNet. } \smallskip
		\label{tab:comparison-cifar-imagenet}
		\begin{scriptsize}
	 \begin{minipage}{0.48\textwidth}
	 		\vspace{0pt}
			\begin{threeparttable}
			\begin{tabular}{*{5}{l}} 	
				\toprule		
				\textbf{Models}   &  \textbf{\scriptsize{Params}}  &  \textbf{\scriptsize{FLOPs}}  &  \textbf{Acc}  &  \textbf{Cost}  \\
				 & \scriptsize{(M)}  & \scriptsize{(M)}  & \scriptsize{(\%)} & \tiny{GPU Days}   \\
				\midrule
				NASNet-A  \citeyp{zoph2017learning}   &  3.3  &  608$^\dagger$   &   97.35  &  2000  \\
				ENAS \citeyp{pham2018efficient}  &  4.6  &  626$^\dagger$ &  97.11  & 0.5    \\	
				%				MdeNAS \citeyp{zheng2019multinomial})  &  3.6  &  599$^\dagger$  &  97.45  &  MDL \\

				DARTS \citeyp{liu2018darts}  &  3.3  &  528$^\dagger$  &  97.00$\pm0.14^\star$  &  0.4  \\ 
				%Proxyless-R \citeyp{cai2018proxylessnas}   &  5.8  &  2.3  &  GD \\  
				SNAS \citeyp{xie2018snas}  &  2.8  &  422$^\dagger$  &  97.15$\pm0.02^\star$  &  1.5\\
				GDAS \citeyp{dong2019searching}  &  3.4  &  519$^\dagger$  &  97.07  & 0.2\\
%				SGAS (Cri.2 avg.) \citeyp{li2019sgas}  &  3.9  &  -  &  97.33$\pm0.21$  &  0.25  & GD\\
				P-DARTS \citeyp{chen2019progressive}  &  3.4  &  532$^\dagger$  &  97.5  &  0.3 \\
				PC-DARTS \citeyp{xu2020pcdarts}  &  3.6  &  558$^\dagger$ &  97.43  &  0.1  \\ 
				DARTS- (best) & 3.5  & 568  &  97.5  & 0.4\\
				\hline
				P-DARTS
				 \citeyp{chen2019progressive}$^\ddagger$   &  3.3$\pm$0.21  &  540$\pm$34  &  97.19$\pm$0.14  &  0.3  \\

%				PC-DARTS \citeyp{xu2020pcdarts} $^\ddagger$  &  3.68$\pm$0.57  &  592$\pm$90  &  97.11$\pm$0.22  &  0.1 &  GD \\ 
				R-DARTS \citeyp{zela2020understanding}  & - &  - &  97.05$\pm$0.21 & 1.6 \\
				%				FairDARTS ( \citeyp{chu2019fair})  &  3.32$\pm$0.46   &  458$\pm$61  &  97.46$\pm0.05$  &  GD \\
				SDARTS-ADV \citeyp{chen2020stabilizing}  & 3.3  & -  &  97.39$\pm$0.02  & 1.3\\
				% DARTS- (avg.) & 3.9  & 624  &  97.36$\pm$0.07  & 0.4\\ % always 1
				DARTS- (avg.) & 3.5$\pm$0.13  & 583$\pm$22  &  97.41$\pm$0.08  & 0.4\\ % linear decay

				\bottomrule
			\end{tabular}
			\begin{tablenotes}
			\footnotesize
			\item[$\dagger$] Based on the provided genotypes.
			\item[$^\ddagger$] 5 independent searches using their released code.
			\item[$\star$]Training the best searched model for several times (\emph{whose average doesn't indicate the stability of the method})
			\end{tablenotes}
			\end{threeparttable}
			\end{minipage}
			\begin{minipage}{0.5\textwidth}
			\vspace{-2pt}
			\begin{threeparttable}
			\begin{tabular}{*{2}{l}*{3}{l}c} 			
				\toprule
				\textbf{Models} & \textbf{FLOPs}  & \textbf{Params} & \textbf{Top-1} & \textbf{Top-5}  & \textbf{Cost} \\
				& \scriptsize{(M)} & \scriptsize{(M)} & \scriptsize{(\%)} & \scriptsize{(\%)} & \scriptsize{(GPU days)}  \\
%				\hline 
%				MobileNetV2(1.4) \citeyp{sandler2018mobilenetv2}   & 585 & 6.9 & 74.7 \\
				\midrule
				AmoebaNet-A \citeyp{real2019regularized} & 555  & 5.1 & 74.5& 92.0& 3150 \\
				MnasNet-92 \citeyp{tan2018mnasnet}  & 388 & 3.9 & 74.79  &92.1& \ \ 3791$^\ddagger$ \\ 	
				FBNet-C \citeyp{wu2018fbnet}   & 375 & 5.5 &  74.9 & 92.3& 9 \\ 
				FairNAS-A \citeyp{chu2019fairnas} &388 & 4.6 & 75.3 & 92.4 & 12 \\
				SCARLET-C \citeyp{chu2019scarletnas}  & 365 & 6.7 & 76.9 & 93.4 & 10 \\
				FairDARTS-D \citeyp{chu2019fair} & 440 & 4.3 & 75.6& 92.6&3 \\
				PC-DARTS \citeyp{xu2020pcdarts} & 597 & 5.3 & 75.8 & 92.7 & 3.8 \\
				\textbf{DARTS- (ours) }& 467 & 4.9 & 76.2 & 93.0 & 4.5\\
%				Proxyless-R \citeyp{cai2018proxylessnas} & 320$^*$ & 4.0  & 74.6  & 8.3 \\  
				%				Proxyless GPU $^\ddagger$ \citeyp{cai2018proxylessnas}  & 465$^*$ & 7.1  & 75.1 & 92.4 \\
				\midrule
				NASNet-A \citeyp{zoph2017learning}  & 564 & 5.3 &74.0 & 91.6& 2000 \\
%				PNAS \citeyp{liu2018progressive} & 588 & 5.1 & 74.2 & 91.9 &225 \\ 
				DARTS \citeyp{liu2018darts} & 574 & 4.7 & 73.3 & 91.3 &0.4 \\
				SNAS \citeyp{xie2018snas}  &522&4.3&72.7 & 90.8&1.5 \\
%				GDAS \citeyp{dong2019searching}  &581&5.3&74.0 & 91.5 & 0.2 \\
%			MdeNAS\citeyp{zheng2019multinomial} & - & 6.1 & 74.5 & 92.1 & 0.16 \\
%								P-DARTS $^{\dagger\dagger}$\citeyp{chen2019progressive}& 577 & 5.1 & 74.9$^{*}$ & 92.3& 0.3   \\  %75.3 & 92.5
				PC-DARTS  \citeyp{xu2020pcdarts} & 586 & 5.3 & 74.9 & 92.2 &0.1 \\ 
				FairDARTS-B \citeyp{chu2019fair}& 541 & 4.8 &75.1 & 92.5 & 0.4 \\

				\midrule
				MobileNetV3 \citeyp{howard2019searching} & 219 & 5.4 &75.2 &92.2&$\approx$3000 \\
%				GhostNet 1.3$\times$ \citeyp{han2020ghostnet} & 226 & 7.3 & 75.7  \\
				MoGA-A \citeyp{chumoga} & 304 & 5.1 & 75.9 & 92.8 & 12 \\
				MixNet-M \citeyp{tan2020mixconv} &360 & 5.0 & 77.0 & 93.3& $\approx$3000\\
				EfficientNet B0 \citeyp{tan2019efficientnet} &390 & 5.3  & 76.3 &93.2 &$\approx$3000 \\
				NoisyDARTS-A$^{\diamond}$ &449 & 5.5 & 77.9 & 94.0 & 12\\
			     	%FairDARTS-C& 386 & 5.3 & 77.2 & 93.5 & 3\\
				%			FairDARTS-D \\
				%			FairDARTS-B-SE$^\star$ & 380 &7.2 & & &3\\
				%			FairDARTS-C-SE$^\star$ & 440 &7.4 &76.5 & 93.2 &3\\
				
				%DARTS+ \citeyp{liang2019darts}$^\ddagger$ & 582 & 5.1 & 76.1 & 92.6 \\
				\textbf{DARTS- (ours)}$^\diamond$ & 470  & 5.5 &  77.8& 93.9 &4.5 \\
				\bottomrule
			\end{tabular}
			\begin{tablenotes}
			\scriptsize
			\item[$\ddagger$] Estimated by \cite{wu2018fbnet}.
			\item[${\diamond}$] SE modules and Swish enabled. 
			\end{tablenotes}
			\end{threeparttable}
			\end{minipage}
		\end{scriptsize}
	\end{center}
	\vskip -0.2 in
\end{table}

\textbf{ImageNet.}
To further verify the efficiency of DARTS-, we directly search on ImageNet in S5 and compare our results with the state-of-the-art models under the mobile setting in Table~\ref{tab:comparison-cifar-imagenet}. The visualization of the architecture is given in Fig~\ref{fig:darts-a-arch-imagnet}. DARTS-A obtains 76.2$\%$ top-1 accuracy on the ImageNet validation dataset. By contrast, direct applying DARTS on this search space only obtains $66.4\%$ \citep{chu2019fair}. Moreover, it obtains 77.8$\%$ top-1 accuracy after being equipped with auto-augmentation \citep{cubuk2018autoaugment} and squeeze-and-excitation \citep{hu2018squeeze}, which are also used in EfficientNet.

\begin{table*}[tb!]
	\setlength{\tabcolsep}{2pt}
	\begin{center}
		\caption{Searching performance on NAS-Bench-201 \citep{dong2020nasbench201}. Our method robustly obtains new SOTA. Averaged on 4 runs of searching. $^{1st}$: first-order, $^{2nd}$: second-order }
		\label{table:nas-bench-201}
		\small
		\begin{tabular}{lr*{6}{c}}
			\toprule
			Method	& Cost  & \multicolumn{2}{c}{CIFAR-10}  & \multicolumn{2}{c}{CIFAR-100}   & \multicolumn{2}{c}{ImageNet16-120}   \\
  & (hours) & valid & test & valid & test & valid & test \\
 \midrule
	DARTS$^{1st}$ \citeyp{liu2018darts} & 3.2 & 39.77$\pm$0.00 & 54.30$\pm$0.00 & 15.03$\pm$0.00 & 15.61$\pm$0.00 & 16.43$\pm$0.00 & 16.32$\pm$0.00 \\
	DARTS$^{2nd}$ \citeyp{liu2018darts} & 10.2 & 39.77$\pm$0.00 & 54.30$\pm$0.00 & 15.03$\pm$0.00 & 15.61$\pm$0.00 & 16.43$\pm$0.00 & 16.32$\pm$0.00 \\
	GDAS \citeyp{dong2019searching}  &  8.7 & 89.89$\pm$0.08 & 93.61$\pm$0.09 & 71.34$\pm$0.04 & 70.70$\pm$0.30 & 41.59$\pm$1.33 & 41.71$\pm$0.98 \\
	SETN \citeyp{dong2019one} & 9.5 & 84.04$\pm$0.28 & 87.64$\pm$0.00 & 58.86$\pm$0.06 & 59.05$\pm$0.24 & 33.06$\pm$0.02 & 32.52$\pm$0.21   \\ 
%	NoisyDARTS \citeyp{chu2020noisy} & 3.2 & 90.26$\pm$0.22 & 93.49$\pm$0.25 & \textbf{71.36$\pm$0.21} & \textbf{71.55$\pm$0.51} & 42.47$\pm$0.00 & 42.34$\pm$0.06 \\
	DARTS- & 3.2 & \textbf{91.03$\pm$0.44}	& \textbf{93.80$\pm$0.40} &	\textbf{71.36$\pm$1.51} &	\textbf{71.53$\pm$1.51}  	& \textbf{44.87$\pm$1.46}	& \textbf{45.12$\pm$0.82} \\ 
	DARTS- (best) & 3.2 & 91.55 &	94.36&73.49&	73.51&	46.37&	46.34 \\
	optimal & n/a & 91.61 & 94.37 & 73.49 & 73.51 & 46.77 & 47.31 \\
			\bottomrule
		\end{tabular}
	\end{center}
	\vskip -0.15in
\end{table*}

\textbf{NAS-Bench-201.}
Apart from standard search spaces, benchmarking with known optimal in a limited setting is also recommended. NAS-Bench-201 \citep{dong2020nasbench201} consists of 15,625 architectures in a reduced DARTS-like search space, where it has 4 internal nodes and 5 operations per node. We compare our method with prior work in Table~\ref{table:nas-bench-201}. We search on CIFAR-10 and look up the ground-truth performance with found genotypes on various test sets. Remarkably, we achieve a new state of the art, the best of which almost touches the optimal.

\textbf{Transfer results on objection detection}
We further evaluate the transfer-ability of our models on down-stream object detection task by replacing the backbone of RetinaNet \citep{lin2017focal} on MMDetection toolbox platform \citep{chen2019mmdetection}. Specifically, with the same training setting as \cite{chu2019fair}, our model achieves  32.5\% mAP on the COCO dataset, surpassing other similar-sized models such as MobileNetV3, MixNet, and FairDARTS. The detailed results are shown in Appendix (Table~\ref{table:darts-coco-retina}).

\subsection{Orthogonal Combination with Other Variants}

Our method can be flexibly adapted to combine with prior work for further improvements. Here we investigate the joint outcome with two methods: P-DARTS  and PC-DARTS.

\label{sec:pdarts-discuss}
\begin{comment}
\begin{wraptable}{r}{5.5cm}
	\vspace{-20pt}
	\setlength{\tabcolsep}{1pt}
	\small
	\centering
	\caption{Comparison of P-DARTS removing the strong prior and combining DARTS-. The results are averaged over 3 independent experiments on C10.} 
	\smallskip\begin{tabular}{cp{1cm}p{1cm}c}
		\toprule
		&  \textbf{Remove Prior (M=2)} &  \textbf{With DARTS-} &  \textbf{Acc ($\%$)}\\
		\midrule
		\multirow{2}*{P-DARTS 
		} & \checkmark & $\times$ &  96.48$\pm$0.55 \\
		~    &  \checkmark & \checkmark &97.28$\pm$0.04 \\
		\bottomrule
	\end{tabular}
	\label{tab:pdarts-darts-}
	\vspace{-10pt}
\end{wraptable}
\end{comment}

\begin{wraptable}{r}{5.5cm}
	\vspace{-20pt}
	\setlength{\tabcolsep}{1pt}
	\small
	\centering
	\caption{We remove the strong constraints on \textit{the number of skip connections as 2 and dropout }(priors) for P-DARTS and compare its performance w/ and w/o DARTS-. } 
	\smallskip\begin{tabular}{ccc}
		\toprule
		\textbf{Method} & \textbf{Setting}& \textbf{Acc (\%)} \\
		\midrule
		P-DARTS & w/o priors  & 96.48$\pm$0.55 \\
		P-DARTS- & w/o priors &97.28$\pm$0.04 \\
		\bottomrule
	\end{tabular}
	\label{tab:pdarts-darts-}
	\vspace{-12pt}
\end{wraptable}

\textbf{Progressive DARTS (P-DARTS).} P-DARTS \citep{chen2019progressive} proposes a progressive approach to search gradually with deeper depths while pruning out the uncompetitive paths. Additionally, it makes use of some handcrafted criteria to address the collapse (the progressive idea itself cannot deal with it), for instance, they impose two strong priors by regularizing the number of skip connections $M$ as 2 as well as dropout. To be fair, we remove such a carefully handcrafted trick and run P-DARTS for several times. As a natural control group, we also combine DARTS- with P-DARTS. We run both experiments for 3 times on CIFAR-10 dataset in Table~\ref{tab:pdarts-darts-}. Without the strong priors, P-DARTS severely suffers from the collapse, where the inferred models contain an excessive number of skip connections.  Specifically, it has a very high test error ($3.42\%$ on average), even worse than DARTS.  However, P-DARTS can benefit greatly from the combination with DARTS-. The improved version (we call P-DARTS-) obtains much higher top-1 accuracy (\textbf{+0.8\%}) on CIFAR-10 than its baseline.

\begin{wraptable}{r}{6.3cm}
	\vspace{-20pt}
	\setlength{\tabcolsep}{1pt}
	\small
	\centering
	\caption{Comparison of PC-DARTS \textit{removing the strong prior }(i.e. channel shuffle) and combining DARTS-. The results are from 3 independent runs on CIFAR-10. The GPU memory cost is on a batch size of 256.} 
	\smallskip\begin{tabular}{lp{1cm}ccc}
		\toprule
		\textbf{Method}&  \textbf{Setting} &  \textbf{Acc} ($\%$) & \textbf{Memory} & \textbf{Cost} \\
		\midrule
		PC-DARTS  & $K=2$ &   97.09$\pm$0.14 & 19.9G & 3.75h \\
		PC-DARTS-    &  $K=2$ & 97.35$\pm0.02$ & 20.8G & 3.41h \\
		\bottomrule
	\end{tabular}
	\label{tab:pcdarts-darts-}
	\vspace{-15pt}
\end{wraptable}

\textbf{Memory Friendly DARTS (PC-DARTS).}
To alleviate the large memory overhead from the whole supernet, PC-DARTS \citep{xu2020pcdarts} selects the partial channel for searching.  The proportion hyper-parameter $K$ needs careful calibration to achieve a good result for specific tasks.  As a  byproduct, the search time is also reduced to 0.1 GPU days ($K$=4).  We use their released code and run repeated experiments across different seeds under the same  settings. 

To accurately evaluate the role of our method, we choose $K$=2 (a bad configuration in the original paper). We make comparisons between the original PC-DARTS and its combination with ours (named PC-DARTS-) in Table~\ref{tab:pcdarts-darts-}. PC-DARTS- can marginally boost the CIFAR-10 top-1 accuracy (+$0.26\%$ on average). The result also confirms that our method can make PC-DARTS less sensitive to its hyper-parameter $K$ while keeping its advantage of less memory cost and run time.

\subsection{Ablation Study}\label{sec:ablation}

\textbf{Robustness to Decay Strategy.} Our method is insensitive to the type of decay policy on $\beta$. We design extra two strategies as comparisons: \emph{cosine} and \emph{step} decay. They both have the similar performance. Specifically, when $\beta$ is scheduled to zero by the cosine strategy, the average accuracy of four searched CIFAR-10 models in S3  is 97.33\%$\pm$0.09, the best is 97.47\%.  The step decay at epoch 45 obtains $97.30\%$ top-1 accuracy on average in the same search space.

\begin{wraptable}{r}{5cm}
\vspace{-23pt}
	\caption{Searching performance on CIFAR-10 in S3 w.r.t the initial linear decay rate $\beta_0$. Each setting is run for three times.}
	%		The first row shows the number of skip connections in the normal cell out of 8 operations. 
	\smallskip
	\centering
	\smallskip\begin{tabular}{l*{2}{c}}
		\toprule
%		$\beta_0$  &  1  & 0.7 & 0.4 & 0.1 & 0 \\
    		%%		$N_{skip}$ & 2 & 3 & 5 & 6.6 & 7.3 \\ 
%		\midrule
%		Error & 2.65$\pm$0.04 & 2.76$\pm$0.16 & 3.04$\pm$0.19 &  3.11$\pm$0.16 & 4.58$\pm$1.30 \\
    				$\beta_0$ & Error (\%) \\
    				\midrule
    				1  & 2.65$\pm$0.04 \\
    				0.7 & 2.76$\pm$0.16 \\
    				0.4 &3.04$\pm$0.19 \\
    				0.1   & 3.11$\pm$0.16\\
    				0   & 4.58$\pm$1.3 \\
		\bottomrule
	\end{tabular}
	\label{tab:beta-sensitiveness}
	\vspace{-20pt}
\end{wraptable}

\textbf{Robustness Comparison on C10 and C100 in S0-S4.}
To verify the robustness, it is required to search several times to report the average performance of derived models \citep{Yu2020Evaluating,yang2020nas}.  As shown in Table~\ref{tab:CNN-standard-space}, Table~\ref{tab:comparison-rdarts-s2-s3-avg} and Table~\ref{tab:comparison-rdarts-s2-s3-best}, our method outperforms the recent SOTAs across several spaces and datasets. Note that SDARTS-ADV  utilizes adversarial training and requires 3$\times$ more search times than ours. Especially, we find a good model in $S_3$ on CIFAR-100 with the lowest top-1 test error  $15.86\%$. The architectures of these models can be found in the appendix.

\begin{table*}[t]
	\centering
	\caption{Comparison in various search spaces. We report the \textbf{lowest error rate} of 3 found architectures. $^\dagger$: under \cite{chen2020stabilizing}'s training settings where all models have 20 layers and 36 initial channels (the best is shown in boldface). $^\ddagger$:  under  \cite{zela2020understanding}'s settings where CIFAR-100 models have 8 layers and 16 initial channels (The best is in boldface and underlined). } \smallskip %The results of compared methods are from R-DARTS and SDARTS. 
	%\resizebox{.86\textwidth}{!}{
	\footnotesize
	\setlength{\tabcolsep}{3pt}
	\begin{tabular}{c*{7}{c}|c*{4}{c}}
		\toprule
		\multicolumn{2}{c}{ \multirow{2}*{\textbf{Benchmark}}} &   \multirow{2}*{\textbf{DARTS}$^\ddagger$}  &  \multicolumn{2}{c}{\textbf{R-DARTS}$^\ddagger$} & \multicolumn{2}{c}{\textbf{DARTS}$^\ddagger$} &  \multirow{2}*{\textbf{Ours}$^{\ddagger}$} & \multirow{2}*{\textbf{PC-DARTS}$^\dagger$} & \multicolumn{2}{c}{\textbf{SDARTS}$^\dagger$} &  \multirow{2}*{\textbf{Ours}$^\dagger$} \\
		\cmidrule(lr){4-5} \cmidrule(lr){6-7} \cmidrule(lr){10-11}
		 & & & DP & L2 & ES & ADA & & & RS & ADV &\\
		
		\midrule
		 \multirow{4}*{C10}& S1  & 3.84 & 3.11 & 2.78 & 3.01 & 3.10 & \textbf{2.68}  & 3.11 & 2.78 & 2.73 & \textbf{2.68} \\
		~  & S2 &  4.85 &  3.48 &  3.31 &  3.26 &  3.35 & \textbf{2.63} & 3.02 &  2.75 &  2.65 &  \textbf{2.63} \\
		%\cline{2-13}
		~     &  S3 &  3.34 &  2.93 &  2.51 &  2.74 &  2.59  &  \textbf{2.42} & 2.51 & 2.53 &  2.49 & \textbf{2.42}  \\
		~ & S4 & 7.20 & 3.58 & 3.56 & 3.71 & 4.84 & \textbf{2.86}  & 3.02 & 2.93 & 2.87 & \textbf{2.86} \\
		\midrule
		\multirow{4}*{C100} & S1 & 29.46 & 25.93 & 24.25  &  28.37 & 24.03 & \textbf{\underline{22.41}} & 18.87 & 17.02 & \textbf{16.88} & 16.92  \\ %SDARTS(v2) 23.51 & \textbf{22.33}
		~ &  S2 &  26.05 &  22.30 &  22.24 &  23.25 &  23.52 &  \textbf{\underline{21.61}} & 18.23 & 17.56 &  17.24  &  \textbf{16.14}   \\ % SDARTS(v2)  22.28 & 20.56 
		%\cline{2-13}
		~     &  S3 &  28.90 &  22.36 &  23.99 &  23.73 &  23.37  &  \textbf{\underline{21.13}} & 18.05 & 17.73 &  17.12   &  \textbf{15.86}\\ %SDARTS(v2) 21.09 & \textbf{21.08}
		~ & S4 & 22.85 & 22.18 & 21.94 & \textbf{\underline{21.26}} & 23.20 & 21.55  & 17.16 & 17.17 & \textbf{15.46} & 17.48   \\%SDARTS(v2) 21.46 & \textbf{21.25}
		\bottomrule
		\end{tabular}
		%}
	\label{tab:comparison-rdarts-s2-s3-best}
	\vskip -0.2 in
\end{table*}

\textbf{Sensitivity Analysis of $\beta$}\quad
The power of the auxiliary skip connection branch can be discounted by setting a lower initial $\beta$. We now evaluate how sensitive our approach is to the value of $\beta$. It's trivial to see that our approach degrades to DARTS when $\beta = 0$. We compare results when searching with $\beta \in \{1,0.7,0.4,0.1,0\}$ in Table~\ref{tab:beta-sensitiveness}, which show that a bigger $\beta_0$ is advantageous to obtain better networks.

\textbf{The choice of auxiliary branch}
Apart from the default skip connection serving as the auxiliary branch, we show that it is also effective to replace it with a learnable 1$\times$1 convolution projection, which is initialized with an identity tensor. The average accuracy of three searched CIFAR-10 models in S3 is 97.25\%$\pm$0.09. Akin to the ablation in \cite{he2016deep}, the projection convolution here works in a similar way as the proposed skip connection. This proves the necessity of the auxiliary branch.

\textbf{Performance with longer epochs. } It is claimed by  \cite{bi2019stabilizing} that much longer epochs lead to better convergence of the supernet, supposedly beneficial for inferring the final models. However, many of DARTS variants fail since their final models are full of skip connections.  We thus evaluate how our method behaves in such a situation.  Specifically, we extend the standard 50 epochs to 150, 200 and we search 3 independent times each for S0, S2 and S3. Due to the longer epochs, we slightly change our decay strategy, we keep $\beta=1$ all the way until for last 50 epochs we decay $\beta$ to 0. Other hyper-parameters are kept unchanged. The results are shown in Table~\ref{tab:long_epoch_search} and the found genotypes are listed in Figure~\ref{fig:c10_s5_decay_best_cells}, \ref{fig:c10_s2_e150_decay_cells}, \ref{fig:c10_s2_e200_decay_cells}, \ref{fig:c10_s3_e150_decay_cells} and \ref{fig:c10_s3_e200_decay_cells}. It indicates that DARTS- doesn't suffer from longer epochs since it has reasonable values for $\#P$ compared with those ($\#P=0$) investigated by \cite{bi2019stabilizing}. Notice S2 and S3 are harder cases where DARTS suffers more severely from the collapse than S0. As a result, \emph{DARTS- can successfully survive longer epochs even in challenging search spaces.} Noticeably, it is still unclear whether longer epochs can truly boost searching performance.  Although we achieve a new state-of-the-art result in S2 where the best model has 2.50\% error rate (previously 2.63\%), it still has worse average performance (2.71$\pm$0.11\%) in S0 than the models searched with 50 epochs (2.59$\pm$0.08\%), and the best model in S3 (2.53\%) is also weaker than before (2.42\%).

\begin{table}[ht]
	\setlength{\tabcolsep}{1pt}
	\caption{Searching performance on CIFAR-10 in S0, S2 and S3 using longer epochs. Following \cite{bi2019stabilizing}, $\#P$ means the number of parametric operators in the normal cell. Averaged on 3 runs of search.}
	\smallskip
	\centering
	\smallskip\begin{tabular}{c|*{3}{c}|*{3}{c}|*{3}{c}}
		\toprule
		Epoch  &  \multicolumn{3}{c|}{S0} & \multicolumn{3}{c|}{S2} & \multicolumn{3}{c}{S3} \\
		 &$\#P$  & \small{Params (M)} & \small{Error (\%)} &$\#P$  & \small{Params (M)} & \small{Error (\%)} &  $\#P$  & \small{Params (M)} & \small{Error (\%)} \\
		\midrule
		150 & 6.6$\pm$1.1 & 3.3$\pm$0.3& 2.74$\pm$0.06 & 6.0$\pm$0.0 & 3.9$\pm$0.3 & 2.58$\pm$0.11 & 6.0$\pm$1.0 & 3.6$\pm$0.3 & 2.55$\pm$0.03 \\
		200 & 7.3$\pm$0.6 & 3.2$\pm$0.3 & 2.71$\pm$0.11 &  8.0$\pm$0.0 & 4.3$\pm$0.1 & 2.65$\pm$0.21 & 7.6$\pm$0.5 & 4.3$\pm$0.2 & 2.66$\pm$0.09 \\
		\bottomrule
	\end{tabular}
	\label{tab:long_epoch_search}
\end{table}

Besides, compared with first-order DARTS with a cost of 0.4 GPU days in S0, Amended-DARTS \citep{bi2019stabilizing}, particularly designed to survive longer epochs, reports 1.7 GPU days even with pruned edges in S0. Our approach has the same cost as first-order DARTS, which is more efficient.

\section{Analysis and Discussions}
\subsection{Failure of Hessian Eigenvalue}\label{sec:failure of eigen}
The maximal Hessian eigenvalue calculated from the validation loss w.r.t $\alpha$ is regarded as an indicator of performance collapse \citep{zela2020understanding,chen2020stabilizing}.  Surprisingly, our method develops a growing eigenvalue in the majority of configurations, which conflicts with the previous observations. We visualize these statistics across different search space and datasets in Figure~\ref{fig:eigen_value} (\ref{sec:failure_eigen}).  Although eigenvalues increase almost monotonously and reach a relatively large value in the end, the final models still have good performance that matches with state-of-the-art ones (see Table~\ref{tab:comparison-rdarts-s2-s3-avg}). 
%It's also interesting to see that the eigenvalue trend in $S_2$ on CIFAR-10 is similar to that of R-DARTS (with high L2 regularization factor), where these values don't increase much during the search stage (the derived model has \textbf{2.79$\pm$0.04\%} test error). 
%{\color{red} there seems no Hessian Eigenvalue reported in the two Tables. I suggest we can report the Hessian Eigenvalue in the Table or give the value in this paragraph}
These models can be mistakenly deemed as bad ones or never visited according to the eigenvalue criteria. Our observations disclose one fatal drawback of these indicator-based approaches: \emph{they are prone to rejecting good models}. Further analysis can be found in \ref{sec:failure_eigen}.
%We suggest more attention should be paid on the cause of the collapse during the searching process, other than its result. Our method belongs to such process-oriented solutions, and we are thus free of inventing a new indicator for the failure modes. 

%\begin{figure}[ht]
%\centering
%\includegraphics[width=0.6\columnwidth]{figures/sampled-eigen-acc.pdf}
%\caption{Growing Hessian eigenvalues don't induce poor performance in DARTS-. Among the sampled five models, the one corresponding to the highest eigenvalue has the best performance.}
%\label{fig:sampled_eigenvalue_acc}
%\end{figure}

\subsection{Validation Accuracy Landscape}

Recent works, R-DARTS \citep{zela2020understanding} and SDARTS \citep{chen2020stabilizing} point that the architectural weights are expected to converge to an optimal point where accuracy is insensitive to perturbations to obtain a stable architecture after the discretization process, i.e., the convergence point should have a smooth landscape. SDARTS  proposes a perturbation-based regularization, which further stabilizes the searching process of DARTS. However, the perturbation regularization disturbs the training procedure and thus misleads the update of architectural weights. 
Different from SDARTS that explicitly smooths landscape by perturbation, DARTS- can implicitly do the same without directly perturbing architectural weights.

\begin{figure*}[ht]
	\centering
%	\vskip -0.1 in
	\subfigure[DARTS]{
		\includegraphics[width=0.23\columnwidth]{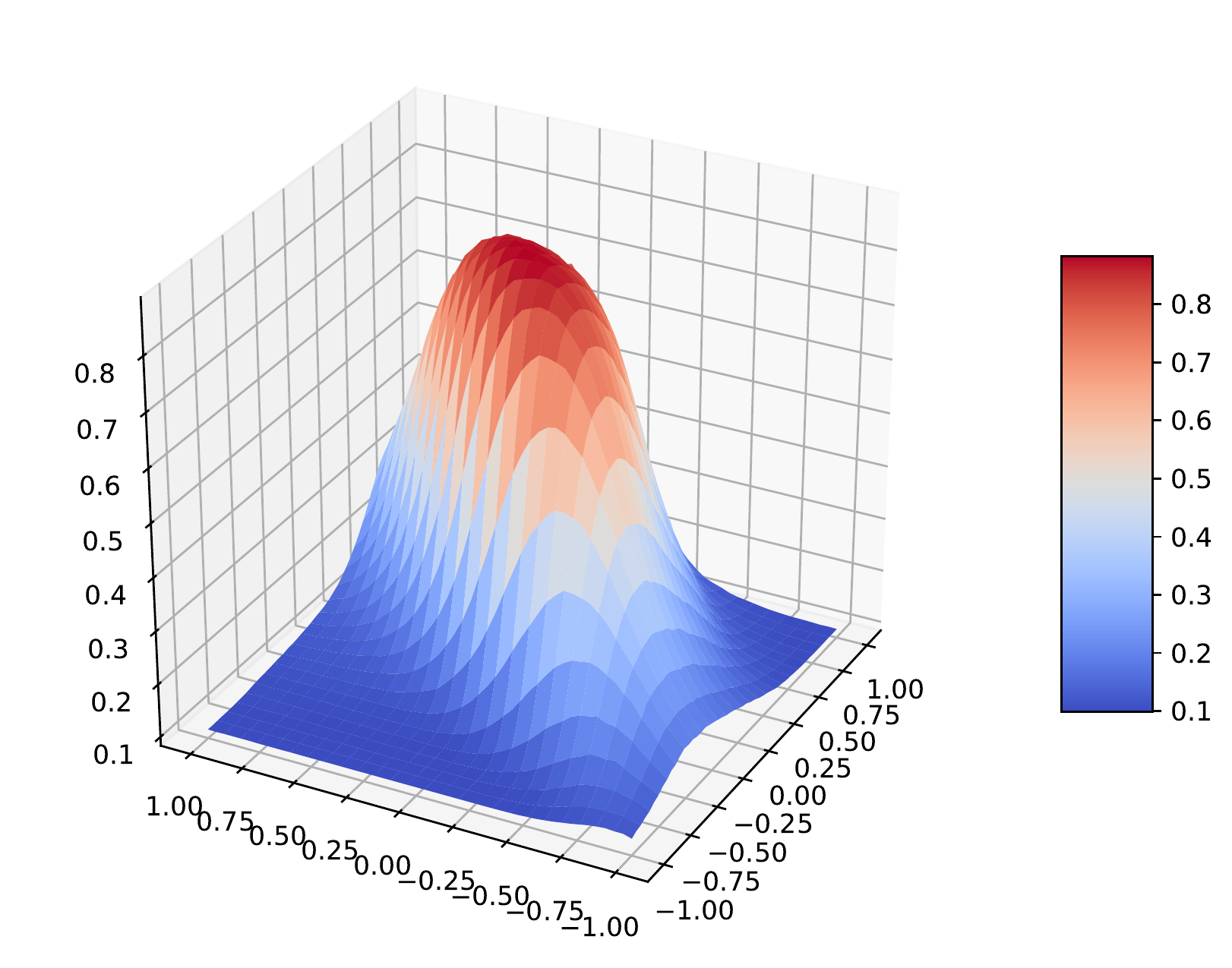} 
	}
	\subfigure[DARTS-]{
		\includegraphics[width=0.23\columnwidth]{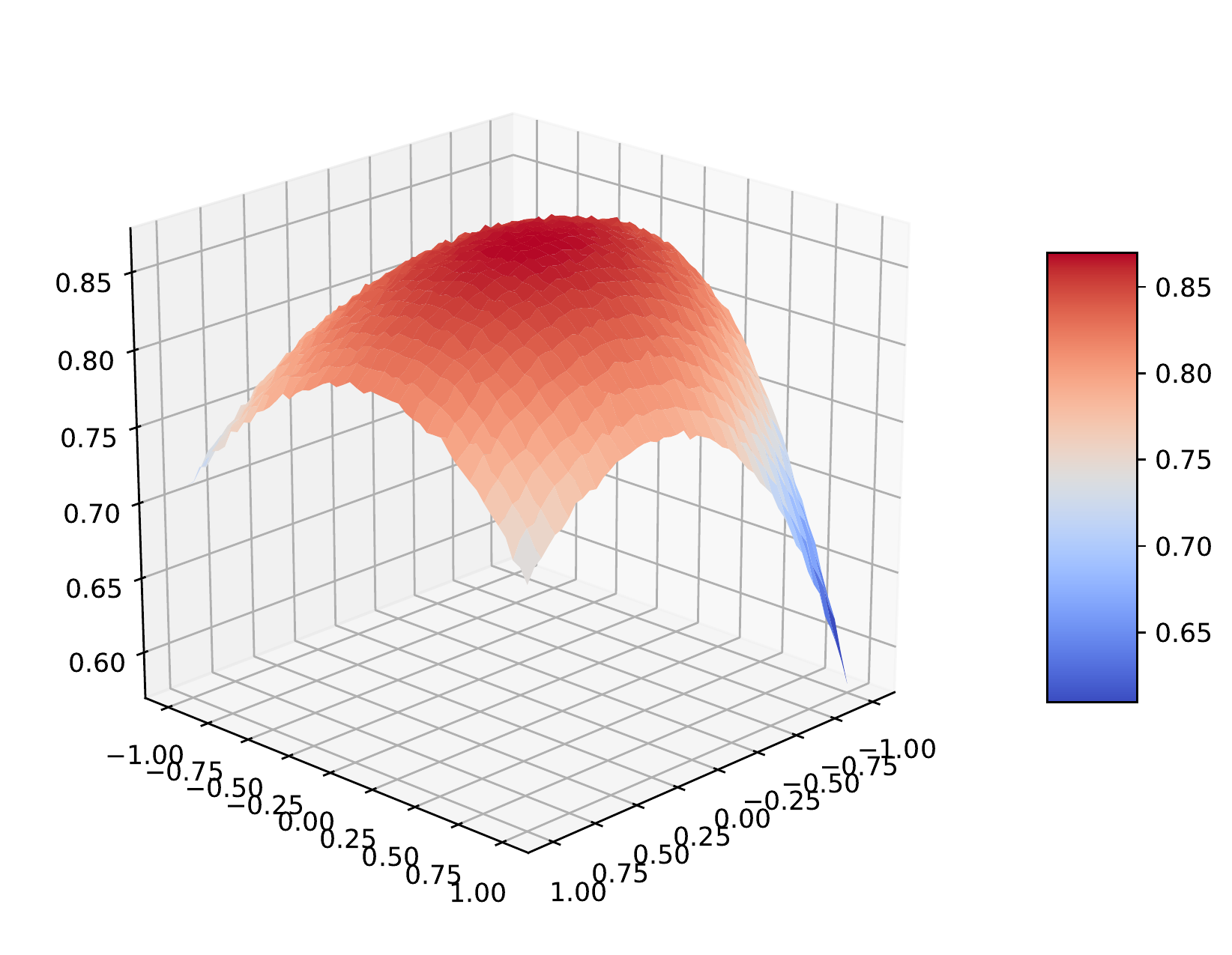}
	}
	\subfigure[DARTS]{
		\includegraphics[width=0.23\columnwidth]{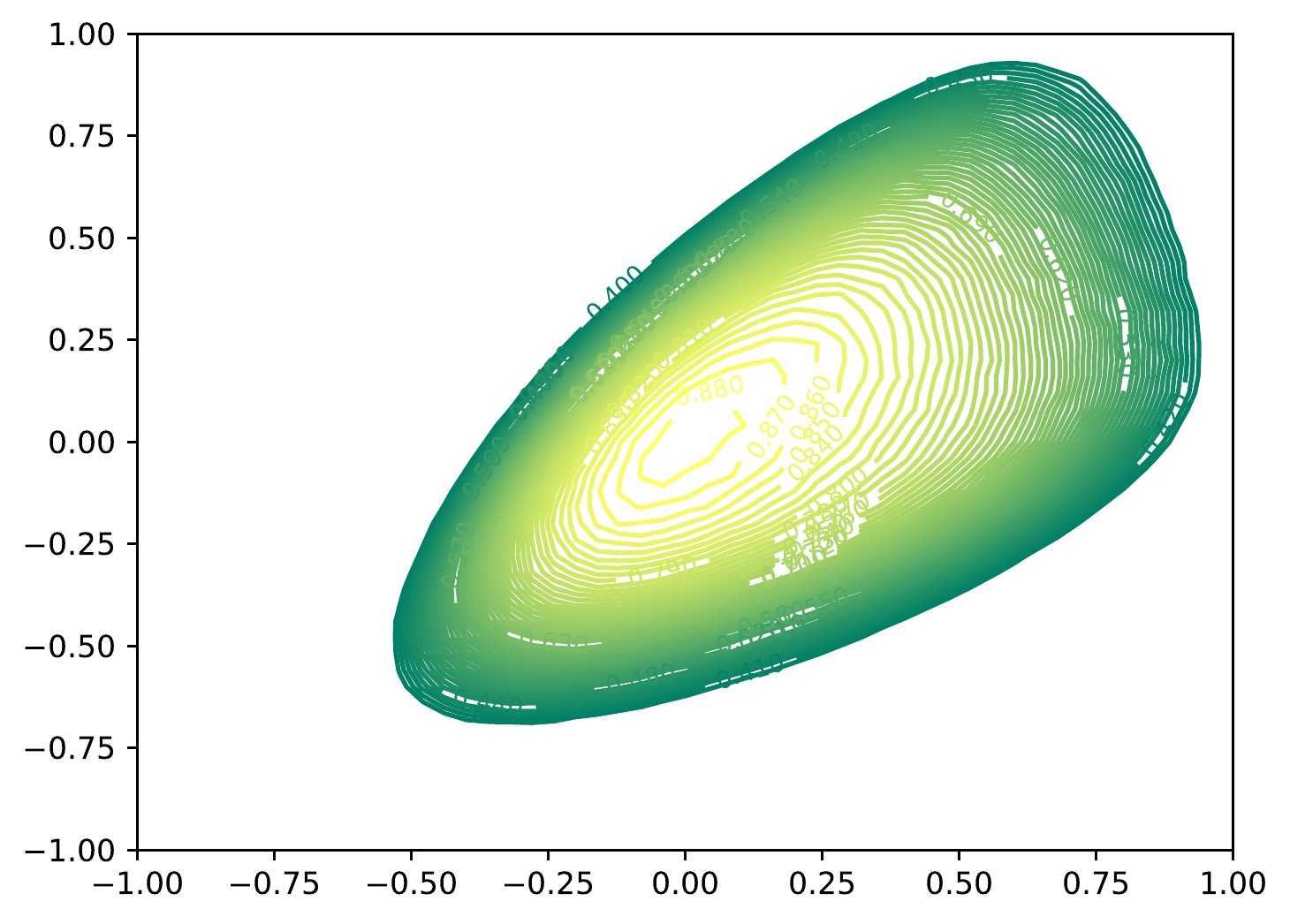} 
	}
	\subfigure[DARTS-]{
		\includegraphics[width=0.23\columnwidth]{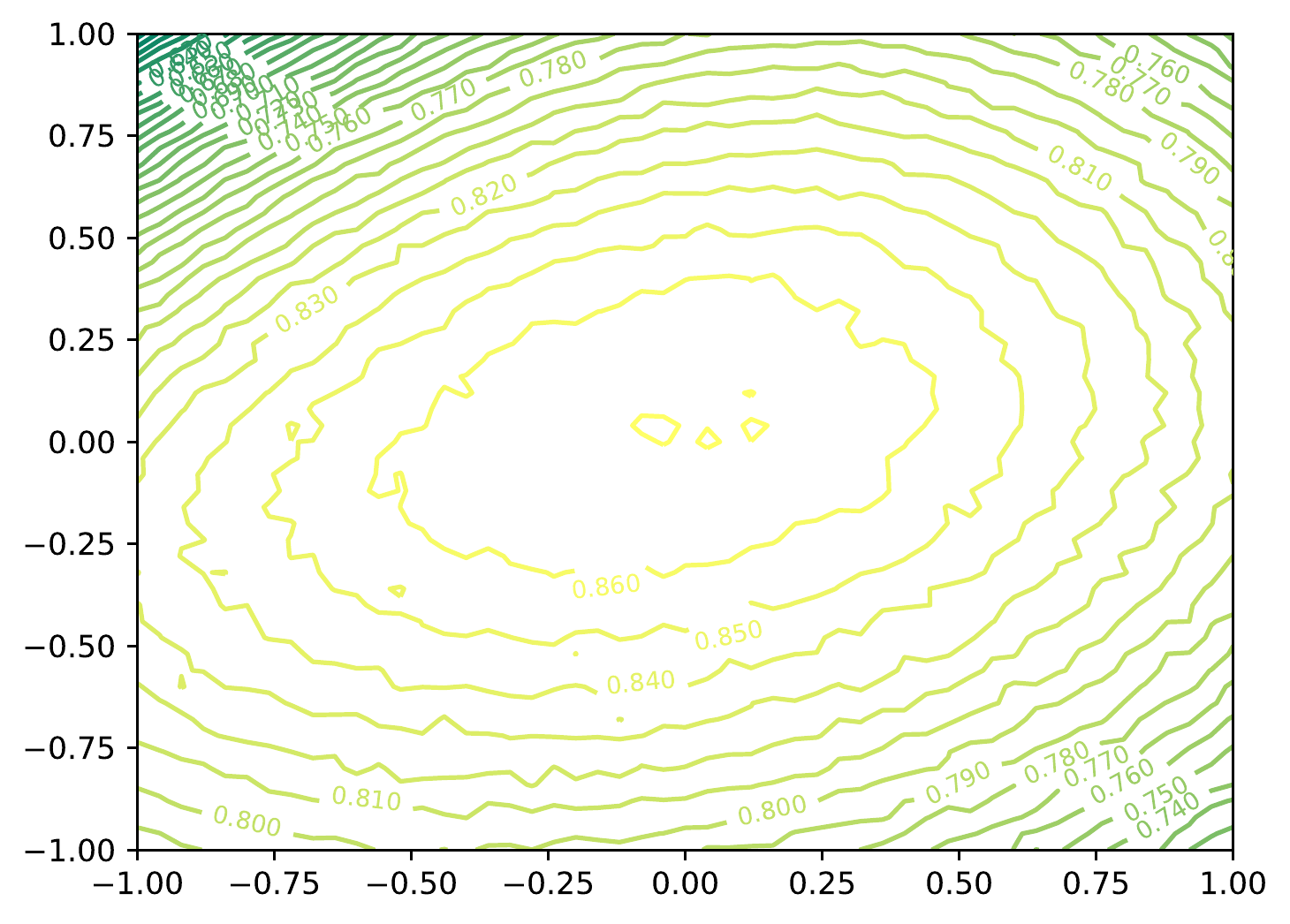}
	}
	\caption{Comparison of the validation accuracy landscape of (a) DARTS and (b) DARTS- w.r.t.  $\alpha$ on CIFAR-10 in S3. Their contour maps are shown respectively in (c) and (d), where we set the step of contour map as 0.1. The accuracy of the derived models are 94.84\% (a,c) and 97.58\% (b,d), while the maximum Hessian eigenvalues are similarly high (0.52 and 0.42)}
	\label{fig:landcape-loss-val}
\end{figure*}
 
To analyze the efficacy of DARTS-, we plot the validation accuracy landscape w.r.t architectural weights $\alpha$, and find that auxiliary connection smooths the landscape and thus stabilizes the searching stage. 
Specifically, we choose two random directions and apply normalized perturbation on $\alpha$ (following \citeauthor{li2018visualizing} \citeyear{li2018visualizing}). As shown in Figure~\ref{fig:landcape-loss-val}, DARTS- is less sensitive to the perturbation than DARTS, and the contour map of DARTS- descends more gently. 

\begin{comment}
\subsection{Trajectory of Model Performance During Search}
We are also curious to know the searching performance trajectory during the whole  process. Specifically, we sample a model  every 10 epochs and train these models from scratch using the same setting as above, the results are shown in Figure~\ref{fig:eigen_value} (b).  The performance of the inferred models continues growing, where the accuracy is boosted from $96.5\%$ to $97.4\%$. This affirms the validity of searching using our method. In contrast, the early-stopping strategies based on  eigenvalues \citep{liang2019darts} would fail in this setting. We argue that the proposed auxiliary skip branch can regularize the overfitting of the supernet, leaving the architectural weights represent the ability of candidate operations. This experiment poses as a counterexample to R-DARTS \citep{zela2020understanding}, where good models can appear albeit Hessian eigenvalues change fast. It again denies the need for the costly indicator.
\end{comment}

\section{Conclusion}
We propose a simple and effective approach named DARTS- to address the performance collapse in differentiable architecture search. Its core idea is to make use of an auxiliary skip connection branch to take over the gradient advantage role from the candidate skip connection operation. This can create a fair competition where the bi-level optimization process can  easily differentiate good operations from the bad. As a result, the search process is more stable and the collapse seldom happens across various search spaces and different datasets. 
Under strictly controlled settings, it steadily outperforms recent state-of-the-art RobustDARTS \citep{zela2020understanding}  with 3$\times$ fewer search cost.
Moreover, our method disapproves of various handcrafted regularization tricks. Last but not least, it can be used stand-alone or in cooperation with various orthogonal improvements if necessary.

This paper conveys two important messages for future research. On the one hand, the Hessian eigenvalue indicator for performance collapse \citep{zela2020understanding,chen2020stabilizing} is not ideal because it has a risk of rejecting good models. On the other hand, handcraft regularization tricks \citep{chen2019progressive} seem more critical to search a good model instead of the proposed methods.  Then what's the solution? In principle, it's difficult to find a perfect indicator of the collapse. Our approach shows the potential to control the search process and doesn't impose limitations or priors on the final model.  We hope more attention be paid in this direction.

\section{Acknowledgement}
This research was supported by Meituan.
%\clearpage
\bibliography{../aaai2020/egbib}
\bibliographystyle{iclr2021_conference}

\clearpage
\appendix
\section{Appendix}
\subsection{Preliminary about DARTS}\label{app:prelim}
In differentiable architecture search \citep{liu2018darts}, a cell-based search space in the form of Directed Acyclic Graph (DAG) is constructed. The DAG has two input nodes from the previous layers, four intermediate nodes, and one output node. There are several paralleling operators (denoted as $\mathcal{O}$) between each two nodes (say $i$, $j$), whose output $\bar{o}^{(i,j)}$ given an input $x$ is defined as,
\begin{equation}\label{eq:darts-node-softmax}
\bar{o}^{(i,j)}(x) = \sum_{o \in \mathcal{O}} \frac{\exp(\alpha_{o}^{(i,j)})}{\sum_{o' \in \mathcal{O}} \exp(\alpha_{o'}^{(i,j)})} o(x)
\end{equation}
It is essentially applying softmax over all operators where each operator is assigned with an architectural weight $\alpha$. A supernet is built on two kinds of such cells, so-called normal cells and reduction cells (for down-sampling). The architectural search is then characterized as a bi-level optimization:
\begin{align}\label{eq:darts-bi-level-opt}
&\min_{\alpha} \quad \mathcal{L}_{val} (w^*(\alpha), \alpha) \\
&s.t. \quad w^*(\alpha) = \arg \min_w \mathcal{L}_{train} (w,\alpha)
\end{align}
This indicates that the training of such a cell-based supernet should be interleaved, where at each step the network weights and architectural weights are updated iteratively. The final model is determined by simply choosing operations with the largest architectural weights.

\subsection{Experiment}

\subsubsection{Training Details}\label{app:train}

\textbf{CIFAR-10 and CIFAR-100.} Table~\ref{tab:comparison-rdarts-s2-s3-avg} gives the averaged performance in reduced search spaces S1-S4, as well with maximum eigenvalues. Table~\ref{tab:comparison-cifar100} reports the CIFAR-100 results in S0.

%\begin{table}[ht]

\begin{table}[ht]
	\caption{Comparison of searched CNN architectures in four reduced search spaces S1-S4 \citep{zela2020understanding} on CIFAR-10 and CIFAR-100. We report the mean$\pm$std of test error over 3 found architectures retrained from scratch, alongside with eigenvalue (EV) that corresponds to the best validation accuracy. We follow the same settings as  \cite{zela2020understanding}.} 
	\smallskip
	\centering
	%\resizebox{.98\columnwidth}{!}{
	%\smallskip
	\small
	\begin{tabular}{ccccccc}
		\toprule
		\multicolumn{2}{c}{\textbf{Benchmark}}  &  \textbf{DARTS} &  \textbf{DARTS-ES} &  \textbf{DARTS-ADA}  &  \textbf{ours } & \textbf{ours (EV)}\\
		\midrule
		\multirow{4}*{C10} & S1 & 4.66$\pm$0.71 & 3.05$\pm$0.07 & 3.03$\pm$0.08 & \textbf{2.76$\pm$0.07} & 0.37$\pm$0.19 \\
		~ &  S2 &   4.42$\pm$0.40 &  3.41$\pm$0.14 &  3.59$\pm$0.31 &  \textbf{2.79$\pm$0.04} & 0.41$\pm$0.08 \\
		%	   \cline{2-7}
		~    &  S3 &   4.12$\pm$0.85 &  3.71$\pm$1.14 &  2.99$\pm$0.34 & \textbf{2.65$\pm$0.04} &0.31$\pm$0.09\\ 
		~ & S4 & 6.95$\pm$0.18 & 4.17$\pm$0.21 & 3.89$\pm$0.67 & \textbf{2.91$\pm$0.04} & 0.30$\pm$0.11 \\
		\midrule
		\multirow{4}*{C100} & S1 & 29.93$\pm$0.41 & 28.90$\pm$0.81 & 24.94$\pm$0.81 & \textbf{23.26$\pm$0.59} & 0.60$\pm$0.15  \\
		~ &  S2 &    28.75$\pm$0.92 &  24.68$\pm$1.43 &  26.88$\pm$1.11  &  \textbf{22.31$\pm$0.65} &0.54$\pm$0.15\\ 
		%		\cline{2-7}
		~    &  S3  &  29.01$\pm$0.24 &  26.99$\pm$1.79 &  24.55$\pm$0.63 &  \textbf{21.47$\pm$0.40} &0.40$\pm$0.03\\ 
		~ & S4 & 24.77$\pm$1.51 & 23.90$\pm$2.01 & 23.66$\pm$0.90 & \textbf{21.75$\pm$0.26} &0.93$\pm$0.15 \\
		\bottomrule
	\end{tabular}
	%}
	\label{tab:comparison-rdarts-s2-s3-avg}
\end{table}

%\begin{wraptable}{R}{.5\columnwidth}
\begin{table}[ht]
\setlength{\tabcolsep}{1pt}
% CIFAR-100
	\caption{Comparison of searched models on CIFAR-100. $^\diamond$: Reported by \cite{dong2019searching}, $^\star$: Reported by \cite{zela2020understanding}, $^\ddagger$:Rerun their code. }
	\label{tab:comparison-cifar100}
	\centering
	\small
			\begin{tabular}{llHll} 	
				\toprule		
				\textbf{Models}   &  \textbf{Params}  &  \textbf{FLOPs} &  \textbf{Error } &  \textbf{Cost}  \\
				 & \scriptsize{(M)}  & \scriptsize{(M)}  & \scriptsize{(\%)} & \scriptsize{GPU Days}   \\
				\midrule
				ResNet \citeyp{he2016deep}   &  1.7 &     &   22.10$^\diamond$  &  -  \\
				AmoebaNet \citeyp{real2019regularized} & 3.1 & &  18.93$^\diamond$ & 3150  \\
				PNAS \citeyp{liu2018progressive} & 3.2 & & 19.53$^\diamond$ & 150 \\
				ENAS \citeyp{pham2018efficient}  &  4.6  &    &  19.43$^\diamond$  & 0.45    \\
				DARTS \citeyp{liu2018darts}  & - & & 20.58$\pm$0.44$^\star$ &  0.4  \\ 
				GDAS \citeyp{dong2019searching}  &  3.4  &    &  18.38  & 0.2 \\

				P-DARTS \citeyp{chen2019progressive}  &  3.6  &   &  17.49$^\ddagger$  &  0.3 \\
				R-DARTS \citeyp{zela2020understanding}  & - &  - &  18.01$\pm$0.26 & 1.6 \\
				DARTS- (avg.) & 3.3& & 17.51$\pm$0.25 & 0.4 \\
				\textbf{DARTS- (best)} &  3.4 &   & 17.16  & 0.4\\
				\bottomrule
			\end{tabular}
\end{table}
%\end{wraptable}

\textbf{ImageNet classification. } For training on ImageNet, we use the same setting as MnasNet \citep{tan2018mnasnet}. To be comparable with EfficientNet \citep{tan2019efficientnet}, we also use squeeze-and-excitation \citep{hu2018squeeze}. Furthermore, we don't include methods trained using large model distillation, because it can boost final validation accuracy marginally. To be fair, we don't use the efficient head in MobileNetV3 \citep{howard2019searching} although it can reduce FLOPs marginally.

\textbf{COCO object detection.} 
All models are trained and evaluated on MS COCO dataset for 12 epochs with a batch size of 16. The initial learning rate is 0.01 and reduced by 0.1 at epoch 8 and 11. 

\begin{table*}[ht]
	\small
	\begin{center}
		\caption{Transfer results on COCO datasets of various drop-in backbones.}
		\label{table:darts-coco-retina}
		\begin{tabular}{*{1}{l}H*{8}{l}}
			\toprule
			\textbf{Backbones} &\textbf{FLOPs}  & \textbf{Params (M)} & \textbf{Acc}  & \textbf{AP} & \textbf{AP$_{50}$} & \textbf{AP$_{75}$} & \textbf{AP$_S$} & \textbf{AP$_M$} & \textbf{AP$_L$} \\
			%\begin{scriptsize}
			%			& (M) & (M) & (\%) &(\%) & (\%)& (\%)&(\%) &(\%) &(\%)  \\%\end{scriptsize} \\
			\midrule
			MobileNetV2 \citeyp{sandler2018mobilenetv2} & 300 & 3.4& 72.0 & 28.3 & 46.7 & 29.3 & 14.8 & 30.7 & 38.1\\
			SingPath NAS \citeyp{stamoulis2019single} & 365 & 4.3 & 75.0 & 30.7 & 49.8 & 32.2 & 15.4 &33.9 & 41.6\\
			MnasNet-A2 \citeyp{tan2018mnasnet} & 340& 4.8 & 75.6 & 30.5 & 50.2 & 32.0 & 16.6 & 34.1 & 41.1\\
			%			FairNAS-B \citep{chu2019fairnas} & 349 & 5.7 & 77.2 & 31.7 & 51.5 & 33.0 & 17.0 & 35.2 & 42.5\\
			%			FBNet-C \citeyp{wu2018fbnet} &375 &5.5 & 74.9\\
			MobileNetV3 \citeyp{howard2019searching} & 219 & 5.4 & 75.2& 29.9 & 49.3 & 30.8 & 14.9 & 33.3 & 41.1\\
			%			EfficientNetB0 \citeyp{tan2019efficientnet} & 390 & 5.3&76.3& \\
			MixNet-M \citeyp{tan2020mixconv} & 360 & 5.0 & 77.0 & 31.3& 51.7 & 32.4& 17.0 & 35.0 & 41.9   \\
			FairDARTS-C \citeyp{chu2019fair} & 386 & 5.3 & 77.2 & 31.9 & 51.9 & 33.0 & 17.4 & 35.3 & 43.0 \\
			\textbf{DARTS-A (Ours)} & 470 & 5.5& 77.8& 32.5& 52.8 & 34.1& 18.0 & 36.1 & 43.4  \\
			
			\bottomrule
			%			FairNAS-A$^{\dagger}$ & 392 & 5.9 & \textbf{77.5} & \textbf{32.4} & \textbf{52.4} & \textbf{33.9} & \textbf{17.2} & \textbf{36.3} & \textbf{43.2}\\
			%			FairNAS-B$^{\dagger}$ & 349 & 5.7 & 77.2 & 31.7 & 51.5 & 33.0 & 17.0 & 35.2 & 42.5\\
			%			FairNAS-C$^{\dagger}$ &325 & 5.6& 76.7& 31.2 & 50.8 & 32.7 & 16.3 & 34.4 & 42.3\\
		\end{tabular}
	\end{center}
	\vskip -0.15in
\end{table*}

\subsubsection{Further Discussions on Failure of EigenValue}\label{sec:failure_eigen}

\begin{figure}[ht]
	\centering
	\subfigure[Trajectory of eigenvalues in S0-S4 (left to right) on CIFAR-10]{
		\includegraphics[width=0.65\columnwidth]{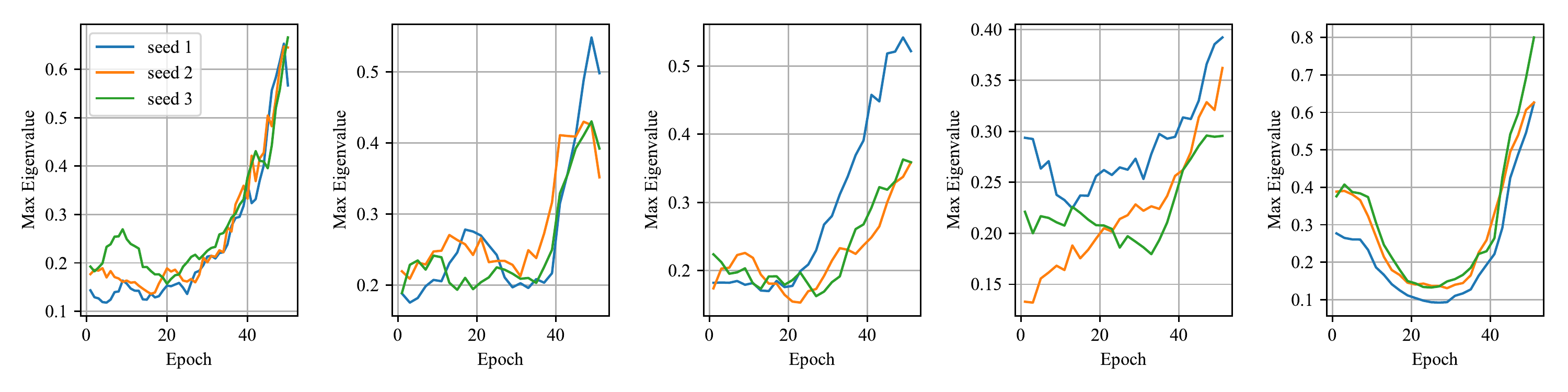}
	}
	\subfigure[Sampled models' performance]{
		\includegraphics[width=0.3\columnwidth]{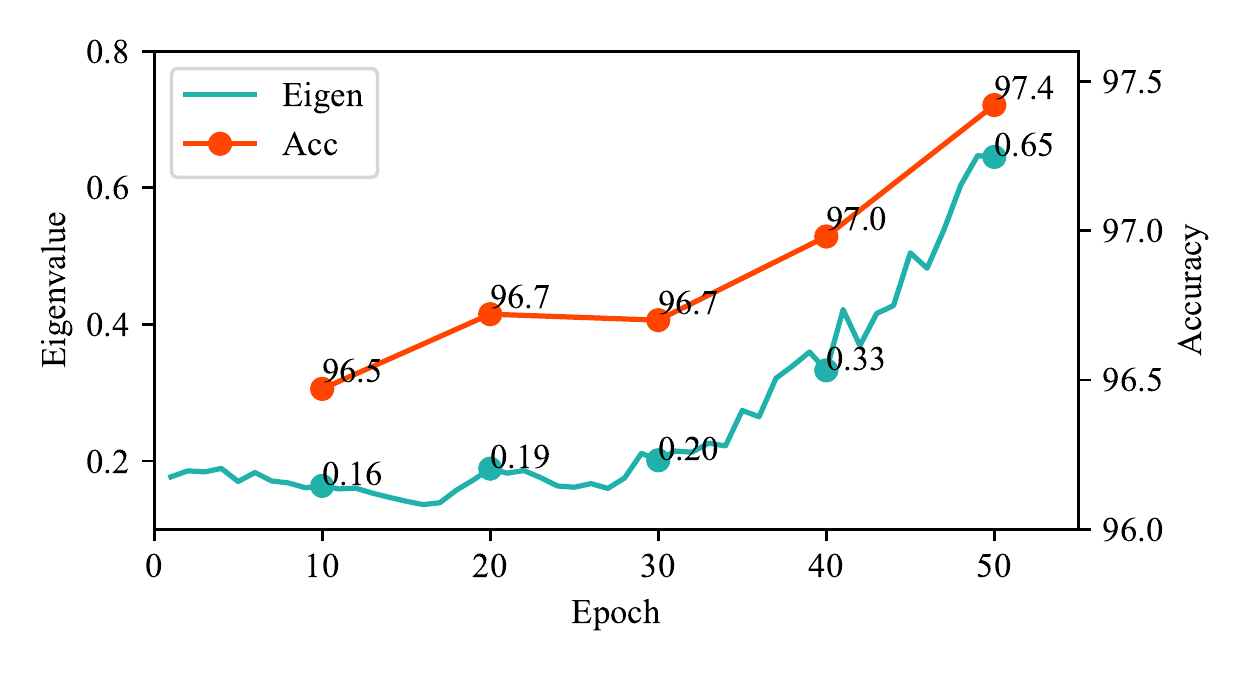}
	}
	\subfigure[Trajectory of eigenvalues in S0-S4 (left to right) on CIFAR-100]{
		\includegraphics[width=0.65\columnwidth]{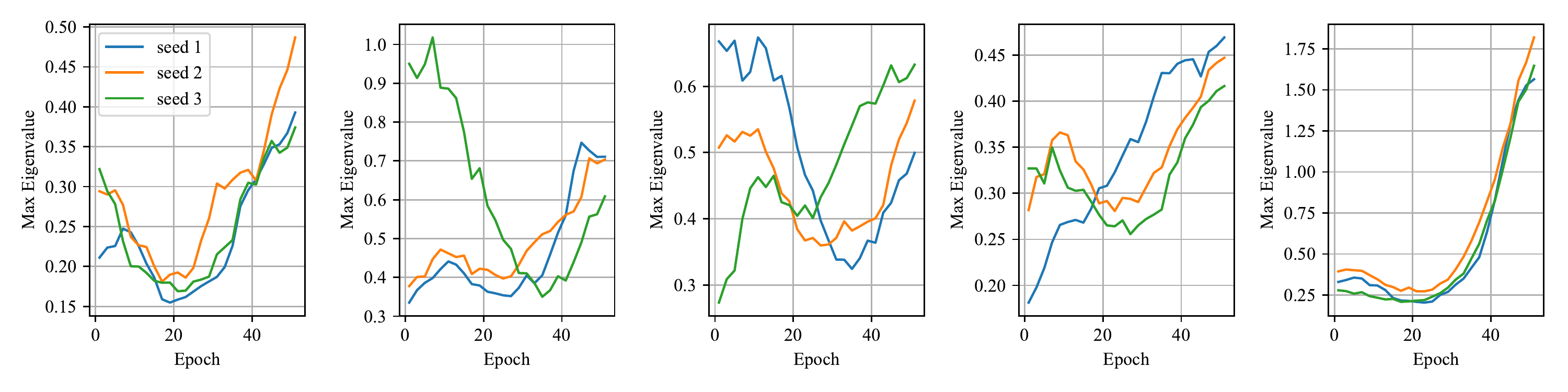}
	}
	\caption{The evolution of maximal eigenvalues of DARTS- when searching in different search spaces S0-S4 on CIFAR-10 (a) and CIFAR-100 (c). We run each experiment 3 times on different seeds. (b) DARTS-'s growing Hessian eigenvalues don't induce poor performance. Among the sampled five models, the one corresponding to the highest eigenvalue has the best performance. This example is done in S0 on CIFAR-10.}
	\label{fig:eigen_value}
\end{figure}

\begin{figure}[ht]
\centering
\includegraphics[width=0.45\columnwidth]{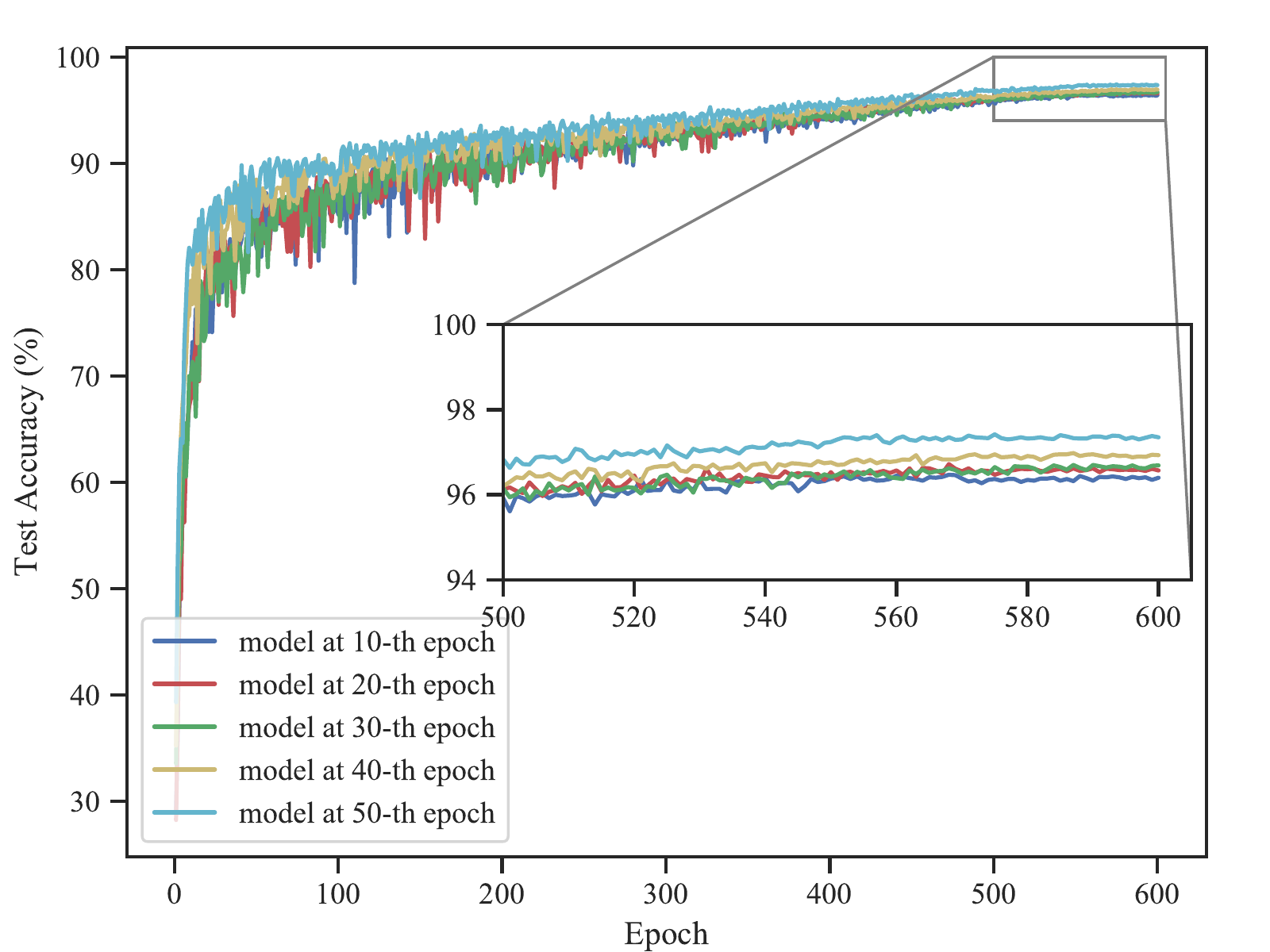}
\caption{Training five models sampled every 10 epochs during DARTS- searching process. See Fig~\ref{fig:eigen_value} (b) for the corresponding eigenvalues.}
\label{fig:sample-model-training-curve}
%\end{wrapfigure}
\end{figure}

To further explore the relationship between the searching performance and Hessian Eigenvalue, we plot the performance trajectory of the searched models in Figure~\ref{fig:eigen_value} (b). Specifically, we sample models every 10 epochs and train these models from scratch using the same setting as above (Figure~\ref{fig:sample-model-training-curve}). The performance of the inferred models continues growing, where the accuracy is boosted from $96.5\%$ to $97.4\%$. This affirms the validity of searching using our method. In contrast, the early-stopping strategies based on eigenvalues \citep{zela2020understanding} would fail in this setting. We argue that the proposed auxiliary skip branch can regularize the overfitting of the supernet, leaving the architectural weights to represent the ability of candidate operations. This experiment poses as a counterexample to R-DARTS, where good models can appear albeit Hessian eigenvalues change fast. It again denies the need for a costly indicator.

\subsubsection{More Ablation Studies}\label{app:ablation}

\begin{figure}[ht]
\centering
\subfigure{
\includegraphics[width=0.35\columnwidth]{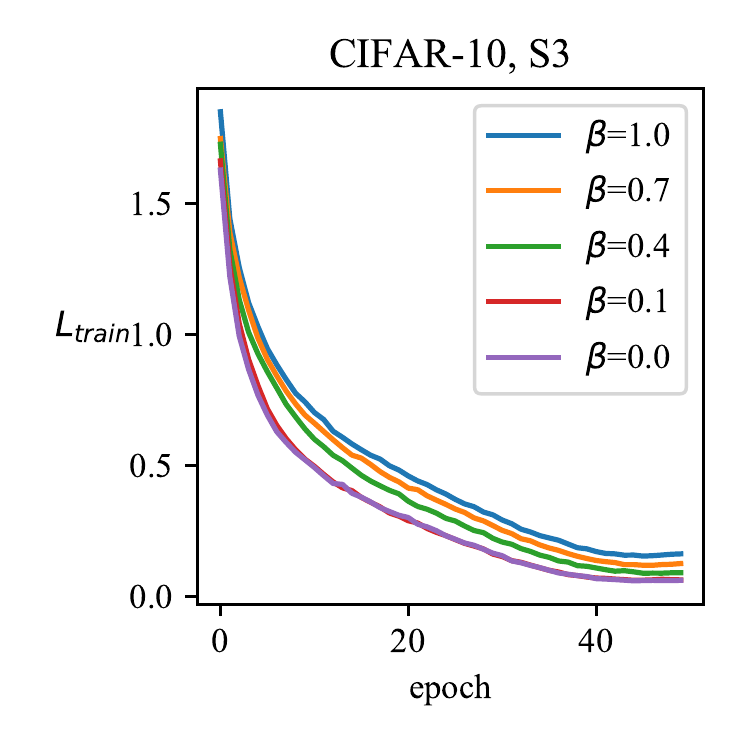}
}
%\vspace{-20pt}
\caption{The training loss curve of the over-parameterized network on CIFAR-10 in S3 with different initial $\beta_0$.}
\label{fig:beta-loss}
\end{figure}

To supplement the sensitivity analysis in Section \ref{sec:ablation}, Figure \ref{fig:beta-loss} shows the training loss curve when initializing $\beta$ with different values.
%\begin{wrapfigure}{l}{0.5\columnwidth}
%\vspace{-20pt}
%\end{wrapfigure}

%\begin{figure}[ht]
%	\centering
%	\subfigure{
%		\includegraphics[width=0.18\columnwidth]{figures/beta-10-loss.pdf}
%	}
%	\subfigure{
%		\includegraphics[width=0.18\columnwidth]{figures/beta-07-loss.pdf}
%	}
%	\subfigure{
%		\includegraphics[width=0.18\columnwidth]{figures/beta-04-loss.pdf}
%	}
%	\subfigure{
%		\includegraphics[width=0.18\columnwidth]{figures/beta-01-loss.pdf}
%	}
%	\subfigure{
%		\includegraphics[width=0.18\columnwidth]{figures/beta-00-loss.pdf}
%	}
%	\caption{Training the over-parameterized network on CIFAR-10 in S3 with different initial $\beta_0$.}
%	\label{fig:beta-loss}
%\end{figure}

\subsection{Loss Landscape}

\begin{figure}[ht]
	\centering
	\vskip -0.2 in
	\subfigure[DARTS]{
		\includegraphics[width=0.23\columnwidth]{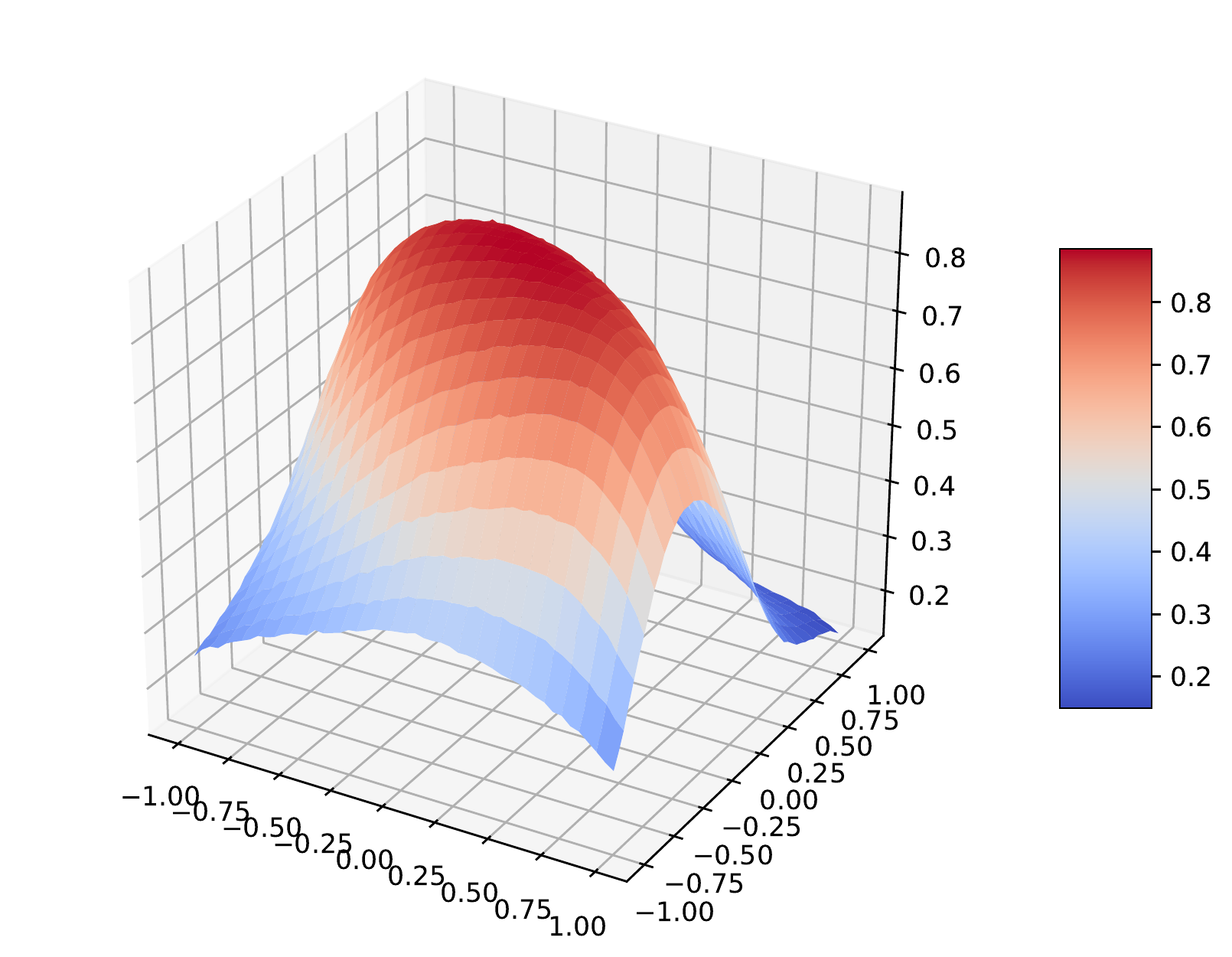} 
	}
	\subfigure[DARTS]{
		\includegraphics[width=0.23\columnwidth]{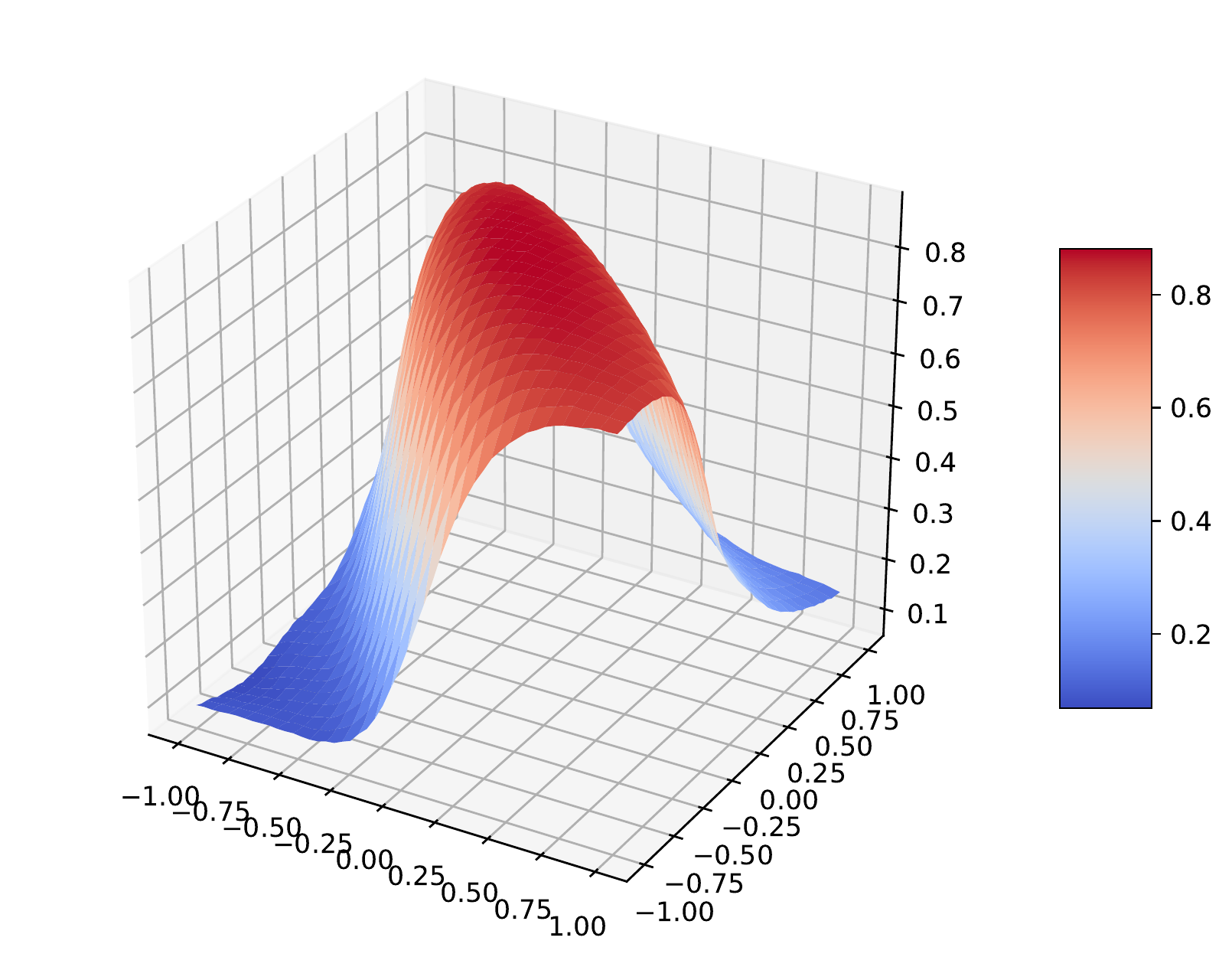}
	}
	\subfigure[DARTS]{
		\includegraphics[width=0.23\columnwidth]{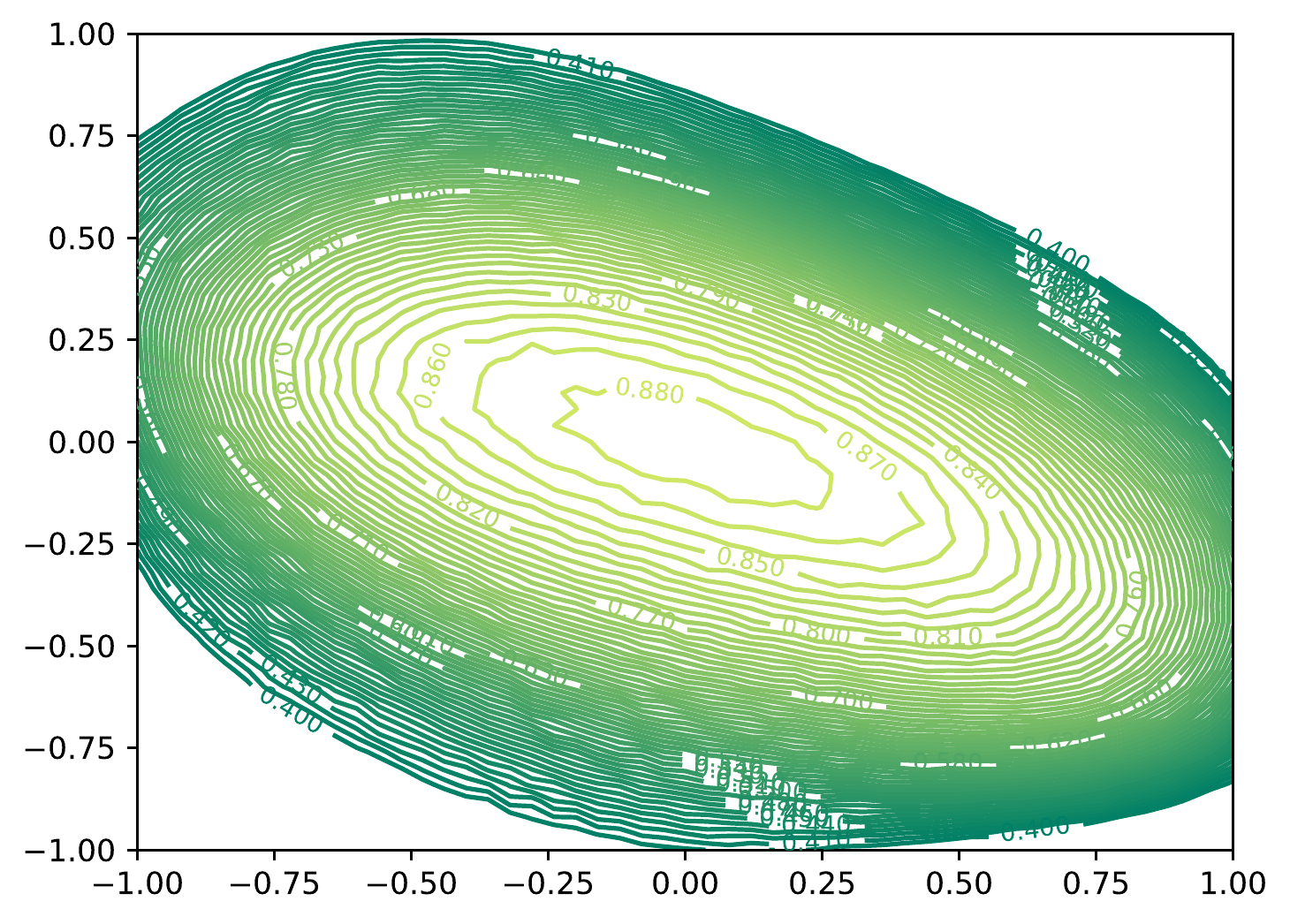} 
	}
	\subfigure[DARTS]{
		\includegraphics[width=0.23\columnwidth]{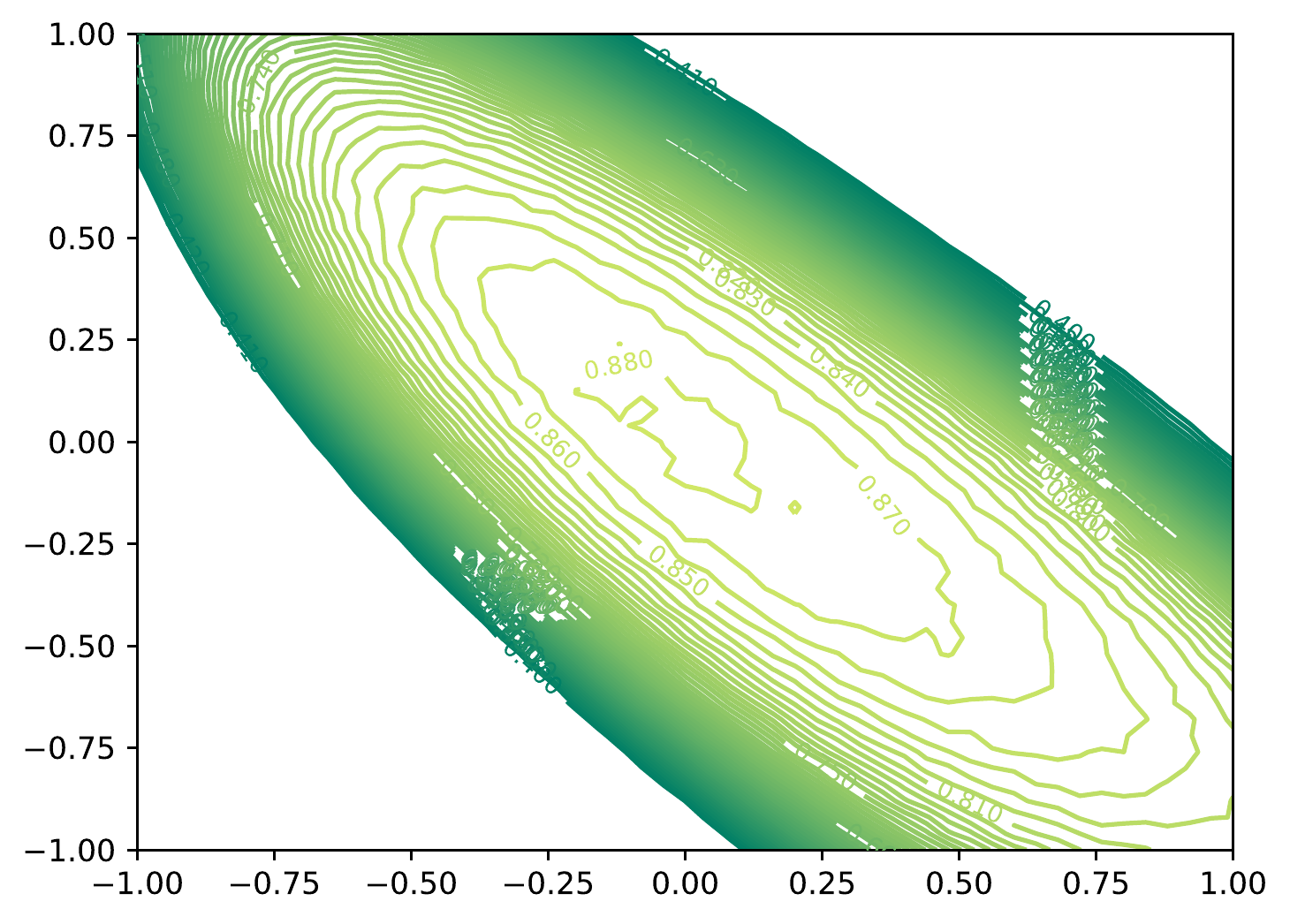}
	}
	\subfigure[DARTS-]{
		\includegraphics[width=0.23\columnwidth]{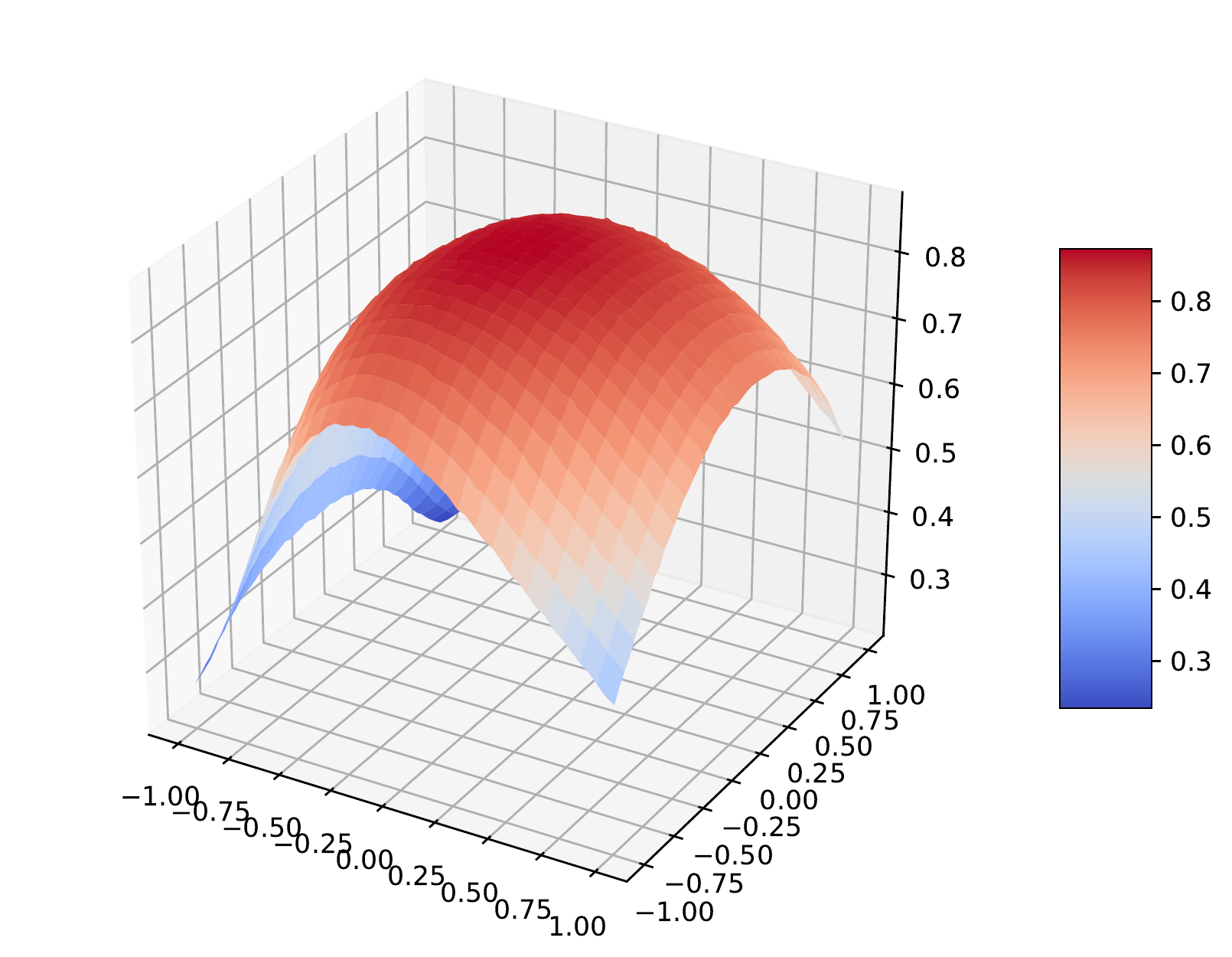} 
	}
	\subfigure[DARTS-]{
		\includegraphics[width=0.23\columnwidth]{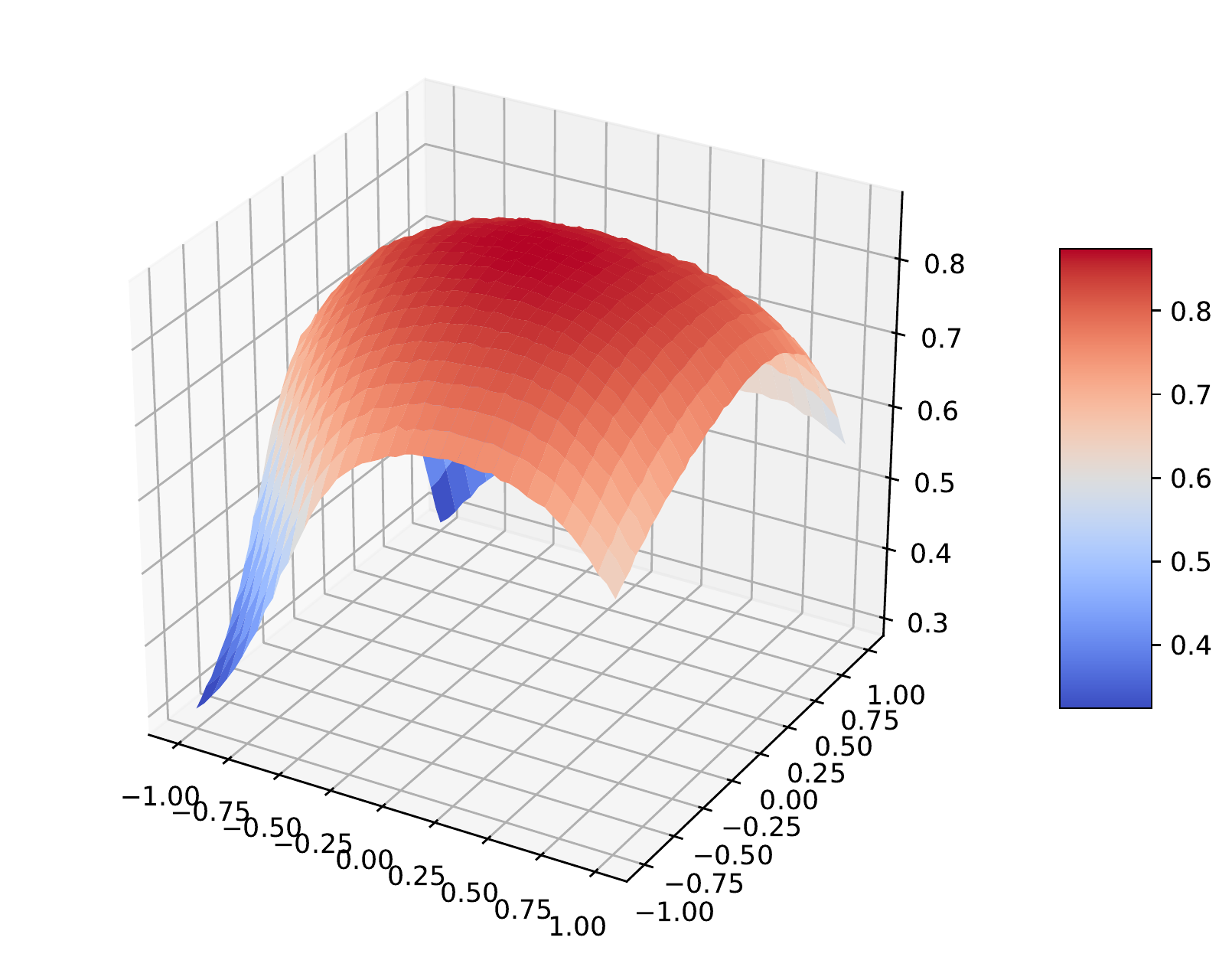}
	}
	\subfigure[DARTS-]{
		\includegraphics[width=0.23\columnwidth]{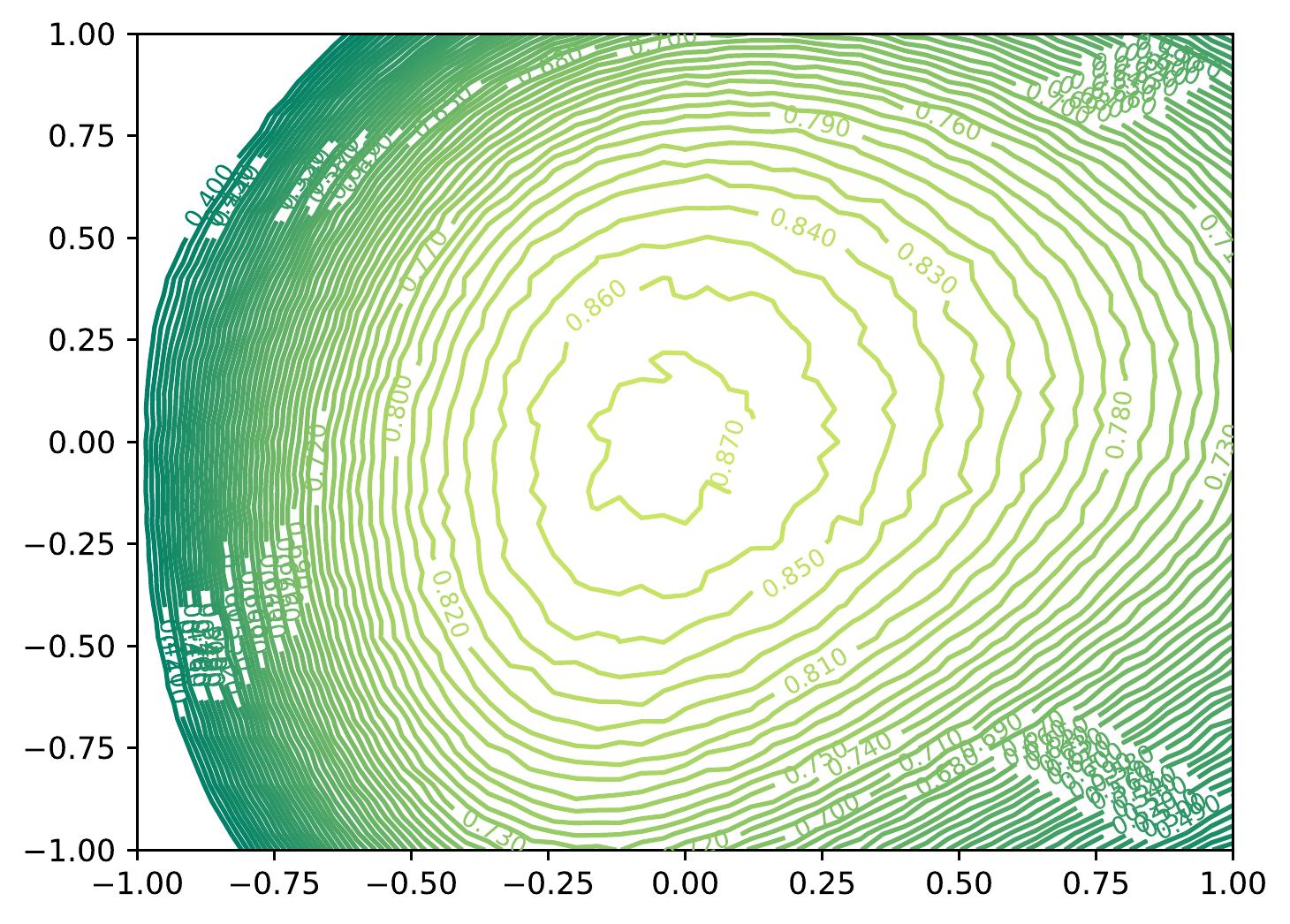} 
	}
	\subfigure[DARTS-]{
		\includegraphics[width=0.23\columnwidth]{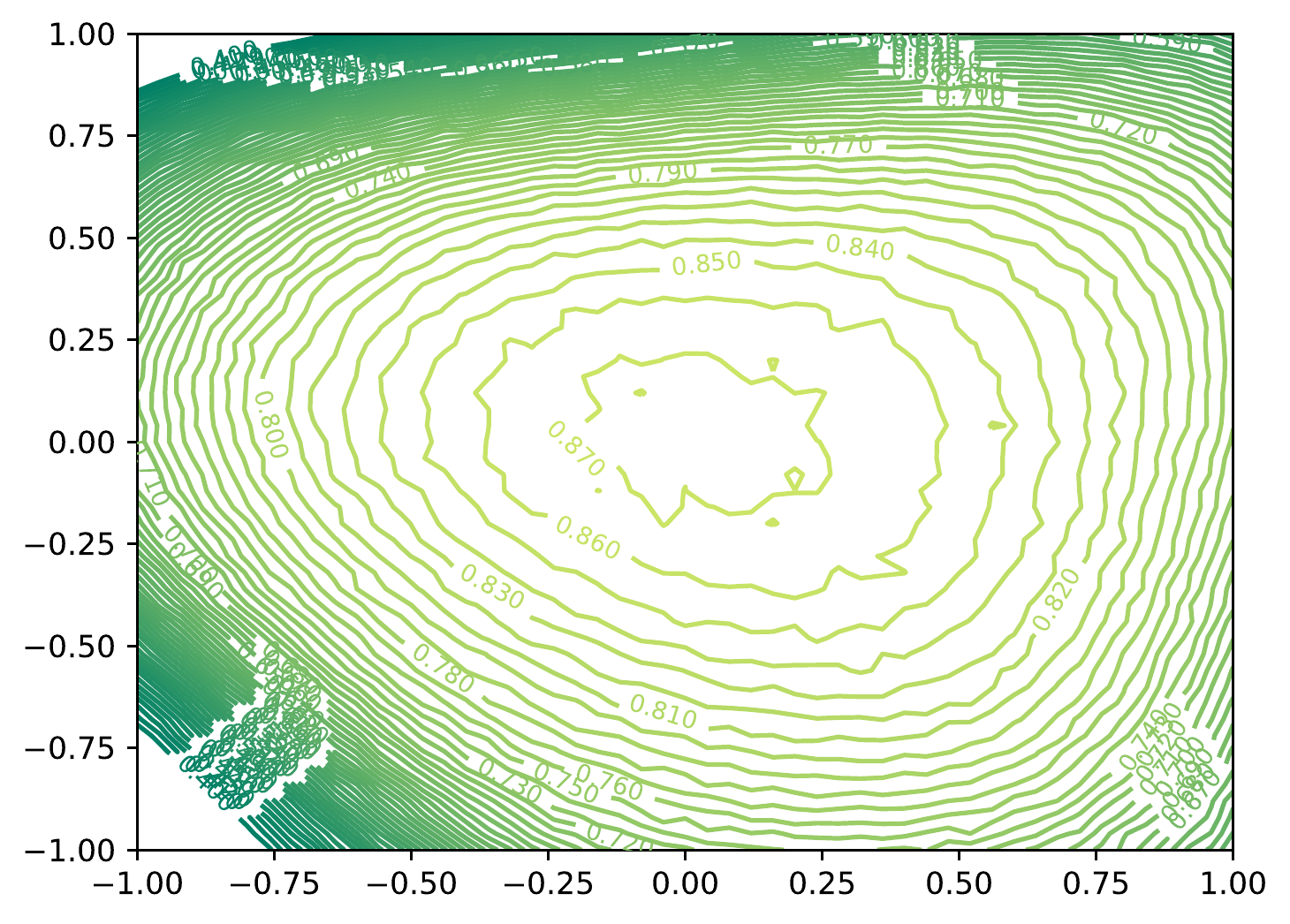}
	}
	\caption{More visualization of validation accuracy landscapes of  DARTS (a,b) and DARTS- (e,f) w.r.t. the architectural weights $\alpha$ on CIFAR-10 in S0. Their contour maps are shown respectively in (c,d) and (g,h). The step of contour map is 0.1. The inferred models by DARTS- have higher accuracies (97.50\%, 97.49\%) than DARTS (97.19\%, 97.20\%).}
	\label{fig:landcape-loss-val-more-s0}
\end{figure}

\begin{figure}[ht]
	\centering
	\vskip -0.2 in
	\subfigure[DARTS]{
		\includegraphics[width=0.23\columnwidth]{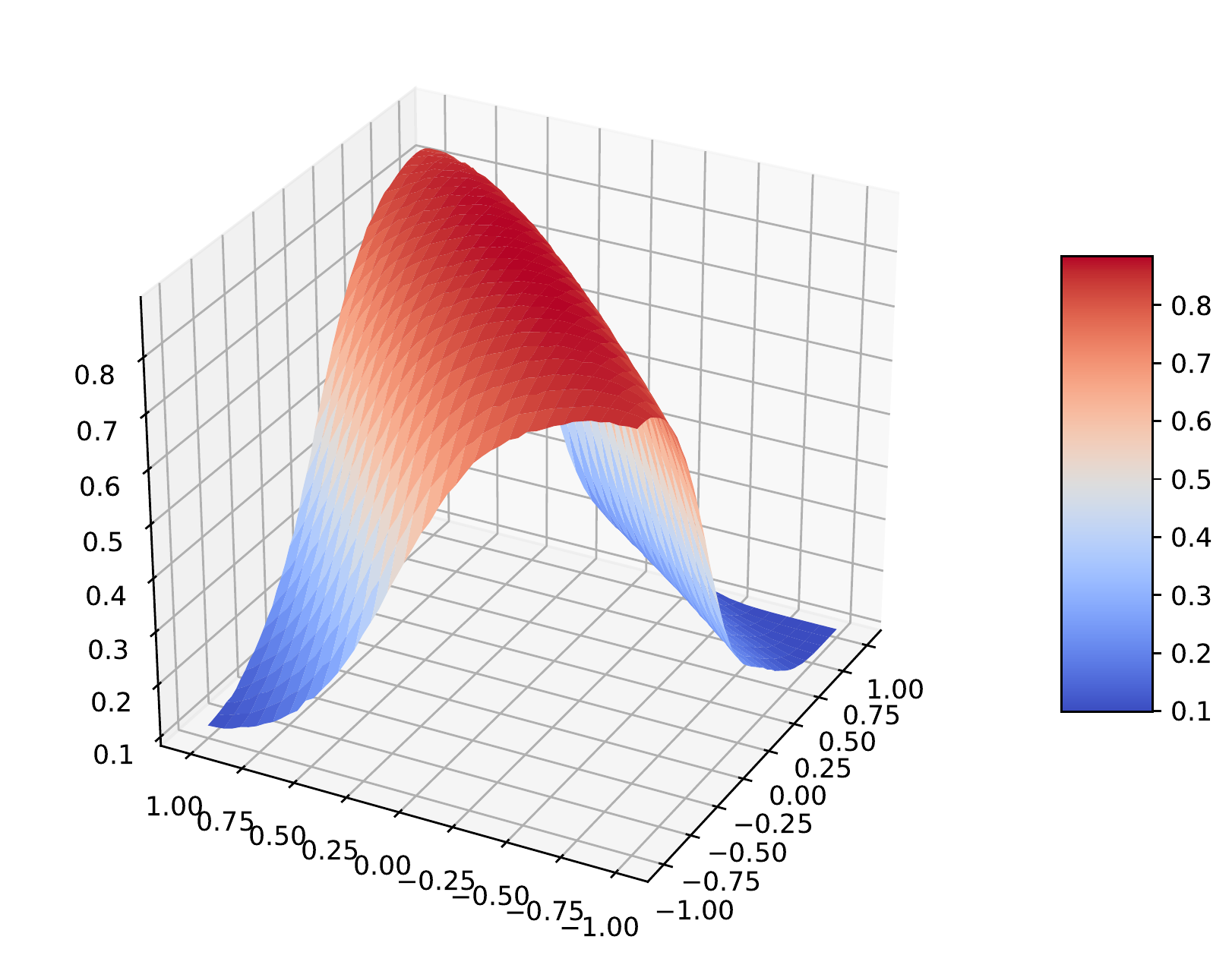} 
	}
	\subfigure[DARTS]{
		\includegraphics[width=0.23\columnwidth]{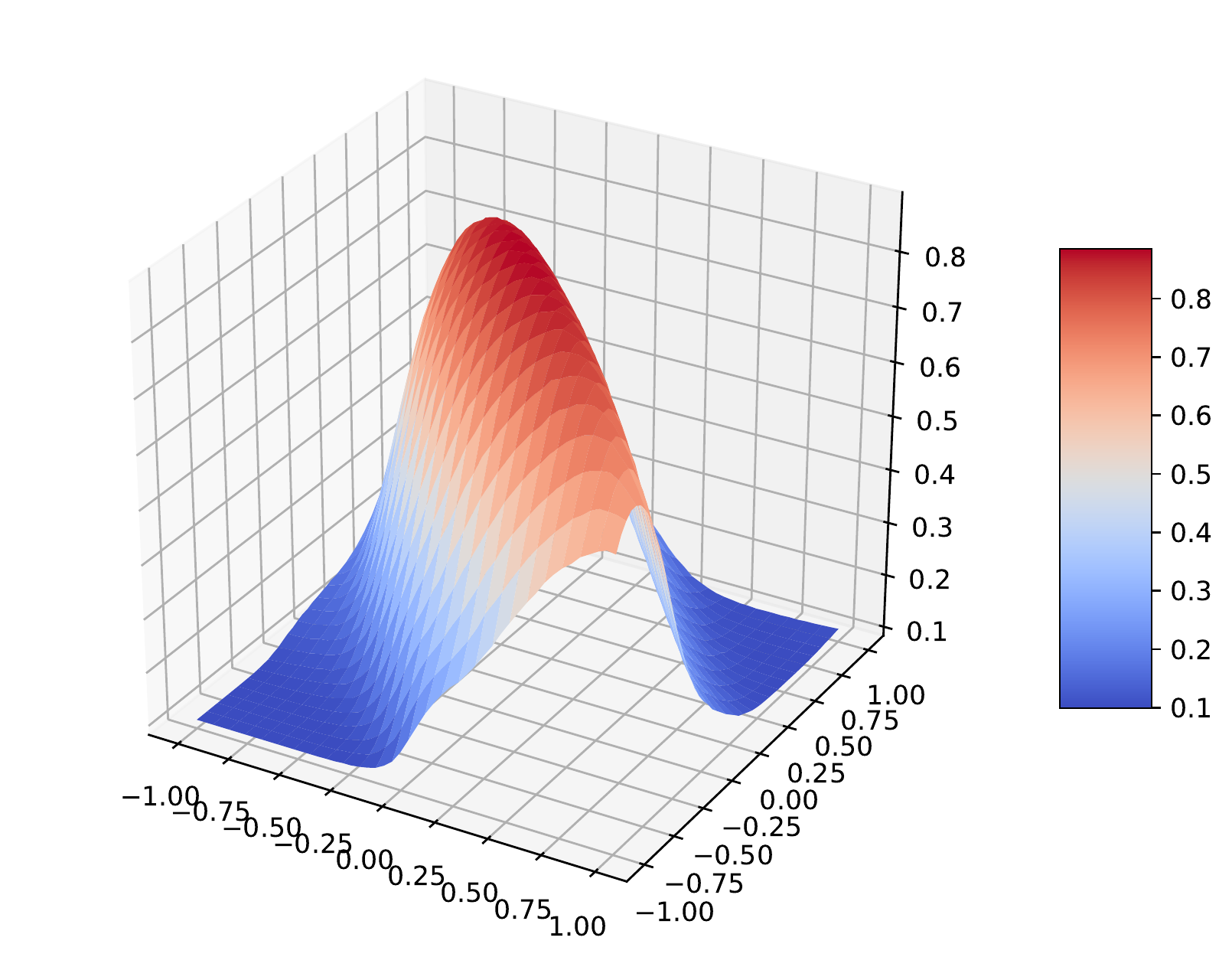}
	}
	\subfigure[DARTS]{
		\includegraphics[width=0.23\columnwidth]{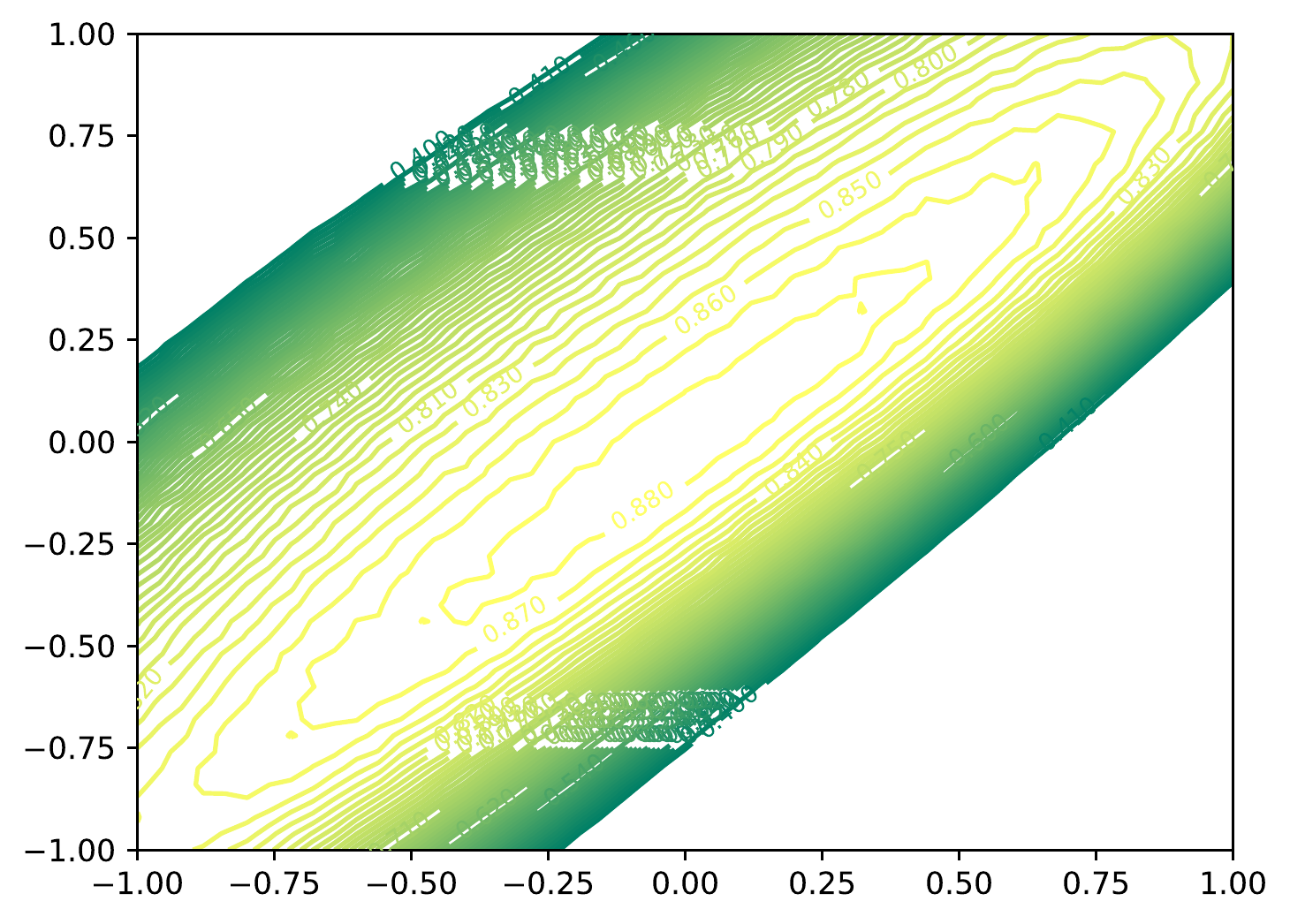} 
	}
	\subfigure[DARTS]{
		\includegraphics[width=0.23\columnwidth]{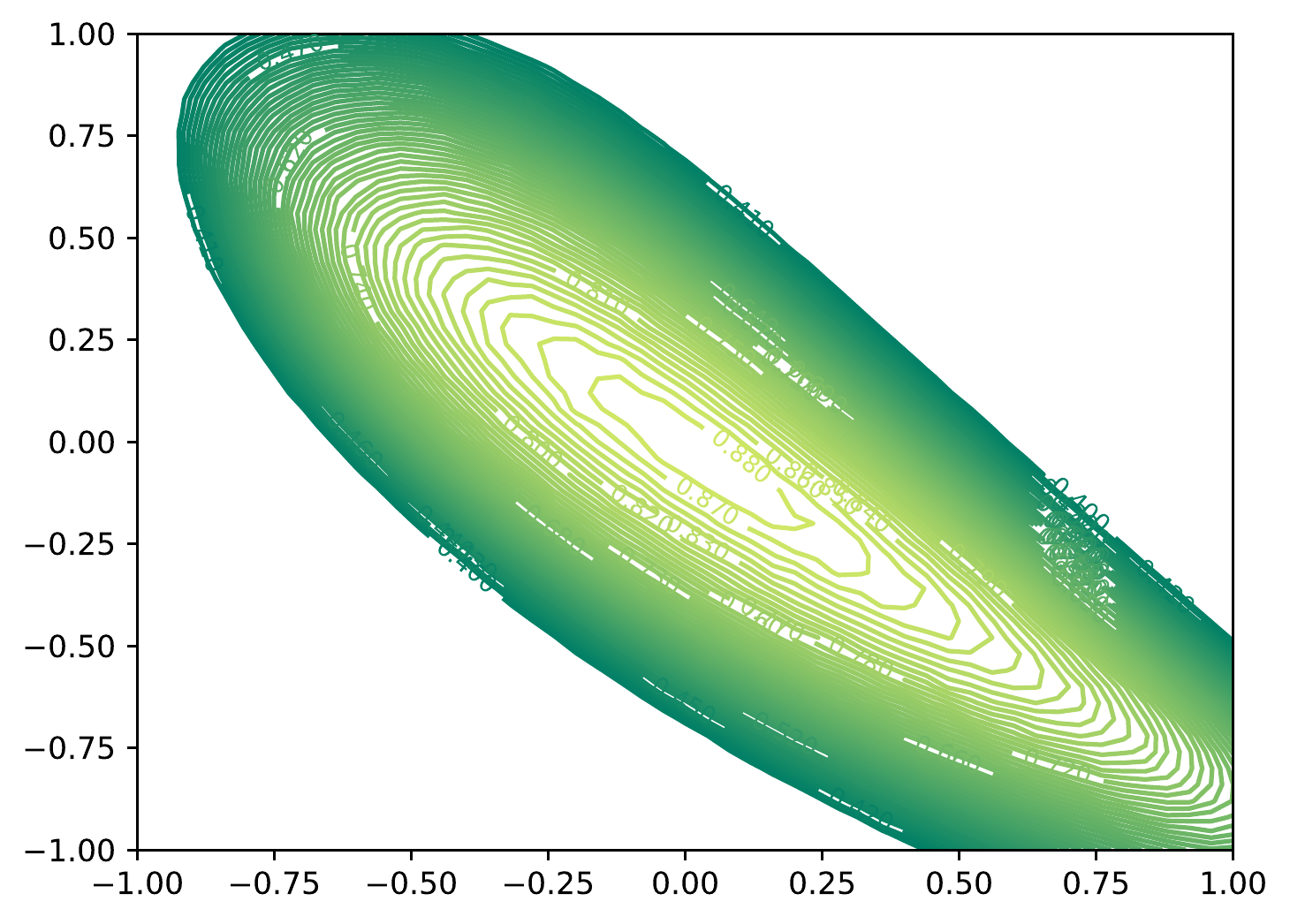}
	}
	\subfigure[DARTS-]{
		\includegraphics[width=0.23\columnwidth]{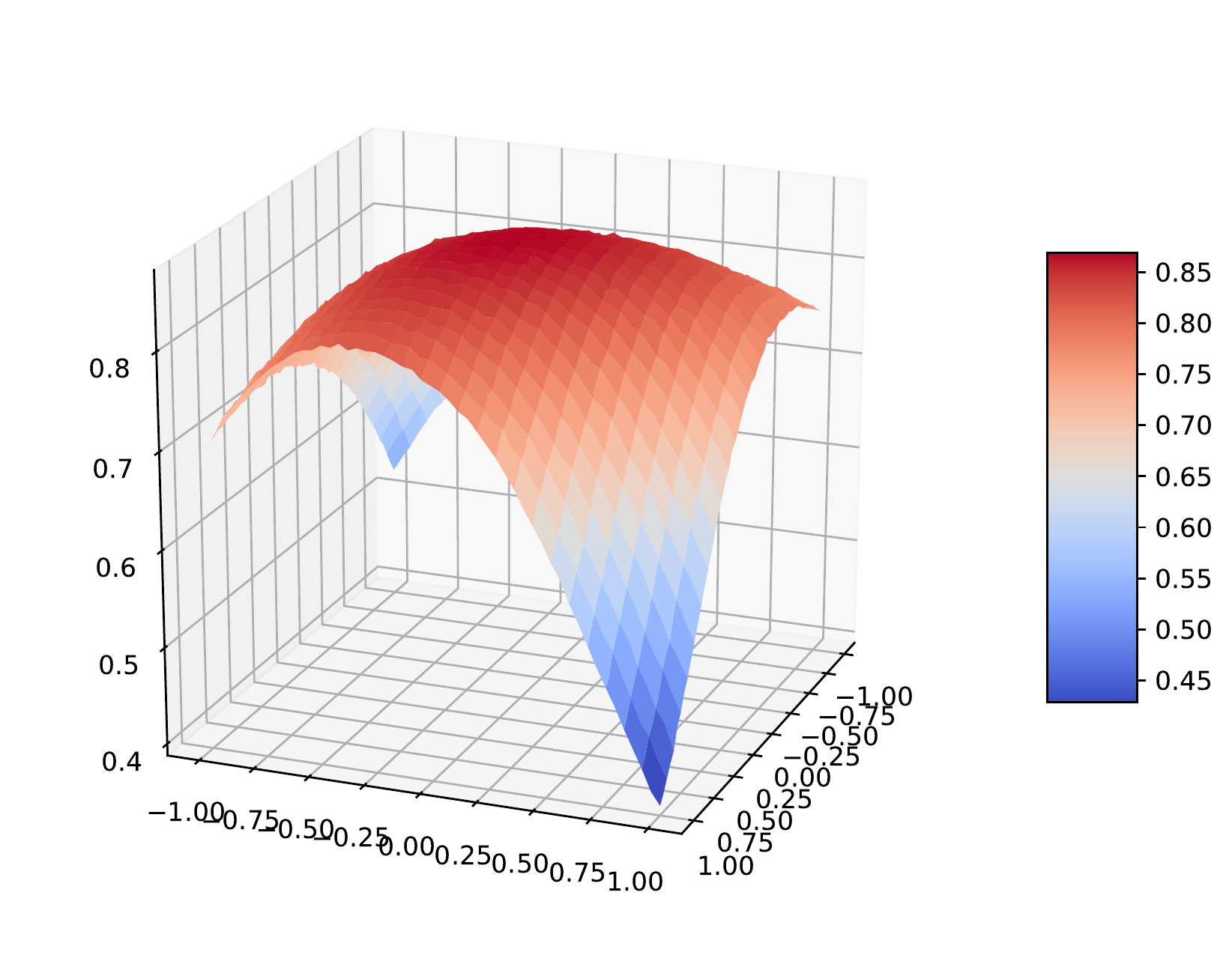} 
	}
	\subfigure[DARTS-]{
		\includegraphics[width=0.23\columnwidth]{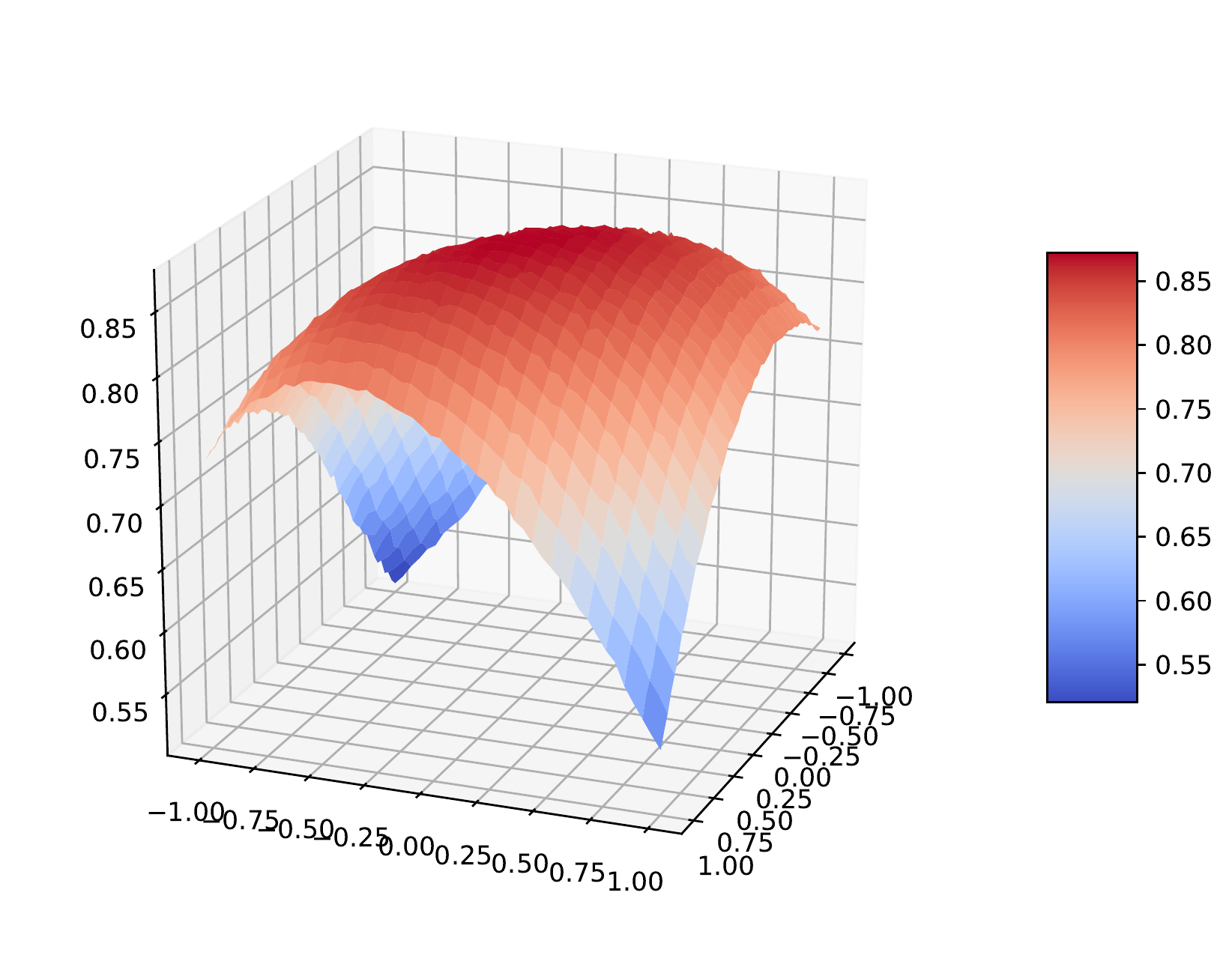}
	}
	\subfigure[DARTS-]{
		\includegraphics[width=0.23\columnwidth]{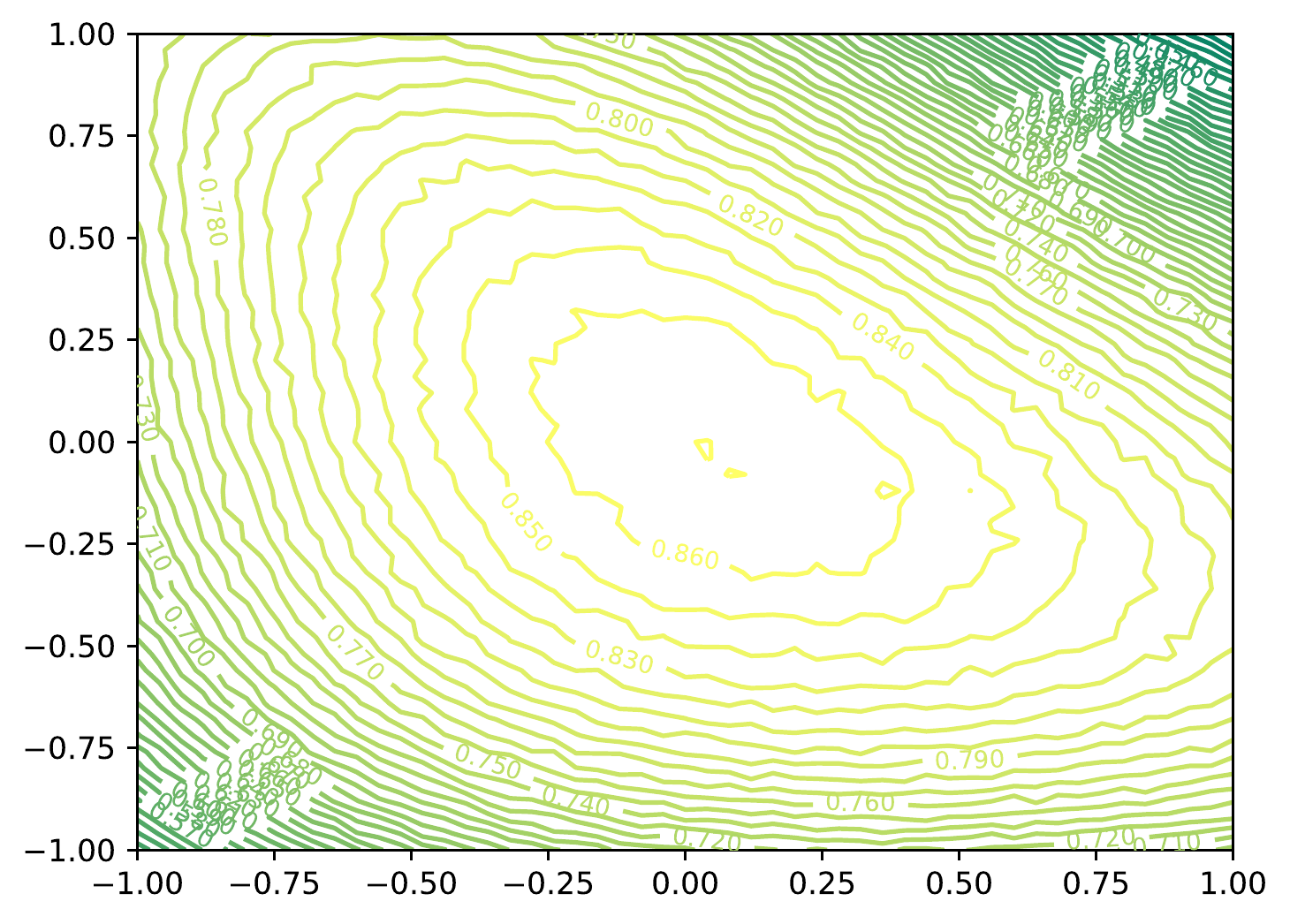} 
	}
	\subfigure[DARTS-]{
		\includegraphics[width=0.23\columnwidth]{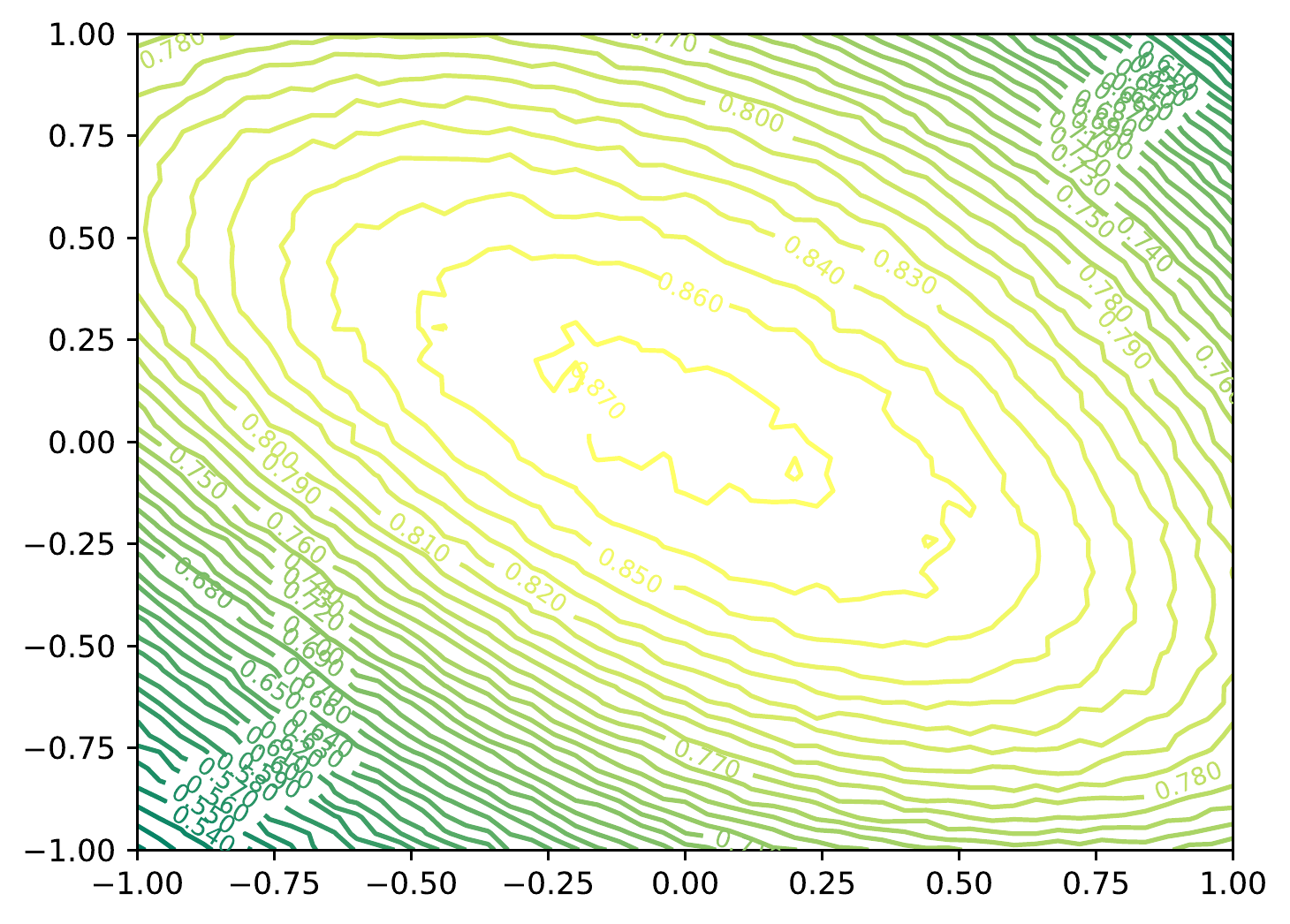}
	}
	\caption{More visualization of validation accuracy landscapes of  DARTS (a,b) and DARTS- (e,f) w.r.t. the architectural weights $\alpha$ on CIFAR-10 in S3. Their contour maps are shown respectively in (c,d) and (g,h). The step of contour map is 0.1.}
	\label{fig:landcape-loss-val-more-s3}
\end{figure}

To show the consistent smoothing capacity of DARTS-, we draw more loss landscapes in S0 and S3 (on several seeds) w.r.t architectural parameters and their contours in Figure \ref{fig:landcape-loss-val-more-s0} and \ref{fig:landcape-loss-val-more-s3}.  Generally, DARTS-'s slopes are more inflated if we consider them as camping tents, which suggest better convergence of the over-parameterized network. 

\subsection{List of Experiments}

We summarize all the conducted experiments with their related figures and tables in Table \ref{tab:list-exp}.

\begin{table*}[ht]
\centering
\caption{List of experiments conducted in this paper}\smallskip
\small
\begin{tabular}{*{7}{l}}
\toprule
Method & Search Space & Dataset &  \multicolumn{2}{c}{Figures}  &  \multicolumn{2}{c}{Tables} \\ 
& &  &  \scriptsize{Main text} &  \scriptsize{Supp.} & \scriptsize{Main text} &  \scriptsize{Supp.} \\
\midrule
DARTS & S0 & CIFAR-10 &  &\ref{fig:landcape-loss-val-more-s0} &  & \\
DARTS & S3 & CIFAR-10 & \ref{fig:landcape-loss-val} & &  & \\
DARTS- & S0 &  CIFAR-10 & & \ref{fig:eigen_value},\ref{fig:sample-model-training-curve},\ref{fig:c10_best_cell},\ref{fig:landcape-loss-val-more-s0}  &\ref{tab:CNN-standard-space},\ref{tab:comparison-cifar-imagenet} & \\
DARTS- & S1-S4 &  CIFAR-10 &  \ref{fig:landcape-loss-val} &\ref{fig:eigen_value},\ref{fig:beta-loss},\ref{fig:c10_best_cell},\ref{fig:landcape-loss-val-more-s3} & \ref{tab:comparison-rdarts-s2-s3-best},\ref{tab:comparison-rdarts-s2-s3-avg} & \ref{tab:beta-sensitiveness} \\
DARTS- & S5 &  ImageNet & & \ref{fig:sample-model-training-curve},\ref{fig:darts-a-arch-imagnet} & \ref{tab:comparison-cifar-imagenet} &\\ 
DARTS- & S5 &  MS COCO & & & \ref{table:darts-coco-retina} &\\
DARTS- & S6 & CIFAR-10 & &  & \ref{table:nas-bench-201} & \\
DARTS- & S1-S4 &  CIFAR-100 & & \ref{fig:eigen_value},\ref{fig:c100_best_cell}   &\ref{tab:comparison-rdarts-s2-s3-best} &  \ref{tab:comparison-rdarts-s2-s3-avg} \\
DARTS- & S0 &  CIFAR-100 & &\ref{fig:eigen_value},\ref{fig:c100_best_cell} & \ref{tab:CNN-standard-space},\ref{tab:comparison-cifar100} &\\
%P-DARTS w/o dropout & S0 & CIFAR-10 & & \ref{fig:c10_p_ndp_cell}  & \ref{tab:pdarts-darts-} &\\
P-DARTS w/o M=2      & S0  & CIFAR-10 & & \ref{fig:c10_p_ndp_cell} & \ref{tab:pdarts-darts-}  &\\ 
P-DARTS w/ auxiliary skip & S0  &CIFAR-10 &  &\ref{fig:c10_p_ask_cell}  & \ref{tab:pdarts-darts-} &\\
PC-DARTS w/o channel shuffling & S0 &CIFAR-10 & & \ref{fig:c10_pc_ns_cell}   & \ref{tab:pcdarts-darts-}  &\\
PC-DARTS w/ auxiliary skip & S0  & CIFAR-10 &  &\ref{fig:c10_pc_ask_cell}   &  \ref{tab:pcdarts-darts-} & \\
\bottomrule
\end{tabular}
\label{tab:list-exp}
\end{table*}

\section{Figures of Genotypes}\label{app:fig-geno}

\begin{figure}[ht]
	\centering
	\vskip -0.2 in
	\subfigure[S0]{
		\includegraphics[width=0.3\columnwidth]{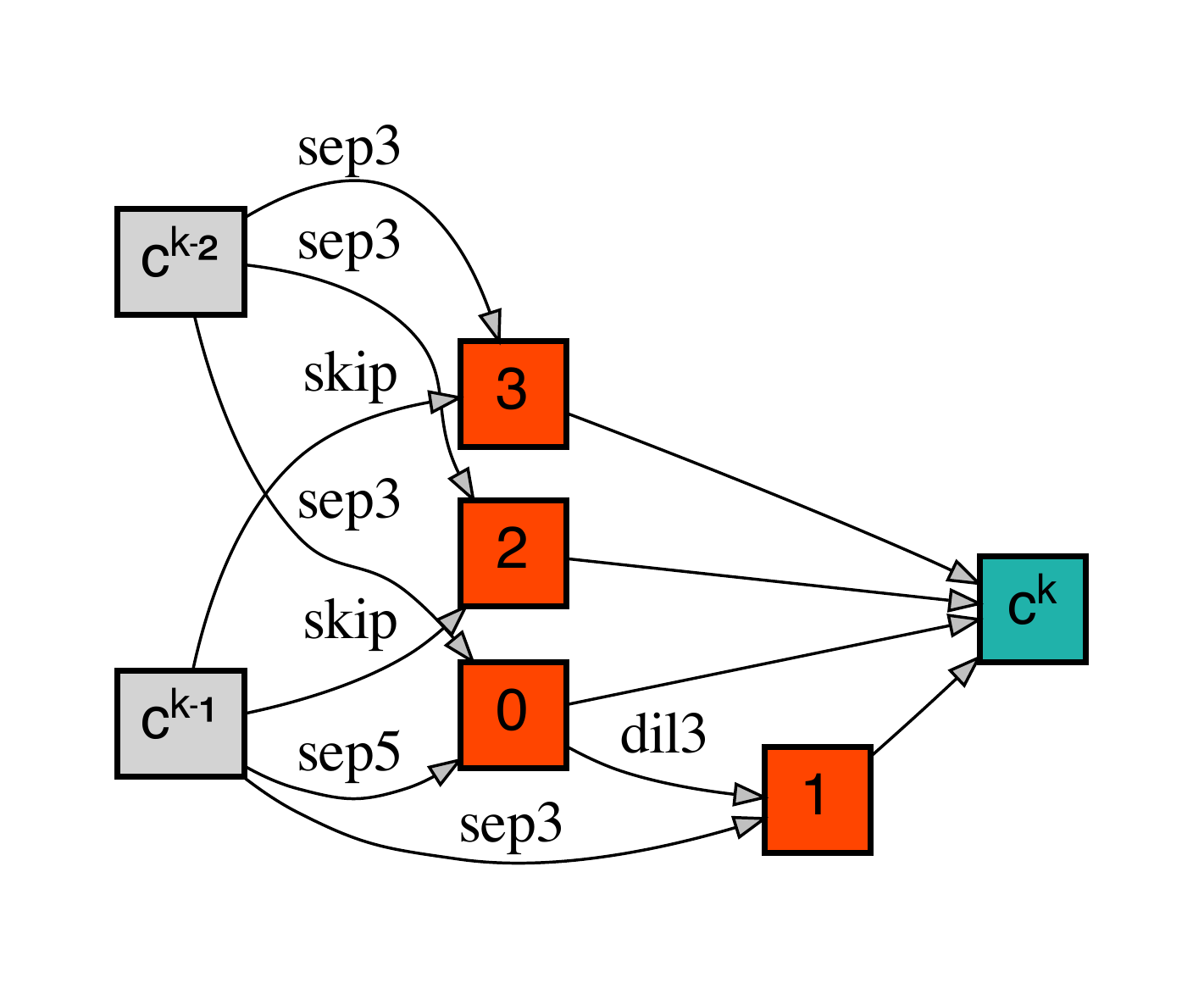} 
		\includegraphics[width=0.4\columnwidth]{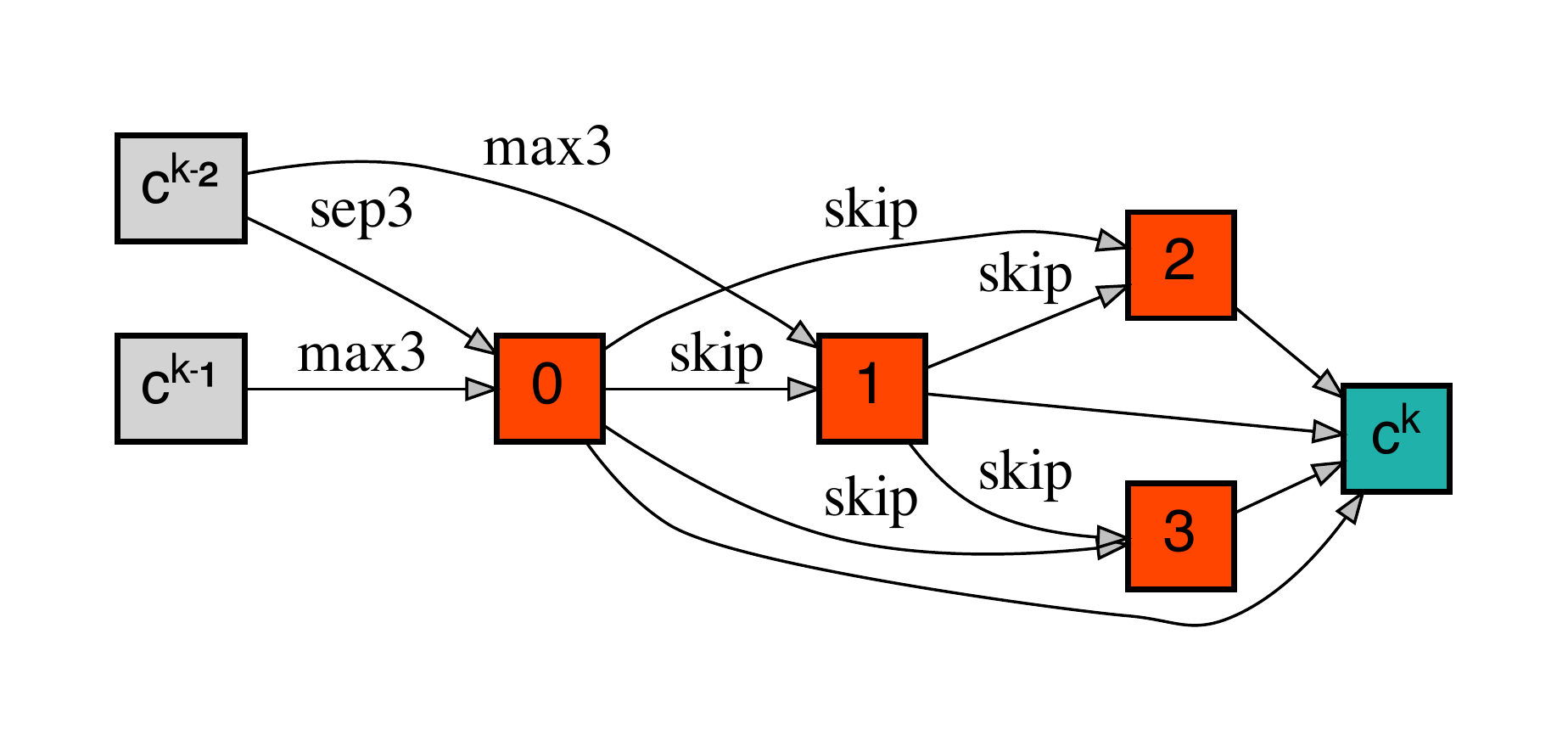}
	}
	\subfigure[S1]{
		\includegraphics[width=0.23\columnwidth]{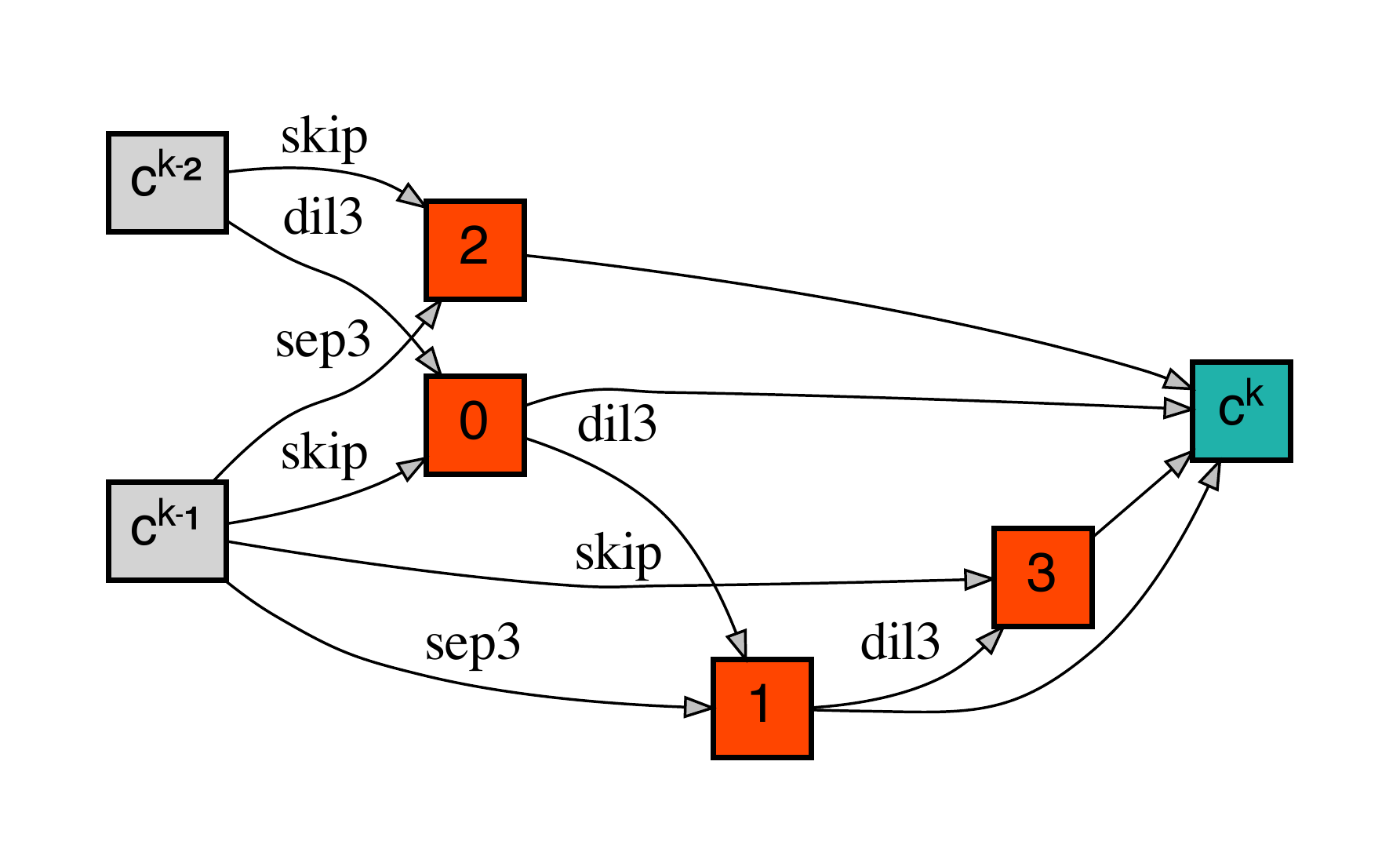} 
		\includegraphics[width=0.23\columnwidth]{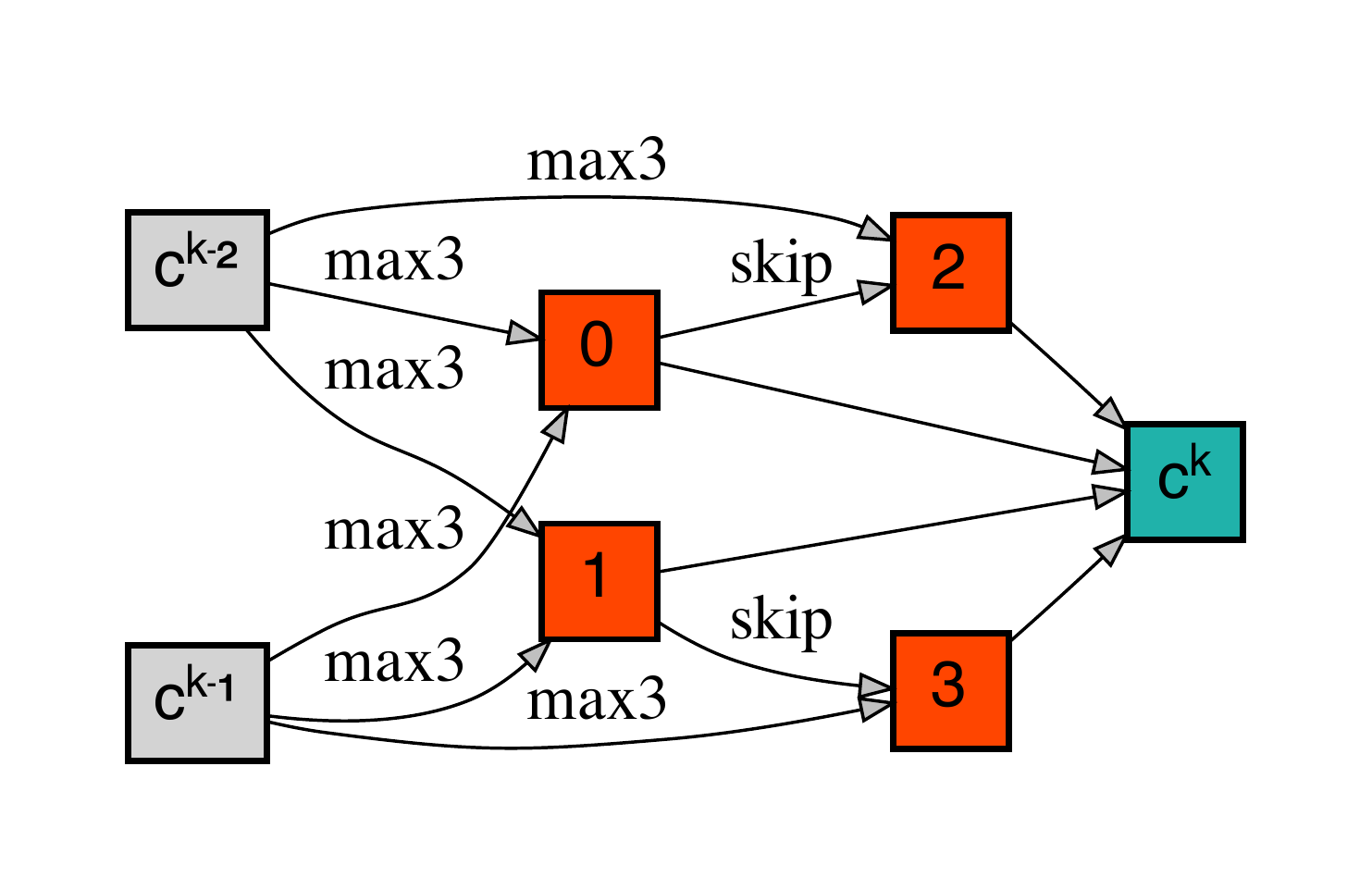}
	}
	\subfigure[S2]{
		\includegraphics[width=0.24\columnwidth]{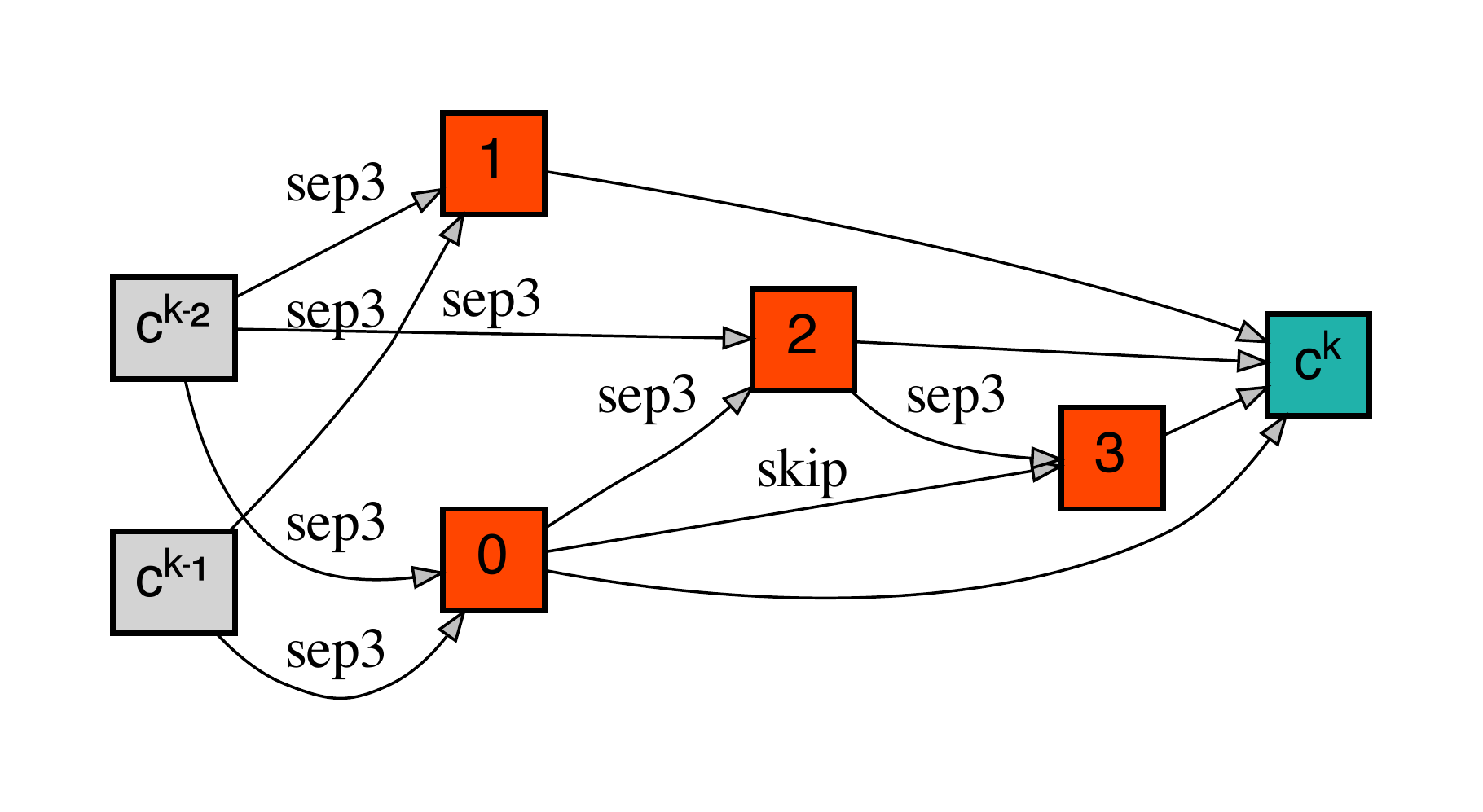} 
		\includegraphics[width=0.25\columnwidth]{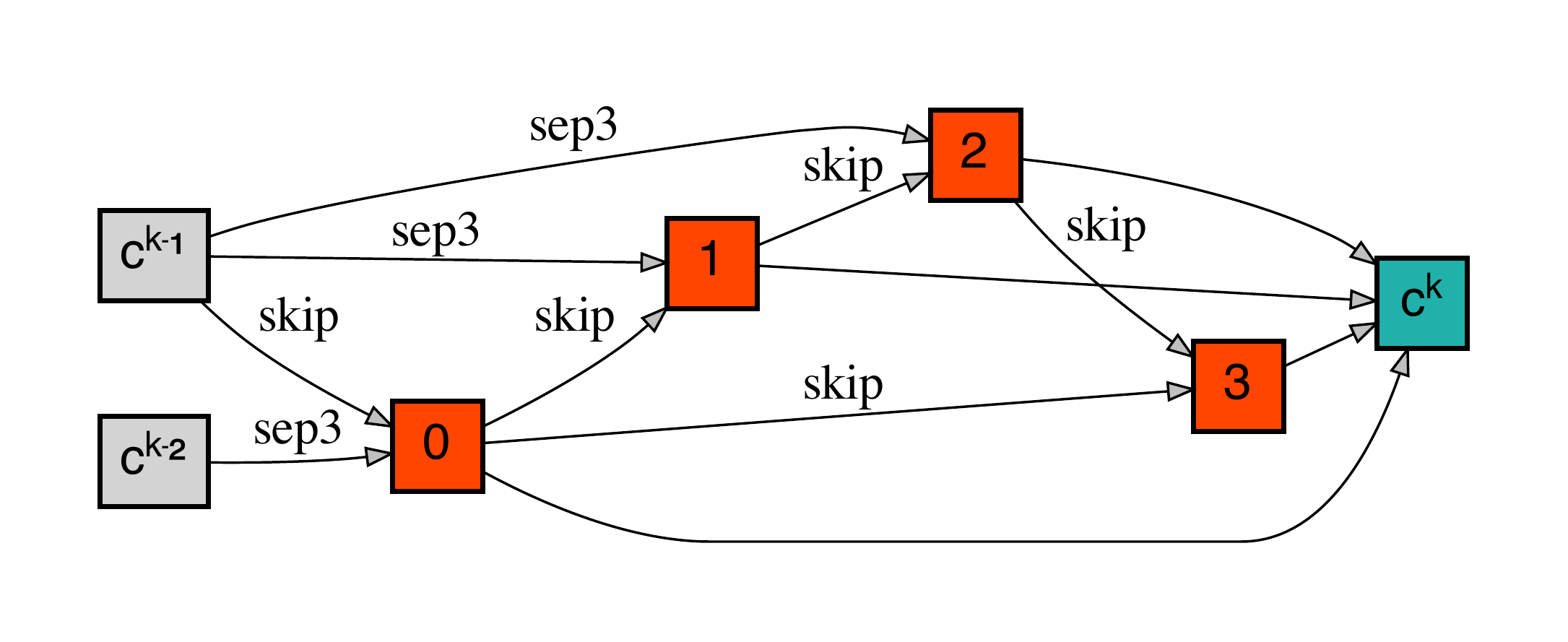}
	}
	\subfigure[S3]{
		\includegraphics[width=0.2\columnwidth]{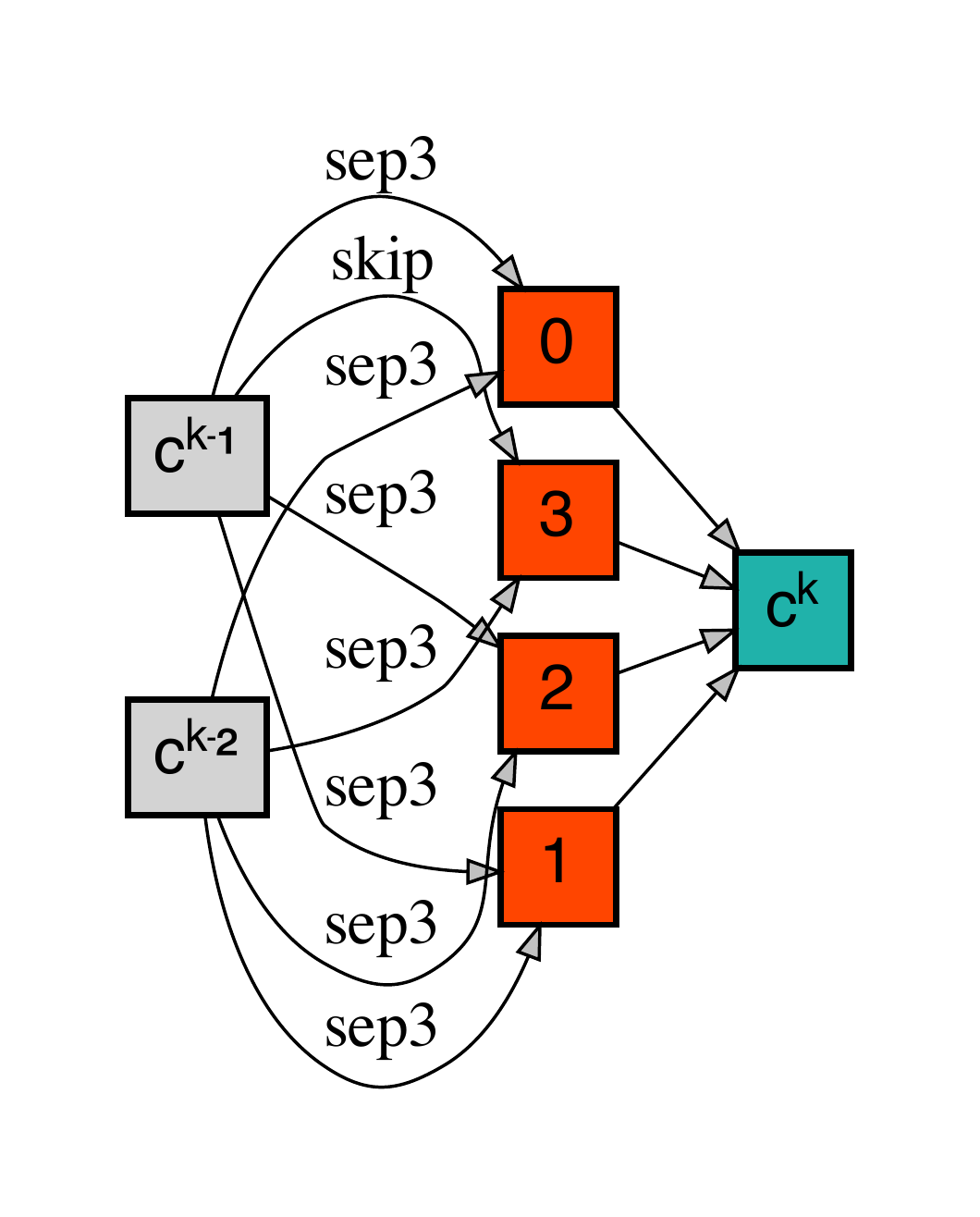} 
		\includegraphics[width=0.25\columnwidth]{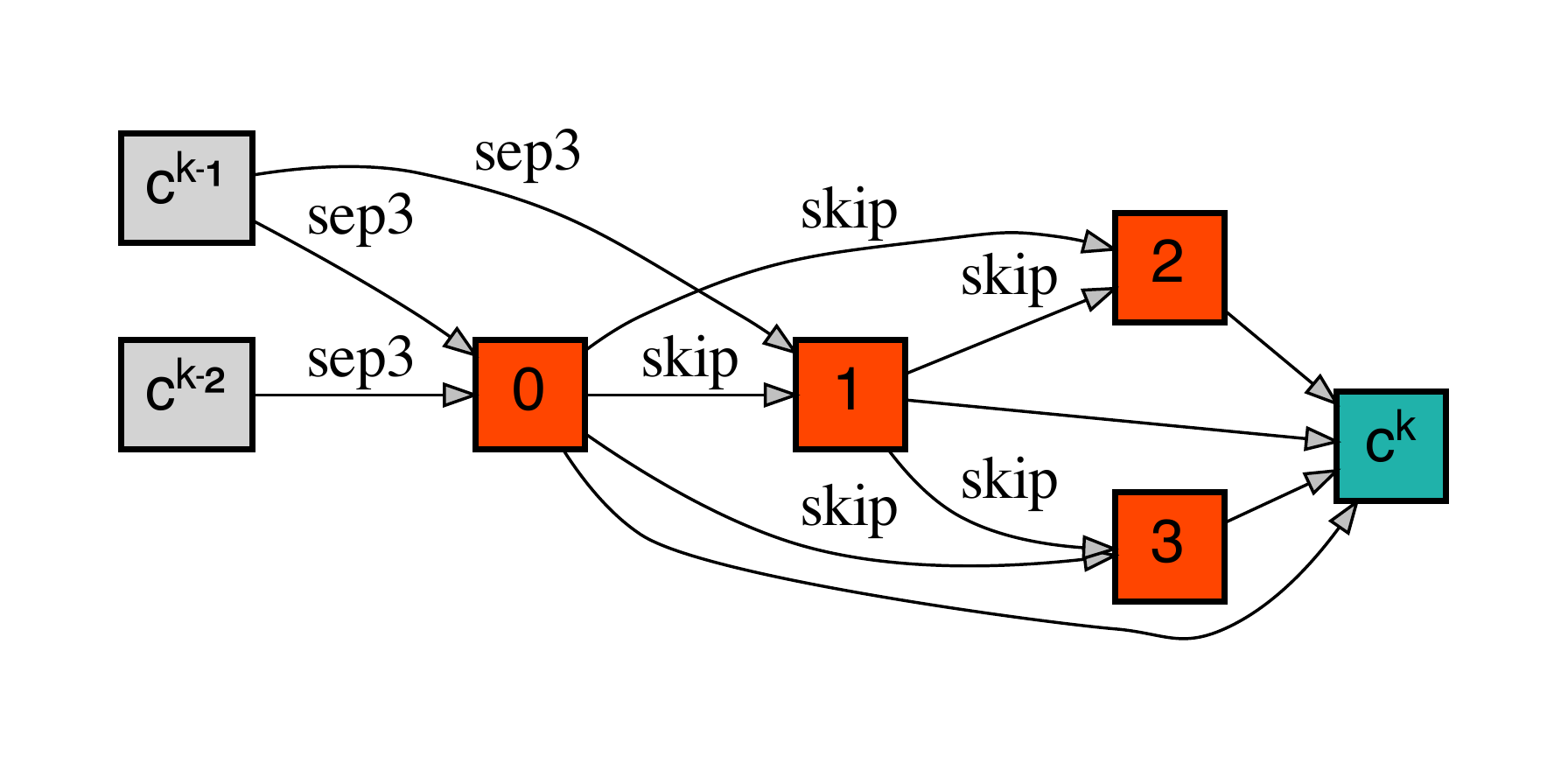}
	}
	\subfigure[S4]{
		\includegraphics[width=0.25\columnwidth]{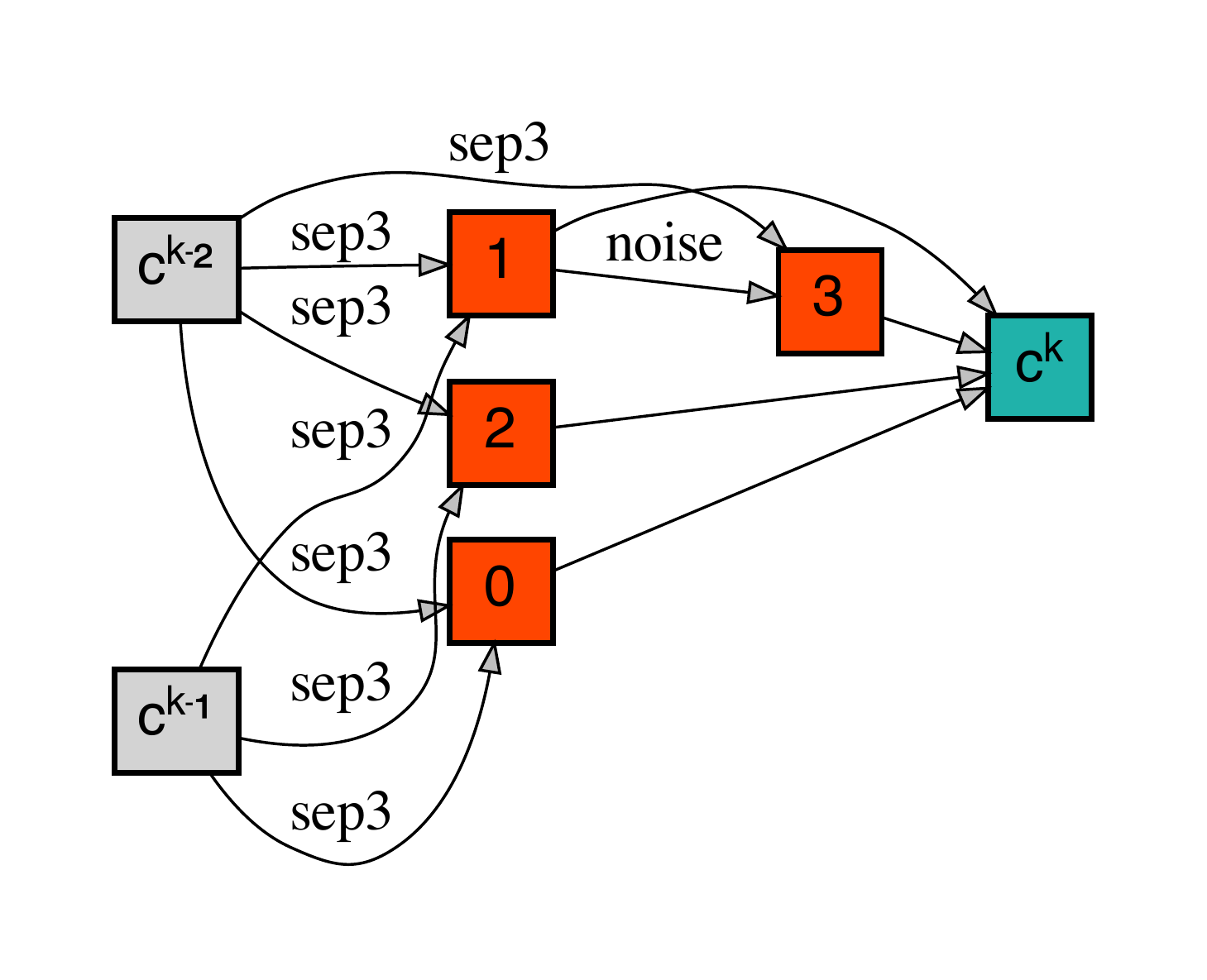} 
		\includegraphics[width=0.2\columnwidth]{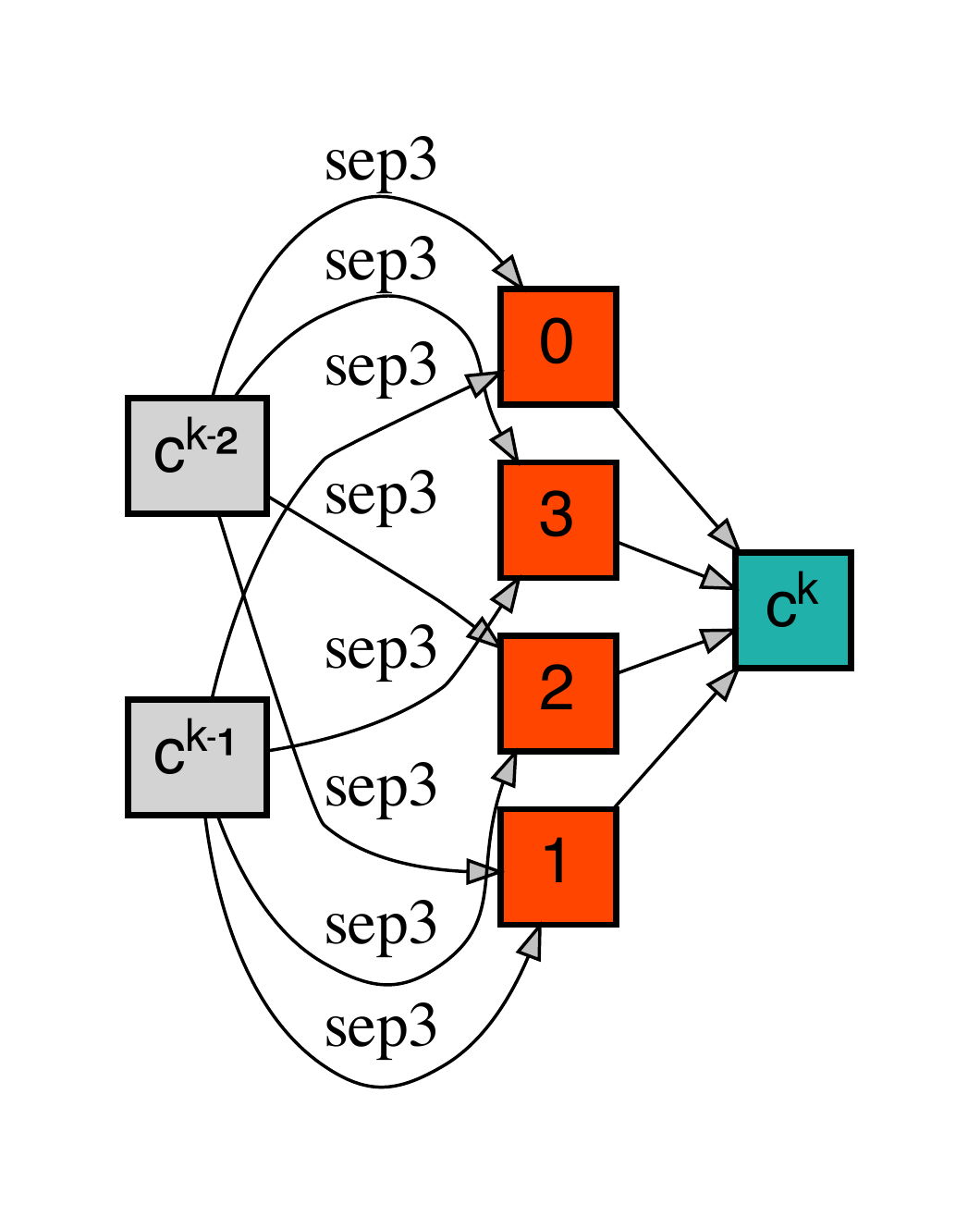}
	}
	\caption{The best found normal cell and reduction cell in search spaces S0-S4 on CIFAR-10 dataset.}
	\label{fig:c10_best_cell}
\end{figure}

\begin{figure*}[ht]
	\centering
	\includegraphics[width=\columnwidth]{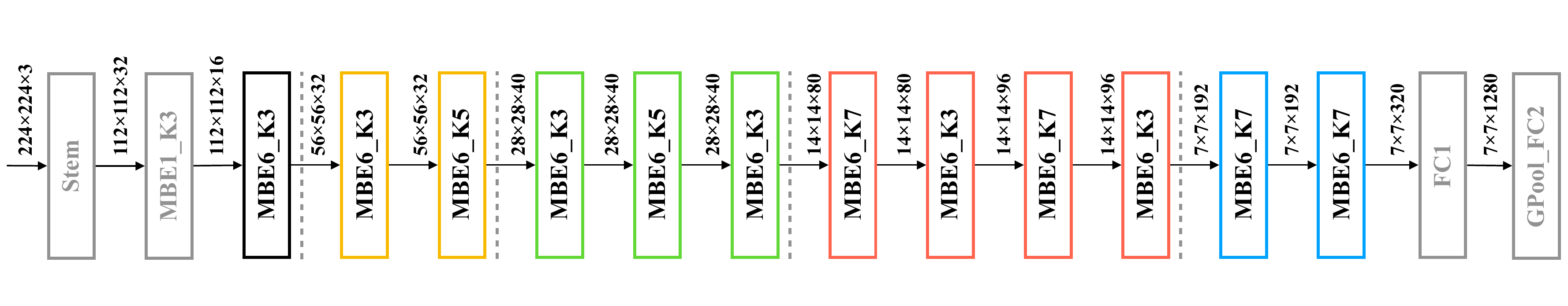} 
	\caption{Architecture of DARTS-A searched on ImageNet dataset.}
	\label{fig:darts-a-arch-imagnet}
\end{figure*}

\begin{figure}[ht]
	\centering
	\vskip -0.2 in
	\subfigure[S0]{
		\includegraphics[width=0.2\columnwidth]{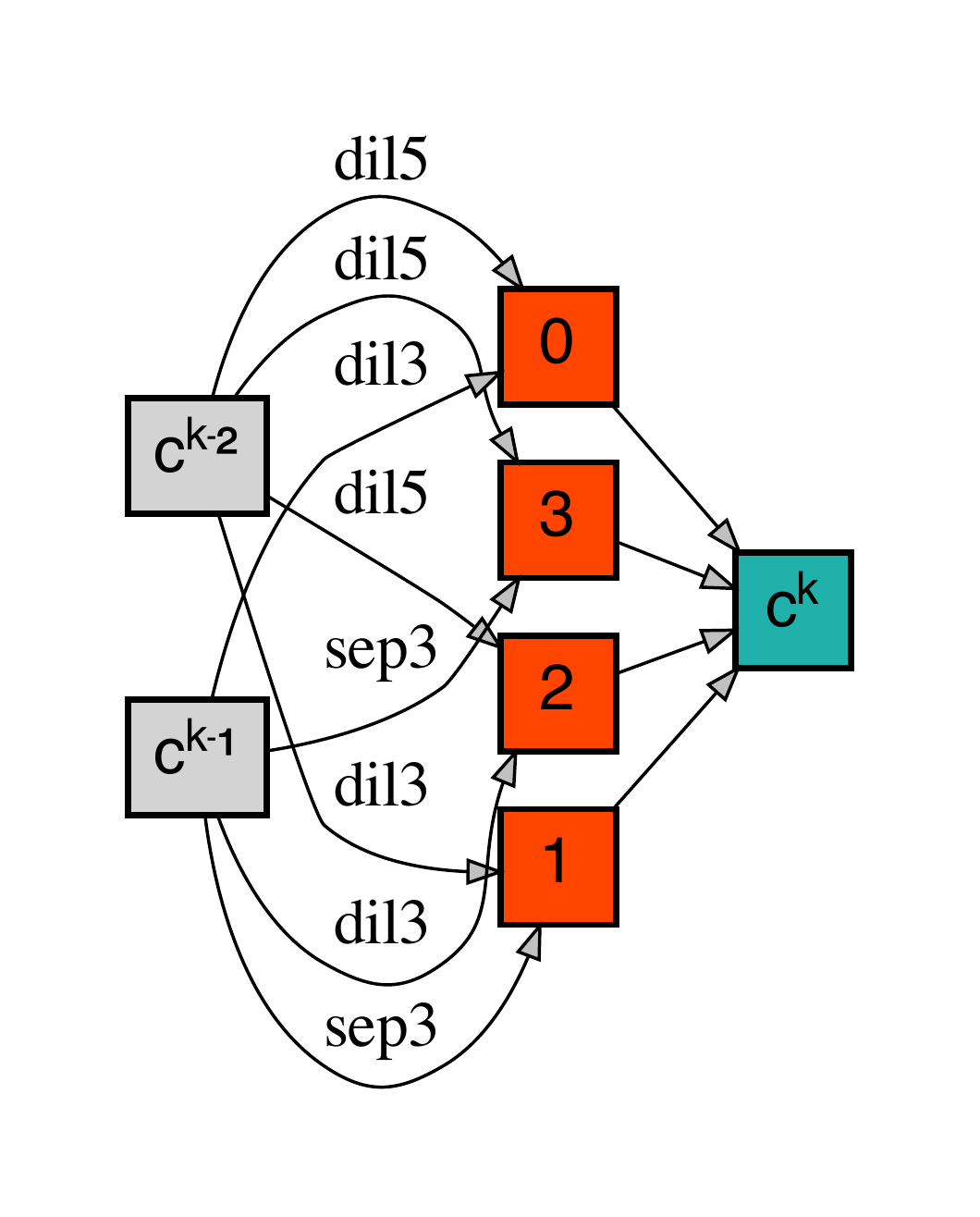} 
		\includegraphics[width=0.4\columnwidth]{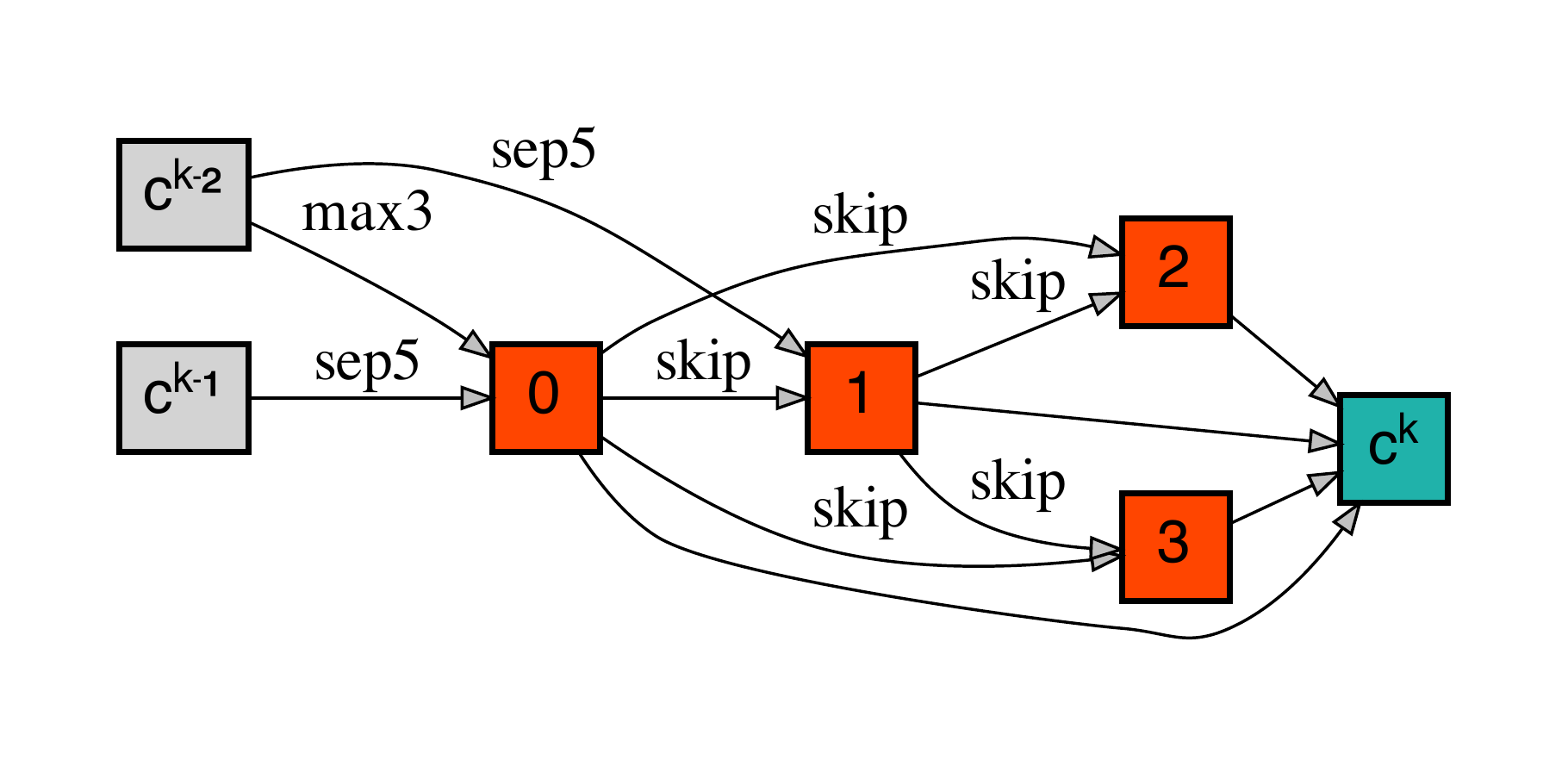}
	}
	\subfigure[S1]{
		\includegraphics[width=0.23\columnwidth]{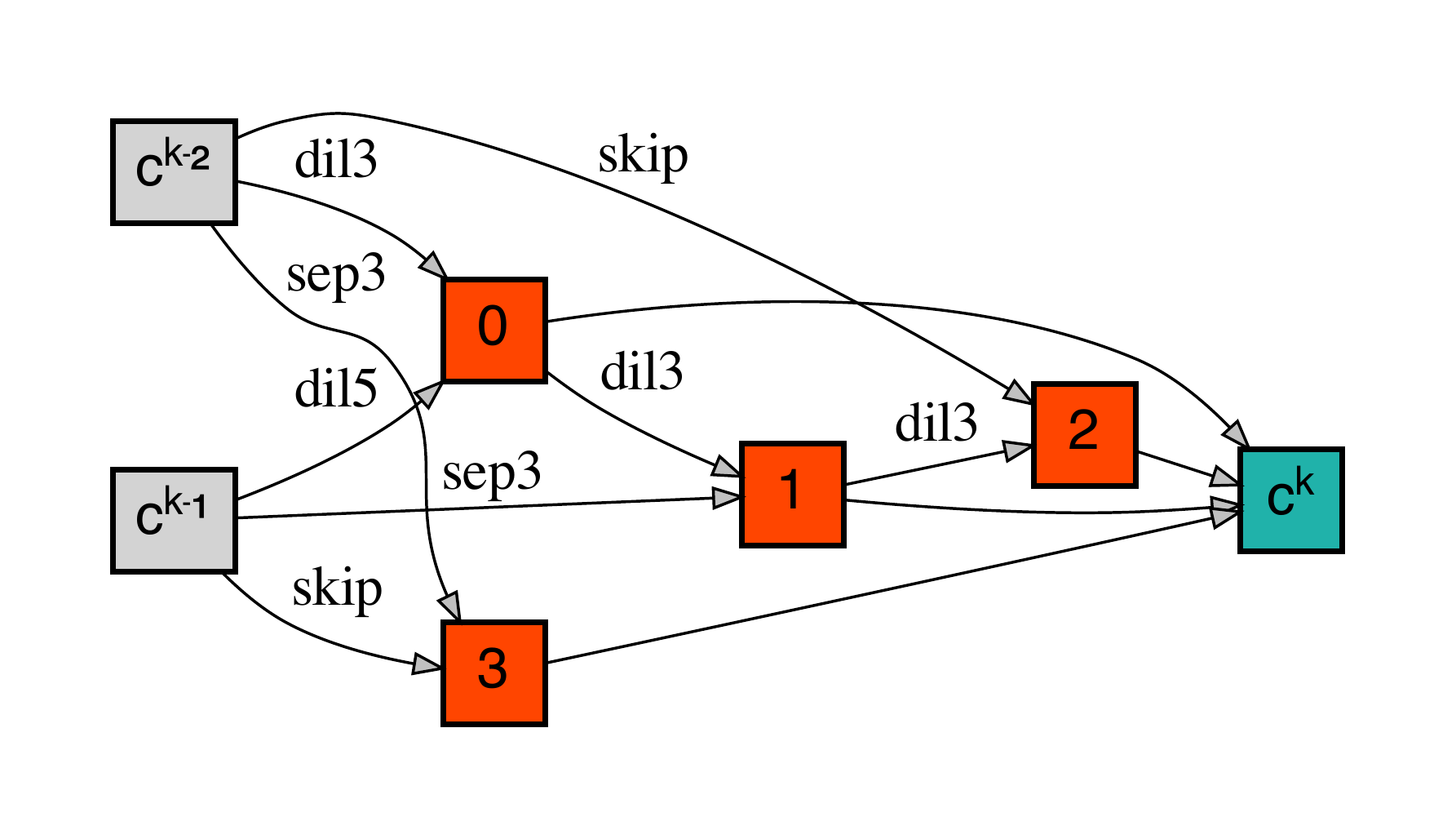} 
		\includegraphics[width=0.23\columnwidth]{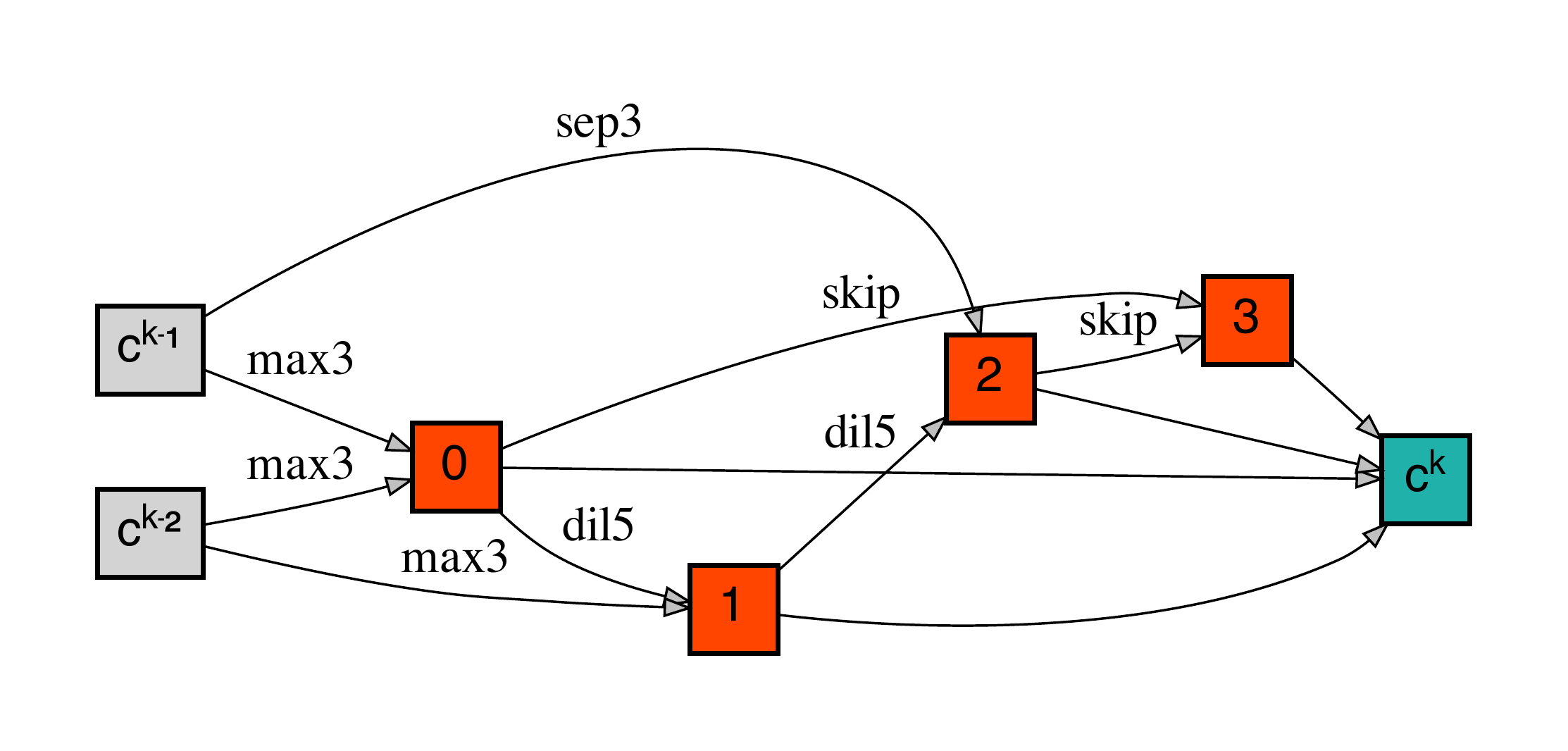}
	}
	\subfigure[S2]{
		\includegraphics[width=0.24\columnwidth]{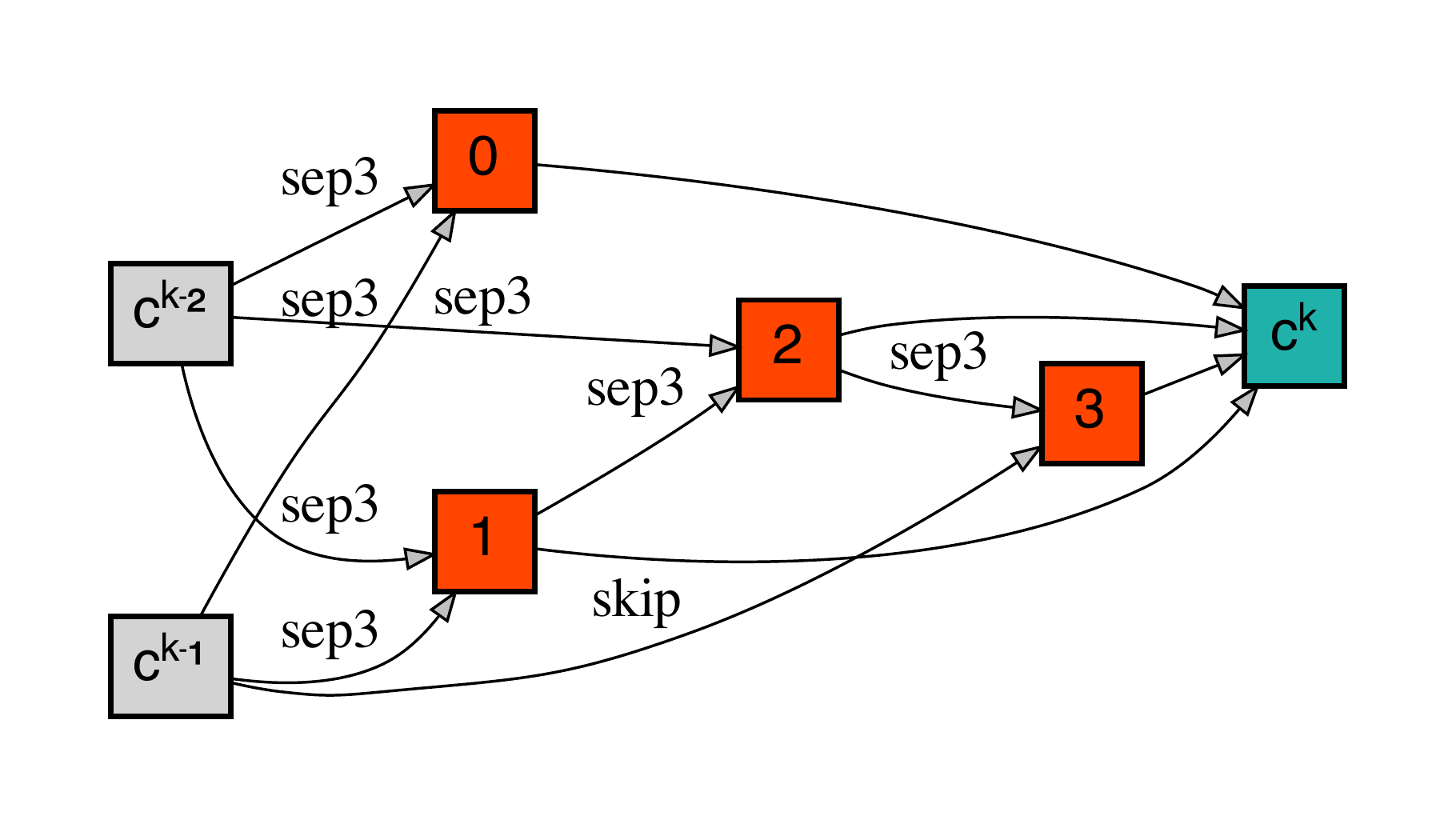} 
		\includegraphics[width=0.25\columnwidth]{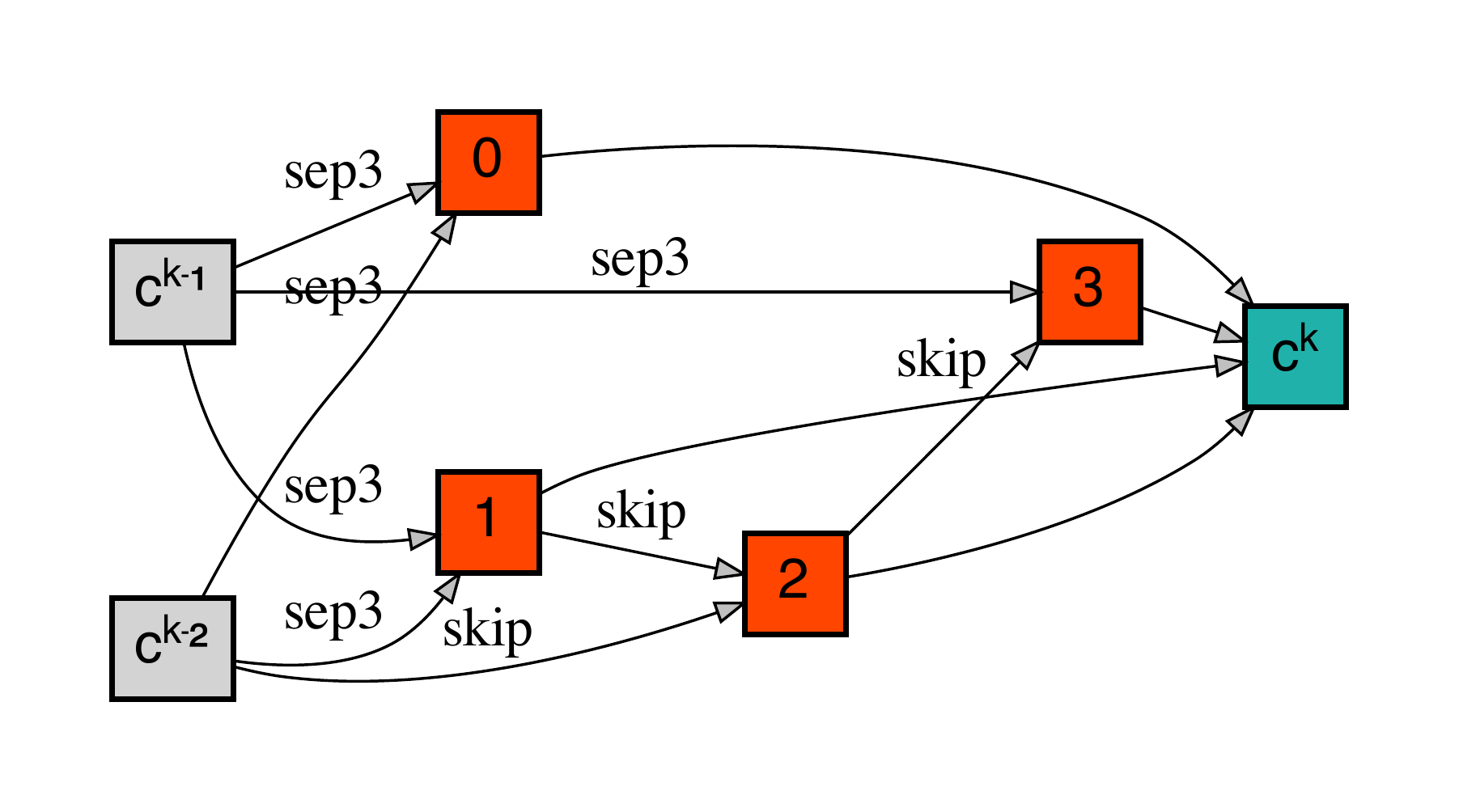}
	}
	\subfigure[S3]{
		\includegraphics[width=0.2\columnwidth]{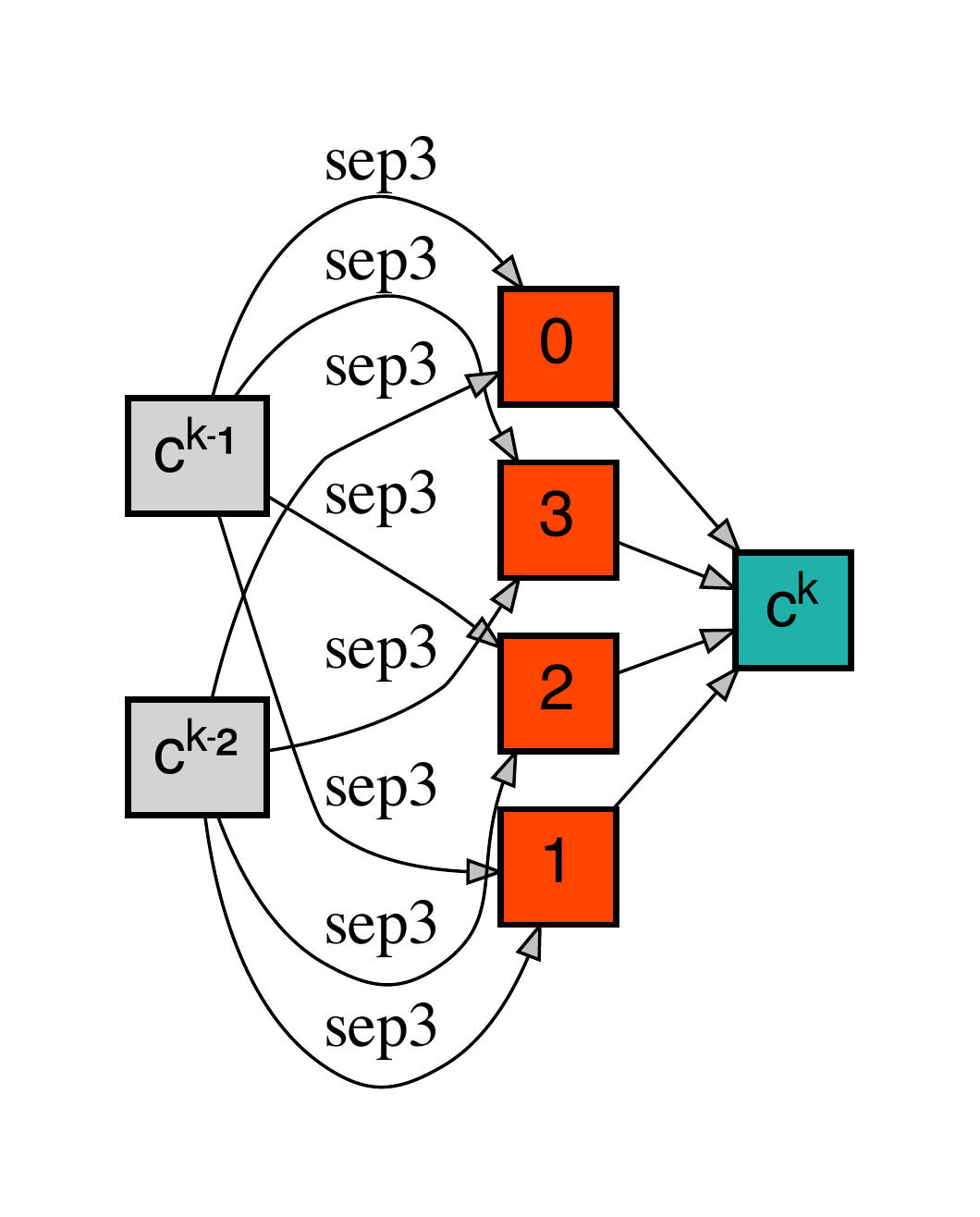} 
		\includegraphics[width=0.25\columnwidth]{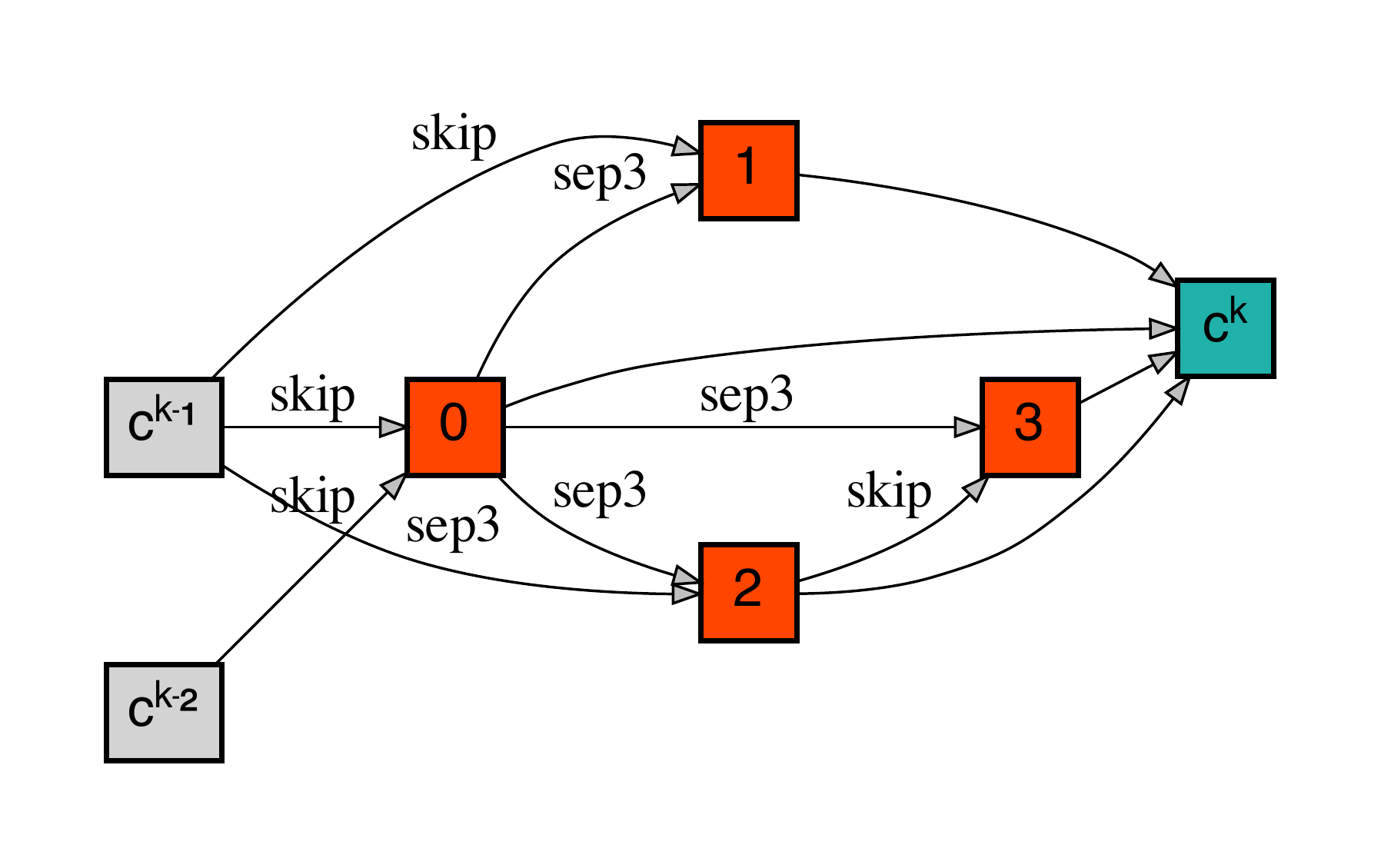}
	}
	\subfigure[S4]{
		\includegraphics[width=0.25\columnwidth]{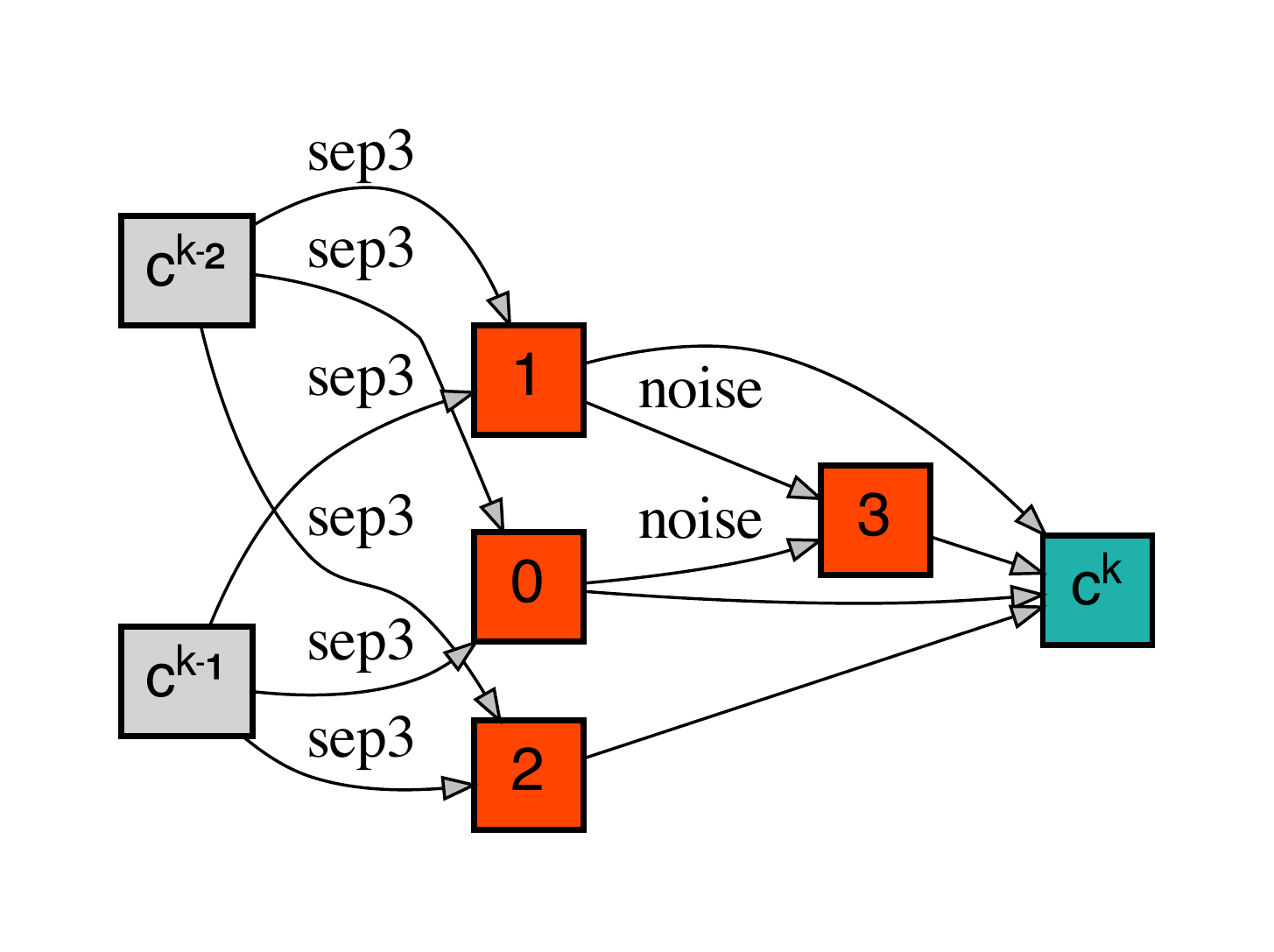} 
		\includegraphics[width=0.2\columnwidth]{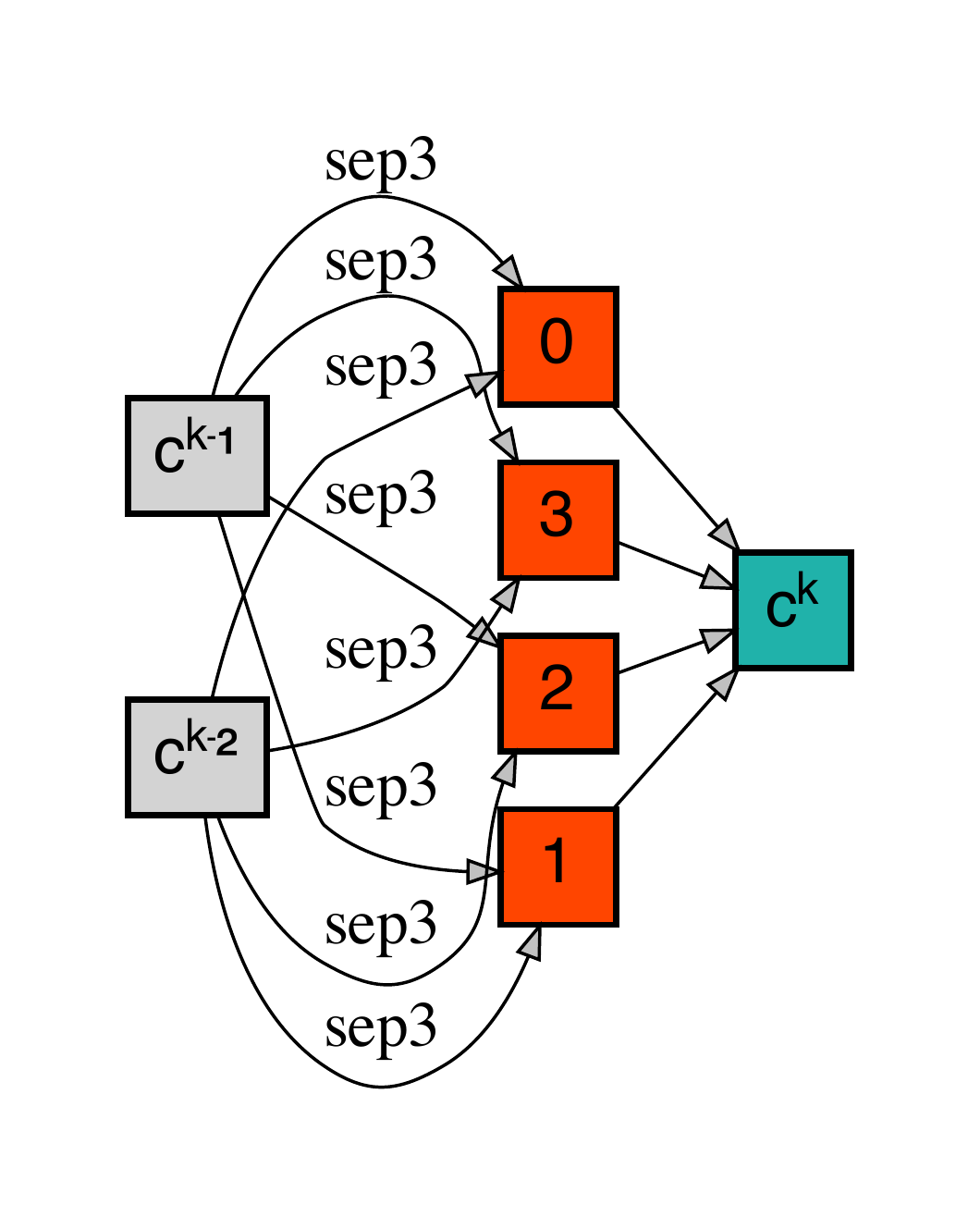}
	}
	\caption{The best found normal cell and reduction cell in search spaces S0-S4 on CIFAR-100 dataset.}
	\label{fig:c100_best_cell}
\end{figure}

\begin{figure*}[ht]
%trim=1cm 0 0 0,
%\captionsetup[subfigure]{labelformat=empty}
\centering
\begin{minipage}{2.6in}
\subfigure[Normal]{
\includegraphics[width=0.45\columnwidth]{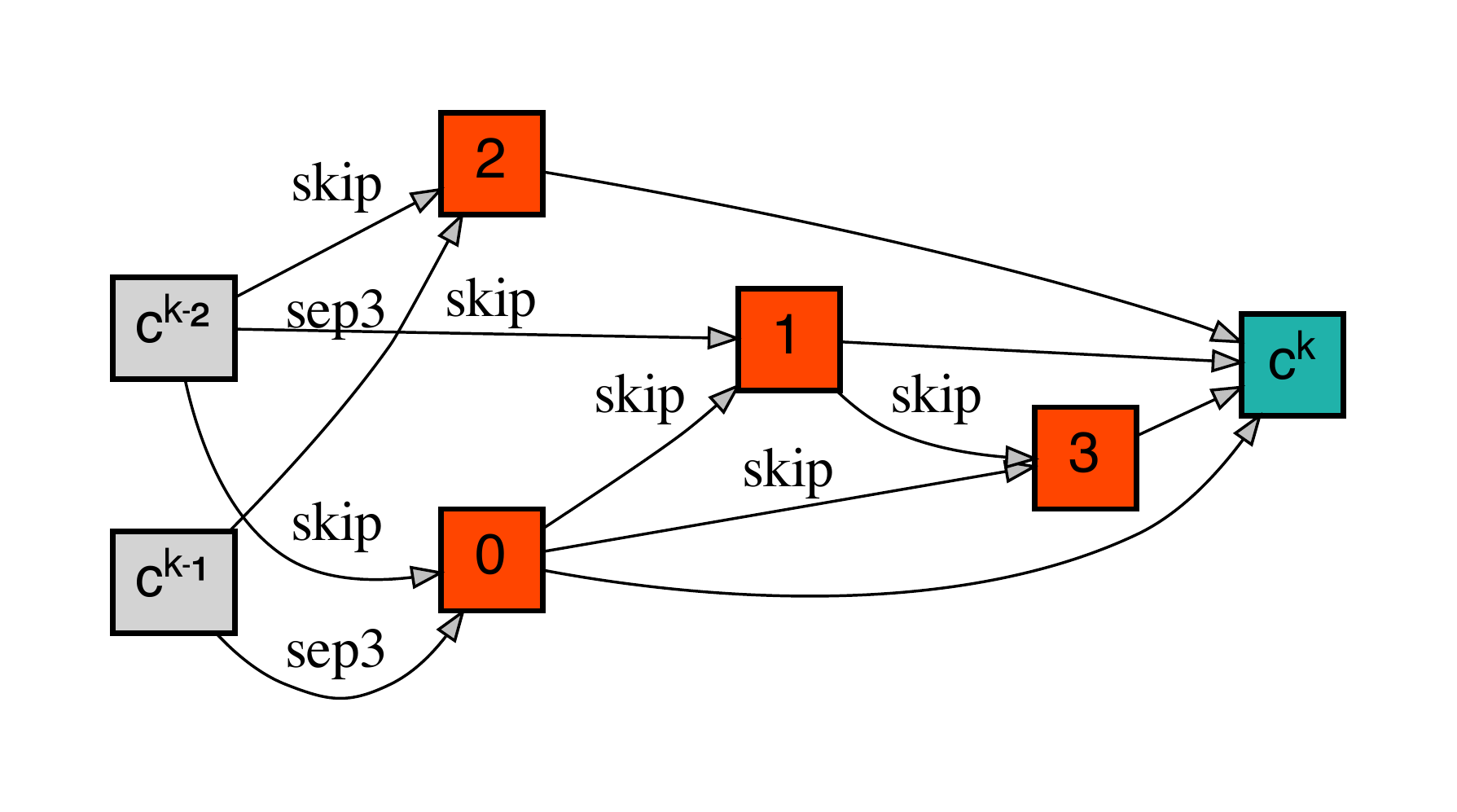} 
}
\subfigure[Reduction]{ 
\includegraphics[width=0.45\columnwidth]{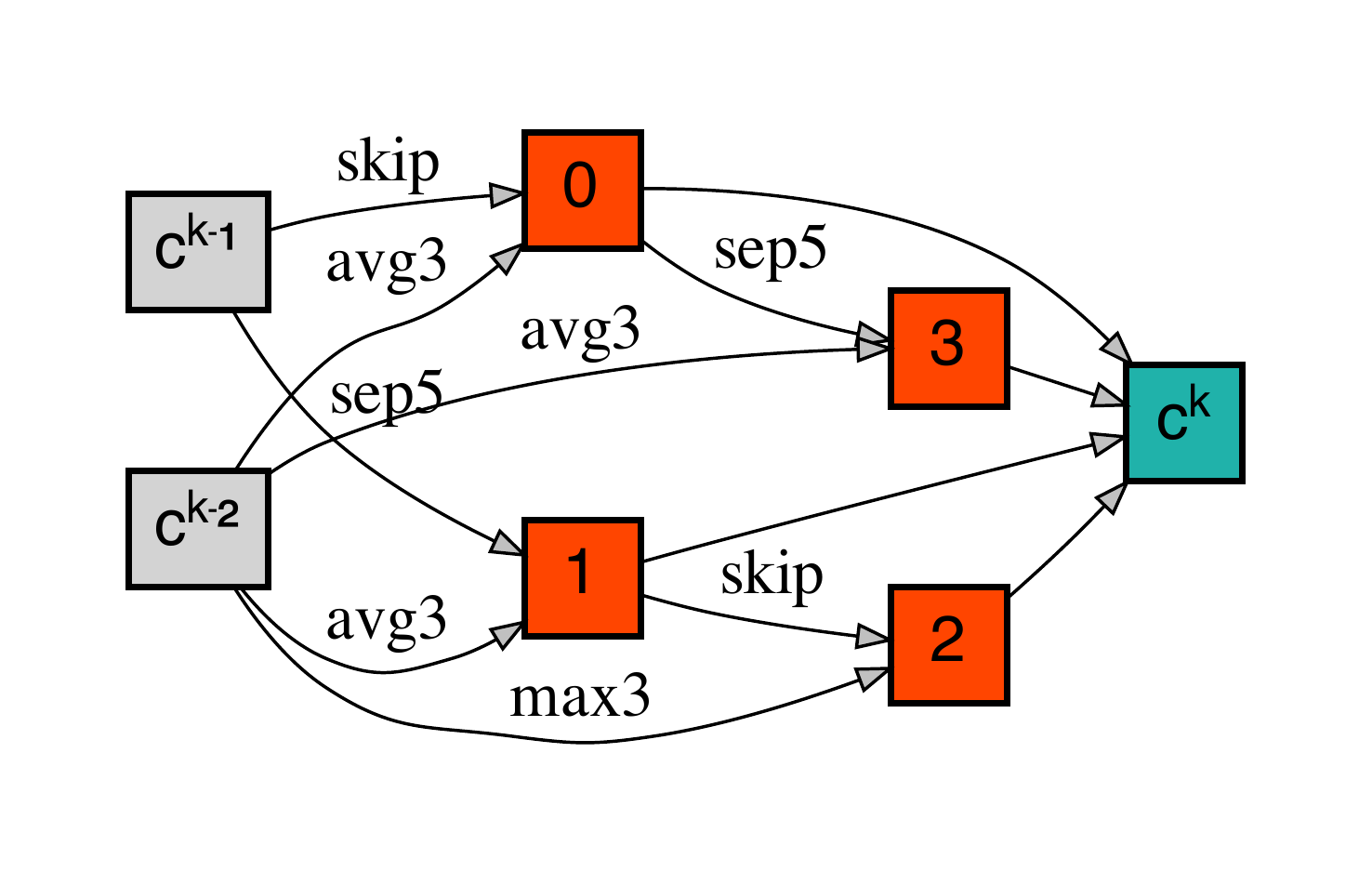}
}
\end{minipage}
\begin{minipage}{2.6in}
\subfigure[Normal]{
\includegraphics[width=0.45\columnwidth]{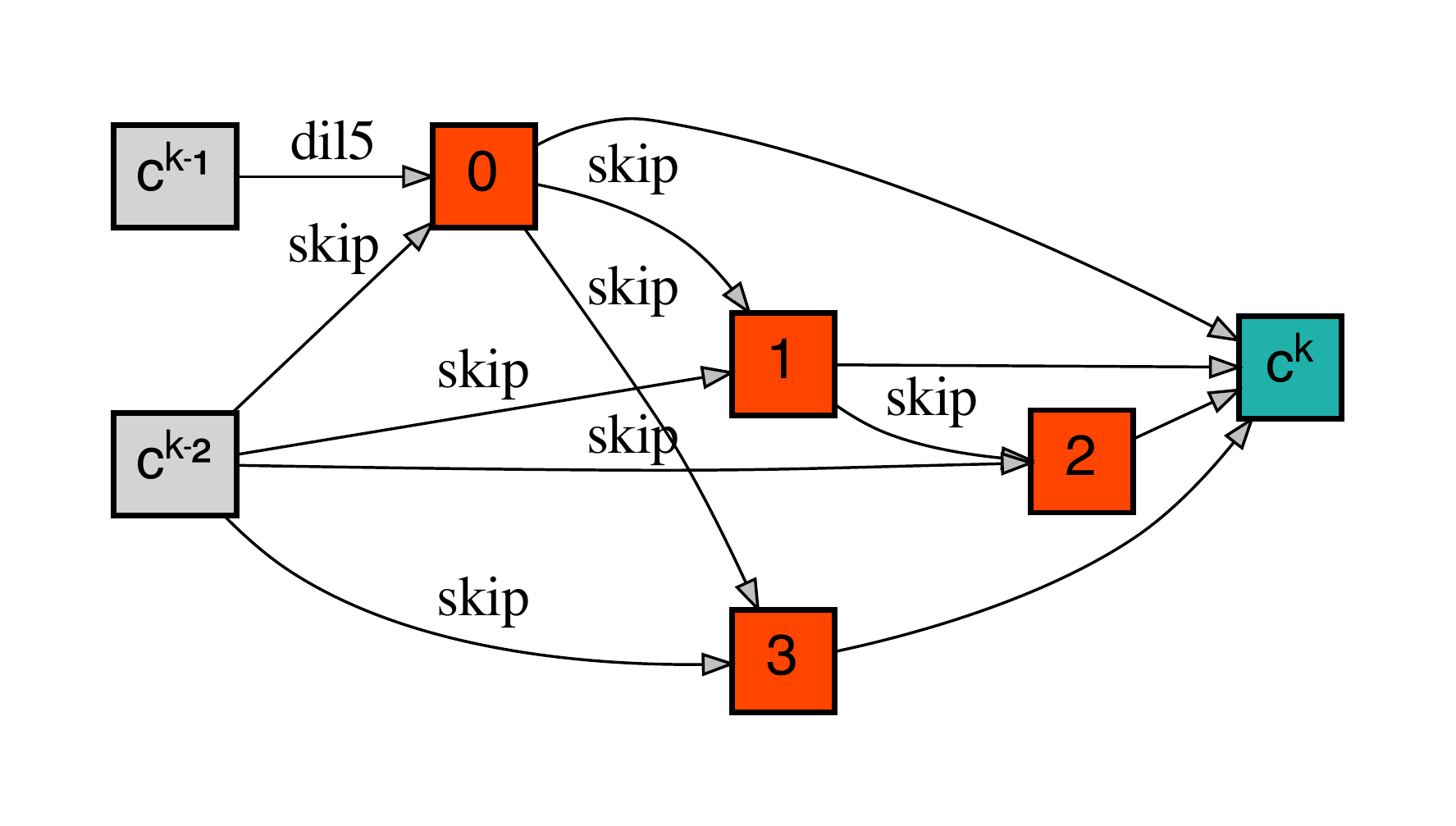} 
}
\subfigure[Reduction]{ 
\includegraphics[width=0.45\columnwidth]{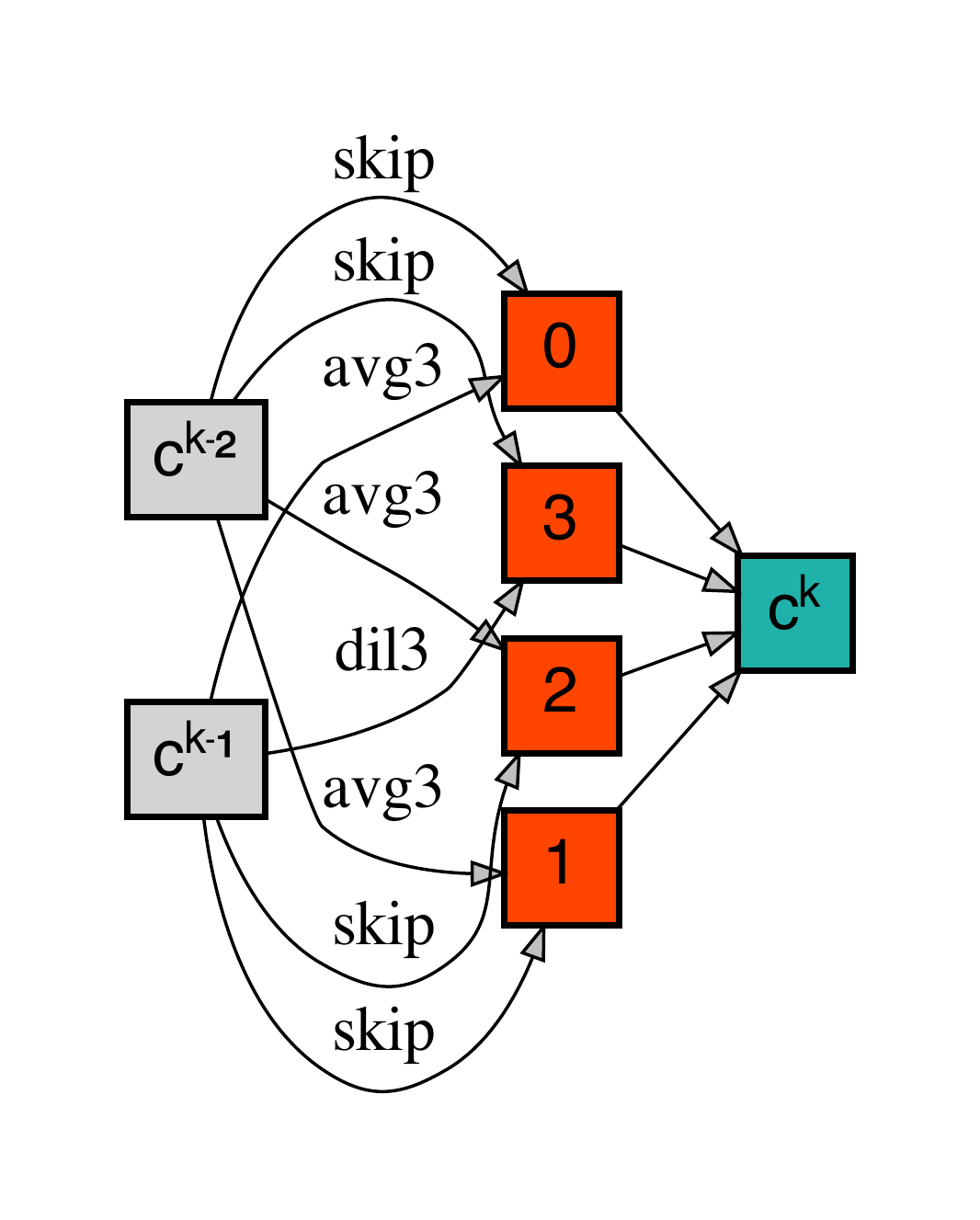}
}
\end{minipage}
\begin{minipage}{2.6in}
\subfigure[Normal]{
\includegraphics[width=0.45\columnwidth]{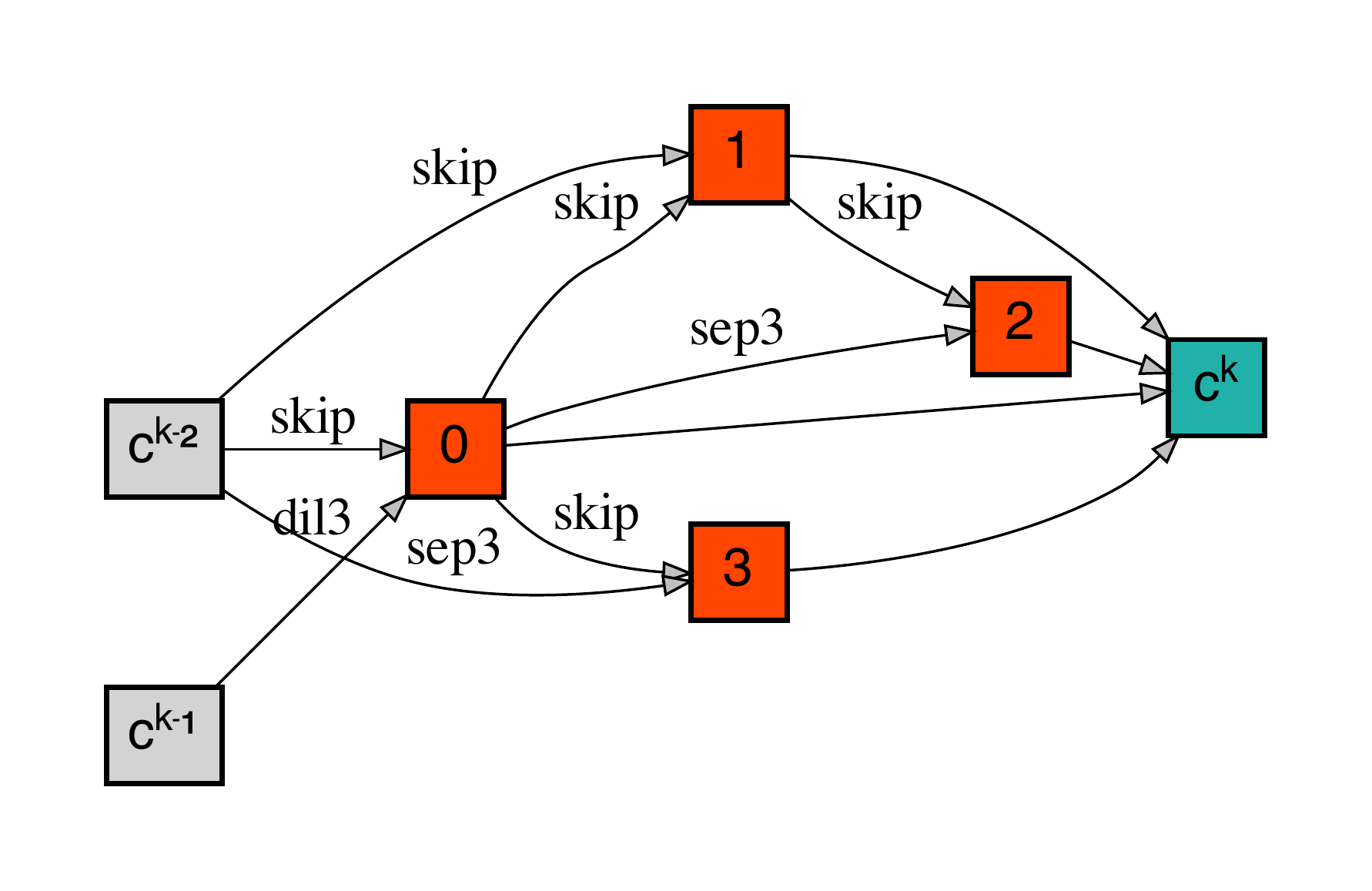} 
}
\subfigure[Reduction]{ 
\includegraphics[width=0.45\columnwidth]{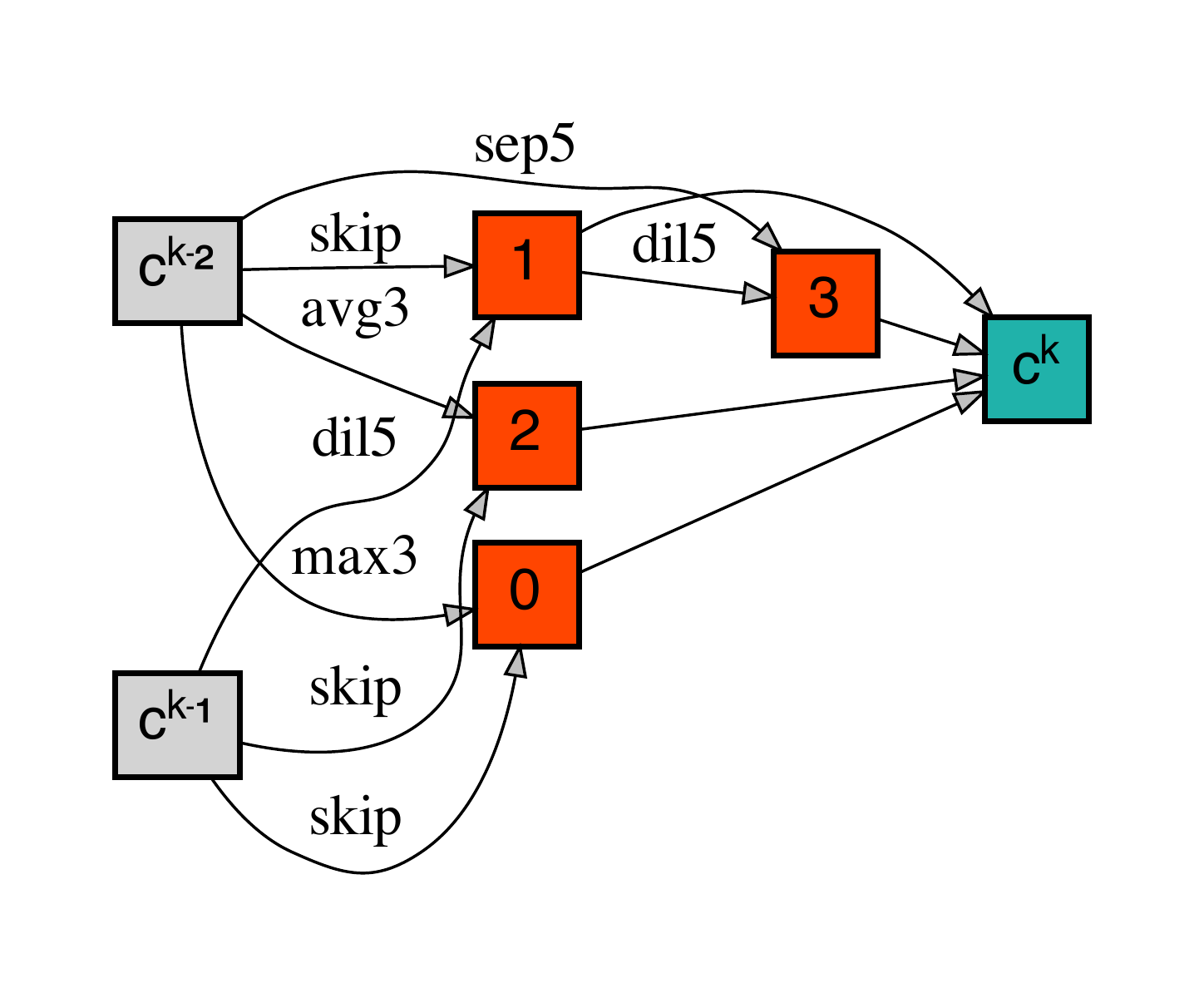}
}
\end{minipage}
\caption{ Found normal cells and reduction cells by P-DARTS \citep{chen2019progressive} without prior (M=2) in the DARTS' standard search space on CIFAR-10 dataset.}
\label{fig:c10_p_ndp_cell}
\end{figure*}

\begin{figure*}[ht]
%trim=1cm 0 0 0,
%\captionsetup[subfigure]{labelformat=empty}
\centering
\begin{minipage}{2.6in}
\subfigure[Normal]{
\includegraphics[width=0.45\columnwidth]{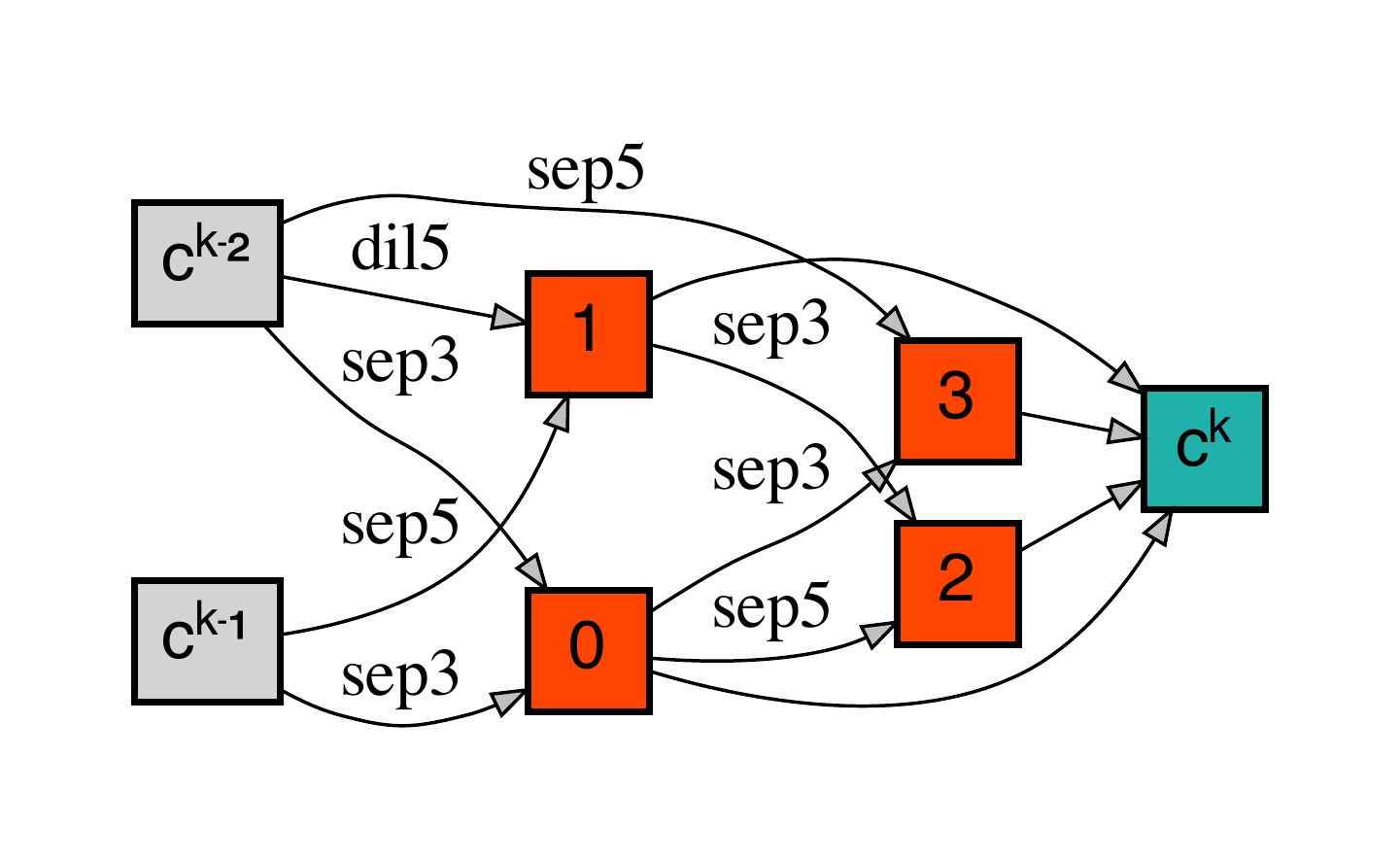} 
}
\subfigure[Reduction]{ 
\includegraphics[width=0.45\columnwidth]{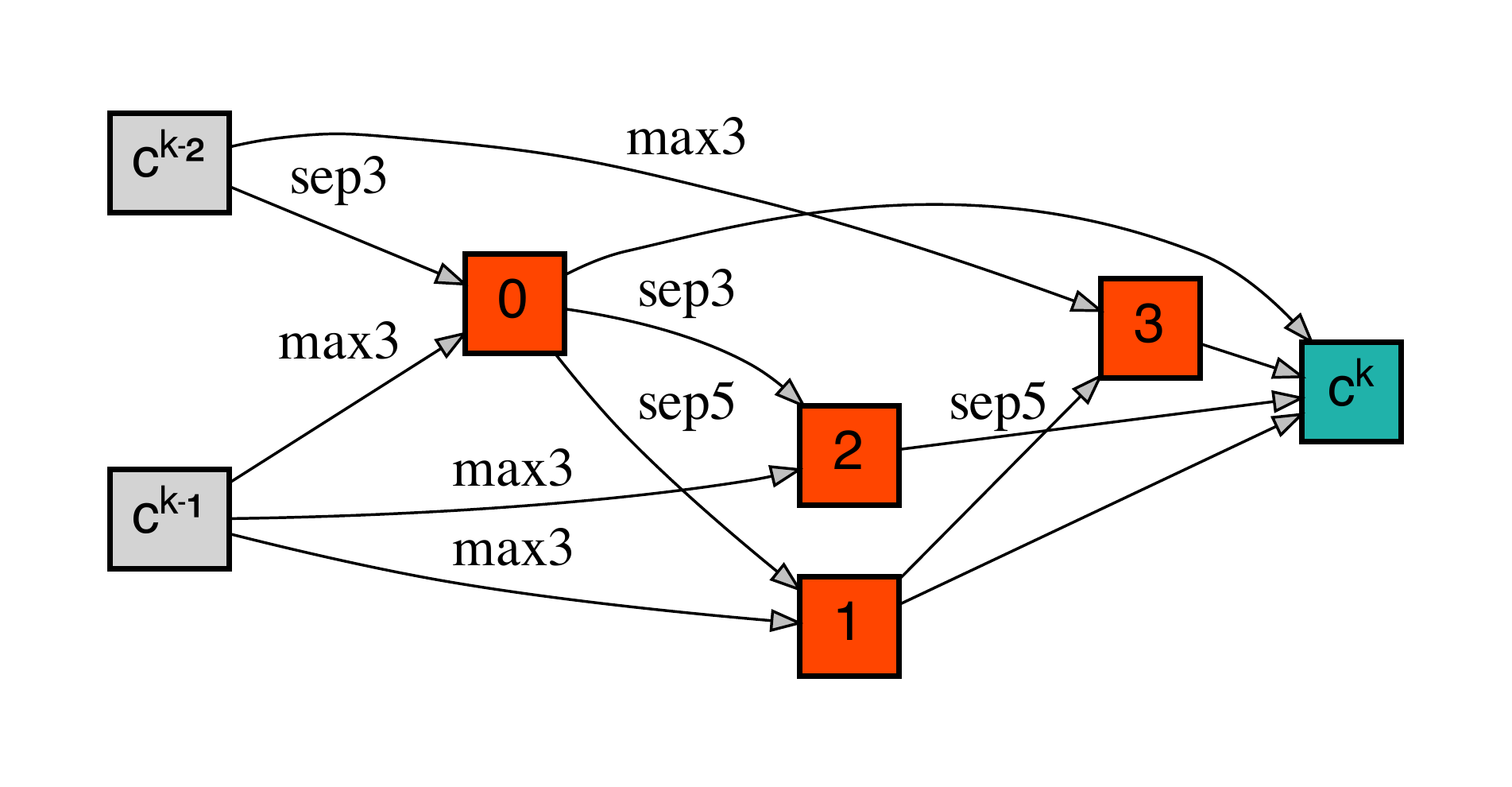}
}
\end{minipage}
\begin{minipage}{2.6in}
\subfigure[Normal]{
\includegraphics[width=0.45\columnwidth]{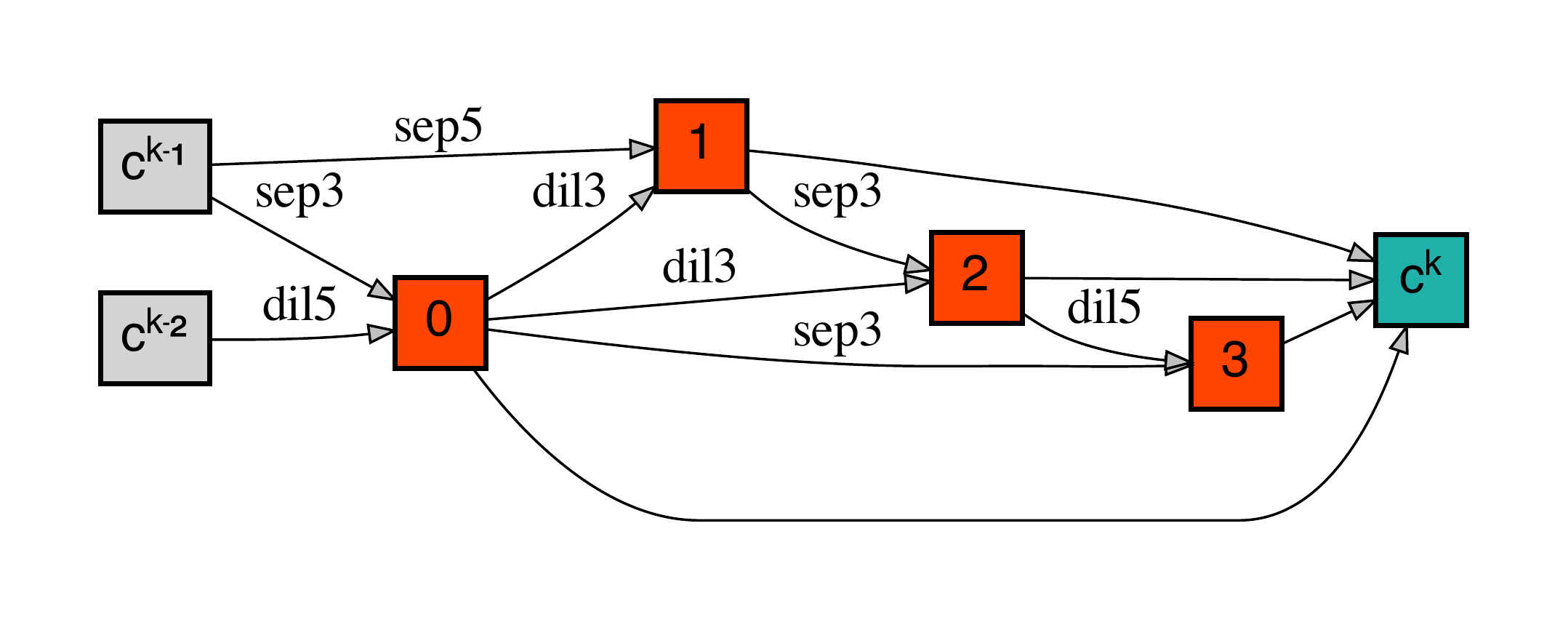} 
}
\subfigure[Reduction]{ 
\includegraphics[width=0.45\columnwidth]{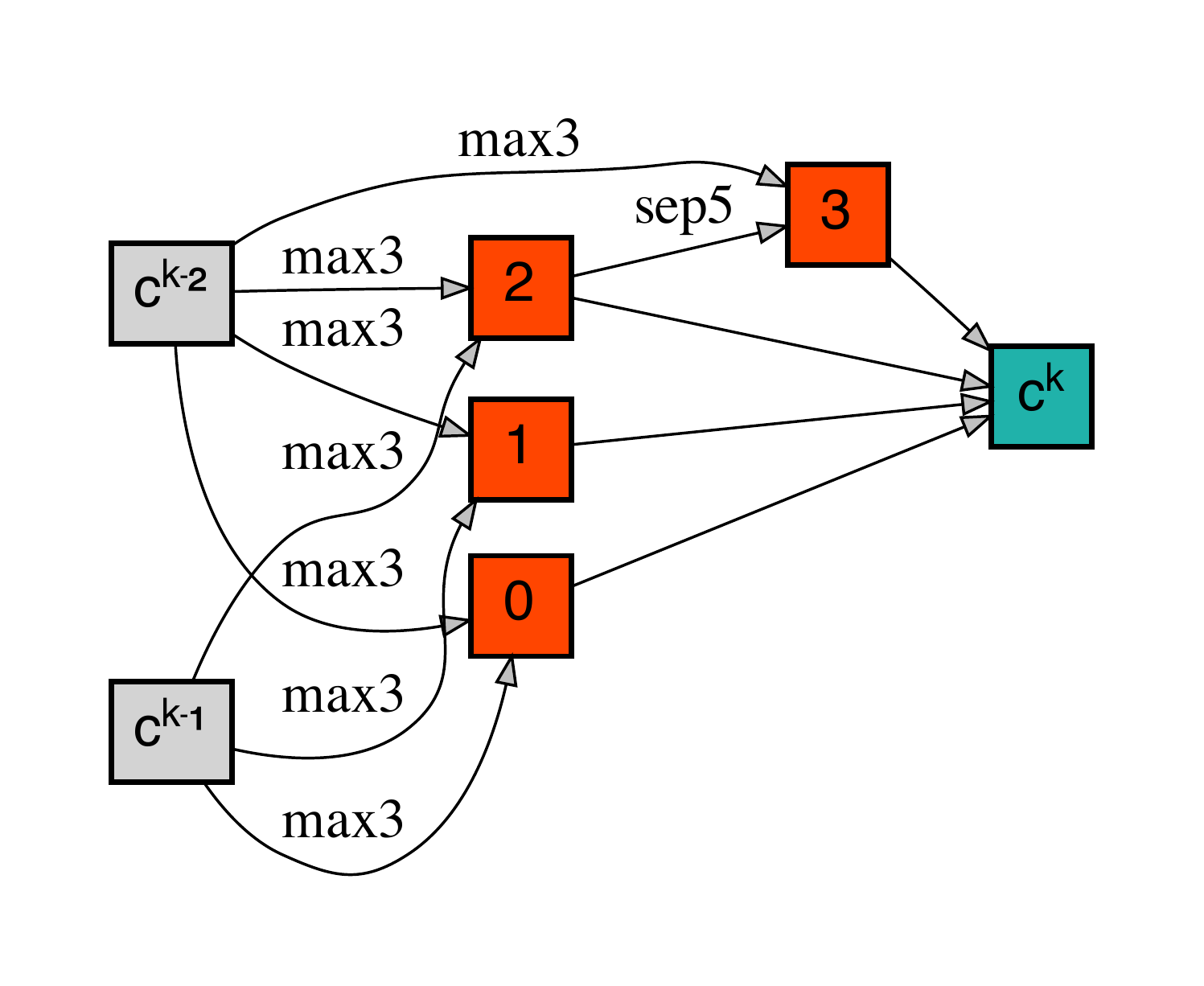}
}
\end{minipage}
\begin{minipage}{2.6in}
\subfigure[Normal]{
\includegraphics[width=0.45\columnwidth]{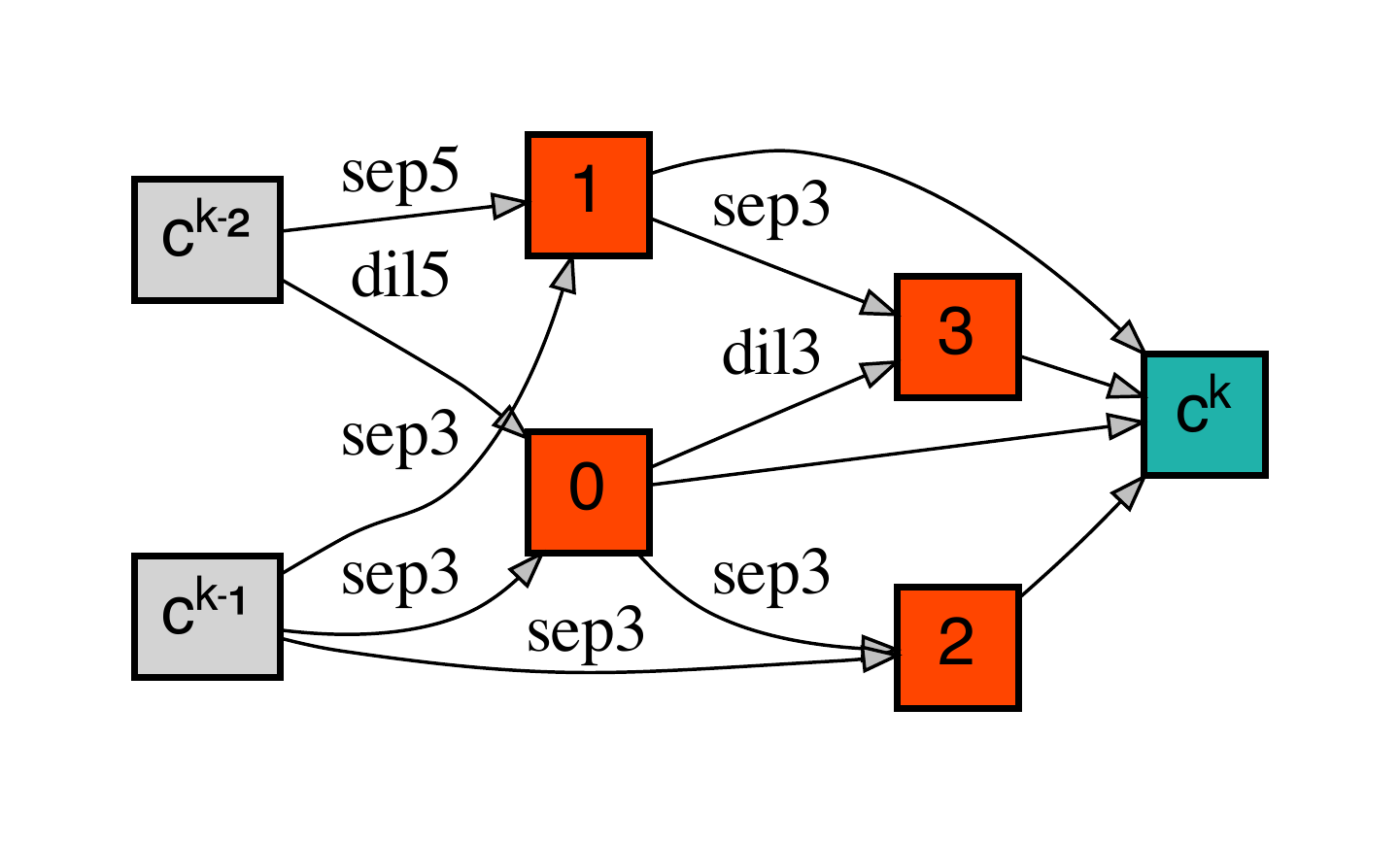} 
}
\subfigure[Reduction]{ 
\includegraphics[width=0.45\columnwidth]{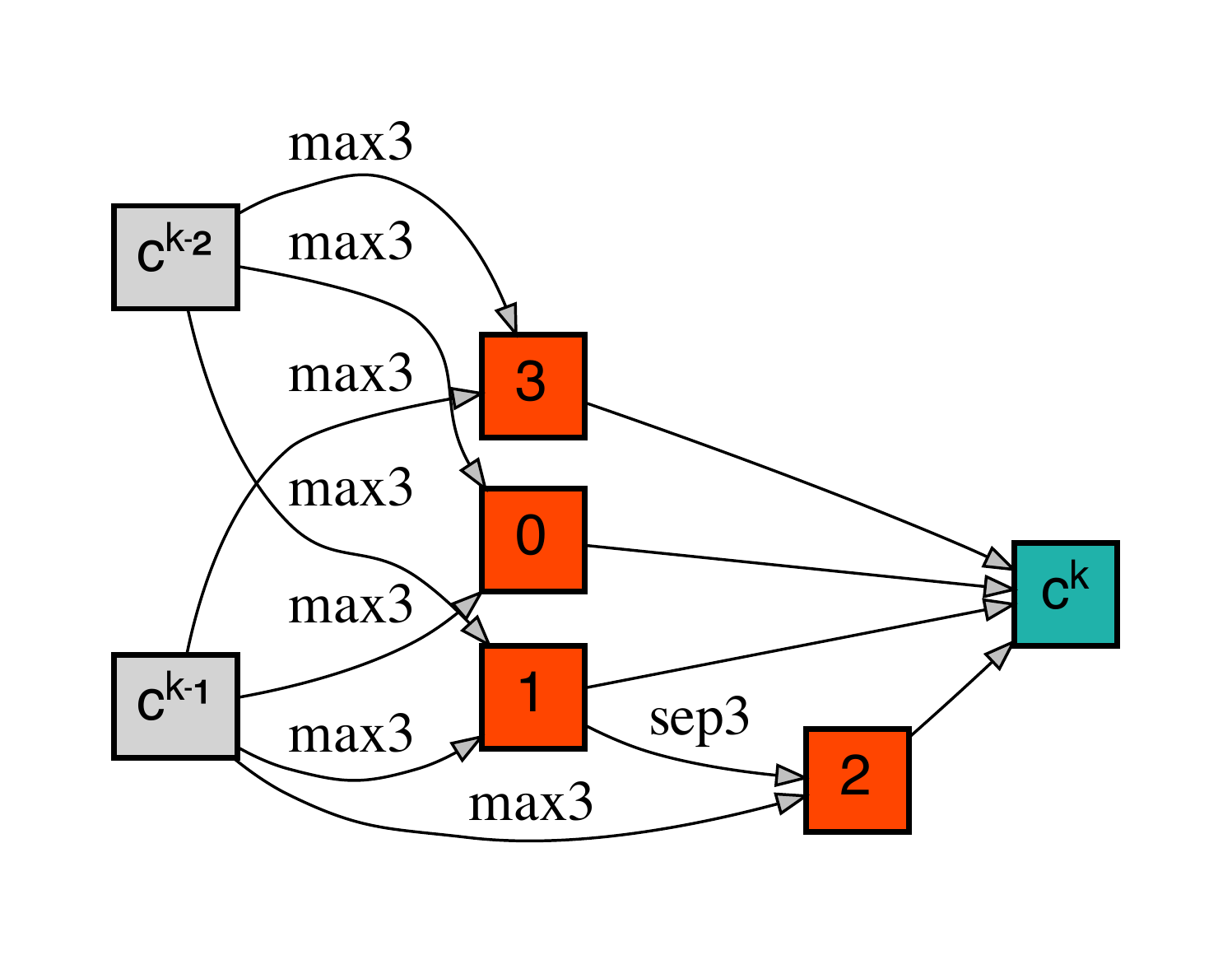}
}
\end{minipage}
\caption{ Found normal cells and reduction cells by P-DARTS \citep{chen2019progressive} with the proposed auxiliary skip connections in the DARTS' standard search space on CIFAR-10 dataset.}
\label{fig:c10_p_ask_cell}
\end{figure*}

\begin{figure*}[ht]
%trim=1cm 0 0 0,
%\captionsetup[subfigure]{labelformat=empty}
\centering
\begin{minipage}{2.6in}
\subfigure[Normal]{
\includegraphics[width=0.45\columnwidth]{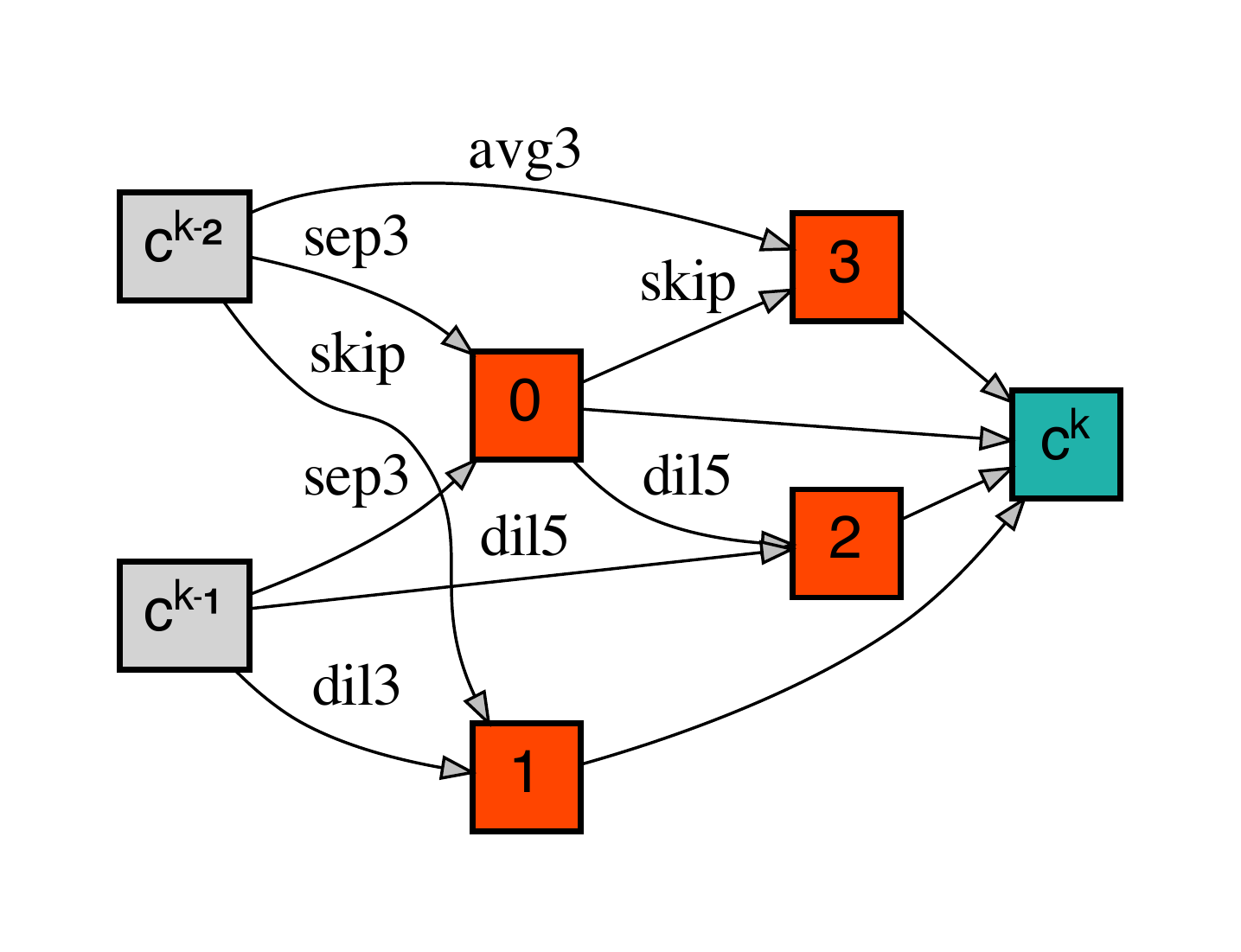} 
}
\subfigure[Reduction]{ 
\includegraphics[width=0.45\columnwidth]{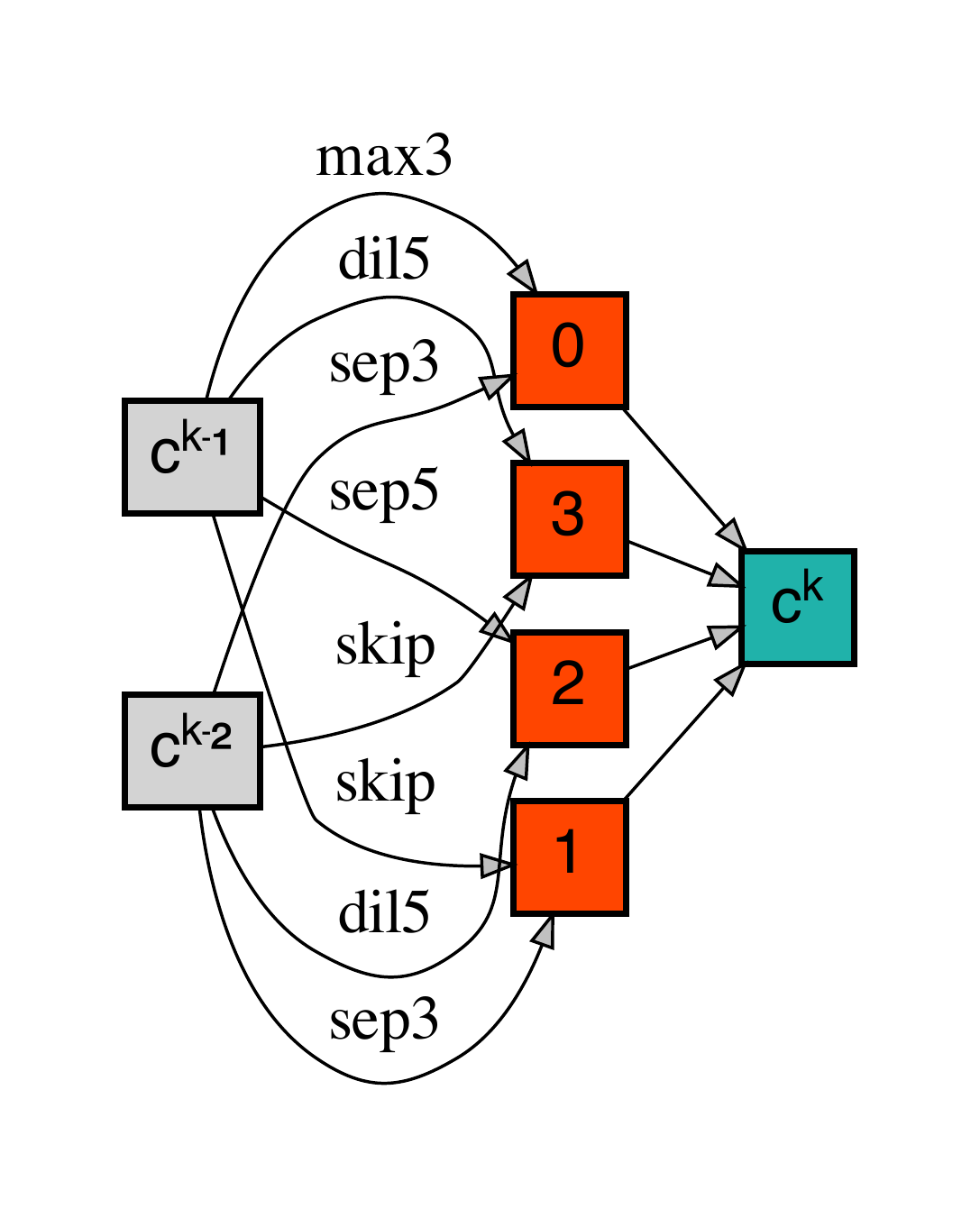}
}
\end{minipage}
\begin{minipage}{2.6in}
\subfigure[Normal]{
\includegraphics[width=0.45\columnwidth]{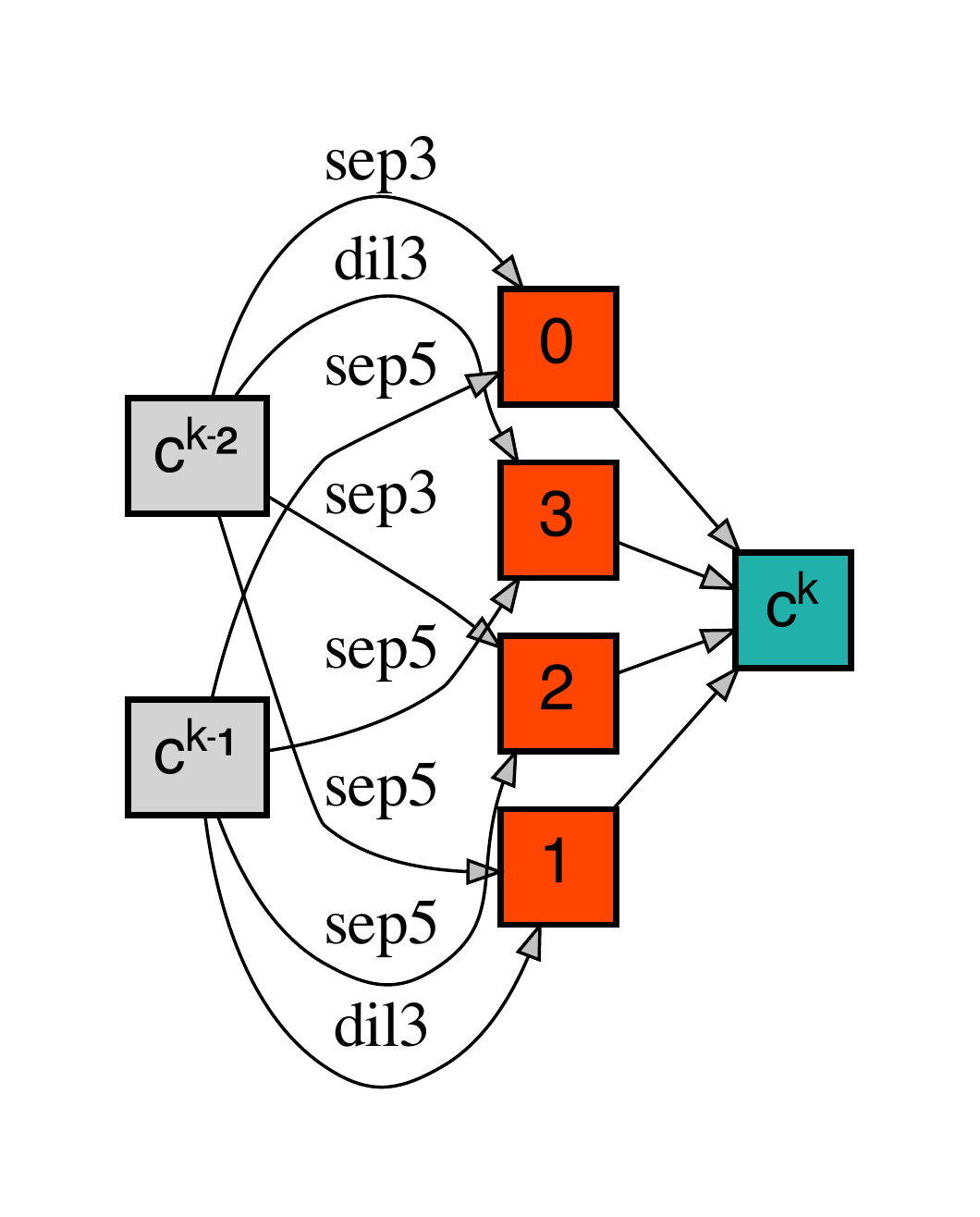} 
}
\subfigure[Reduction]{ 
\includegraphics[width=0.45\columnwidth]{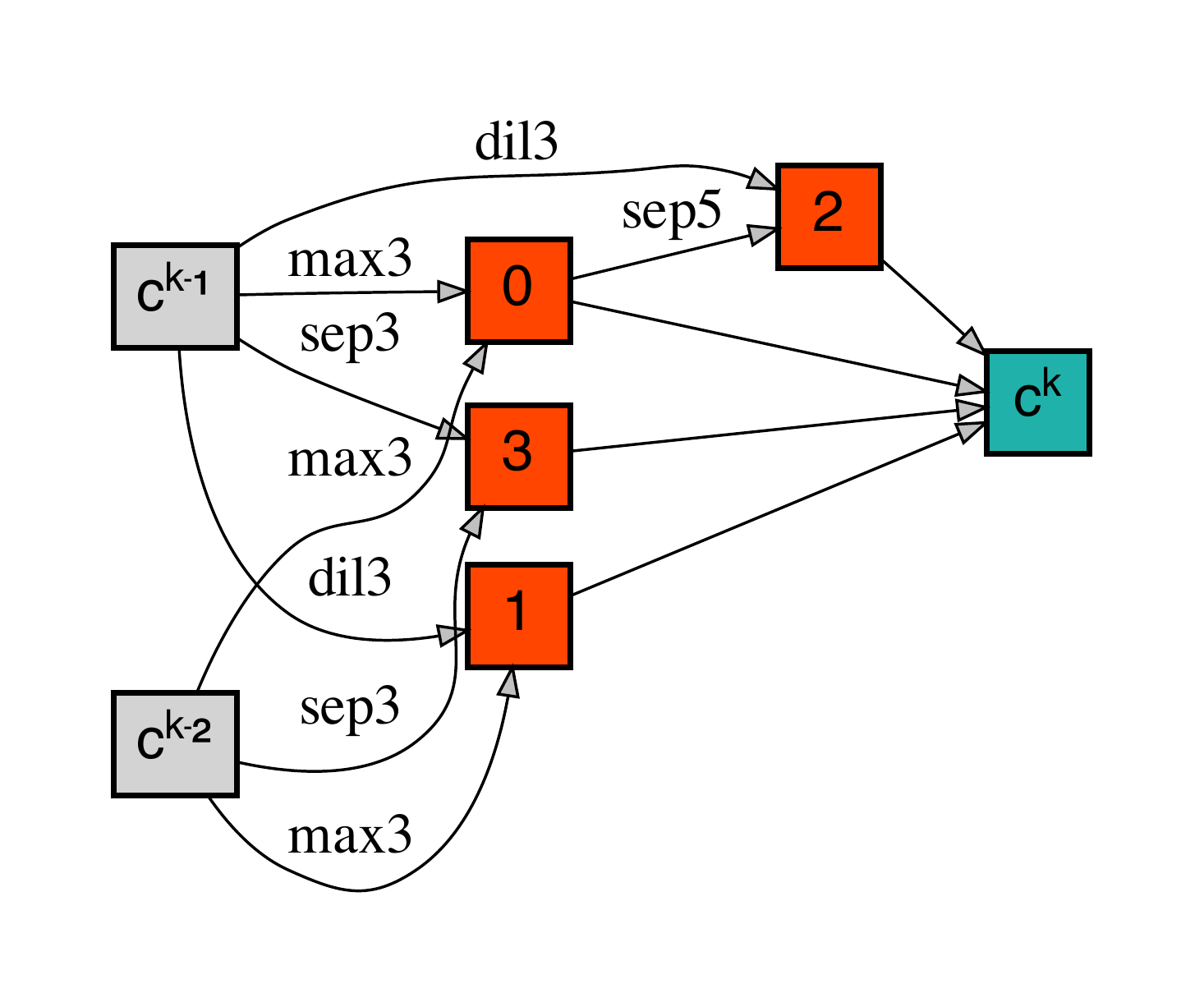}
}
\end{minipage}
\begin{minipage}{2.6in}
\subfigure[Normal]{
\includegraphics[width=0.45\columnwidth]{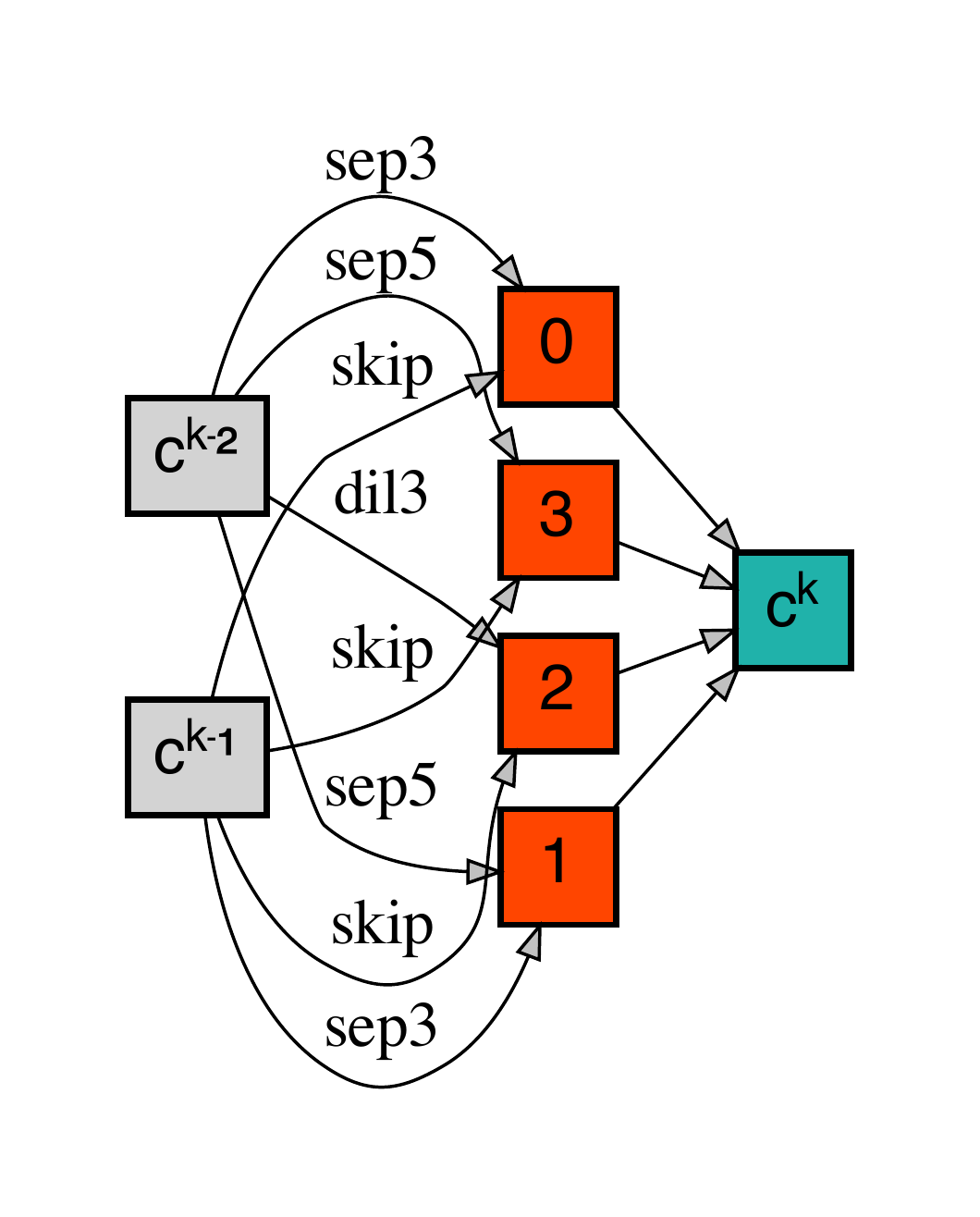} 
}
\subfigure[Reduction]{ 
\includegraphics[width=0.45\columnwidth]{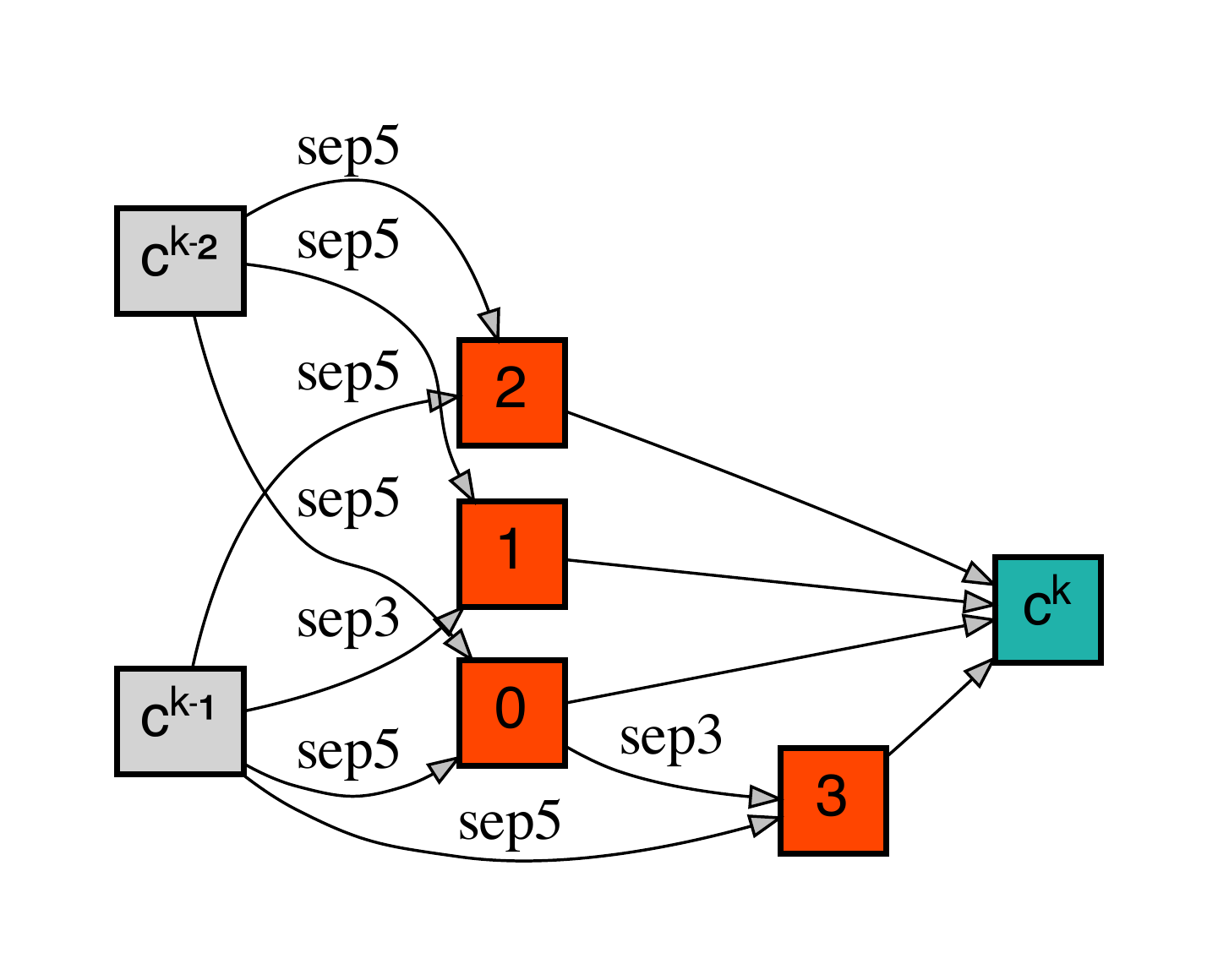}
}
\end{minipage}
\caption{ Found normal cells and reduction cells by PC-DARTS \citep{xu2020pcdarts} without channel shuffling in the DARTS' standard search space on CIFAR-10 dataset.}
\label{fig:c10_pc_ns_cell}
\end{figure*}

\begin{figure*}[ht]
%trim=1cm 0 0 0,
%\captionsetup[subfigure]{labelformat=empty}
\centering
\begin{minipage}{2.7in}
\subfigure[Normal]{
\includegraphics[width=0.45\columnwidth]{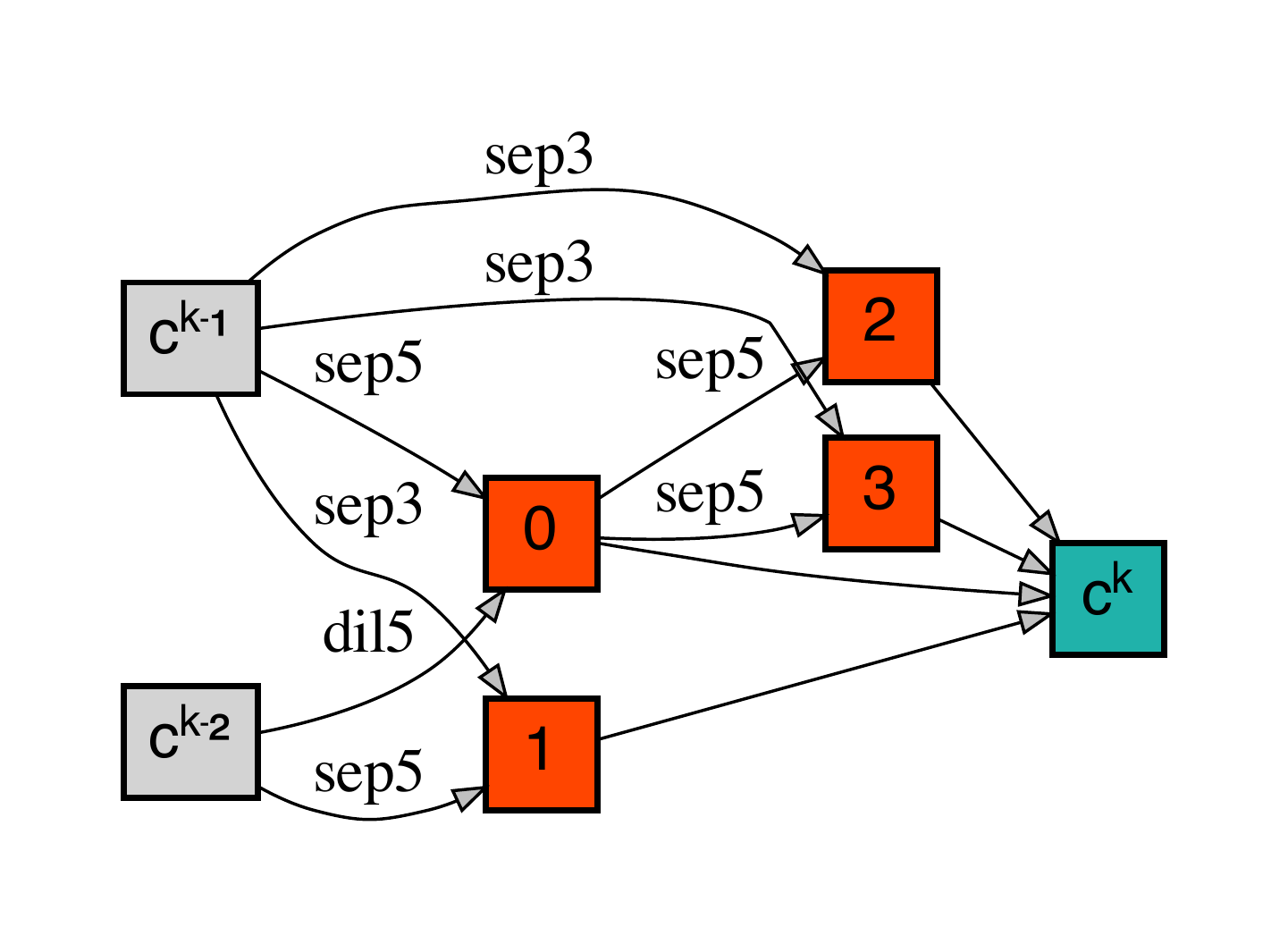} 
}
\subfigure[Reduction]{ 
\includegraphics[width=0.45\columnwidth]{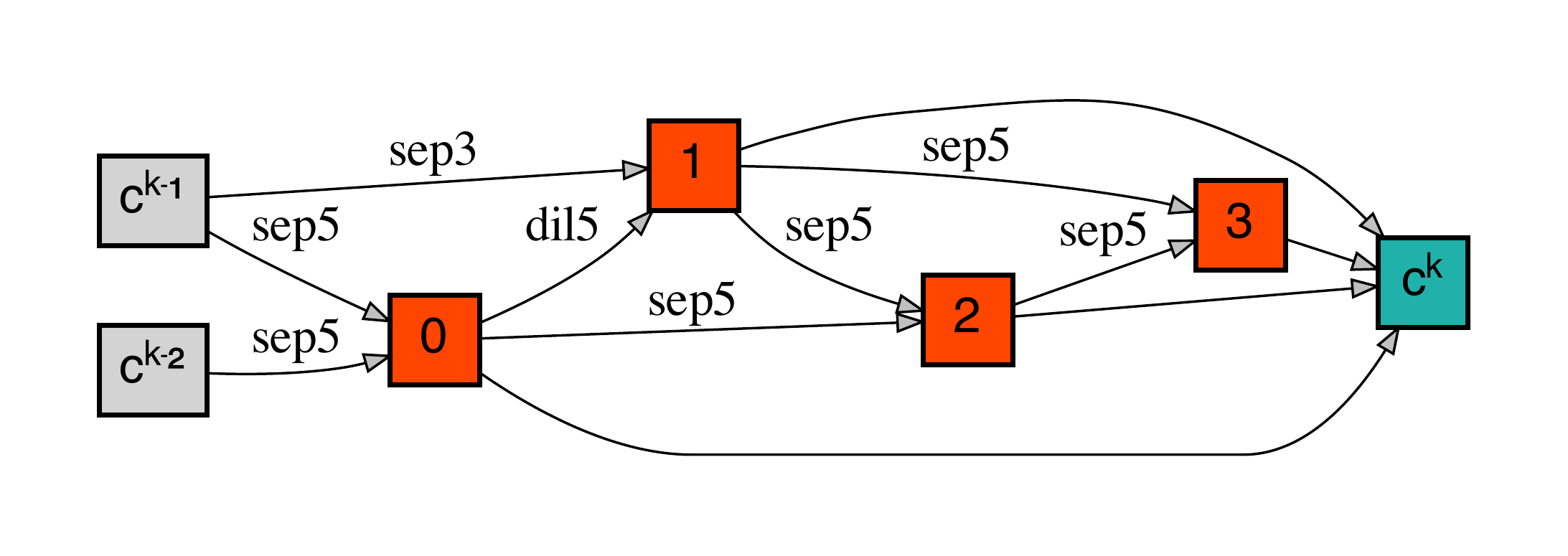}
}
\end{minipage}
\begin{minipage}{2.7in}
\subfigure[Normal]{
\includegraphics[width=0.45\columnwidth]{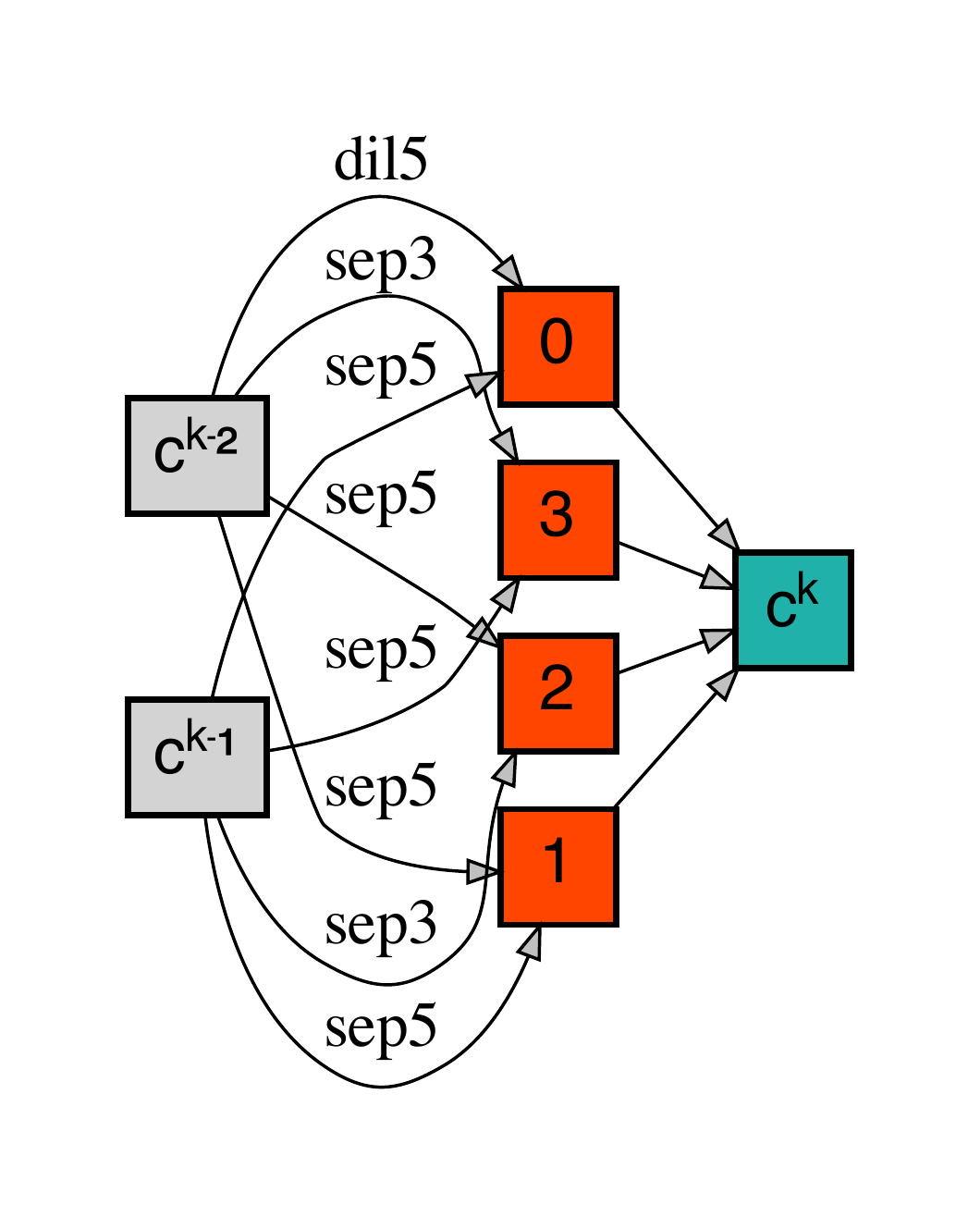} 
}
\subfigure[Reduction]{ 
\includegraphics[width=0.45\columnwidth]{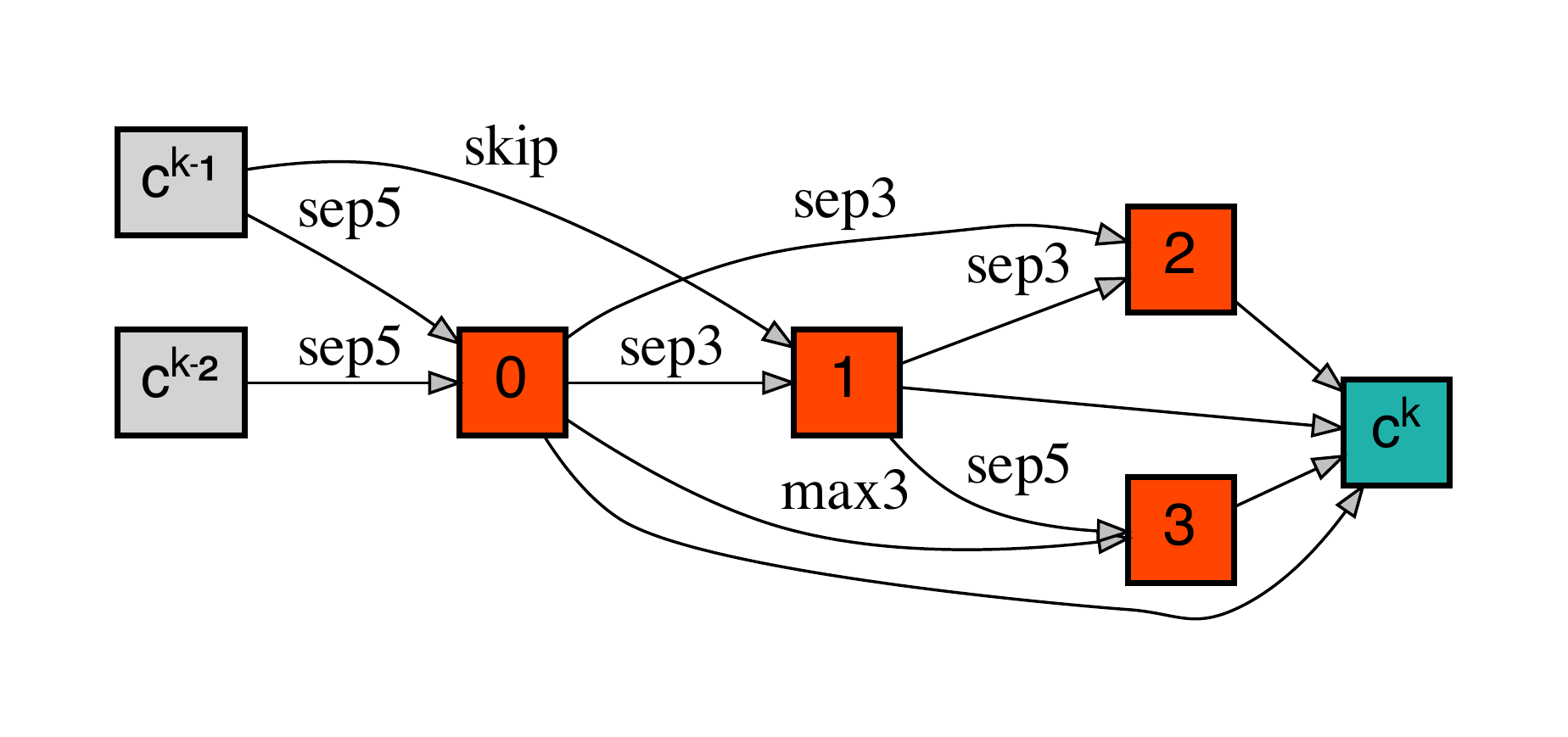}
}
\end{minipage}
\begin{minipage}{2.6in}
\subfigure[Normal]{
\includegraphics[width=0.45\columnwidth]{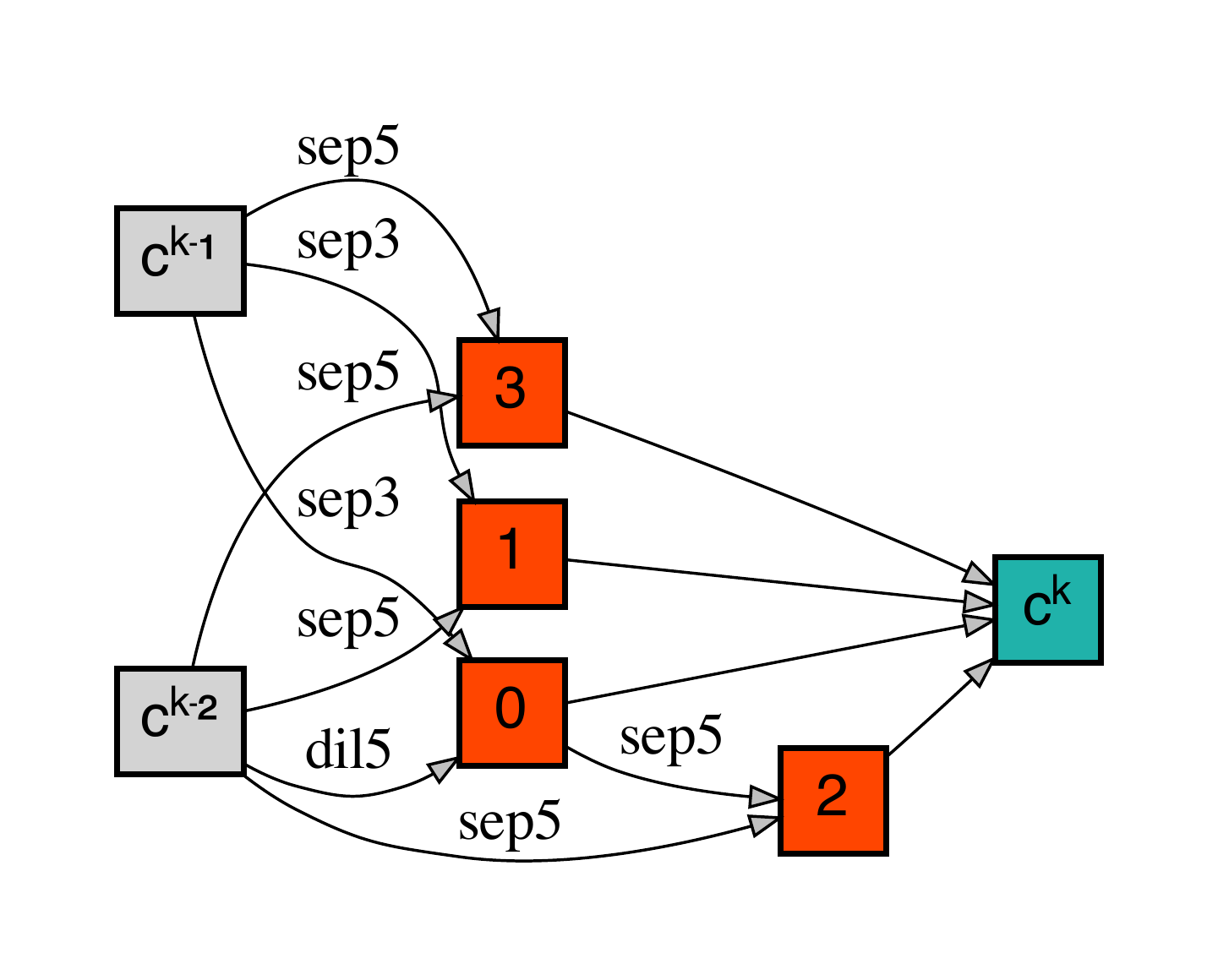} 
}
\subfigure[Reduction]{ 
\includegraphics[width=0.45\columnwidth]{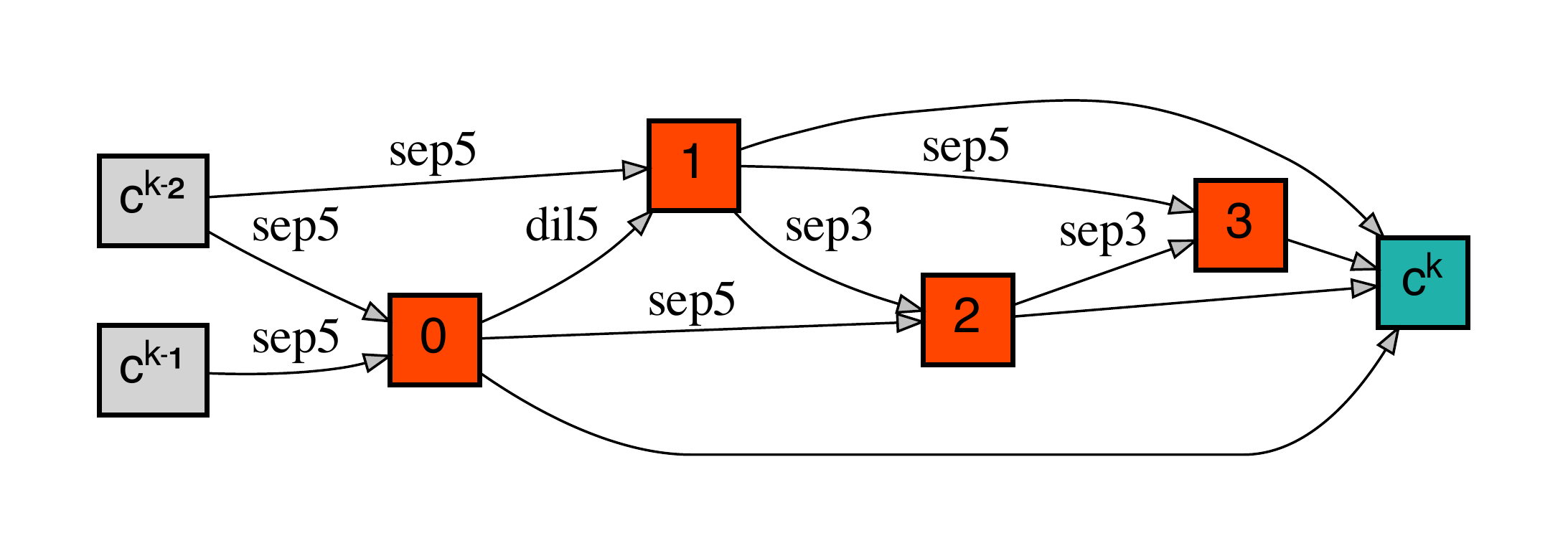}
}
\end{minipage}
\caption{ Found normal cells and reduction cells by PC-DARTS \citep{xu2020pcdarts} with the proposed auxiliary skip connections in the DARTS' standard search space on CIFAR-10 dataset.}
\label{fig:c10_pc_ask_cell}
\end{figure*}

\begin{figure*}[ht]
%trim=1cm 0 0 0,
%\captionsetup[subfigure]{labelformat=empty}
\centering
\begin{minipage}{2.6in}
\subfigure[Normal]{
\includegraphics[width=0.45\columnwidth]{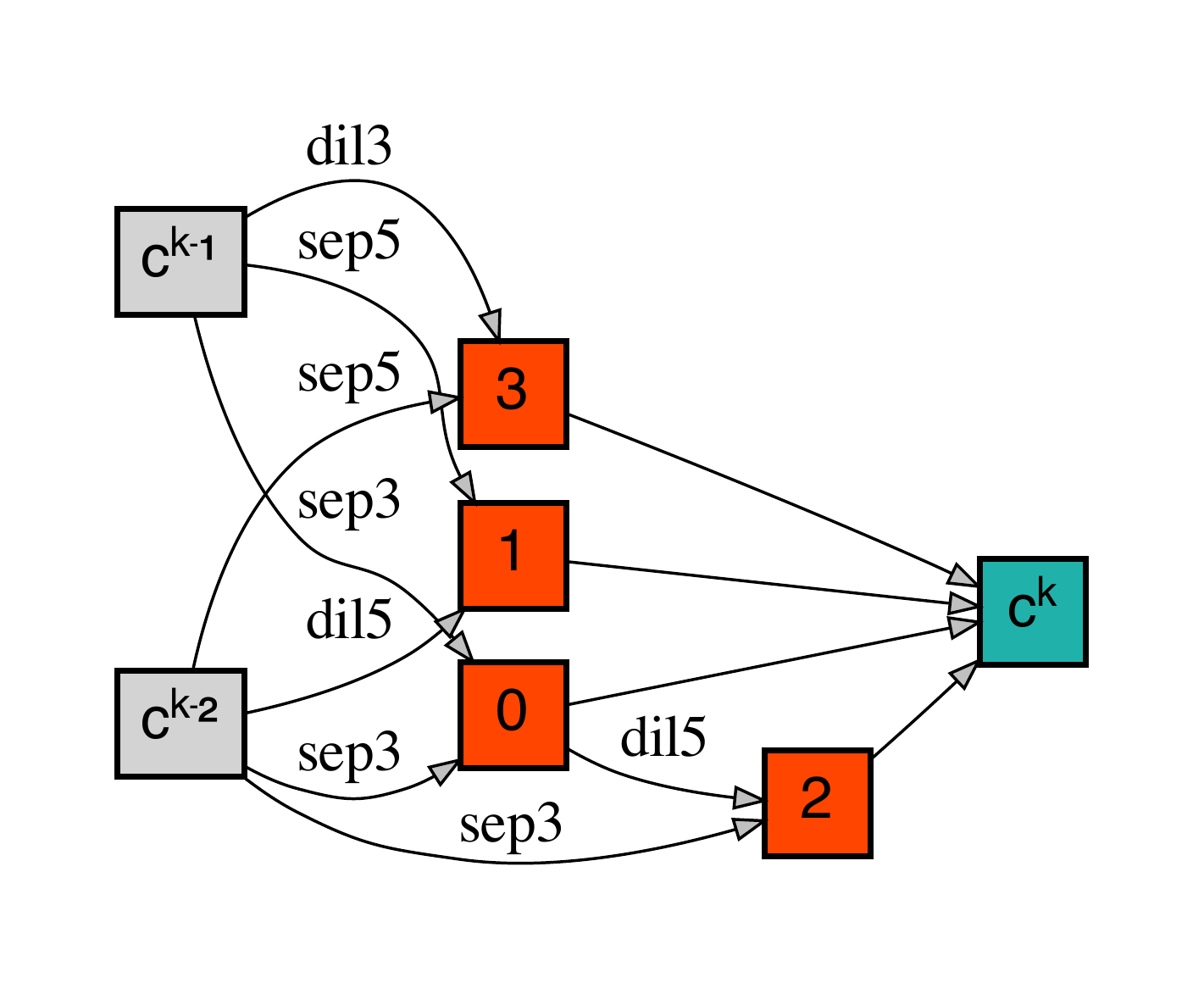} 
}
\subfigure[Reduction]{ 
\includegraphics[width=0.45\columnwidth]{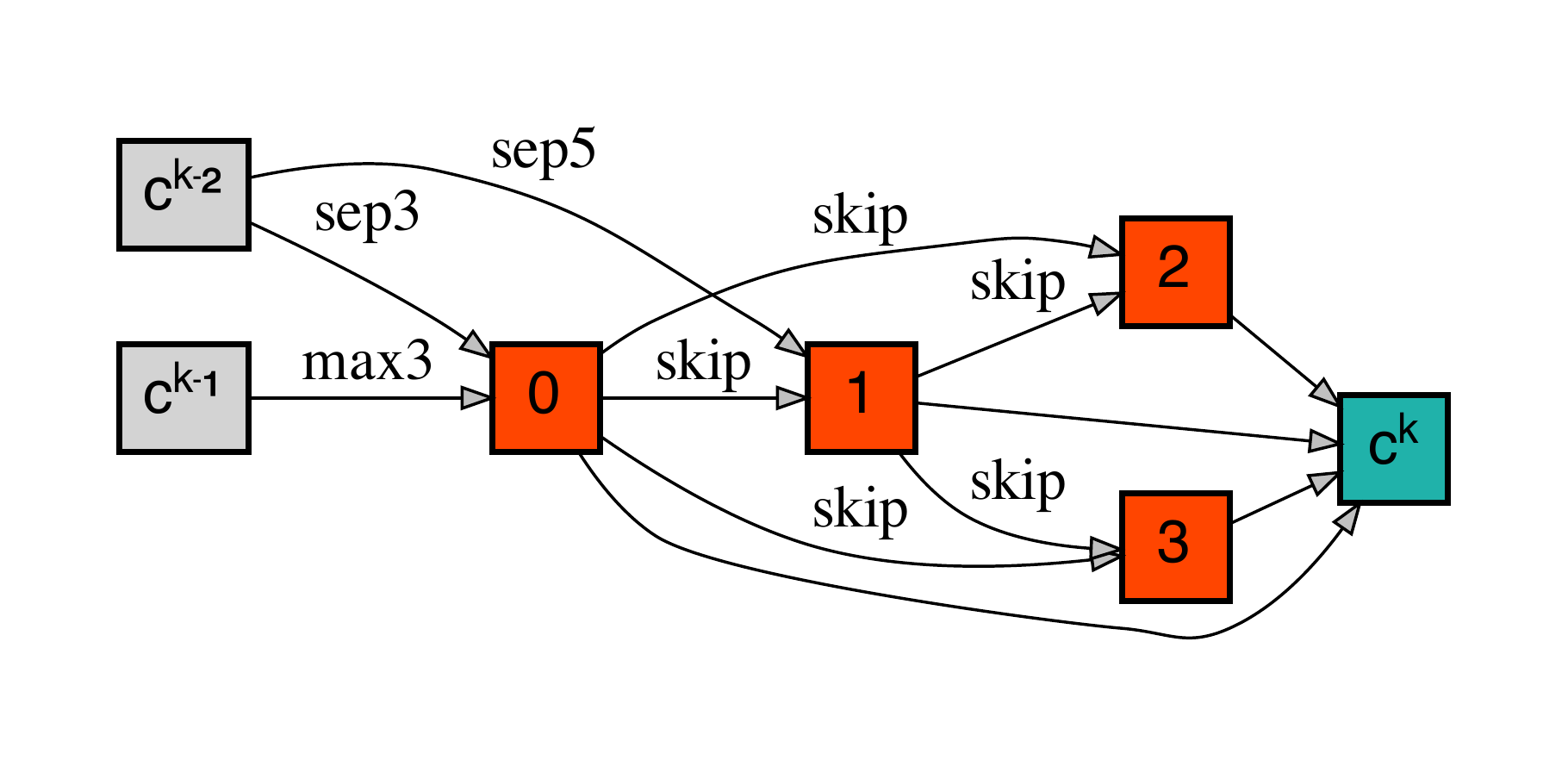}
}
\end{minipage}
\begin{minipage}{2.6in}
\subfigure[Normal]{
\includegraphics[width=0.45\columnwidth]{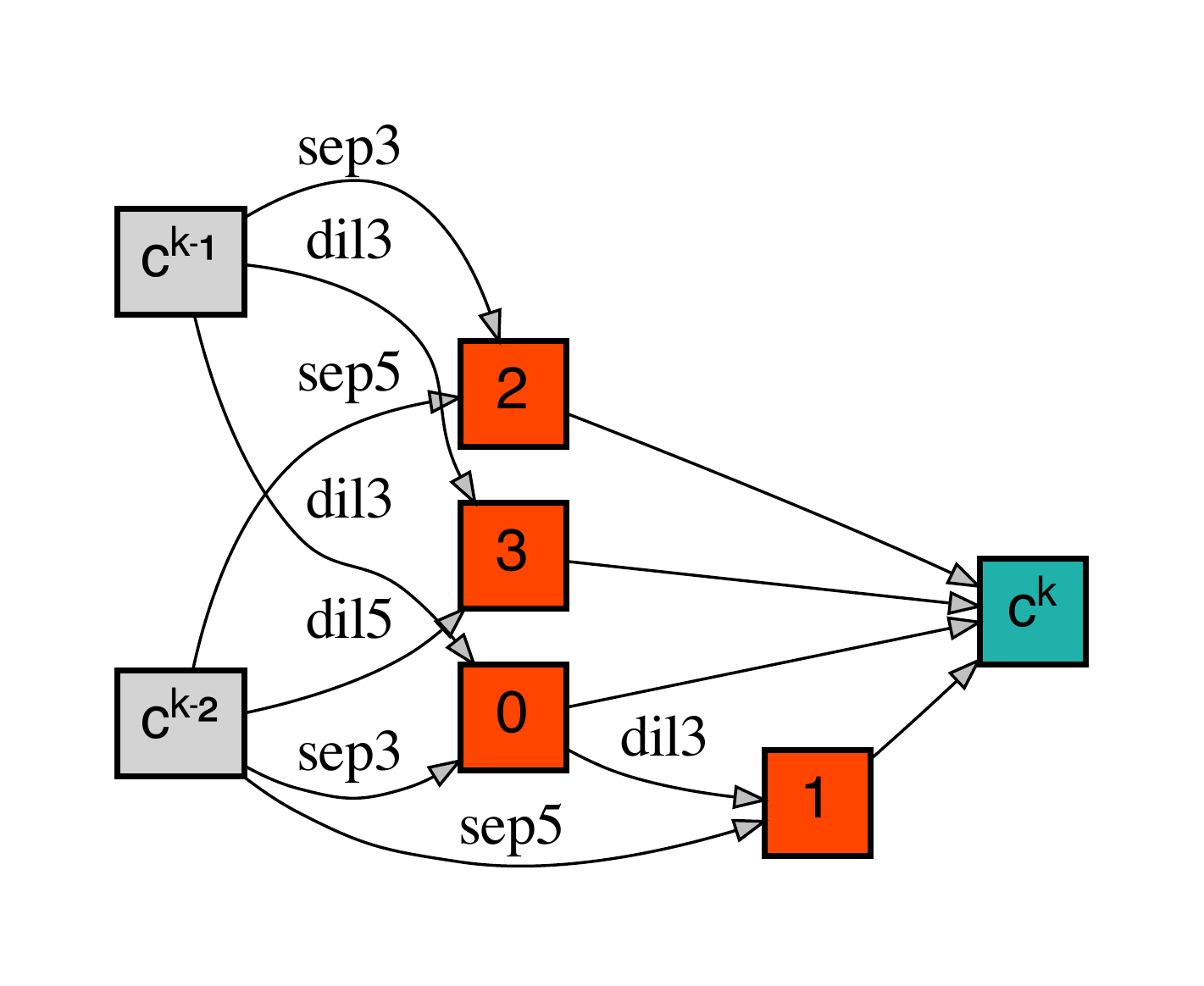} 
}
\subfigure[Reduction]{ 
\includegraphics[width=0.45\columnwidth]{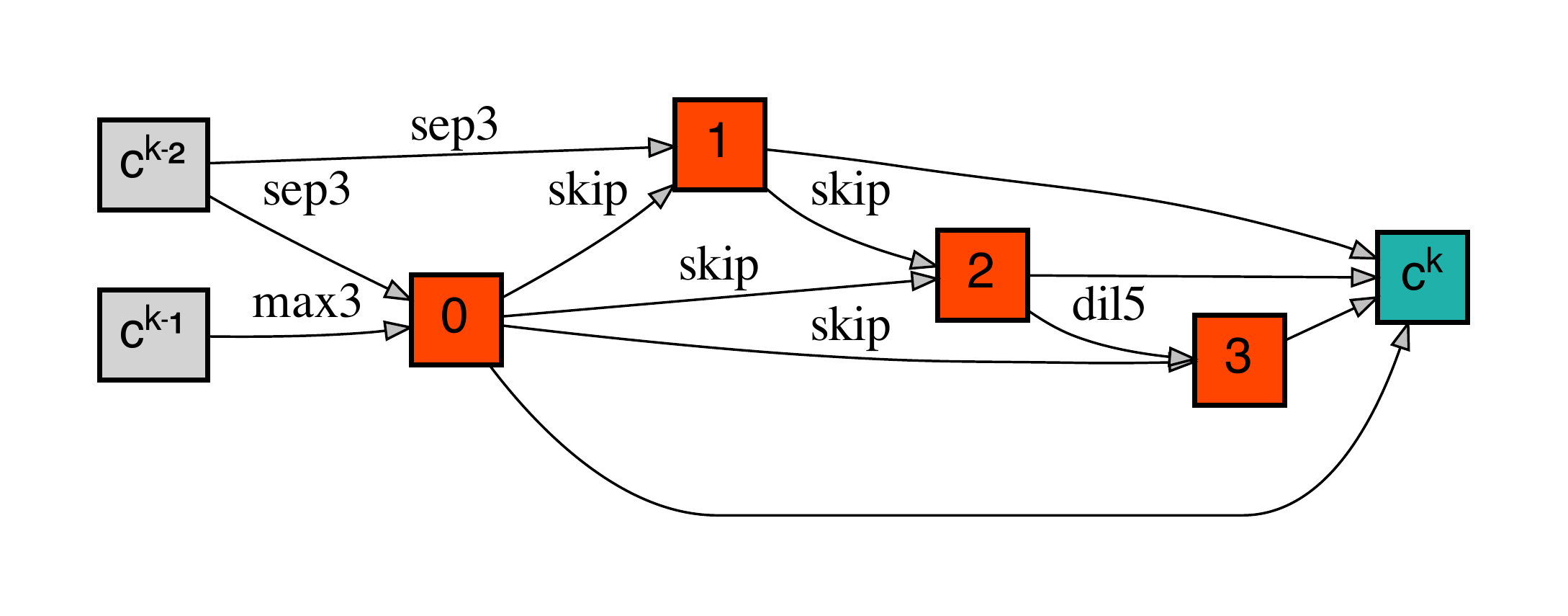}
}
\end{minipage}
\begin{minipage}{2.6in}
\subfigure[Normal]{
\includegraphics[width=0.45\columnwidth]{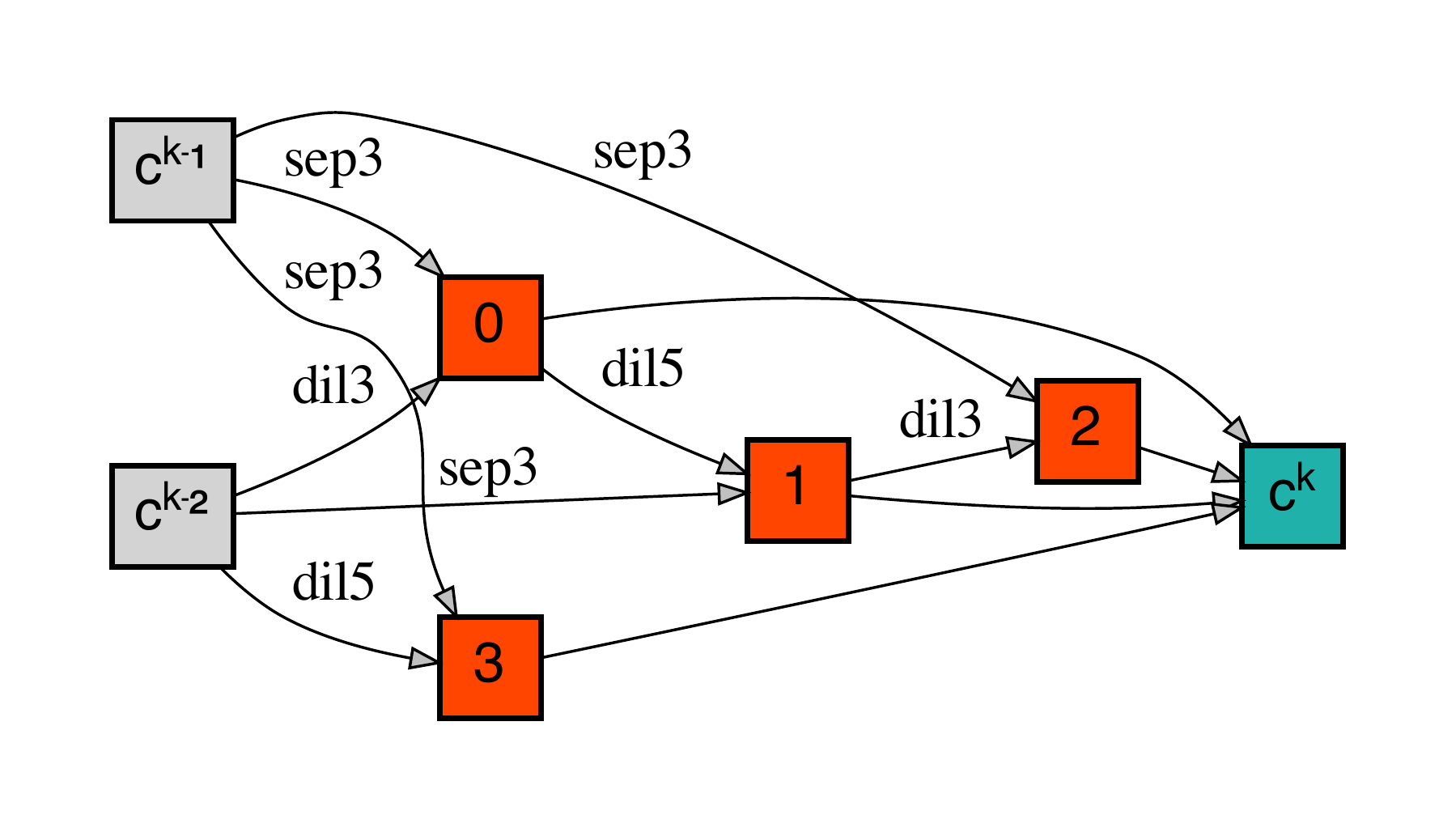} 
}
\subfigure[Reduction]{ 
\includegraphics[width=0.45\columnwidth]{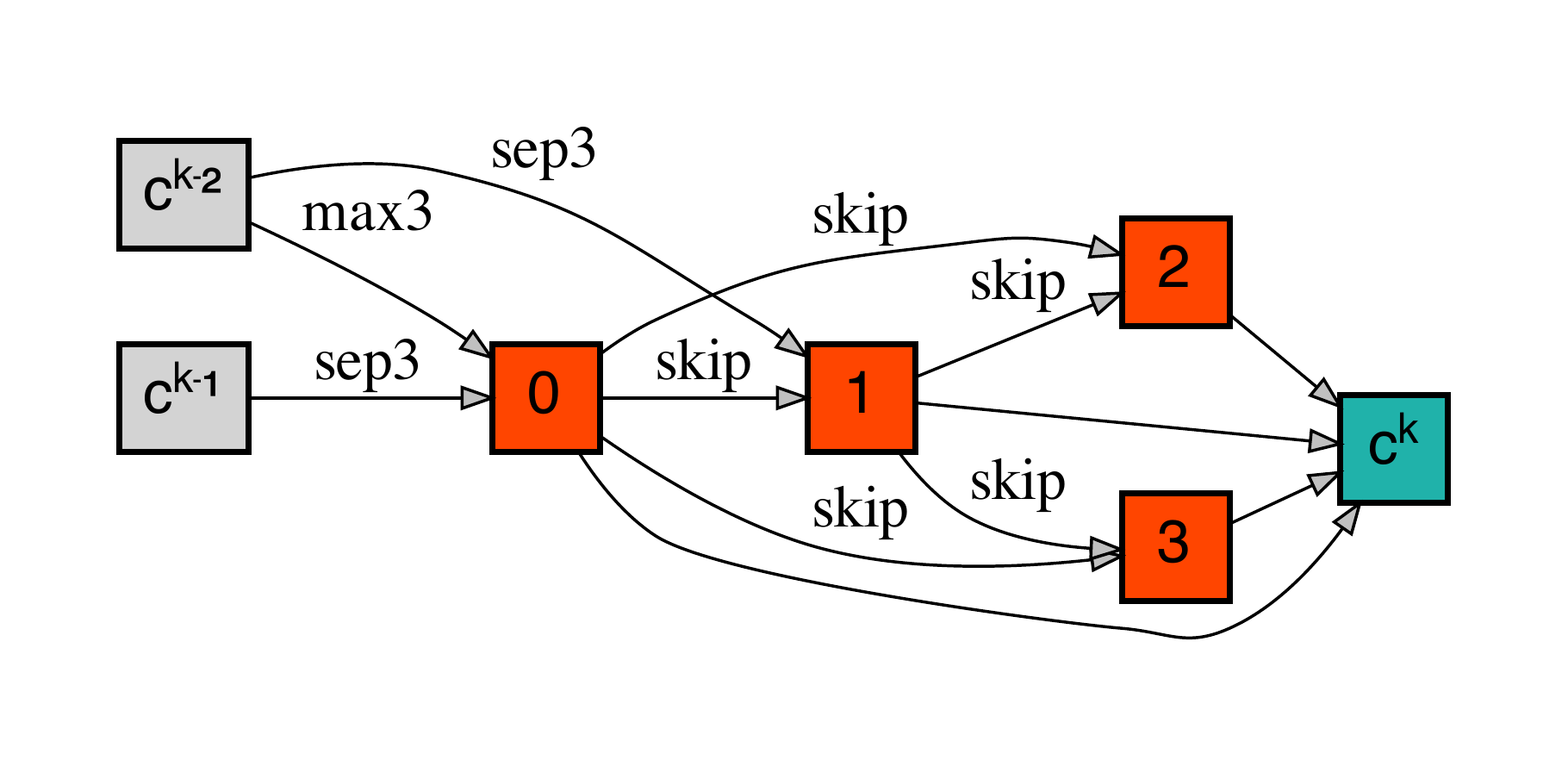}
}
\end{minipage}
\caption{Keep $\beta_{skip}=1$ throughout the DARTS- searching in the DARTS' standard search space on CIFAR-10 dataset.}
\label{fig:c10_nodecay_cells}
\end{figure*}

\begin{figure*}[ht]
%trim=1cm 0 0 0,
%\captionsetup[subfigure]{labelformat=empty}
\centering
\begin{minipage}{2.6in}
\subfigure[Normal]{
\includegraphics[width=0.35\columnwidth]{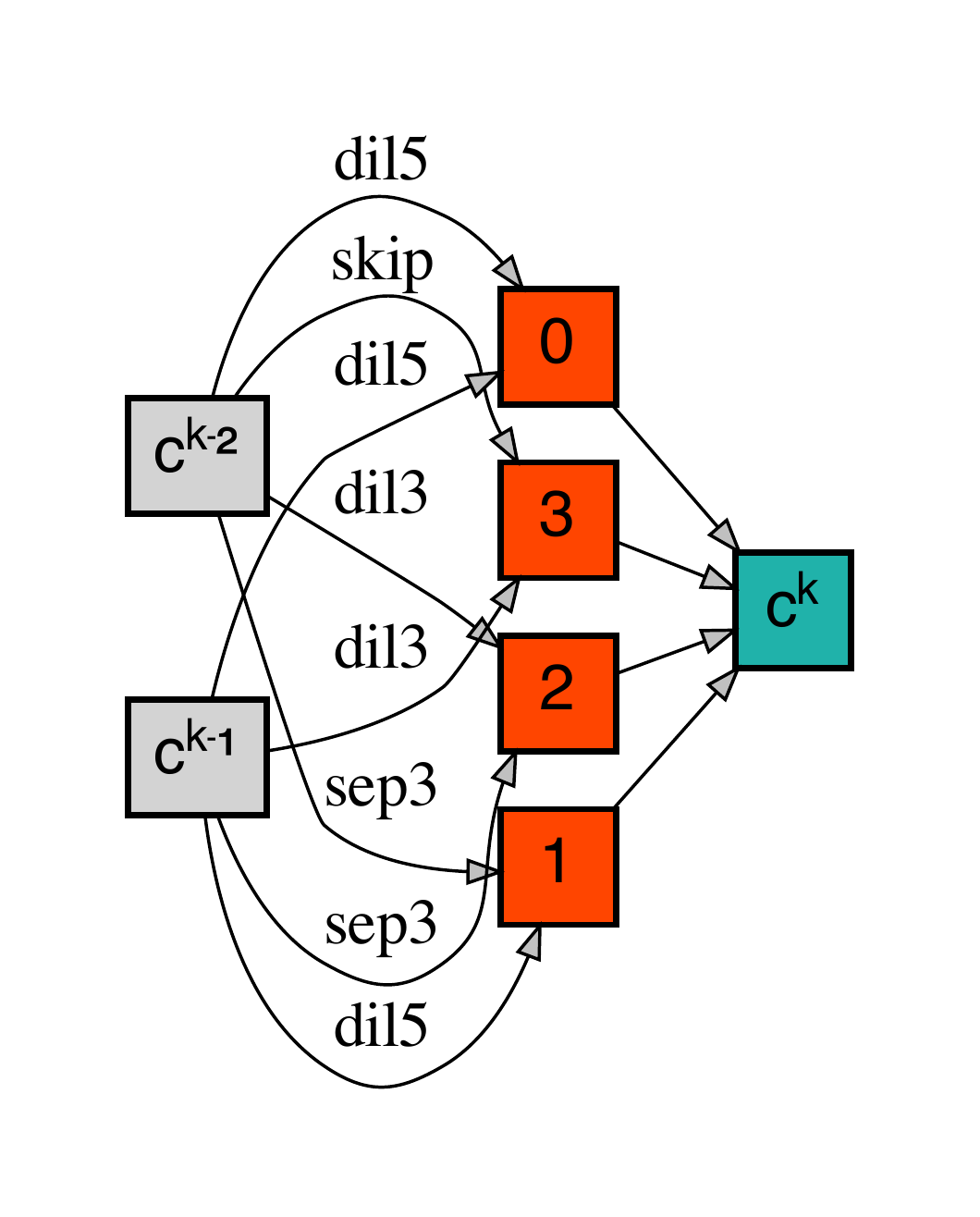} 
}
\subfigure[Reduction]{ 
\includegraphics[width=0.55\columnwidth]{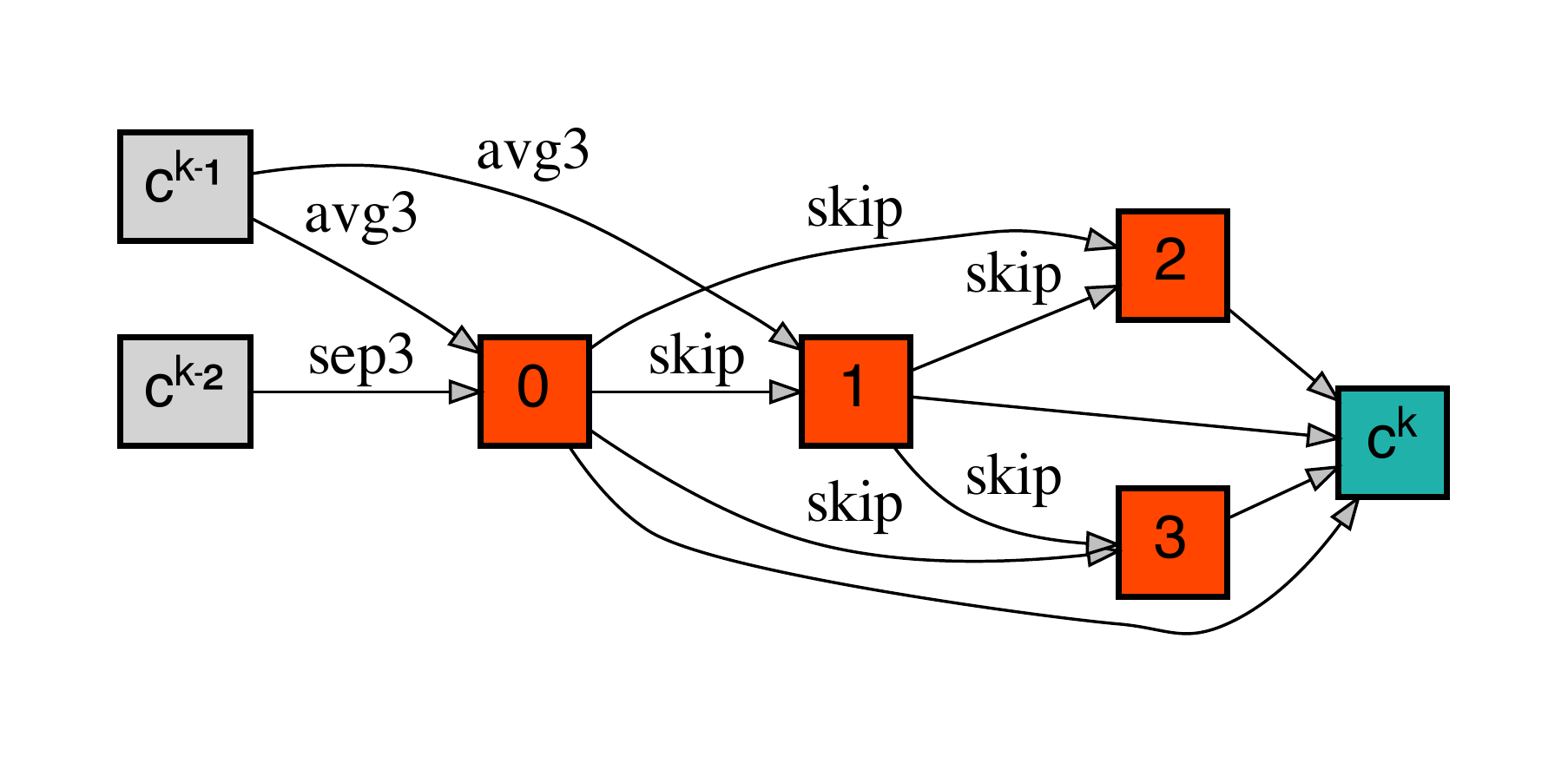}
}
\end{minipage}
\begin{minipage}{2.6in}
\subfigure[Normal]{
\includegraphics[width=0.35\columnwidth]{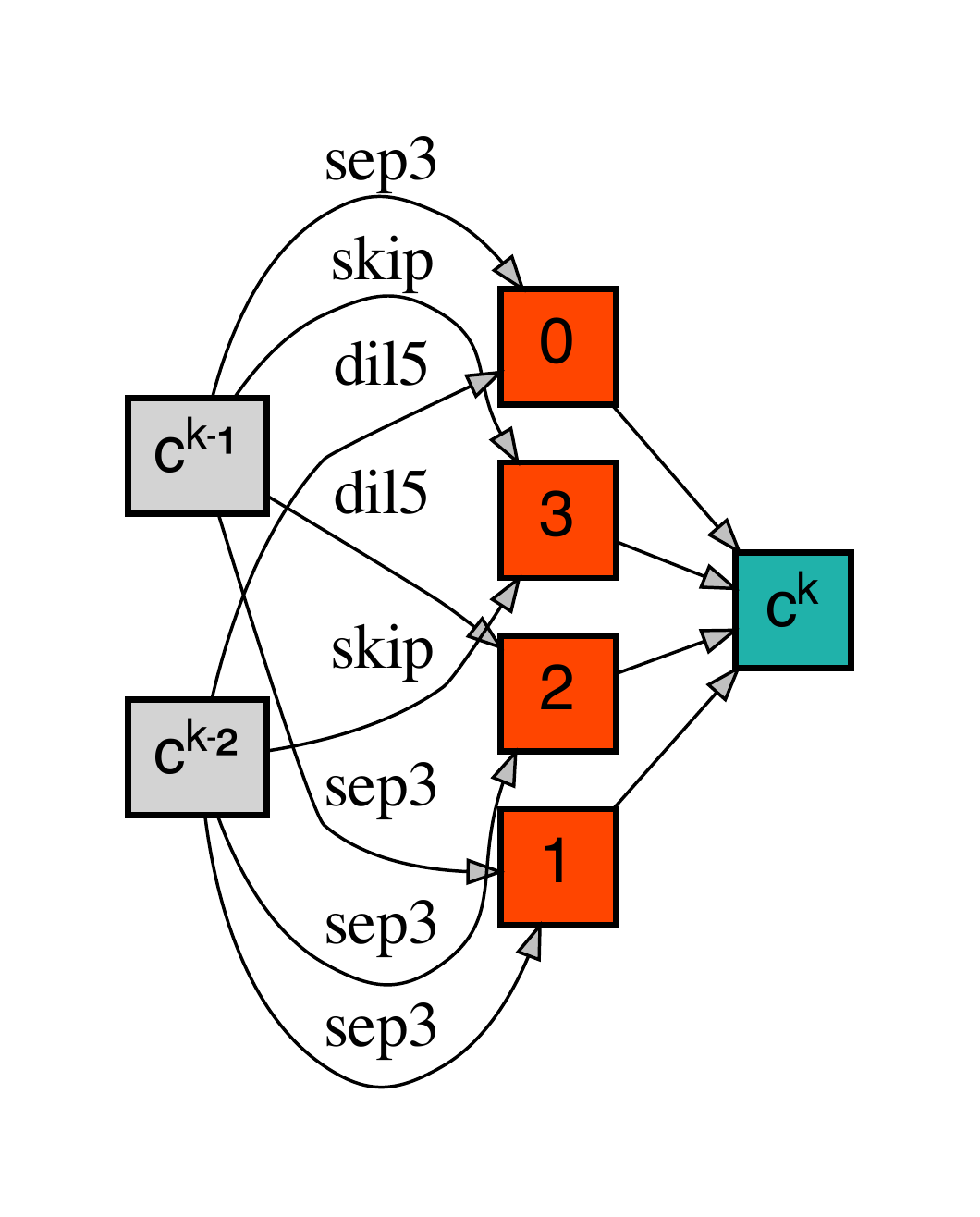} 
}
\subfigure[Reduction]{ 
\includegraphics[width=0.55\columnwidth]{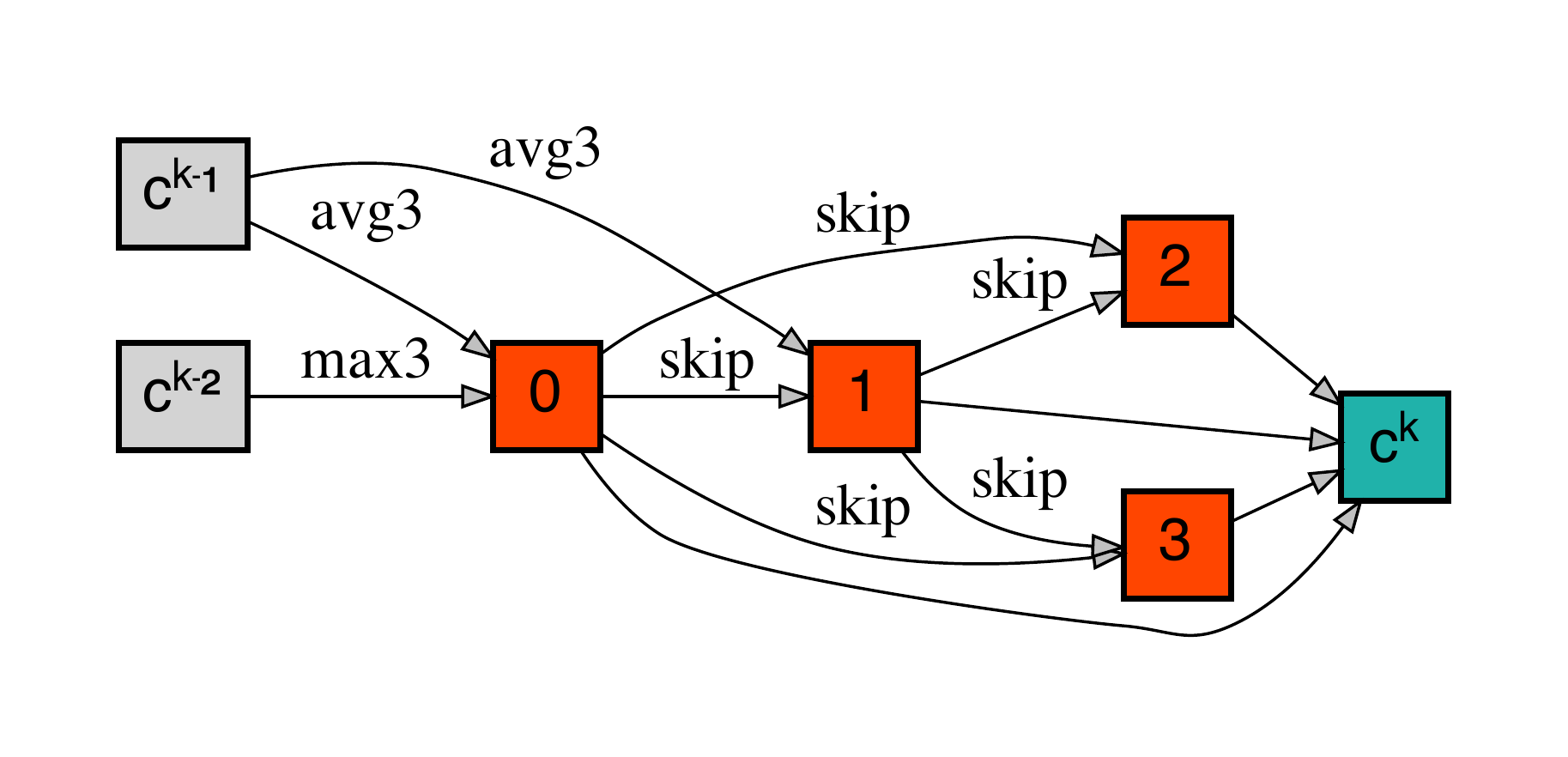}
}
\end{minipage}
\caption{Best cells found when decaying $\beta_{skip}$ in the last 50 epochs during the DARTS- searching for 150 and 200 epochs respectively in the DARTS search space on CIFAR-10.}
\label{fig:c10_s5_decay_best_cells}
\end{figure*}

\begin{figure*}[ht]
\centering
\begin{minipage}{2.6in}
\subfigure[Normal]{
\includegraphics[width=0.45\columnwidth]{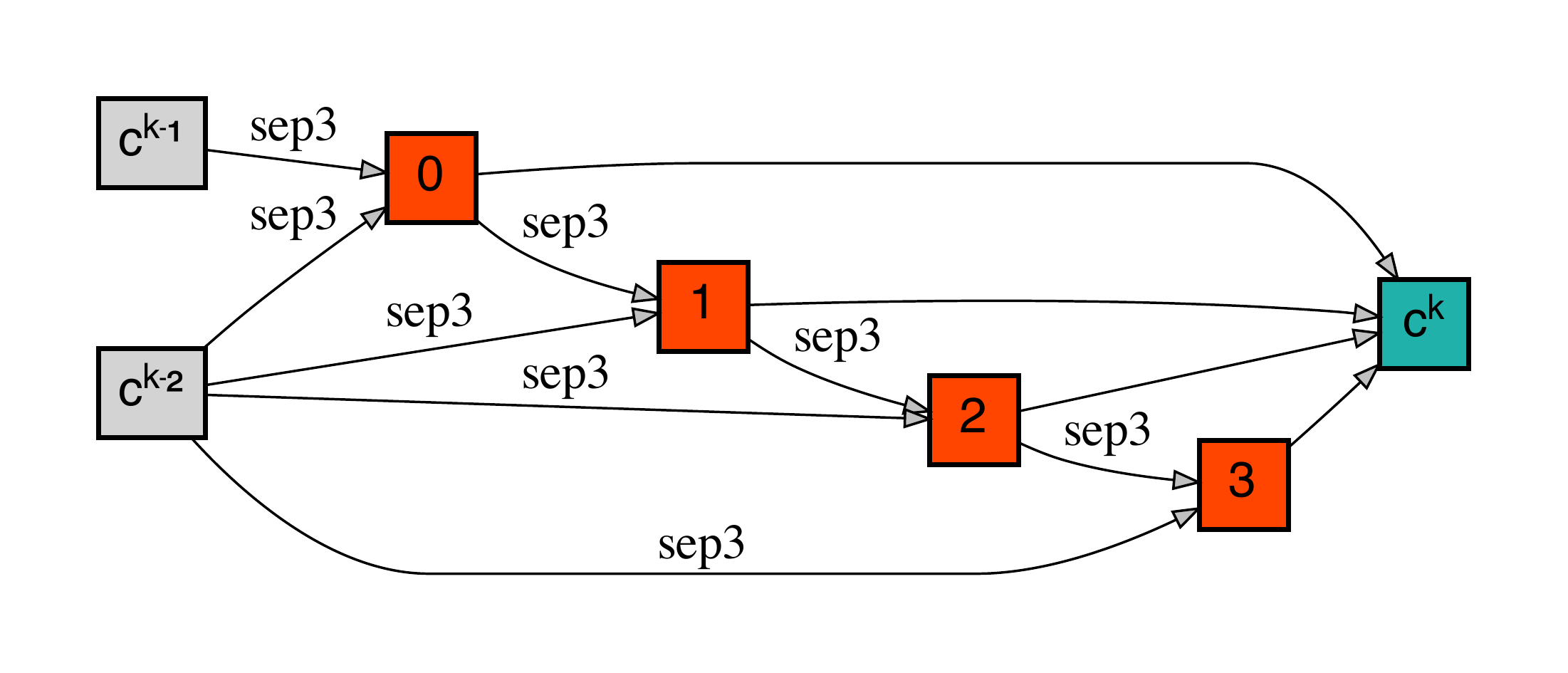} 
}
\subfigure[Reduction]{ 
\includegraphics[width=0.45\columnwidth]{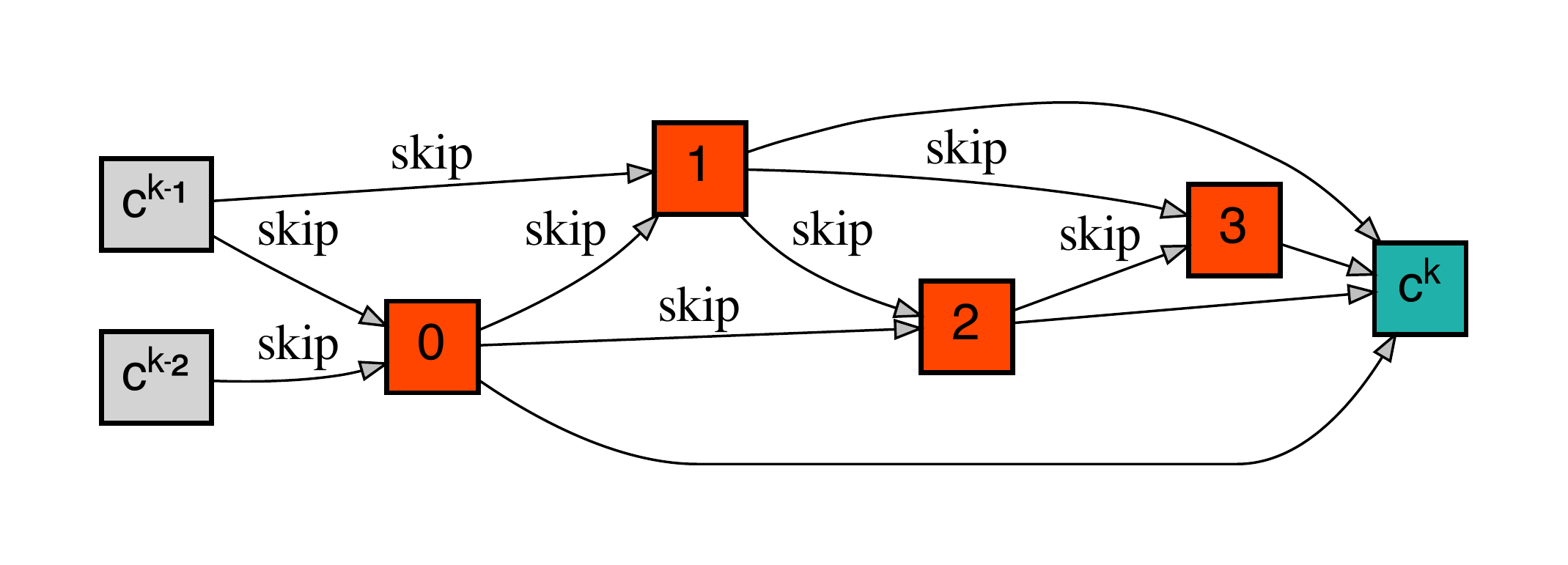}
}
\end{minipage}
\begin{minipage}{2.6in}
\subfigure[Normal]{
\includegraphics[width=0.45\columnwidth]{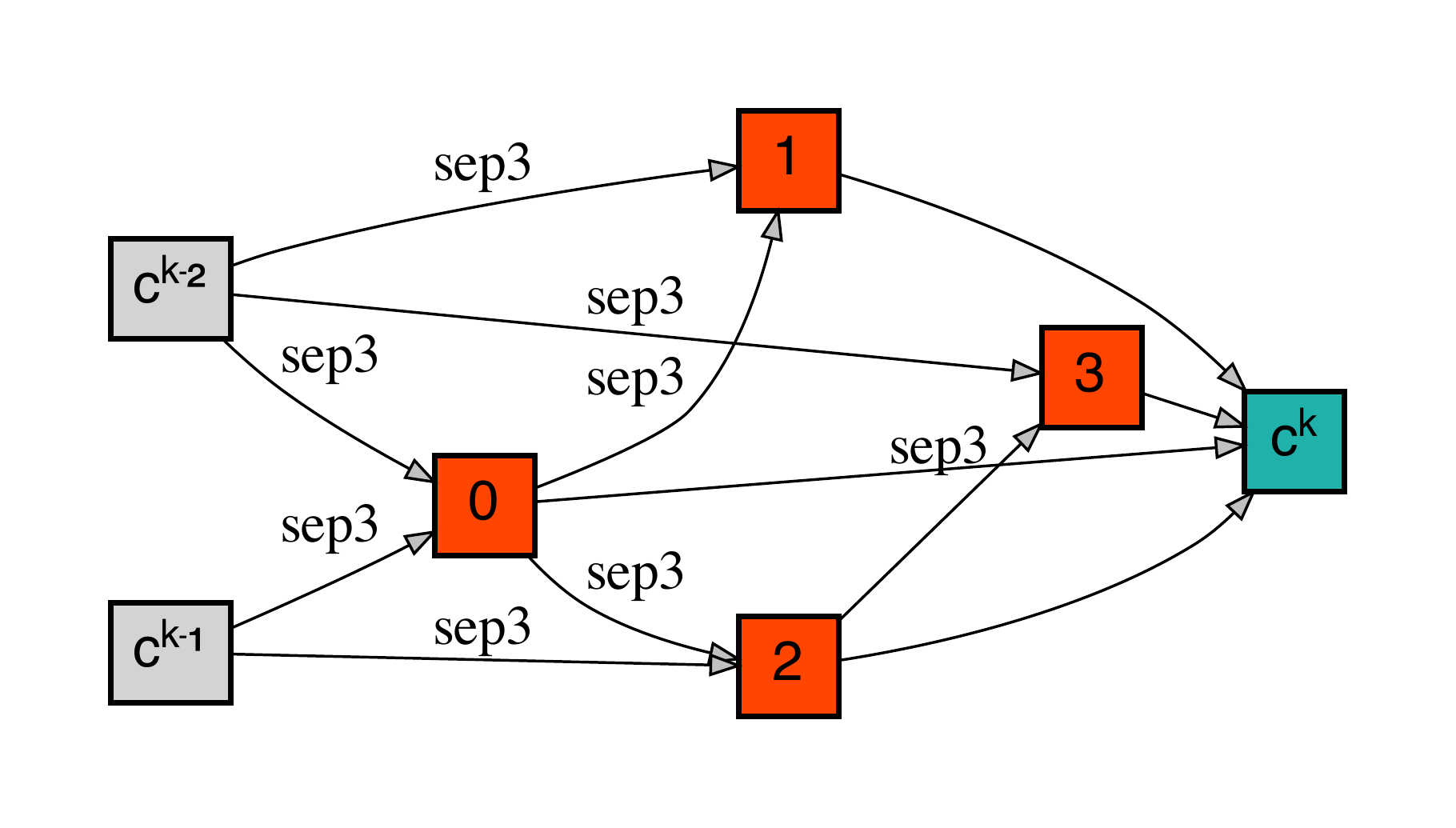} 
}
\subfigure[Reduction]{ 
\includegraphics[width=0.45\columnwidth]{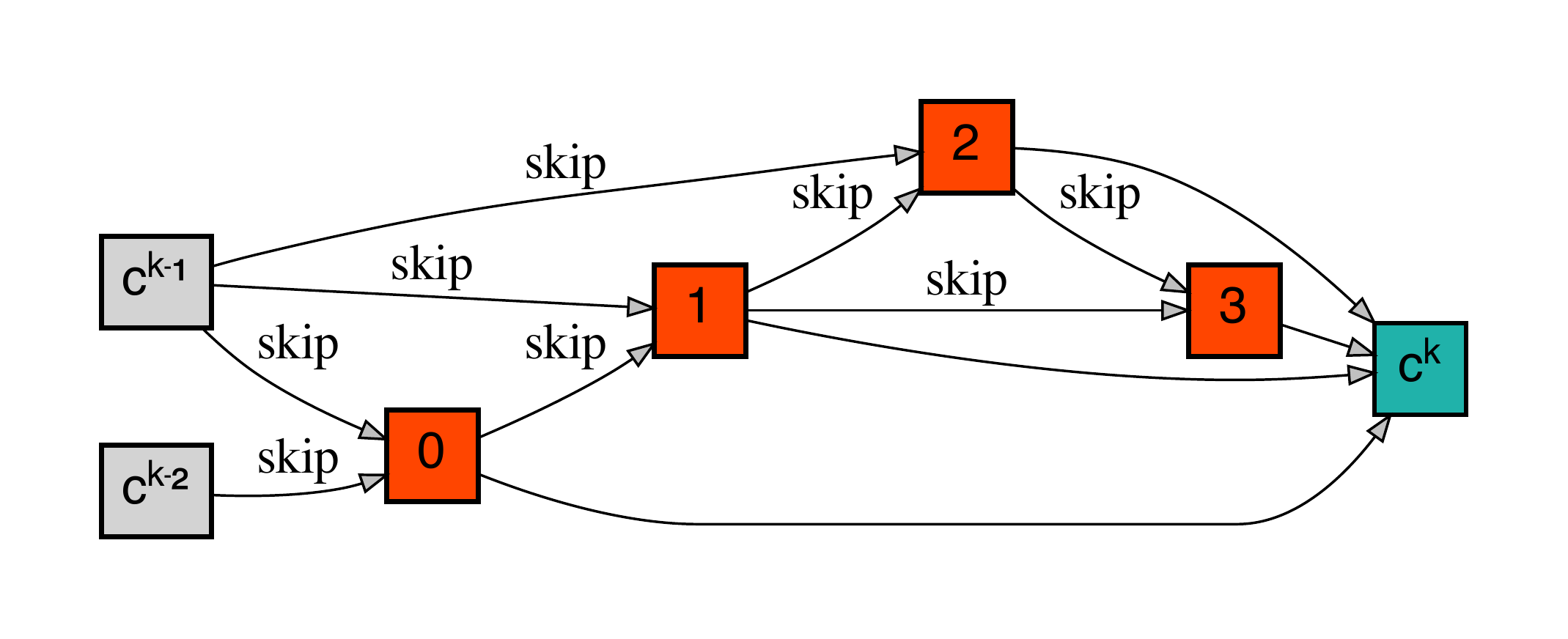}
}
\end{minipage}
\begin{minipage}{2.6in}
\subfigure[Normal]{
\includegraphics[width=0.45\columnwidth]{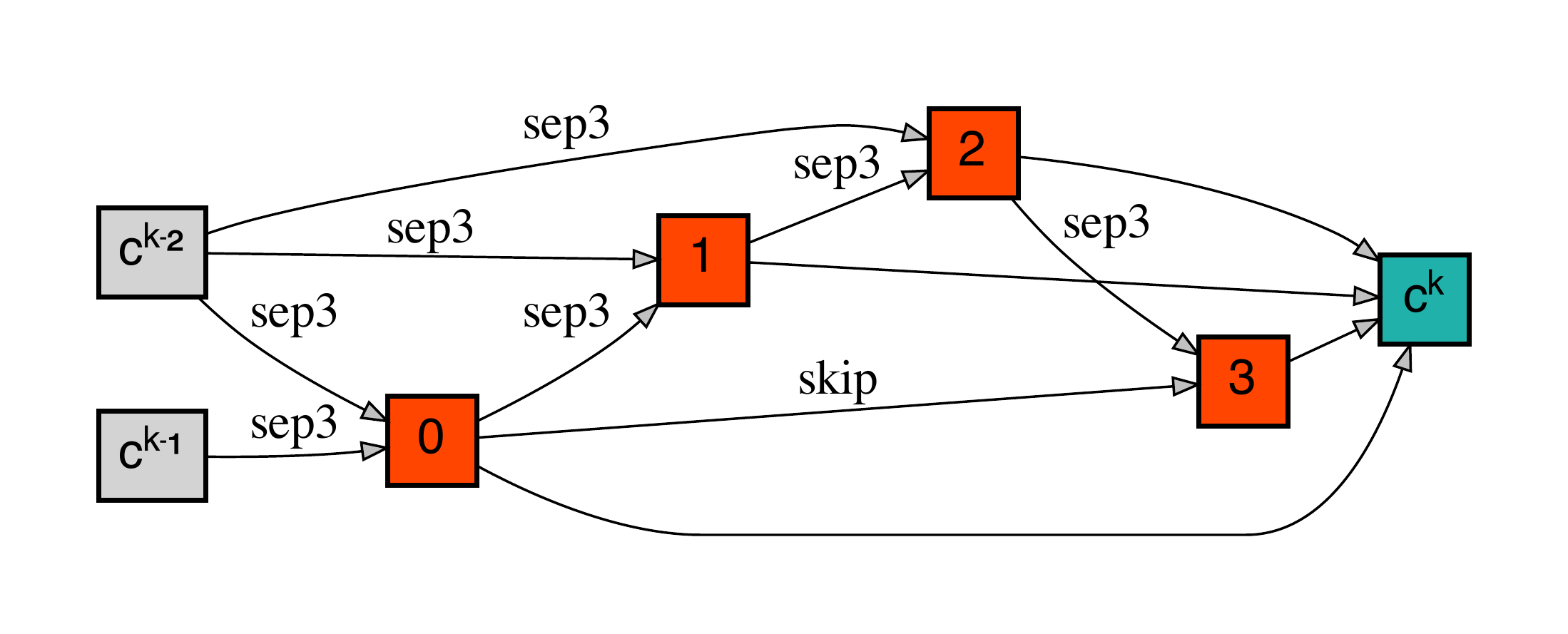} 
}
\subfigure[Reduction]{ 
\includegraphics[width=0.45\columnwidth]{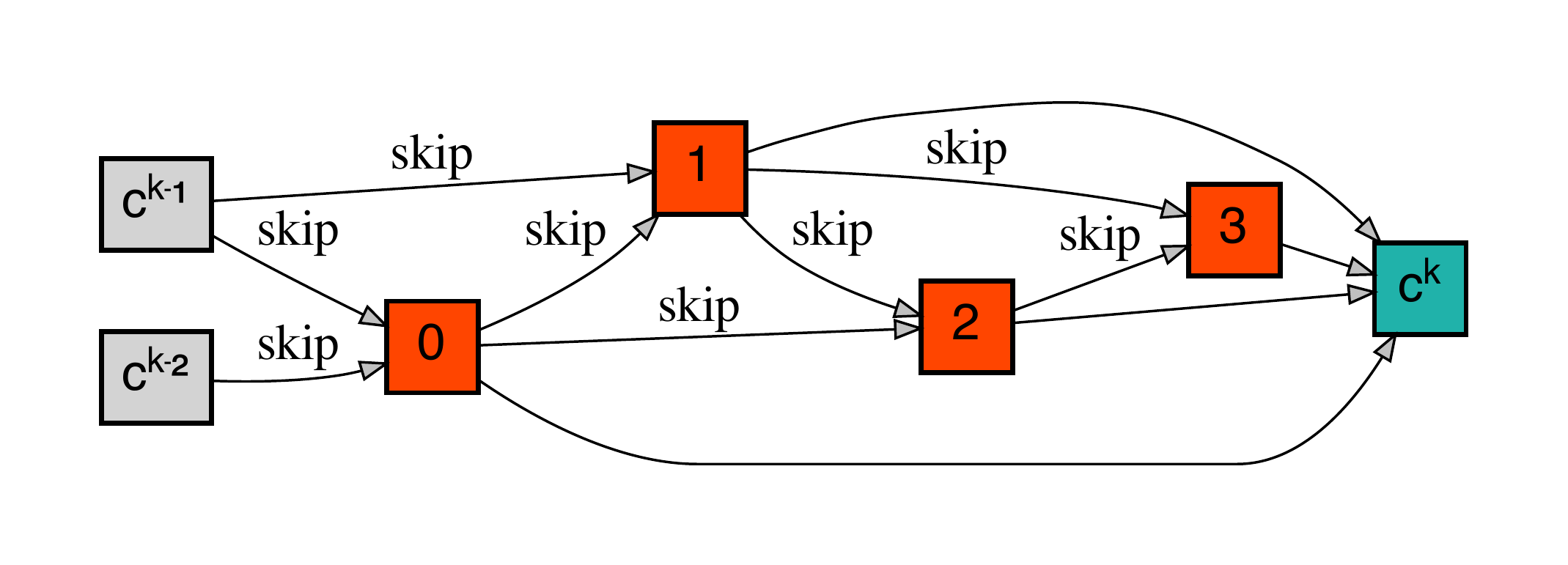}
}
\end{minipage}
\caption{Decaying $\beta_{skip}$ in the last 50 epochs during the DARTS- searching for 150 epochs in S2 on CIFAR-10 dataset.}
\label{fig:c10_s2_e150_decay_cells}
\end{figure*}

\begin{figure*}[ht]
\centering
\begin{minipage}{2.6in}
\subfigure[Normal]{
\includegraphics[width=0.45\columnwidth]{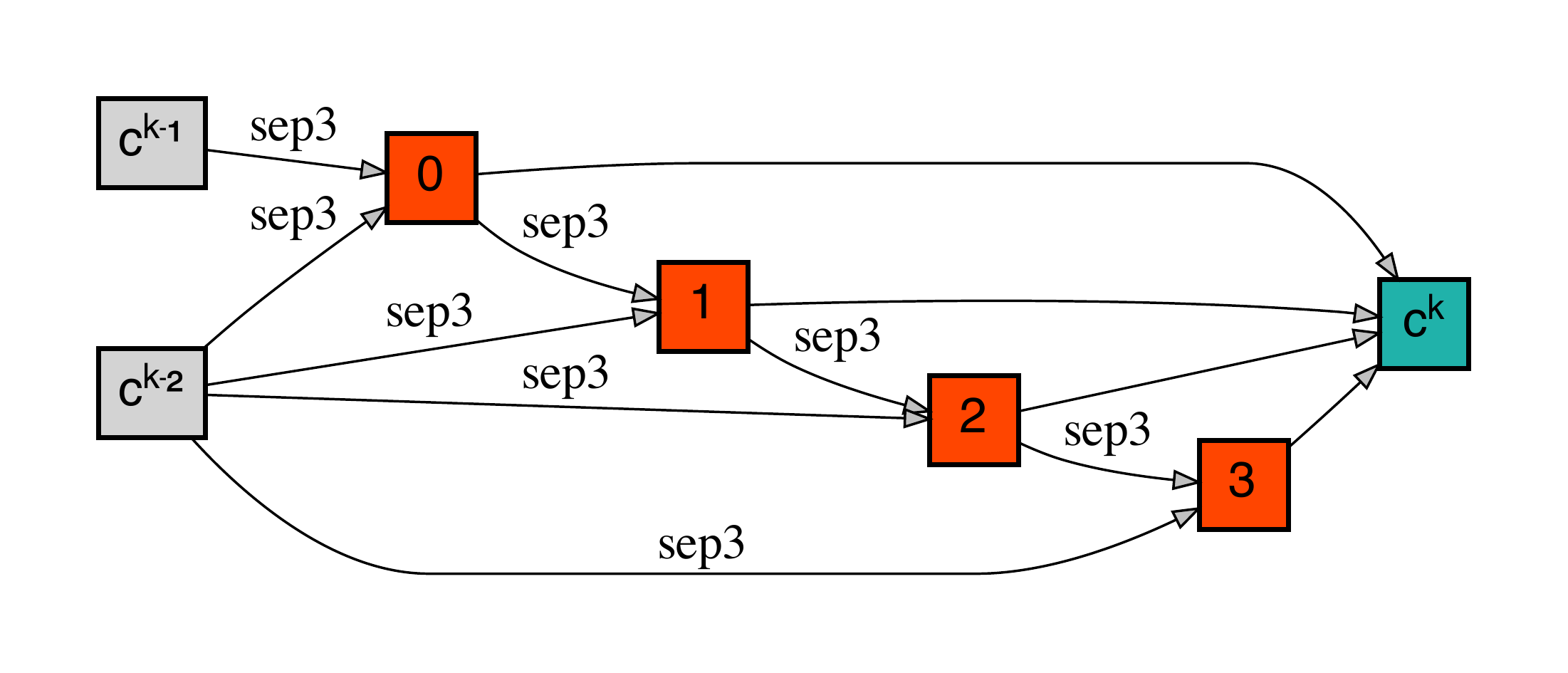} 
}
\subfigure[Reduction]{ 
\includegraphics[width=0.45\columnwidth]{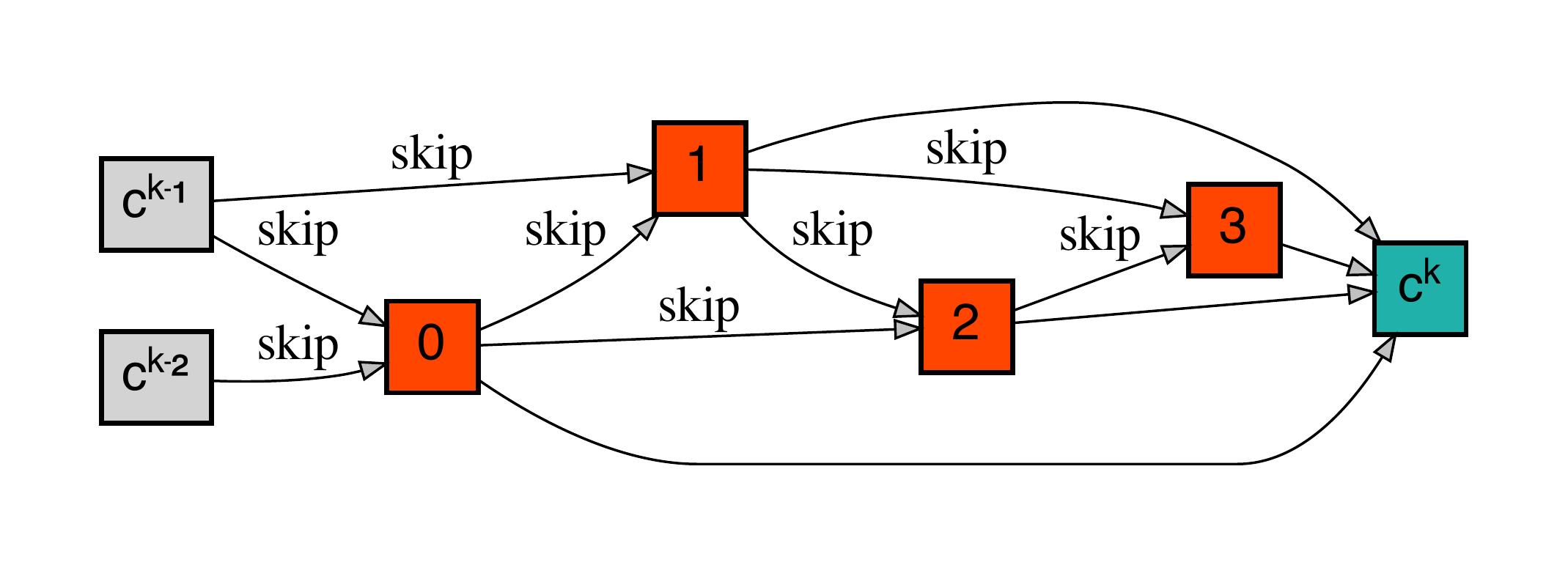}
}
\end{minipage}
\begin{minipage}{2.6in}
\subfigure[Normal]{
\includegraphics[width=0.45\columnwidth]{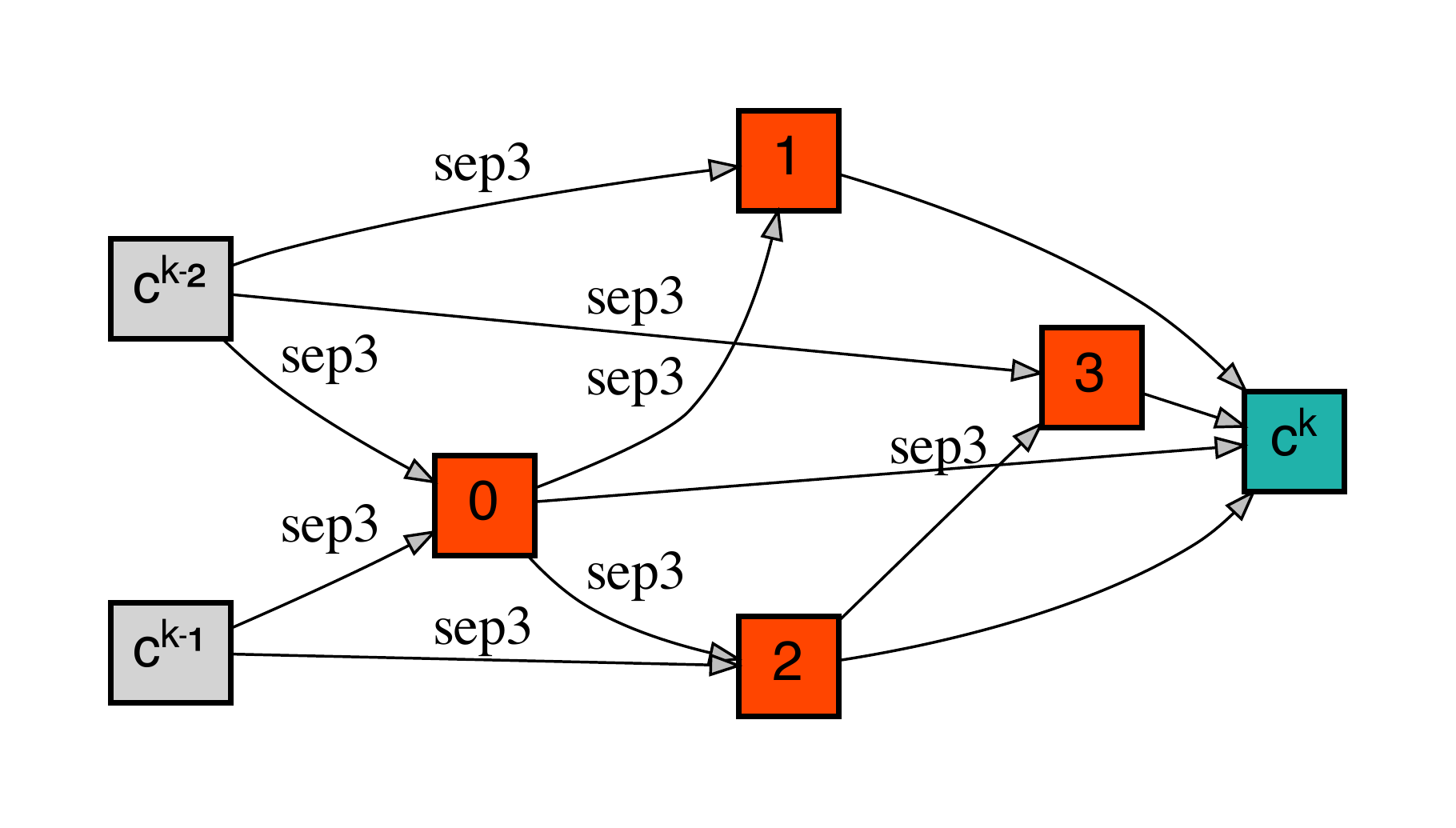} 
}
\subfigure[Reduction]{ 
\includegraphics[width=0.45\columnwidth]{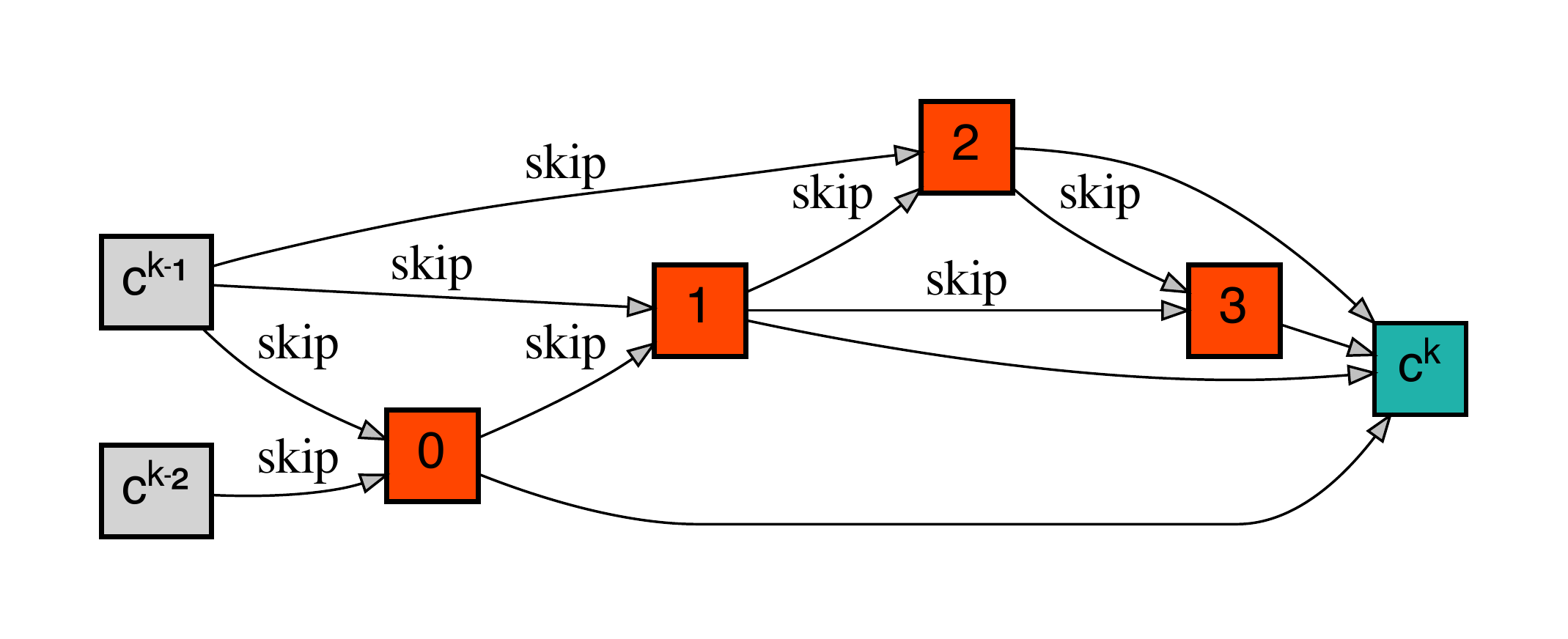}
}
\end{minipage}
\begin{minipage}{2.6in}
\subfigure[Normal]{
\includegraphics[width=0.45\columnwidth]{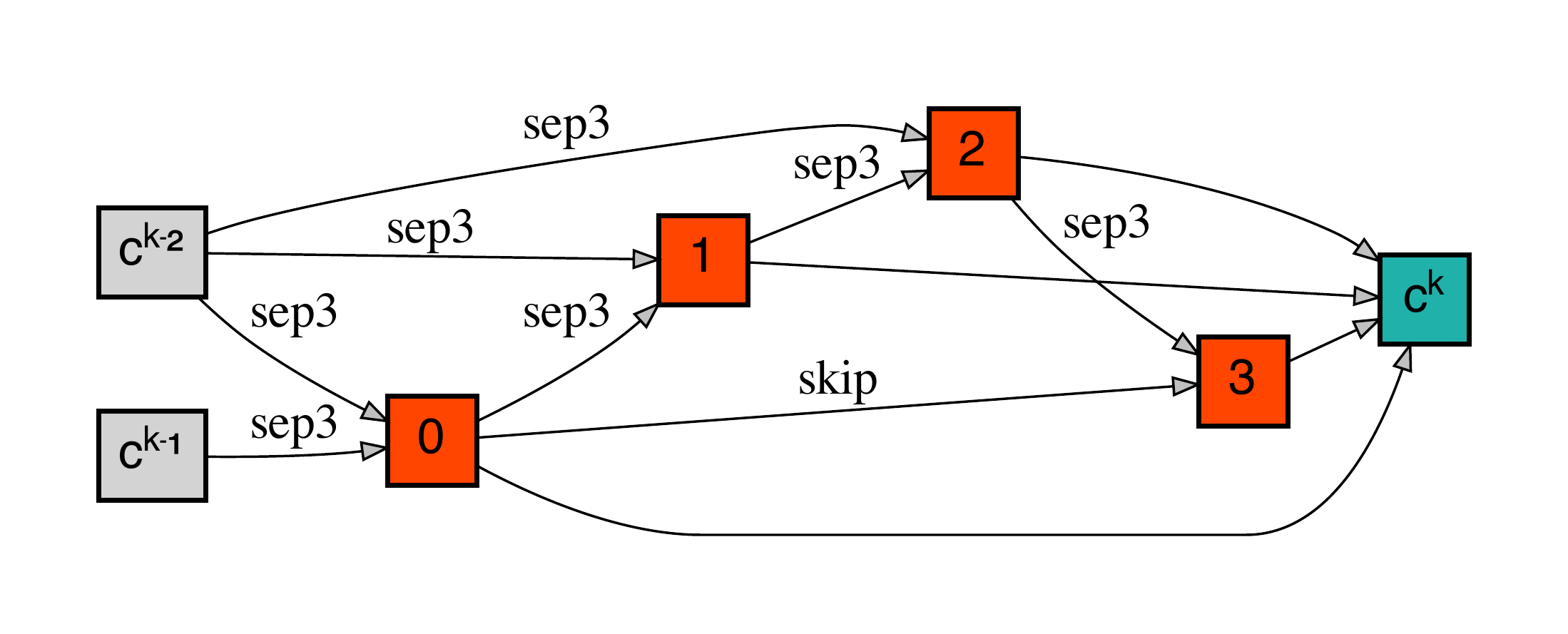} 
}
\subfigure[Reduction]{ 
\includegraphics[width=0.45\columnwidth]{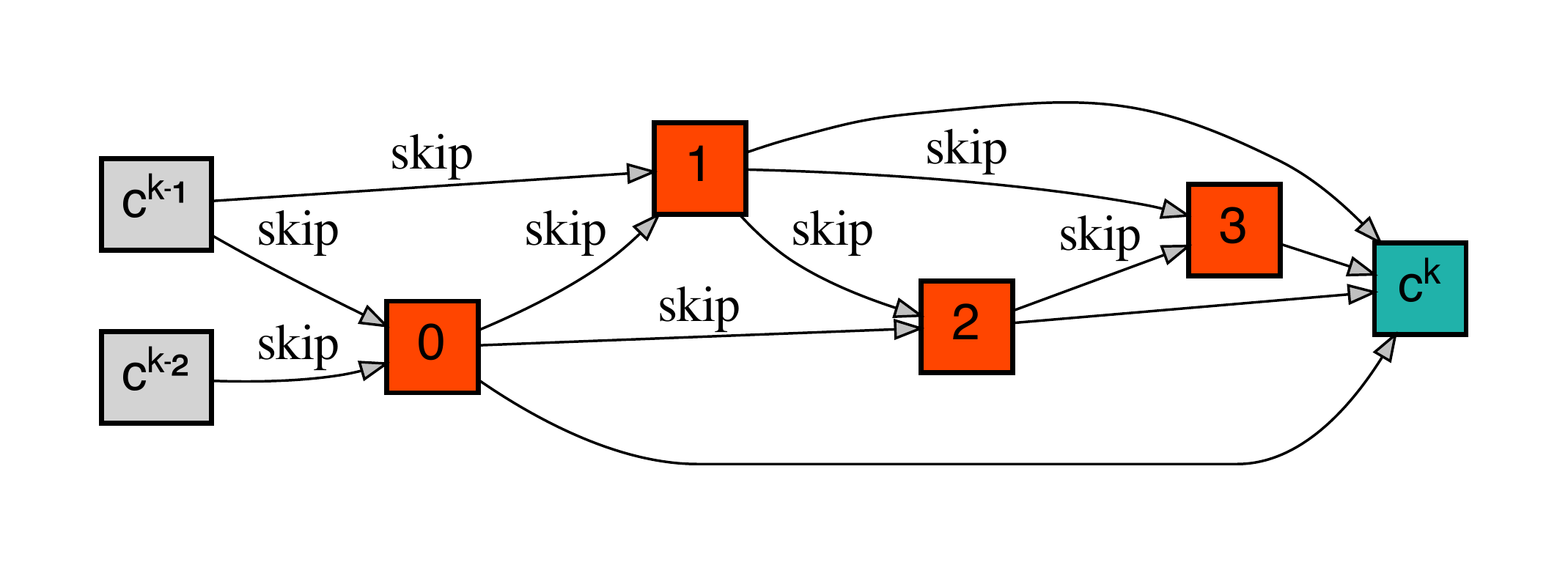}
}
\end{minipage}
\caption{Decaying $\beta_{skip}$ in the last 50 epochs during the DARTS- searching for 200 epochs in S2 on CIFAR-10 dataset.}
\label{fig:c10_s2_e200_decay_cells}
\end{figure*}

\begin{figure*}[ht]
\centering
\begin{minipage}{2.6in}
\subfigure[Normal]{
\includegraphics[width=0.35\columnwidth]{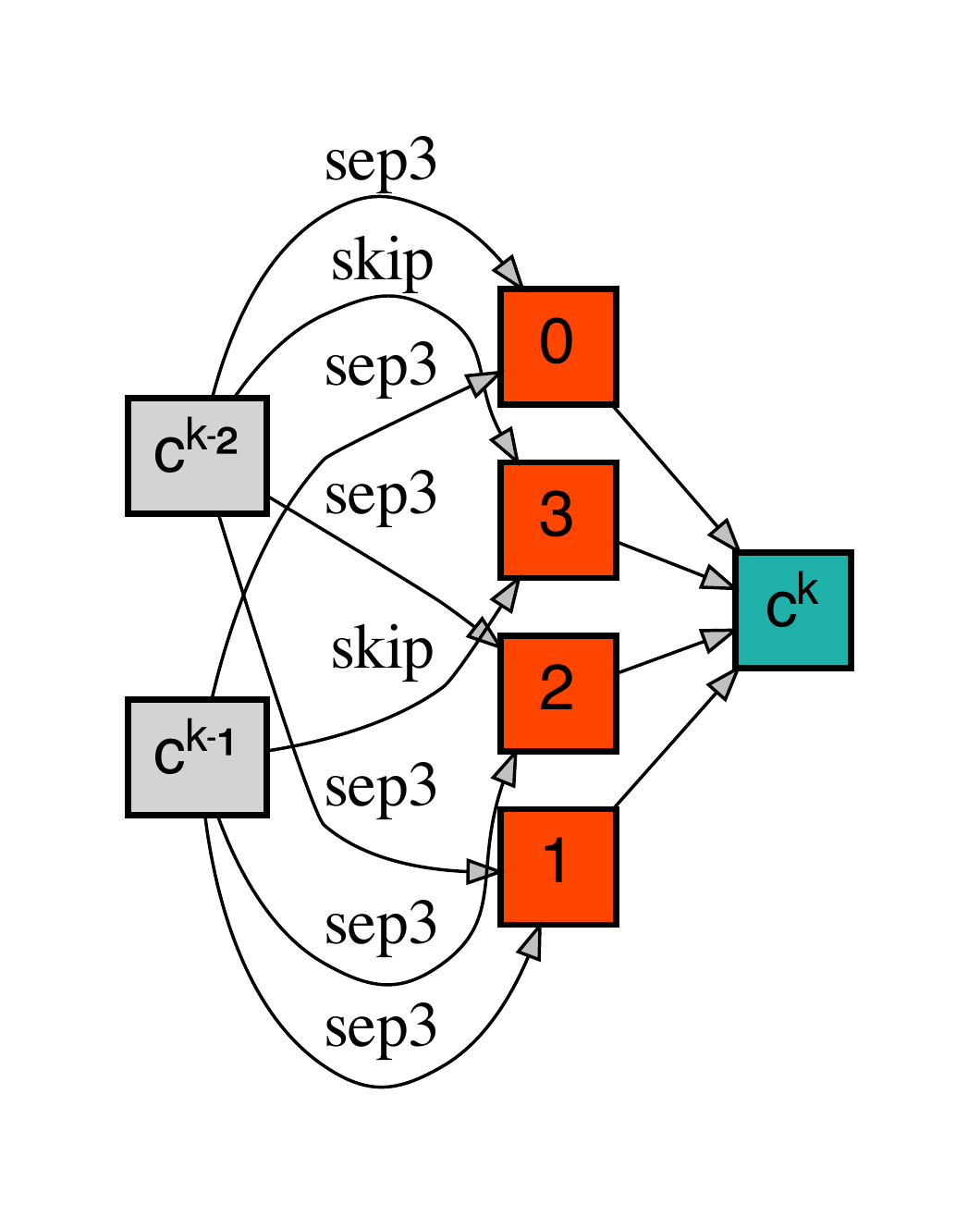} 
}
\subfigure[Reduction]{ 
\includegraphics[width=0.55\columnwidth]{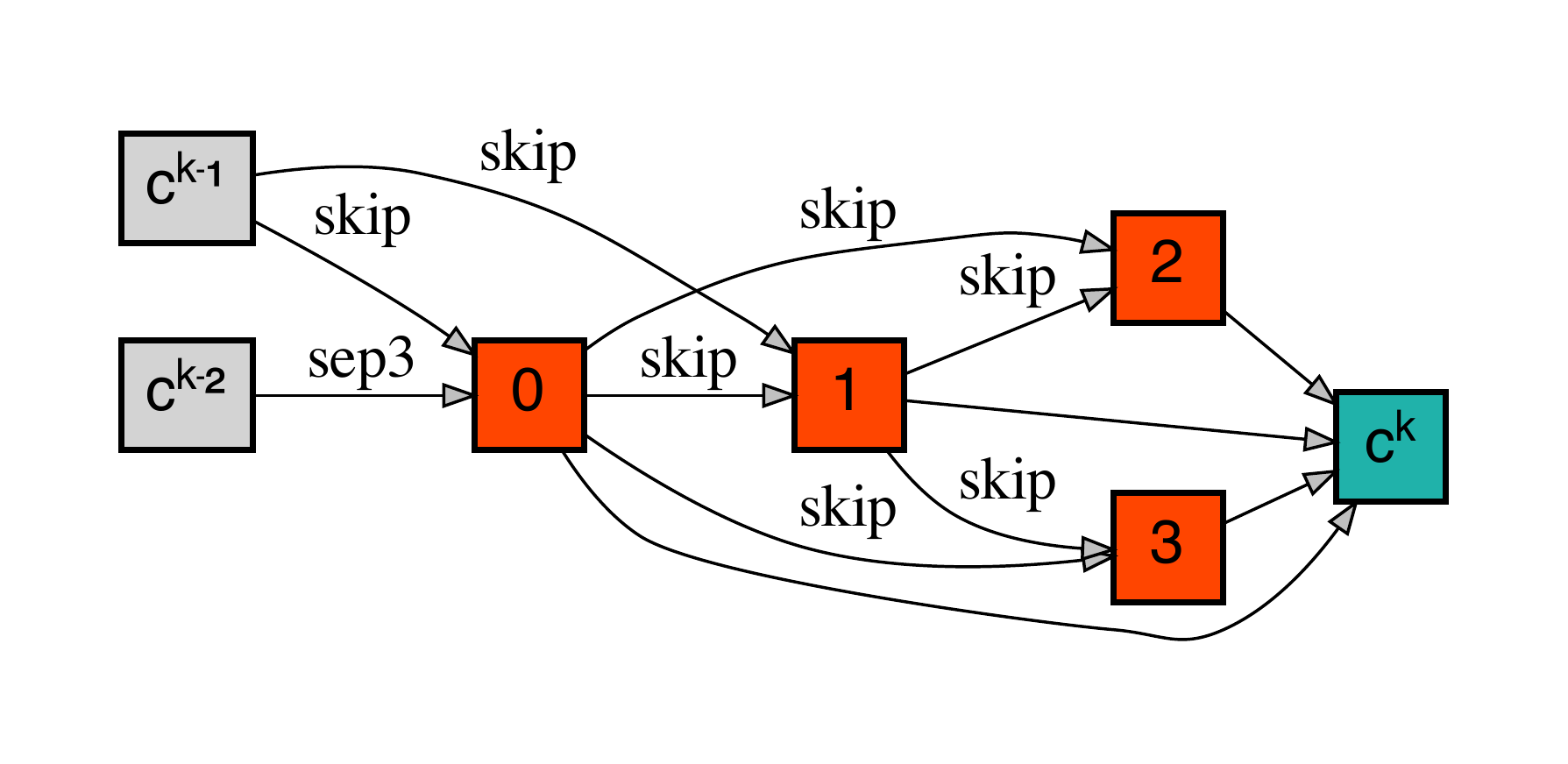}
}
\end{minipage}
\begin{minipage}{2.6in}
\subfigure[Normal]{
\includegraphics[width=0.35\columnwidth]{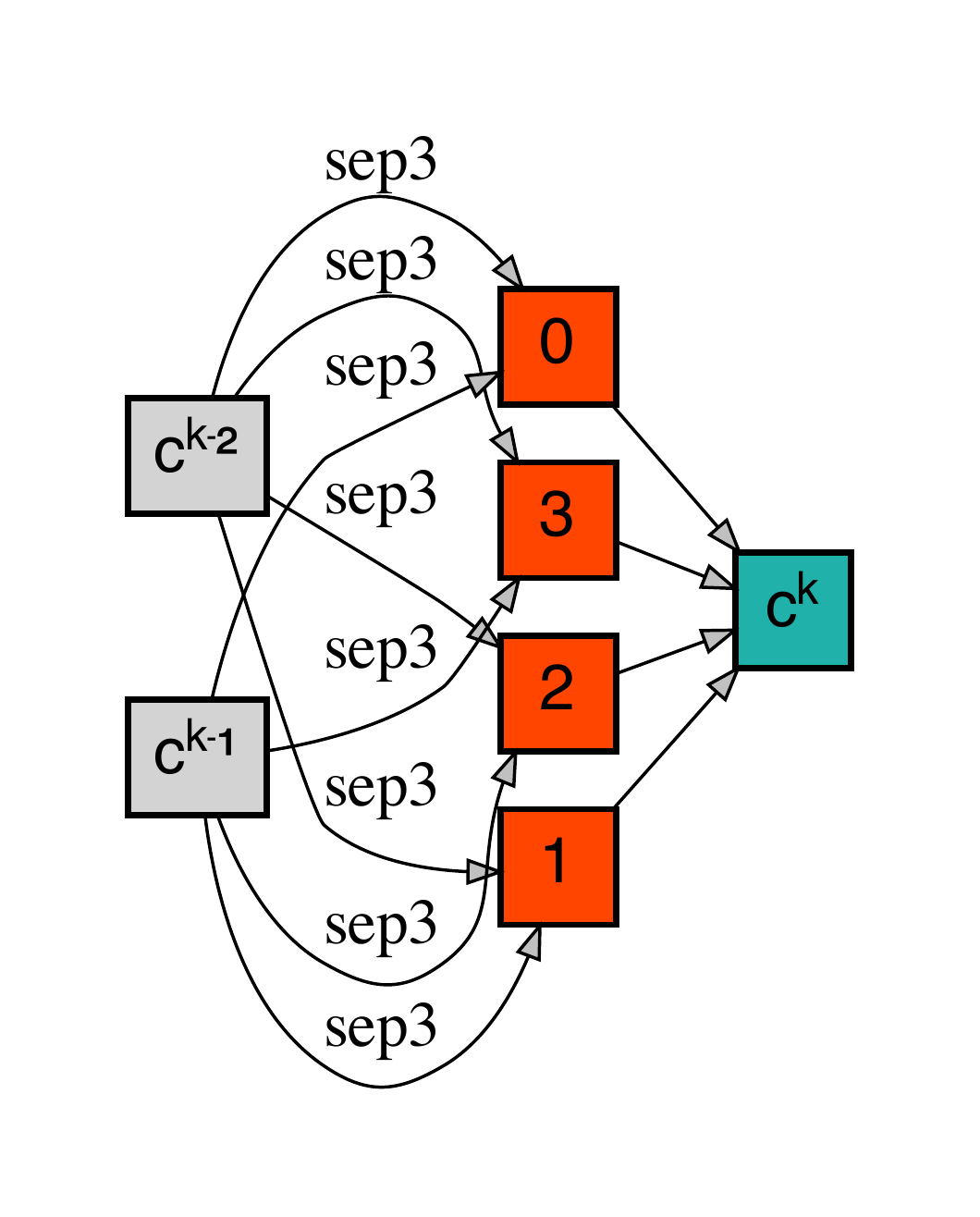} 
}
\subfigure[Reduction]{ 
\includegraphics[width=0.55\columnwidth]{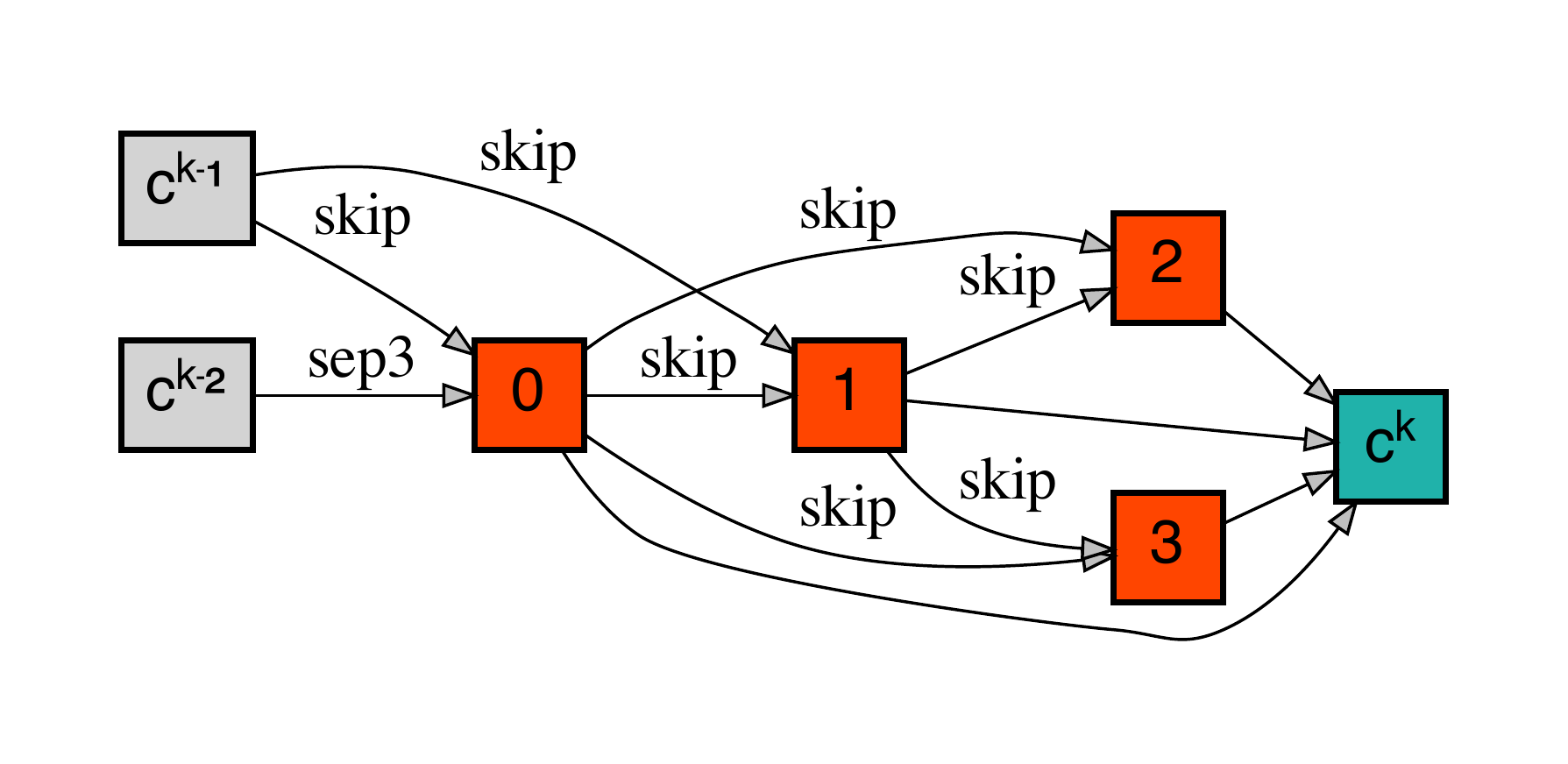}
}
\end{minipage}
\begin{minipage}{2.6in}
\subfigure[Normal]{
\includegraphics[width=0.35\columnwidth]{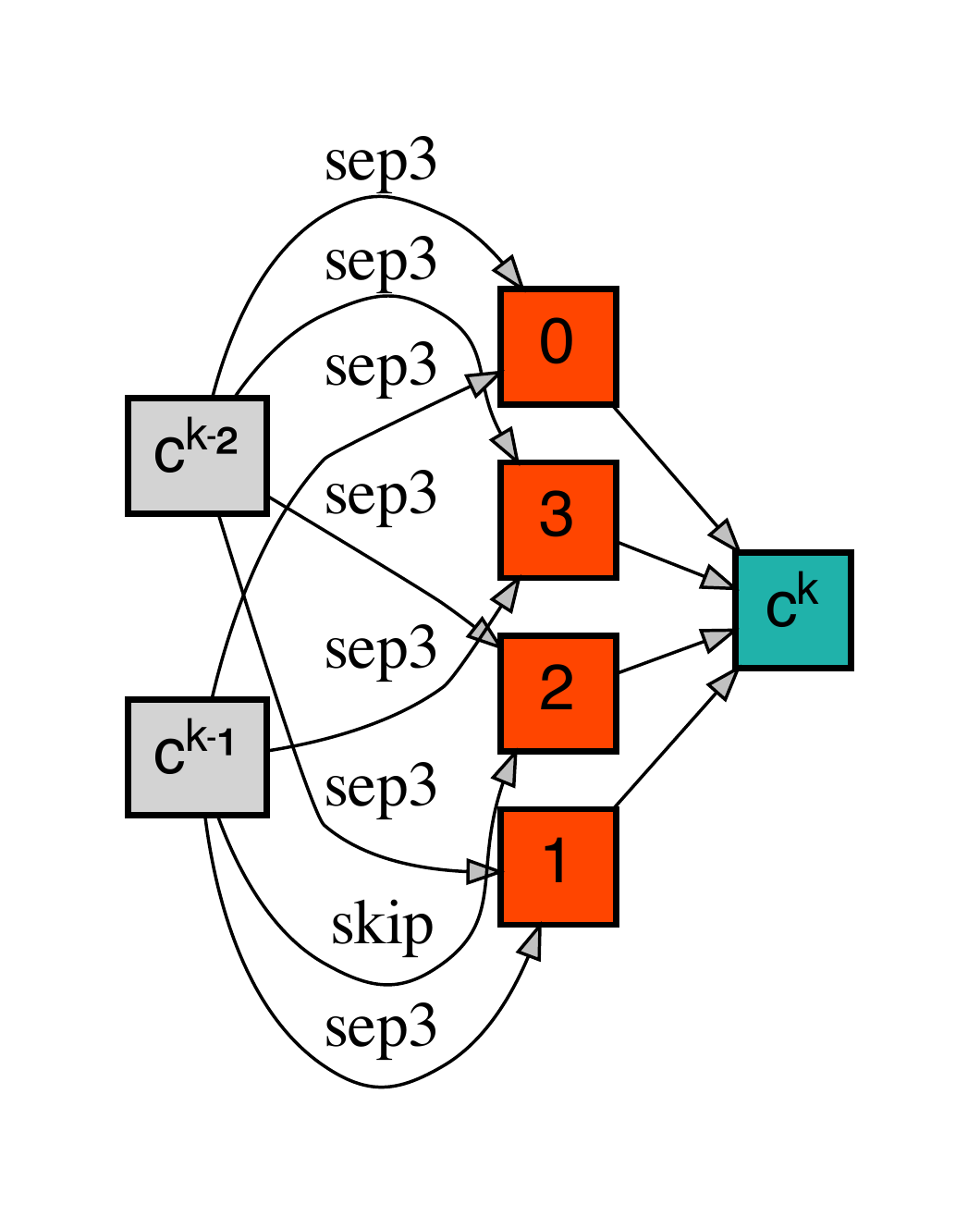} 
}
\subfigure[Reduction]{ 
\includegraphics[width=0.55\columnwidth]{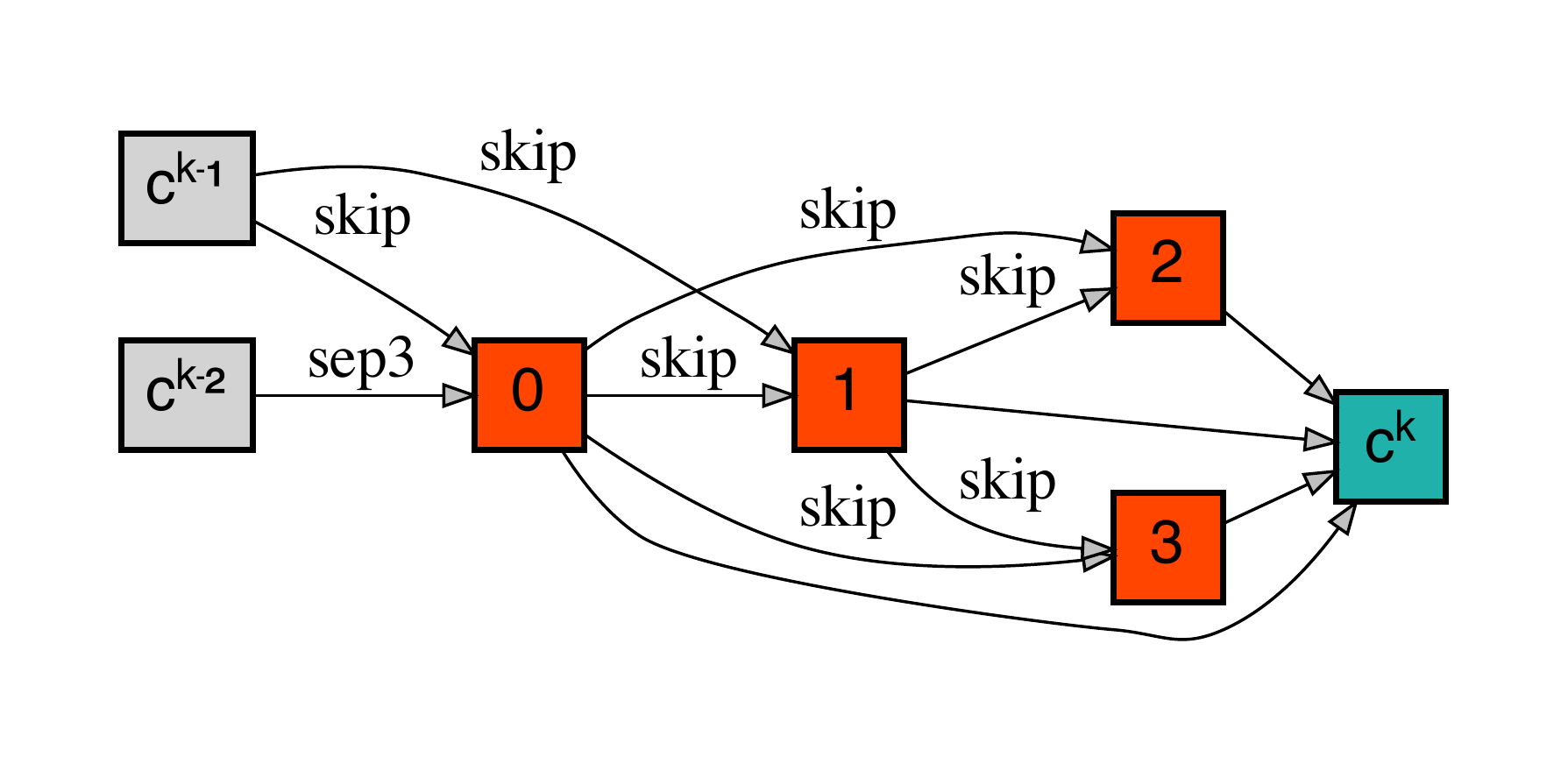}
}
\end{minipage}
\caption{Decaying $\beta_{skip}$ in the last 50 epochs during the DARTS- searching for 150 epochs in S3 on CIFAR-10 dataset.}
\label{fig:c10_s3_e150_decay_cells}
\end{figure*}

\begin{figure*}[ht]
\centering
\begin{minipage}{2.6in}
\subfigure[Normal]{
\includegraphics[width=0.35\columnwidth]{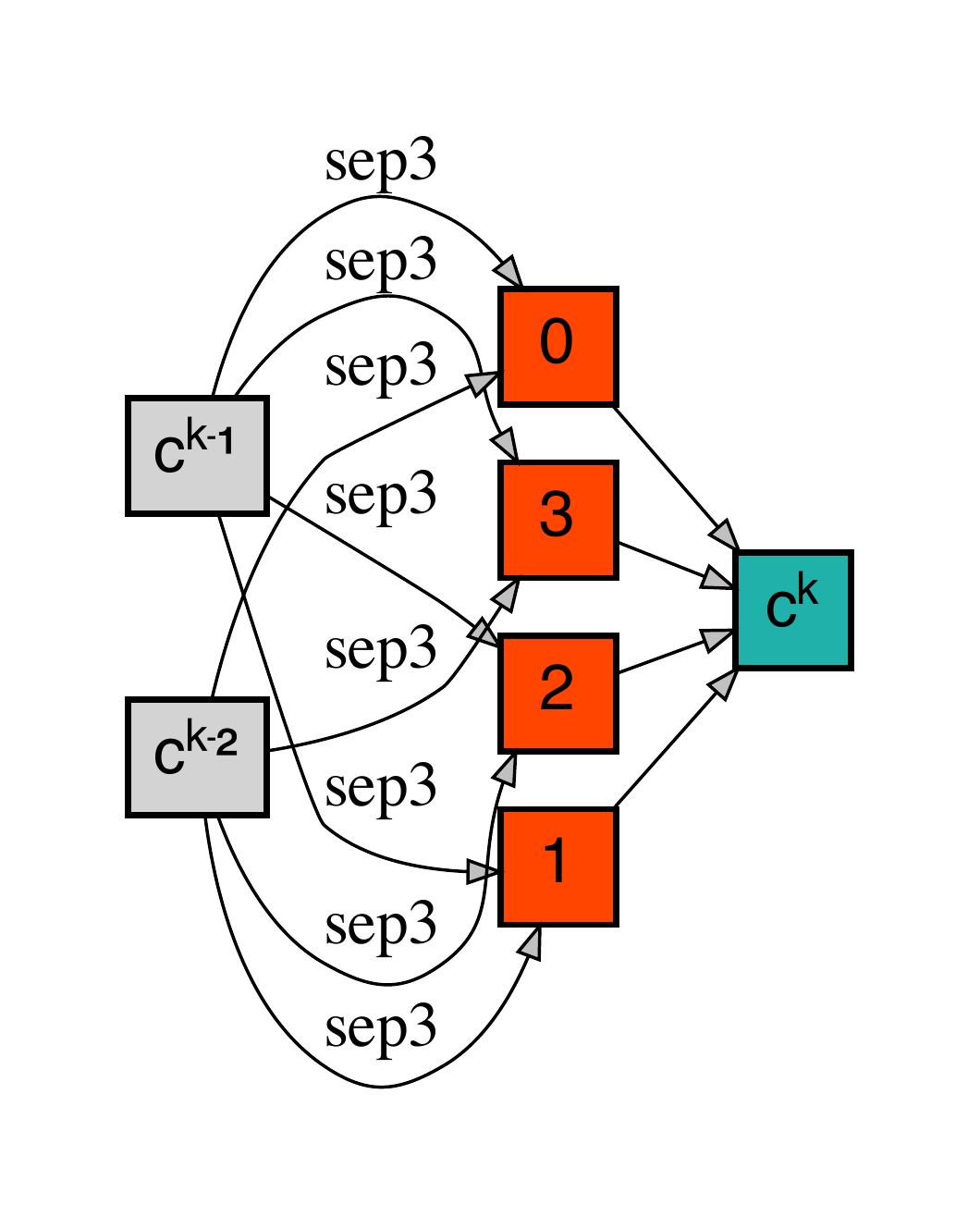} 
}
\subfigure[Reduction]{ 
\includegraphics[width=0.55\columnwidth]{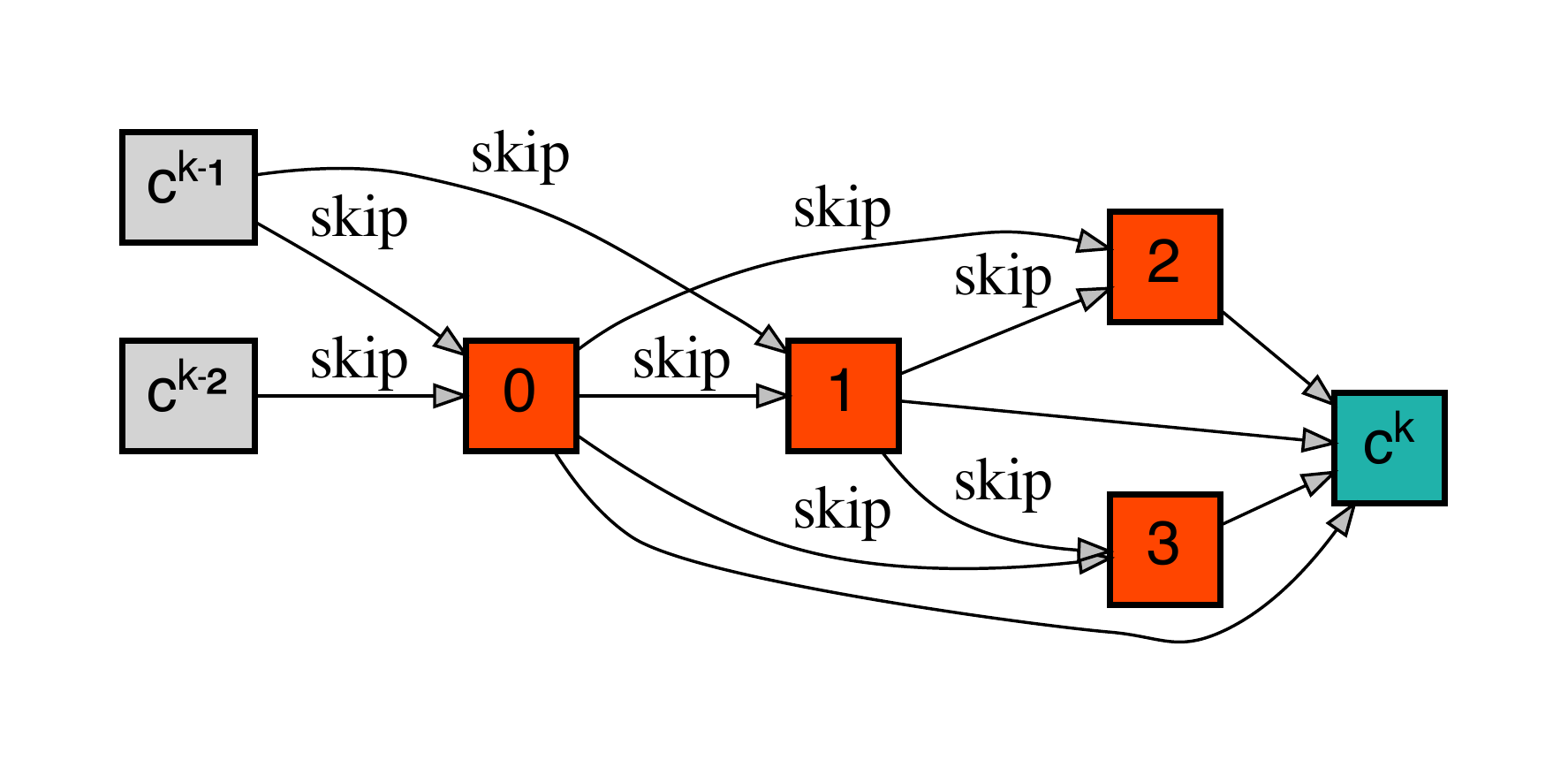}
}
\end{minipage}
\begin{minipage}{2.6in}
\subfigure[Normal]{
\includegraphics[width=0.35\columnwidth]{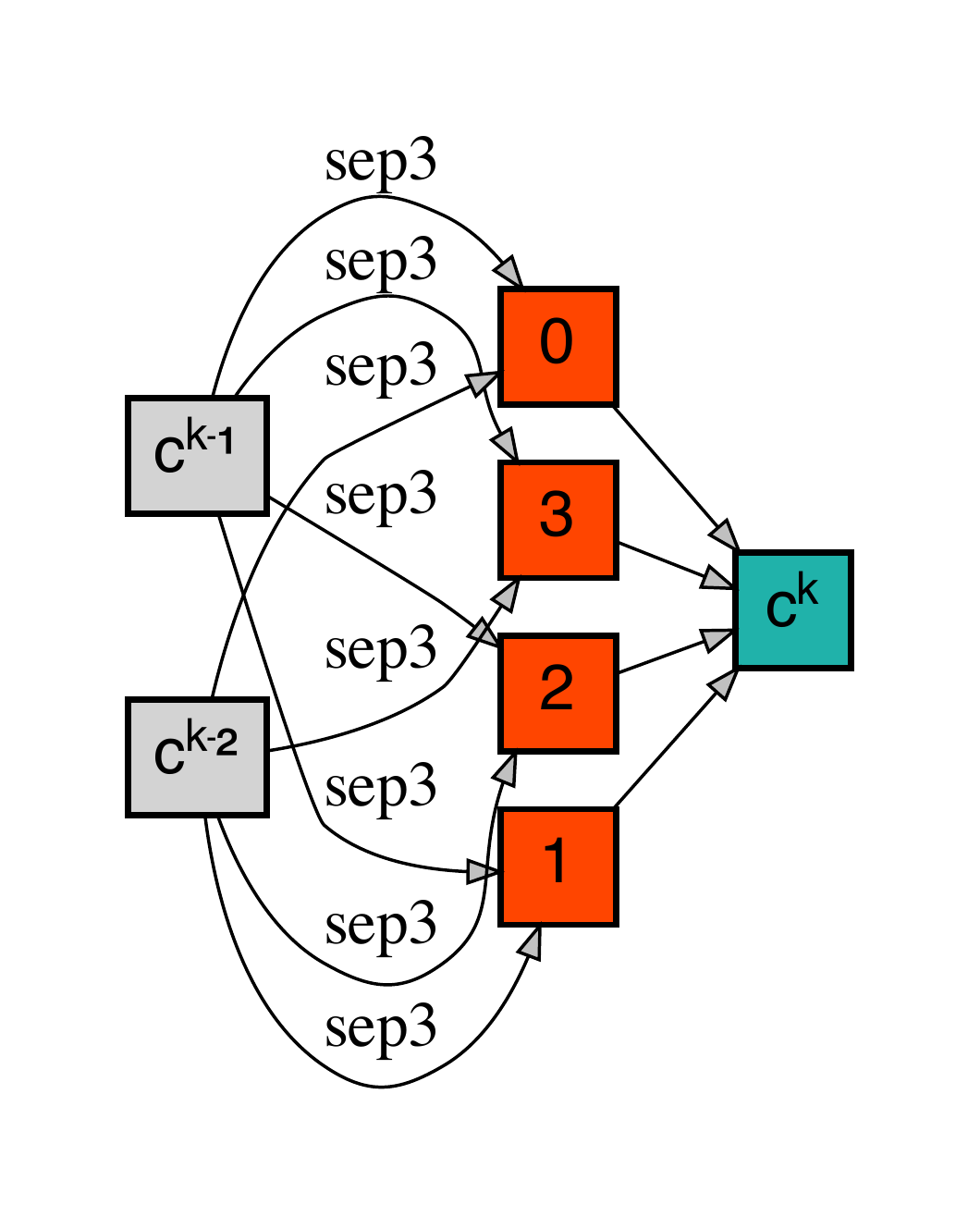} 
}
\subfigure[Reduction]{ 
\includegraphics[width=0.55\columnwidth]{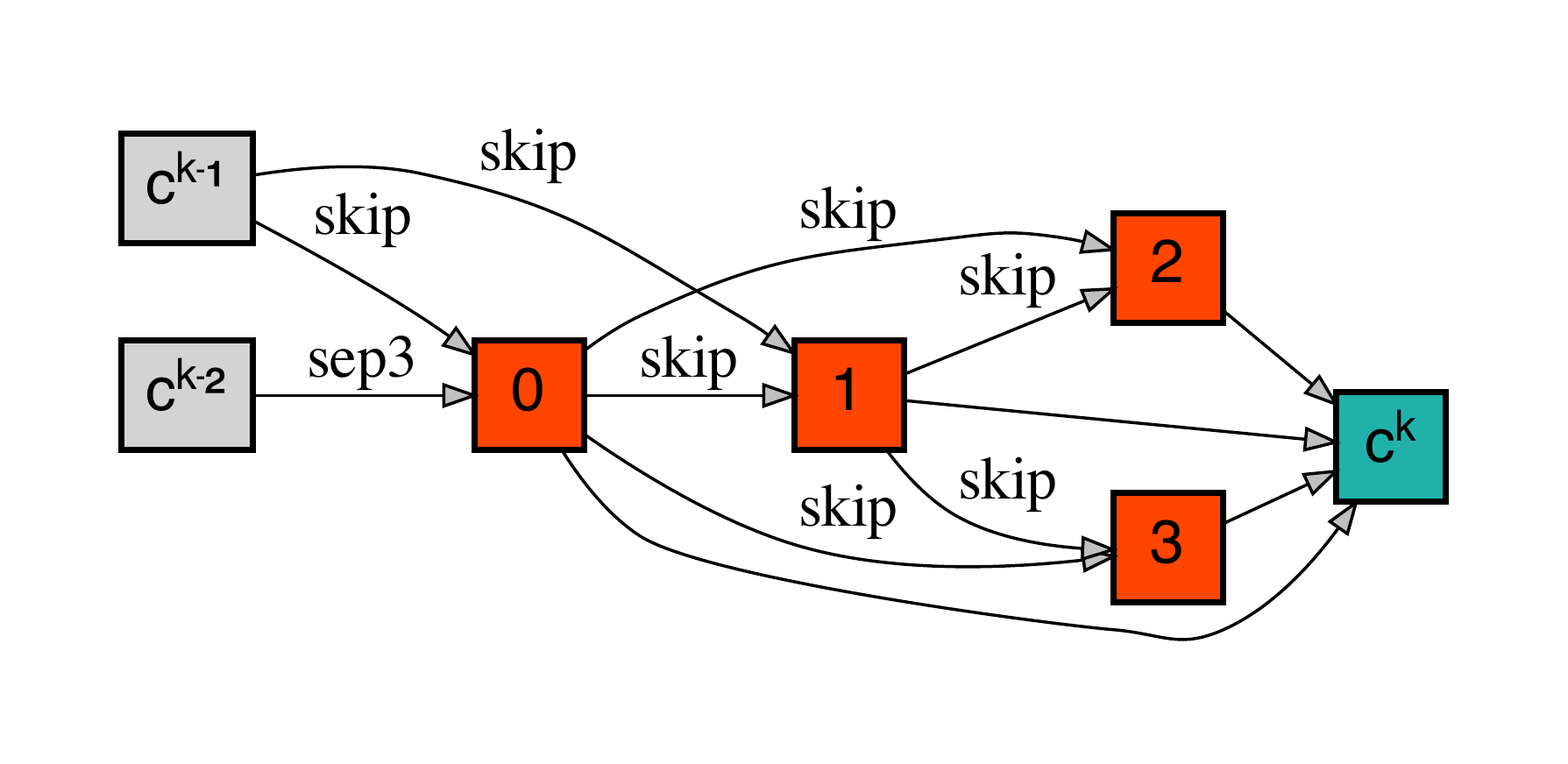}
}
\end{minipage}
\begin{minipage}{2.6in}
\subfigure[Normal]{
\includegraphics[width=0.35\columnwidth]{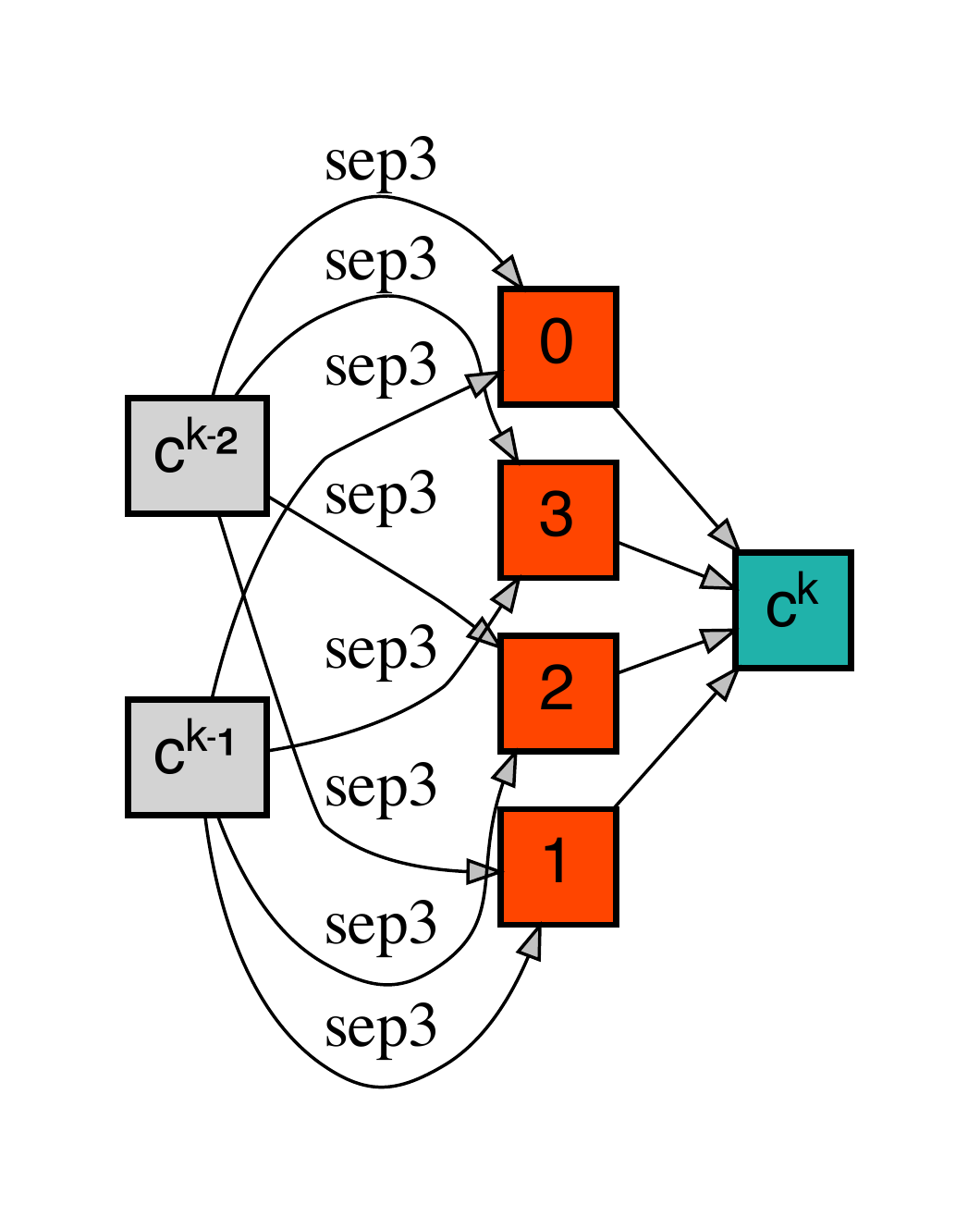} 
}
\subfigure[Reduction]{ 
\includegraphics[width=0.55\columnwidth]{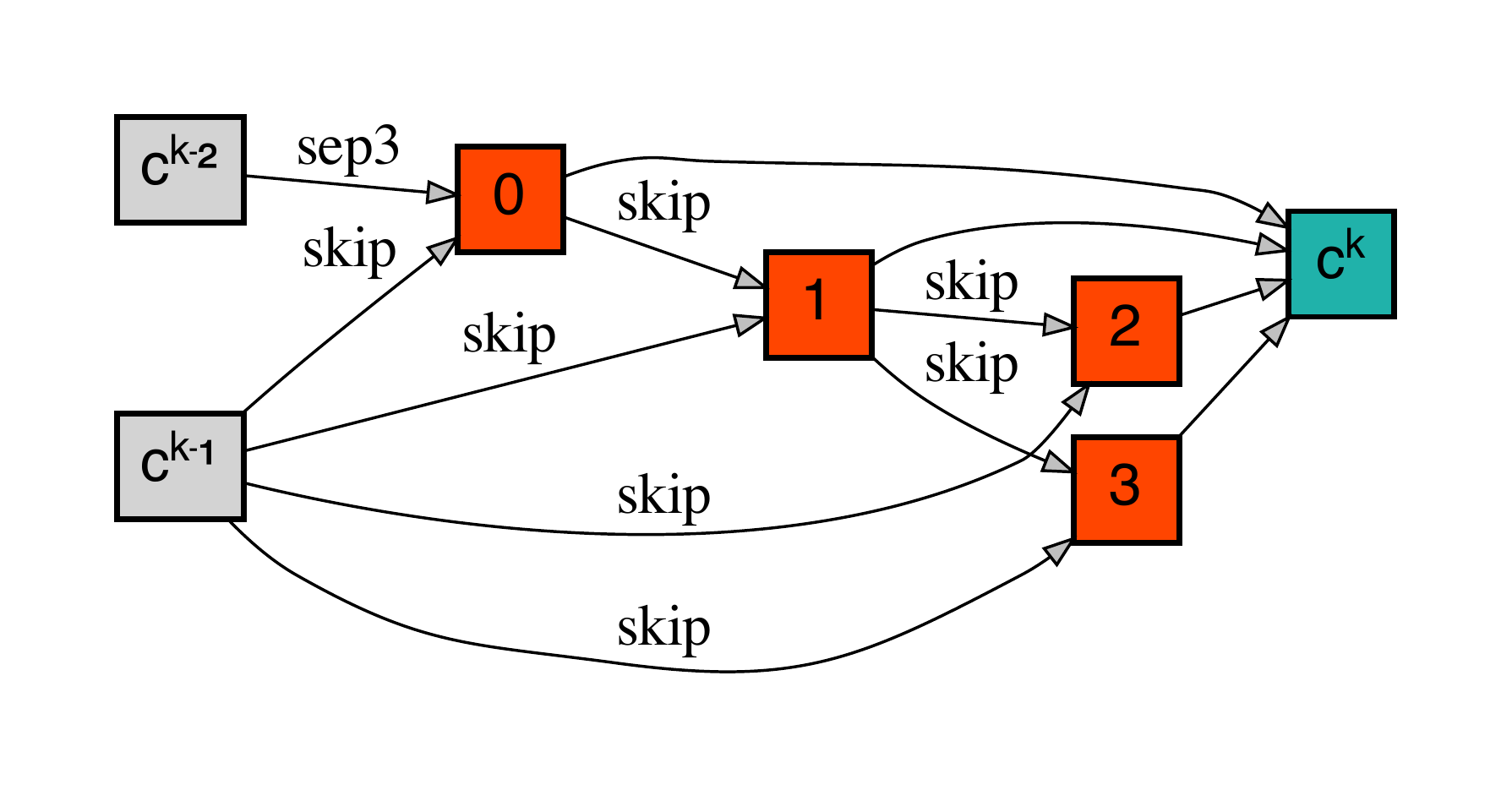}
}
\end{minipage}
\caption{Decaying $\beta_{skip}$ in the last 50 epochs during the DARTS- searching for 200 epochs in S3 on CIFAR-10 dataset.}
\label{fig:c10_s3_e200_decay_cells}
\end{figure*}

\end{document}